%% file: main.tex
\documentclass[runningheads]{llncs}

\usepackage{accv}

\usepackage{accvabbrv}
\usepackage{wrapfig}

\input{math_commands.tex}

\usepackage{graphicx}
\usepackage{placeins}
\usepackage[english]{babel}
\usepackage{url}
\usepackage{algorithm}
\usepackage{algpseudocode}
\usepackage{xcolor}
\usepackage{array}
\usepackage{multirow}
\usepackage{booktabs} 
\usepackage{amssymb}
\usepackage{amsmath}
\usepackage{caption}
\usepackage{changepage}
\usepackage{booktabs}
\usepackage{colortbl}
\usepackage{color, colortbl}
\usepackage{arydshln}
\usepackage{booktabs}
\usepackage{hyperref}

\definecolor{mygray}{rgb}{.95,.95,.95}
\newcommand{\dec}[1]{\ensuremath{_{\text{\textcolor{magenta}{(-#1)}}}}}
\newcommand{\inc}[1]{\ensuremath{_{\text{\textcolor{blue}{(+#1)}}}}}
\definecolor{Gray}{gray}{0.9}

\newcolumntype{x}[1]{>{\arraybackslash\hspace{0pt}}p{#1}}

\usepackage[accsupp]{axessibility}  

\usepackage{orcidlink}

\begin{document}

\title{ObjectCompose: Evaluating Resilience of Vision-Based Models on Object-to-Background Compositional Changes} 


\titlerunning{ObjectCompose}

\author{Hashmat Shadab Malik\thanks{Equal contribution.}\inst{1}\orcidlink{0000-0001-6197-3840} \and
Muhammad Huzaifa$^\star$\inst{1}\orcidlink{0009-0007-9250-8096} \and
Muzammal Naseer\inst{2}\orcidlink{0000-0001-7663-7161} \and
Salman Khan\inst{1,3}\orcidlink{0000-0002-9502-1749} \and
Fahad Shahbaz Khan\inst{1,4}\orcidlink{0000-0002-4263-3143} }

\authorrunning{H. Malik et al.}

\institute{ Mohamed bin Zayed University of AI \and
Center of Secure Cyber-Physical Security Systems \and
Australian National University \and
 Linköping University  
\\
\email{hashmat.malik@mbzuai.ac.ae}, 
\email{muhammad.huzaifa@mbzuai.ac.ae}, 
\email{muhammadmuzammal.naseer@ku.ac.ae},
\email{salman.khan@mbzuai.ac.ae},
\email{fahad.khan@mbzuai.ac.ae}
}

\maketitle

\begin{abstract}

Given the large-scale multi-modal training of recent vision-based models and their generalization capabilities, understanding the extent of their robustness is critical for their real-world deployment.
In this work, our goal is to evaluate the resilience of current vision-based models against diverse object-to-background context variations. The majority of robustness evaluation methods have introduced synthetic datasets to induce changes to object characteristics (viewpoints, scale, color) or utilized image transformation techniques (adversarial changes, common corruptions) on real images to simulate shifts in distributions. Recent works have explored leveraging large language models and diffusion models to generate changes in the background. However, these methods either lack in offering control over the changes to be made or distort the object semantics, making them unsuitable for the task. Our method, on the other hand, can induce diverse object-to-background changes while preserving the original semantics and appearance of the object. To achieve this goal, we harness the generative capabilities of text-to-image, image-to-text, and image-to-segment models to automatically generate a broad spectrum of object-to-background changes. We induce both natural and adversarial background changes by either modifying the textual prompts or optimizing the latents and textual embedding of text-to-image models. This allows us to quantify the role of background context in understanding the robustness and generalization of deep neural networks. We produce various versions of standard vision datasets (ImageNet, COCO), incorporating either diverse and realistic backgrounds into the images or introducing color, texture, and adversarial changes in the background. We conduct thorough experimentation and provide an in-depth analysis of the robustness of vision-based models against object-to-background context variations across different tasks. Our code and evaluation benchmark will be available at \href{https://github.com/Muhammad-Huzaifaa/ObjectCompose}{\color{Magenta}{{https://github.com/Muhammad-Huzaifaa/ObjectCompose}}}.


\end{abstract}

\section{Introduction}
\label{sec:intro}


Deep learning-based vision models have achieved significant improvement in diverse vision tasks. However, the performance on  static held-out datasets does not capture the diversity of different object background compositions present in the real world. 
Previous works have shown that vision models are vulnerable to a variety of image alterations, including common corruptions (e.g., snow, fog, blur) \cite{hendrycks2019benchmarking,moayeri2022comprehensive}, domain shifts (e.g., paintings, sketches, cartoons)\cite{he2016deep,hendrycks2021many}, and changes in viewpoint (e.g., pose, shape, orientation) \cite{chang2015shapenet,idrissi2022imagenet,bordes2023pug}. Additionally, carefully designed perturbations can be added to images to create adversarial examples that are imperceptible to humans but can fool the decision-making of vision models \cite{szegedy2013intriguing,goodfellow2014explaining}.

\begin{figure}[t]
    \centering
    \includegraphics[width=\linewidth]{
    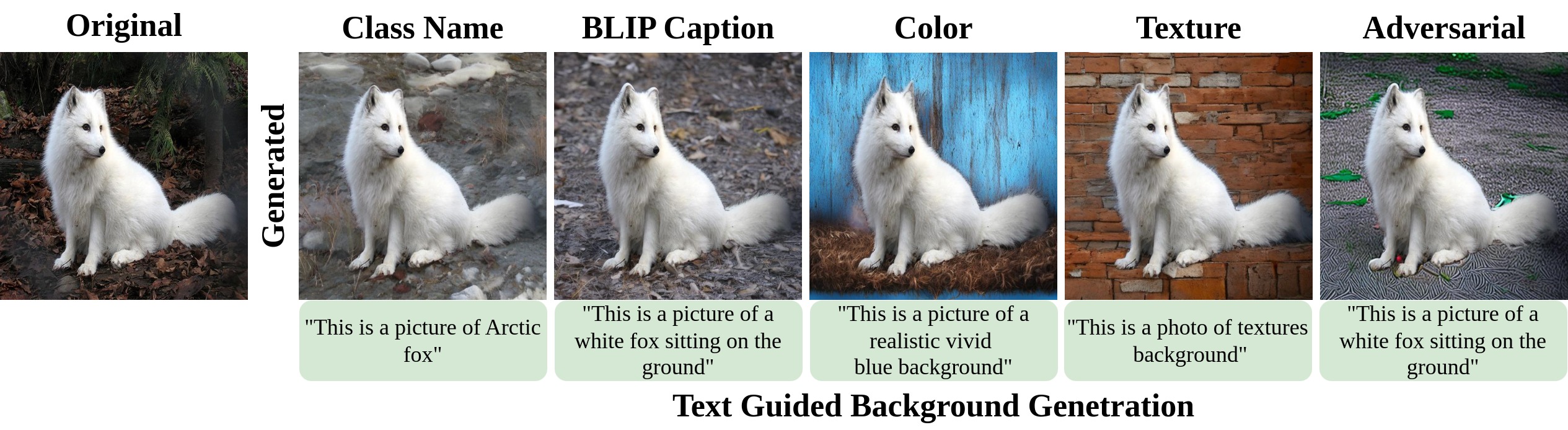
    }
    \caption{\small Image-to-background variations generated by our method, with each column representing a specific background based on the prompt below.}
    \label{motivation}
\end{figure}

Several approaches have been proposed to improve the out-of-distribution robustness of vision models. To achieve adversarial robustness, models are typically trained on adversarial examples\cite{madry2017towards}, and various augmentation policies were proposed to improve non-adversarial robustness of models\cite{zhang2017mixup,cubuk2018autoaugment,yun2019cutmix,hendrycks2021many}. More recently, the computer vision field has seen the emergence of large-scale pretraining of both vision \cite{oquab2023dinov2,kirillov2023segment} and vision-language models \cite{radford2021learning,sun2023eva,li2023blip}. Trained on large-scale datasets and multiple modalities, these models demonstrate promising performance on non-adversarial distribution shifts. Consequently, several works \cite{zhou2022learning, khattak2023maple} have adapted these models for downstream tasks by utilizing learnable prompts to preserve the rich feature space learned during pre-training.

To evaluate the vision-based models on different distribution shifts, numerous benchmarking datasets, comprising either synthetic or altered real images have been proposed. While synthetic datasets \cite{chang2015shapenet, johnson2017clevr, gondal2019transfer} offer more control over desired changes (background, shape, size, viewpoint), most of them capture only simple shape objects within a controlled environment. On the other hand, several studies  \cite{hendrycks2019benchmarking, moayeri2022comprehensive} employ coarse-grained image manipulations on the available ImageNet dataset \cite{deng2009imagenet}. However, the coarse-grained transformations used do not encompass the diverse changes that can be induced in real images.

The main motivation of our work is to understand how object-to-background compositional changes in the scene impact uni/multi-modal model performance. Recent works\cite{prabhu2023lance, li2023imagenet} have focused on leveraging existing foundational models to forge new ways to evaluate the resilience of uni/multi-modal vision models. In \cite{prabhu2023lance}, large language models and text-to-image diffusion models are used for generating diverse semantic changes in real images. However, their method employs the prompt-to-prompt method\cite{hertz2022prompt} for image editing, allowing limited word changes in the textual prompt to preserve object semantics. We also observe that it suffers from object distortion due to absence of strong guidance between the object and background during image editing. In \cite{li2023imagenet}, diffusion models are used for background editing in real images, and  ImageNet-E(diting) dataset is introduced for benchmarking. However, their use of a frequency-based loss for guiding the generation process of diffusion models limit the control to attribute changes in the background. This imposes limitations on the type of background-compositions that can be achieved.

In this work, we develop a framework to investigate the resilience of vision models to diverse object-to-background changes. Leveraging the complementary strengths of image-to-text \cite{li2023blip}, image-to-segment \cite{kirillov2023segment}, and text-to-image \cite{ho2020denoising, rombach2022high} models, our approach better handles complex background variations. We preserve object semantics (Figure \ref{motivation}) by conditioning the text-to-image diffusion model on object boundaries and textual descriptions from image-to-segment and image-to-text models. We guide the diffusion model by adding the desired textual description or optimizing its latent visual representation and textual embedding for generating  diverse natural and adversarial background changes. Additionally, we produce datasets with varied backgrounds from subsets of ImageNet \cite{deng2009imagenet} and COCO \cite{lin2015microsoft}, facilitating the evaluation of uni-modal and multi-modal models. Our contributions are as follows:
\begin{itemize}
    \item We propose \textsc{ObjectCompose}, an automated approach to introduce diverse background changes to real images, allowing us to evaluate resilience of modern vision-based models against object-to-background context.
    \item  Our proposed background changes yield an average performance drop of $13.64\%$ on classifier models compared to the baseline method, and a substantial drop of $68.71\%$ when exposed to adversarial changes (see Table \ref{tab:base_class_comparison}). 
    \item Object detection and segmentation models, which incorporate object-to-background context, display reasonably better robustness to background changes than classification models (see Table \ref{tab:AP} and Figure \ref{detr}). 
    \item Models trained on large-scale datasets with scalable and stable training show better robustness against background changes (see Figure \ref{fig:dinov2} and Table \ref{tab:base_zs_comparison}).
\end{itemize}

\section{Related Work}

\textbf{Common Corruptions.} In \cite{zhu2016object}, different datasets are curated by separating foreground and background elements using ImageNet-1k bounding boxes. They found that models could achieve high object classification performance even when the actual object was absent. Similarly, \cite{rosenfeld2018elephant} demonstrate that subtle changes in object positioning could significantly impact the detector's predictions, highlighting the sensitivity of these models to spatial configurations.
A related approach by \cite{shetty2019not} focuses on co-occurring objects within an image and investigates if removing one object affected the response of the target model toward another. 
\cite{xiao2020noise} analyze the models' reliance on background signals for decision-making by training on various synthetic datasets.
\cite{hendrycks2019benchmarking} benchmark the robustness of classifiers against common corruptions and perturbations like fog, blur, and contrast variations. In subsequent work, \cite{hendrycks2021natural} introduce ImageNet-A dataset, filtering natural adversarial examples from a subset of ImageNet to limit spurious background cues.  Also, \cite{hendrycks2021many} introduce the ImageNet-R dataset, which comprises various renditions of object classes under diverse visual representations such as paintings, cartoons, embroidery, sculptures, and origami. Similarly, \cite{moayeri2022comprehensive} introduce the RIVAL10 dataset to study Gaussian noise corruptions in the foreground, background, and object attributes.

 \textbf{Viewpoint Changes.} \cite{chang2015shapenet, gondal2019transfer, alcorn2019strike} introduce a large-scale 3D shape datasets to study object scale and viewpoints variations. In a similar vein, \cite{johnson2017clevr} introduce a synthetic dataset of rendered objects to aid in diagnostic evaluations of visual question-answering models. Later works have made strides in addressing the realism gap. \cite{barbu2019objectnet} utilize crowd-sourcing to control rotation, viewpoints, and backgrounds of household objects, while  \cite{idrissi2022imagenet}  provide more fine-grained annotations for variations on the ImageNet validation set. In a recent development, \cite{bordes2023pug} released PUG dataset rendered using  Unreal Engine under diverse conditions, including varying sizes, backgrounds, camera orientations, and light intensities. While these methods offer control over changing several attributes in images, they lack in realism and are not suitable for our primary goal of studying object-to-background context in real images. In contrast, our proposed framework can generate a wide range of object-to-background compositional changes that can influence the models performance.

 \textbf{Adversarial and Counterfactual Manipulations.} Researchers have uncovered that subtle, carefully designed alterations to an image, imperceptible to human observers, have the ability to deceive deep learning models \cite{szegedy2013intriguing,goodfellow2014explaining,kurakin2016adversarial}. These perturbations, constructed using gradient-based methods, serve as a worst-case analysis in probing the model's robustness within specified distance norm metrics ($l_2$ or $l_\infty$). 
Another strategy entails applying unbounded perturbations to specific image patches, thereby conserving object semantics while inducing model confusion \cite{sharma2022adversarial,fu2022patch}. Several studies leverage generative models to create  adversarial alterations in images.  \cite{christensen2022assessing, chen2023diffusion} utilise diffusion model and GANs to introduce global adversarial perturbations in the image with strong constraint to semantic changes in order to preserve the original layout of the scene. More recent works\cite{li2023imagenet, prabhu2023lance} are more closely related to our goal of evaluating vision-based models on object-to-background compositional changes in the scene. In \cite{prabhu2023lance}, LANCE framework is proposed which utilises fine-tuned large language models to get the modified textual prompt for editing of  attributes in real images. However, this framework is not ideal for studying object-to-background compositional changes since the prompt-to-prompt\cite{hertz2022prompt} based image editing method often leads to global changes in the scene, often altering the object semantics. This necessitates hyper-parameter tuning of parameters used for prompt-to-prompt editing, leading to  generation of multiple edited image versions and selecting the one most faithful to the original image. In \cite{li2023imagenet}, using already available masks of ImageNet dataset\cite{gao2022large}, diffusion model is utilised to alter the background of images by varying its texture. A complexity loss based on gray-level co-occurence matrix\cite{haralick1973textural} of the image is used during the denoising process to vary the complexity of the background. A concurrent work\cite{zhang2024imagenet} evaluates the resilience of models on synthetic images where both the object and background are generated using a diffusion model. In contrast to previous works, our method induces natural/adversarial background variations in real images through textual guidance and optimization of latent space of the diffusion model, all while maintaining the integrity of the object semantics. Our method can be applied to standard vision datasets to generate diverse background variations, providing a robust benchmark for evaluating vision-based models.

\section{Method}

\begin{figure}[t]
    \centering
    \includegraphics[width=\linewidth]{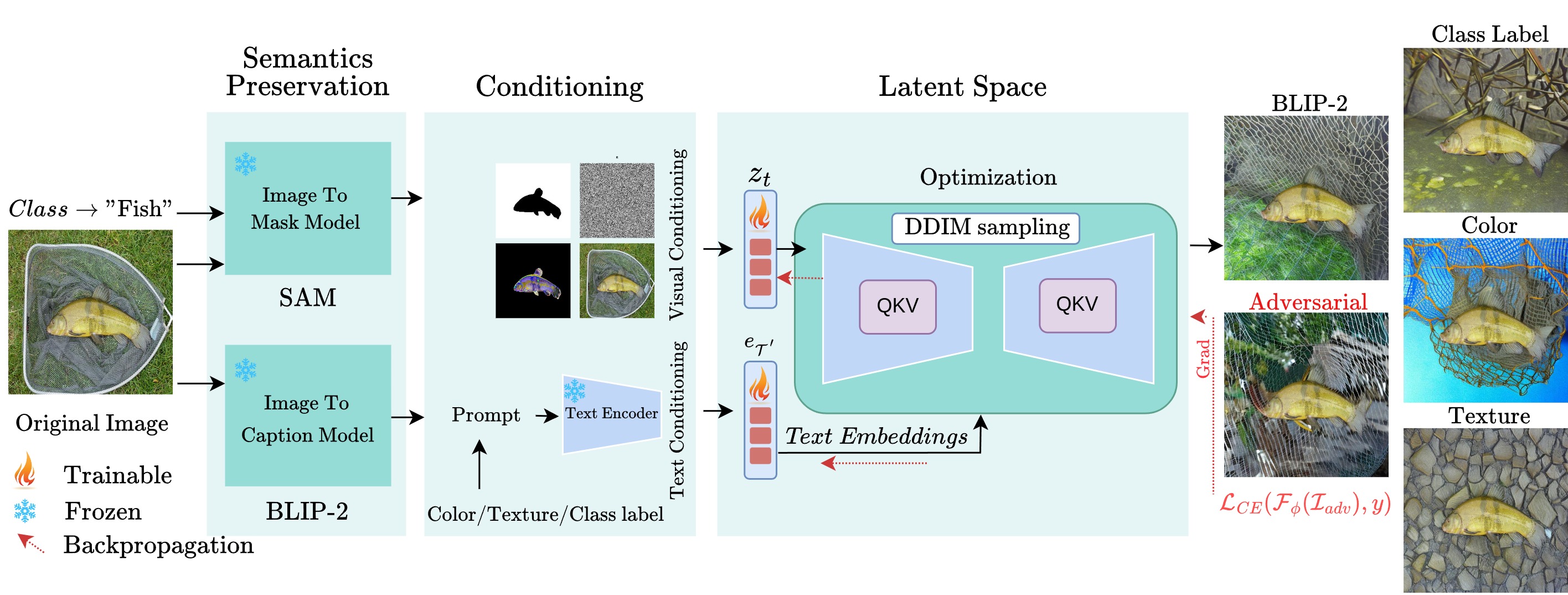}
    \caption{\textsc{ObjectCompose} uses an inpainting-based diffusion model to generate counterfactual backgrounds. The object mask is obtained from SAM using the class label as a prompt. The segmentation mask and original image caption (from BLIP-2) are fed into the diffusion model. For adversarial examples, both the latent and conditional embeddings are optimized during denoising.}
    \label{Block}
\end{figure}

We introduce \textsc{ObjectCompose}, a method for generating diverse language-guided object-to-background compositional changes to evaluate the resilience of vision models. \textsc{ObjectCompose} leverages the complementary strengths of image-to-segment and image-to-text models to guide object-preserving diffusion for natural and adversarial background variations (Figure \ref{Block}). Our automated approach generates datasets under varying distribution shifts, useful for benchmarking vision and vision-language models.

In Section \ref{prelim}, we outline the preliminaries of the foundational models used. In Section \ref{our_approach}, we detail our method.

\subsection{Preliminaries}
\label{prelim}

\textbf{Diffusion Models.} Diffusion models have significantly advanced in generating high-quality images and refining them based on textual guidance. During training, noisy versions $\mathcal{I}_t$ of the clean image $\mathcal{I}$ are input to the model $\epsilon_{\theta}$ at various time steps $t$, with the goal of learning the noise added at each step. Training consists of two stages: in the forward process \emph{(first stage)}, Gaussian noise from a normal distribution $\mathcal{N}(0,I)$ is incrementally added to $\mathcal{I}$ according to a variance schedule $(\beta_t: t=1,..., T)$. Using reparameterization, the noisy image at any time step is:

\begin{equation}
    \mathcal{I}_{t} = \sqrt{\bar{\alpha_{t}}}\mathcal{I} + \sqrt{1 - \bar{\alpha_{t}}}\epsilon \quad \epsilon \sim \mathcal{N}(0,I)
\end{equation}

Here, $\alpha_t = 1-\beta_t$ and $\bar{\alpha_t} =\prod_{s=1}^{t}\alpha_s$. As $T \rightarrow \infty$, $\bar{\alpha}_T \rightarrow 0$, meaning $\mathcal{I}_T \sim \mathcal{N}(0,I)$ and all information from $\mathcal{I}$ is lost. Diffusion models are typically conditioned on $t$, class label $\bm{y}$, or textual description $\mathcal{T}$, with recent extensions incorporating image $\mathcal{I}$ and its mask for image editing tasks \cite{rombach2022high,saharia2022palette}. 

In the reverse process \emph{(second stage)}, the model $\epsilon_{\theta}$ learns to approximate the Gaussian parameters at each time step for the reverse conditional distribution. The objective $L_t$ minimizes the error between the predicted and actual noise at each time step:

\begin{equation}
     L_t = ||\epsilon - \epsilon_{\theta}^{t}(\mathcal{I}_{t},e_{\mathcal{T}}, \psi)||^{2} 
\end{equation}

where $e_{\mathcal{T}}$ is the embedding of the conditional guidance, and $\psi$ represents any additional conditioning, such as masks or scene layouts.

\textbf{Foundational Models.} BLIP-2 \cite{li2023blip} introduces an efficient vision-language pre-training approach using a lightweight Querying Transformer (QFormer) to bridge the gap between pre-trained vision and large language models (LLMs). Images are processed by a vision encoder, with relevant features extracted via QFormer and passed to the LLM to generate descriptive captions.

Recently, Segment Anything Model (SAM)\cite{kirillov2023segment}, an image segmentation model, was introduced that undergoes pre-training on an extensive dataset of high-quality images.  SAM uses prompts—such as points, boxes, masks, or text—to identify objects in images. The image is encoded by a transformer-based encoder, and the extracted features, combined with prompt embeddings, are processed by a lightweight decoder to produce the segmentation mask.

\subsection{Object-to-Background Compositional Changes}
\label{our_approach}

In order to generate object-to-background compositional changes without altering object semantics our method consists of an Object-to-Background Conditioning module to provide strong visual guidance to the text-to-image diffusion model. In the next stage, we condition the diffusion model on the textual prompt to introduce desired background changes or optimize the latent representation and textual embedding in order to generate adversarial backgrounds.

\textbf{Preserving Object Semantics.}
We propose an Object-to-Background Conditioning Module denoted as $\mathcal{C}$, which takes the input image $\mathcal{I}$ and the provided label $\bm{y}$  as inputs, and returns both the textual prompt $\mathcal{T}$ describing the scene and mask $\mathcal{M}$ encapsulates the object in the image:
\begin{equation}
\label{eq:conditionin}
   \mathcal{C}(\mathcal{I}, y) = \mathcal{T}, \mathcal{M} 
\end{equation}
Our conditioning module leverages a promptable segmentation model called SAM \cite{kirillov2023segment} denoted by $\mathcal{S}$. By passing the class information $\bm{y}$ and the image $\mathcal{I}$ to the model $\mathcal{S}(\mathcal{I}, \bm{y})$, we obtain the object mask $\mathcal{M}$. Simultaneously, to acquire a description for the image scene, we utilize BLIP-2 \cite{li2023blip2}, an image-to-text model denoted as $\mathcal{B}$ to get the necessary prompt $\mathcal{T}_{\mathcal{B}}$ describing the scene, thereby providing object-to-background context information.
\begin{equation}
\label{eq:conditioning}
   \mathcal{B}(\mathcal{I}) = \mathcal{T}_{\mathcal{B}}\quad; \quad  \mathcal{S}(\mathcal{I}, \bm{y}) = \mathcal{M}
\end{equation}

The mask $\mathcal{M}$ and the textual prompt $\mathcal{T}_{\mathcal{B}}$ serve as conditioning inputs for the subsequent stage, where we employ a diffusion model to generate diverse background variations. This methodical integration of segmentation and language comprehension offers fine-grained control over image backgrounds while upholding object semantics, leading to refined object-centric image manipulations. It's worth noting that we have the flexibility to choose any desired textual prompt $\mathcal{T}$, and are not confined to using $\mathcal{T}_{\mathcal{B}}$ as the textual condition.

\textbf{Background Generation.} Once we've obtained both visual and textual information $(\mathcal{T}, \mathcal{M})$ from our conditioning module, we employ a diffusion model that has been trained for inpainting tasks, which has additional conditioning $\psi$ comprising of the image $\mathcal{I}$ and its corresponding mask $\mathcal{M}$. The denoising operation takes place in the latent space instead of the image pixel space, which is facilitated through the use of a variational autoencoder that provides the mapping between images and their respective latent representations. During the denoising stage, starting with a standard normal Gaussian noise latent $z_{t}$, the diffusion model calculates the estimated noise $\hat{\epsilon}_{\theta}^{t}$ to be removed from the latent at time step $t$ using a linear combination of the noise estimate conditioned on the textual description  $\epsilon_{\theta}^{t}(z_{t},e_{\mathcal{T}},i, m)$ and the unconditioned estimate $\epsilon_{\theta}^{t}(z_{t},i, m)$:

\begin{equation}
\label{eq:cfg}
\hat{\epsilon}_{\theta}^{t}(z_{t},e_{\mathcal{T}},i, m) = \epsilon_{\theta}^{t}(z_{t},i, m) + \lambda \left( \epsilon_{\theta}^{t}(z_{t},e_{\mathcal{T}}, i, m) - \epsilon_{\theta}^{t}(z_{t},i, m) \right)
\end{equation}

Here, $(i, m)$ represents the representation of the original image $\mathcal{I}$ and its corresponding mask $\mathcal{M}$ in the latent space. The guidance scale $\lambda$  determines how much the unconditional noise estimate $\epsilon_{\theta}(z_{t},i, m)$ should be adjusted in the direction of the conditional estimate $\epsilon_{\theta}(z_{t},e_{\mathcal{T}}, i, m)$ to closely align with the provided textual description $\mathcal{T}$(see Appendix \ref{sec:prompt}). In this whole denoising process, the mask $\mathcal{M}$ generated from our conditioning module guides the image alterations to the background of the object,  while the textual description $\mathcal{T}$ contains information for the desired background change.

Our method also handles adversarial background changes by optimizing the conditioned visual and textual latents $z_t$ and $e_{\mathcal{T}}$ through a discriminative model $\mathcal{F}_{\phi}$ to craft adversaries.
 For generating adversarial examples the goal of the attacker is to craft perturbations $\delta$ that when added to clean image $\mathcal{I}$ with class label $\bm{y}$, result in an adversarial image $\mathcal{I}_{adv} = \mathcal{I} + \delta$  which elicits an incorrect response from a classifier model $\mathcal{F}_{\phi}$ i.e.,  $ \mathcal{F}_{\phi}(\mathcal{I}_{adv}) \neq \bm{y}$, where $\phi$ are the model parameters. Usually in pixel-based perturbations, $\delta$ is bounded by a norm distance, such as $l_2$ or $l_{\infty}$ norm to put a constraint on pixel-level changes done to preserve the semantics of the image. However, in our setting, the control on the amount of perturbation added is governed by the textual and visual latent passed to the diffusion model. In our method (see Algo. \ref{Attack-algo}), we use the discriminative model $\mathcal{F}_{\phi}$ to guide the diffusion model $\epsilon_{\theta}$ to generate adversarial examples by optimizing its latent representations $z_t$ and $e_{\mathcal{T}}$:

\begin{equation}
\label{eq:adv_loss}
   \max_{z_t, e_{\mathcal{T}}} \mathcal{L}_{adv} = \mathcal{L}_{CE}(\mathcal{F}_{\phi}(\mathcal{I}_{adv}), \bm{y}) 
\end{equation}

where $\mathcal{L}_{CE}$ is the cross-entropy loss, $e_{\mathcal{T}}$ is textual embedding and $z_t$ is the denoised latent at time step $t$.  $\mathcal{I}_{adv}$ represents the image generated by the diffusion model after it has been denoised using DDIM \cite{song2020denoising}, a  deterministic sampling process in which the latent update is formulated as:

\begin{equation}
\label{eq:diffusion}
z_{t-1} = \sqrt{\bar{\alpha}_{t-1}} \left( \frac{z_t - \sqrt{1 - \bar{\alpha}_{t}} \hat{\epsilon}_{\theta}^{t}}{\sqrt{\bar{\alpha}_{t}}} \right) + \sqrt{1 - \bar{\alpha}_{t-1}} \hat{\epsilon}_{\theta}^{t}, \quad  \mathit{t = T, \ldots, t-1, \ldots, 1}
\end{equation}

Our proposed unconstrained adversarial objective $\mathcal{L}_{adv}$ would lead to unrestricted changes in the image background while object semantics are preserved by using the mask conditioning from $\mathcal{S}$.

\section{Experimental Protocols}
\textbf{Dataset Preparation.} For classification, we initially gathered 30k images from the ImageNet validation set \cite{deng2009imagenet}, which are correctly classified with high success rate using an ensemble of models; ViT-T, ViT-S \cite{dosovitskiy2020image}, Res-50, Res-152 \cite{he2016deep}, Dense-161 \cite{huang2017densely}, Swin-T, and Swin-S \cite{liu2021swin}. In order to create a high-quality dataset for our object-to-context variation task, we remove image samples where the boundary between foreground and background is not distinct, e.g.,  "mountain tent" where the mountain might appear in the background of the tent. This processing results in 15k images.
Then for foreground semantic preservation, we utilize a compute-efficient variant of SAM, known as FastSAM \cite{zhao2023fast} with class labels as prompts to generate segmentation masks of the foreground object. However, FastSAM encounter challenges in accurately segmenting objects in all images. To address this, we selected images where the mask-creation process demonstrated exceptional accuracy and generated a clear separation between the object of interest and its background. This meticulous selection process yield a curated dataset comprising 5,505 images, representing a subset of 582 ImageNet classes. We refer to this dataset as \texttt{ImageNet-B}. Due to the computational cost involved in adversarial background optimization and running baseline methods,  we select a subset of 1000 images from  500 classes of \texttt{ImageNet-B} by sampling two images from each class for comparison.  We refer to this dataset as $\texttt{ImageNet-B}_{1000}$.  Rest of our experiments are performed on the full \texttt{ImageNet-B} dataset.

For object detection, we manually filtered 1,127 images from the COCO 2017 validation set \cite{lin2015microsoft}, ensuring a clear distinction between foreground objects and background, referred to as \texttt{COCO-DC}. This dataset, containing multiple objects per image, is used for both detection and classification. For classification, models are trained on the COCO train dataset using the label of the object with the largest mask region and evaluated on our generated dataset. Additional details and dataset comparisons are provided in Appendix \ref{sec:Dataset}.

\noindent \textbf{Diffusion Parameters.}  We use the pre-trained Inpaint Stable Diffusion v2\cite{rombach2022high} as our text-to-image model and set the guidance parameter $\lambda$ to $7.5$, and use the DDIM sampling \cite{song2020denoising} with $T=20$ timesteps. We craft adversarial examples on $\texttt{ImageNet-B}_{1000}$ using Res-50\cite{he2016deep} as the classifier model and maximize the adversarial loss $\mathcal{L}_{adv}$ shown in Eq.\ref{eq:adv_loss} for $30$ iterations. For \texttt{COCO-DC}, we maximize the loss in the feature space of the model. Both the text embedding $e_{\mathcal{T}}$ of the prompt $\mathcal{T}$(initialized with $\mathcal{T}_{\mathcal{B}}$) and denoised latent $z_t$ are optimized from denoising time step $t=4$ using  AdamW \cite{loshchilov2017decoupled} with a learning rate of $0.1$.

\noindent \textbf{Vision Models.}
We conducted evaluations for the classification task using a diverse set of models. 
\emph{\textbf{a)} Natural ImageNet Training:} We evaluate seven naturally ImageNet-trained vision transformers and convolutional neural networks (CNNs). Specifically we use  ViT-T, ViT-S\cite{dosovitskiy2020image}, Res-50, Res-152\cite{he2016deep}, Dense-161\cite{huang2017densely}, Swin-T, and Swin-S\cite{liu2021swin}.
\emph{\textbf{b)} Adversarial ImageNet Training:} We also evaluate adversarial ImageNet-trained models including ResAdv-18, ResAdv-50, and WideResAdv-50 at various perturbation budget of $\ell_\infty$ and $\ell_2$ \cite{salman2020adversarially}. 
\emph{\textbf{c)} Multimodal Training:} Additionally, we explored seven vision language foundational models within CLIP\cite{radford2021learning} and  EVA-CLIP\cite{sun2023eva}. 
 \emph{\textbf{d)} Stylized ImageNet Training:} We evaluate the DeiT-T and DeiT-S models trained on a stylized version of the ImageNet dataset \cite{naseer2021intriguing, geirhos2018imagenet}.
 \emph{\textbf{e)} Self-Supervised Training:} We evaluate the performance of Dinov2 models with registers\cite{oquab2023dinov2,darcet2023vision} which are trained in a self-supervised manner on a large-scale curated dataset LVD-142M, and subsequently fine-tuned on ImageNet. 
\emph{\textbf{f)} Segmentation and Detection:} We evaluate Mask-RCNN for segmentation and object detection respectively using our proposed background-to-object variations. Evaluations on FastSAM\cite{zhao2023fast} and DETR\cite{carion2020end} are reported in Appendix \ref{sec:FastSAM} and \ref{sec:detection results}. \emph{\textbf{g)} Image Captioning:} We also evaluate the robustness of a recent image captioning model BLIP-2\cite{li2023blip2}, using our generated dataset. For the task \emph{\textbf{a)}}, and \emph{\textbf{b)}}, we provide comparison with the baseline methods on $\texttt{ImageNet-B}_{1000}$ and report results on $\texttt{ImageNet-B}$ in the Appendix \ref{sec:prompt_evaluation}.

\noindent \textbf{Evaluation Metrics:}
We use the top-1 accuracy (\%), Intersection Over Union (IoU), Average Precision(AP) and Recall(AR), and CLIP text similarity score for classification, segmentation, object detection, and captioning tasks, respectively.

\noindent \textbf{Background Conditioning.} To induce background variations, we use the following text prompt templates: Class Label: "\textit{A picture of a class}" where \textit{class} is the image's class name; Caption: "\textit{captions from BLIP-2}"; Color: "\textit{A picture of \_\_\_ background}" where \_\_\_ is red, green, blue, or colorful; Texture: "\textit{A picture of \_\_ background}" with \_\_\_ replaced by textured, rich textures, colorful textures, distorted textures; Adversarial: "\textit{captions from BLIP-2}" with updated prompts after optimization. For ImageNet-E\cite{li2023imagenet}, default values of $\lambda$ are employed to regulate the strength of texture complexity. For LANCE\cite{prabhu2023lance}, we use the default prompt to generate background variations via a large language model. We report the worst-performing prompt across colors and textures, with detailed analysis in Appendix \ref{sec:prompt_evaluation}.

\begin{table*}[t]
\fontsize{8pt}{8pt}\selectfont
\centering
\caption{
\small Resilience evaluation of vision models on $\texttt{ImageNet-B}_{1000}$(Top-1 (\%) accuracy). Our natural object-to-background changes, including color and texture, perform favorably against state-of-the-art methods. Furthermore, our adversarial object-to-background changes show a significant drop in performance across vision models.}
\resizebox{1\linewidth}{!}{%
\begin{tabular}{lcccccccl}
\toprule

\multirow{2}{*}{Datasets} & 
\multicolumn{4}{c}{\textbf{ViT}} & 
\multicolumn{3}{c}{\textbf{CNN}}
\\  \cmidrule(lr){2-5}
\cmidrule(lr){6-8}
& ViT-T & ViT-S & Swin-T & Swin-S & Res-50 &Res-152& Dense-161 & \cellcolor{gray!20}Average\\
 \midrule
Original & 95.5 & 97.5  &97.9& 98.3& 98.5  & 99.1 & 97.2&\cellcolor{gray!20}97.71 \\
\midrule
 ImageNet-E ($\lambda$=-20) & 91.3 & 94.5   & 96.5 & 97.7 & 96.0 & 97.6\ & 95.4& \cellcolor{gray!20}95.50\dec{2.21}\\
 ImageNet-E ($\lambda$=20) & 90.4 & 94.5   & 95.9 & 97.4 & 95.4 & 97.4\ & 95.0& \cellcolor{gray!20}95.19\dec{-2.52}\\
 ImageNet-E ($\lambda_{adv}=20$) & 82.8 & 88.8   & 90.7 & 92.8 & 91.6 & 94.2\ & 90.4& \cellcolor{gray!20}90.21\dec{7.50}\\
 LANCE  & 80.0 & 83.8   & 87.6 & 87.7 & 86.1 & 87.4\ & 85.1& \cellcolor{gray!20}85.38\dec{12.33}\\

 \midrule
Class label & 90.5 & 94.0 & 95.1 & 95.4 & 96.7  & 96.5 & 94.7&\cellcolor{gray!20}94.70\dec{3.01}\\
BLIP-2 Caption & 85.5 & 89.1 & 91.9 & 92.1 & 93.9 & 94.5 &90.6 &\cellcolor{gray!20}91.08\dec{6.63} \\
Color & 67.1 & 83.8 & 85.8 & 86.1 & 88.2  & 91.7 & 80.9&\cellcolor{gray!20}\textbf{83.37}\dec{14.34} \\
Texture & 64.7 & 80.4 & 84.1 & 85.8 & 85.5  & 90.1 & 80.3&\cellcolor{gray!20}\textbf{81.55}\dec{16.16} \\
Adversarial & 18.4 & 32.1 & 25.0 & 31.7 & 2.0  & 14.0 & 28.0&\cellcolor{gray!20}\textbf{21.65}\dec{76.06} \\

\bottomrule
\end{tabular}%
}
\label{tab:base_class_comparison}
\end{table*}

\subsection{Comparison with Baseline Methods}
\label{Results}

\begin{figure}[t!]
\begin{minipage}{\textwidth}

\centering

\begin{minipage}{0.16\textwidth}
  \centering
  \includegraphics[ trim= 5mm 15mm 5mm 5mm, clip, width=\linewidth , keepaspectratio]{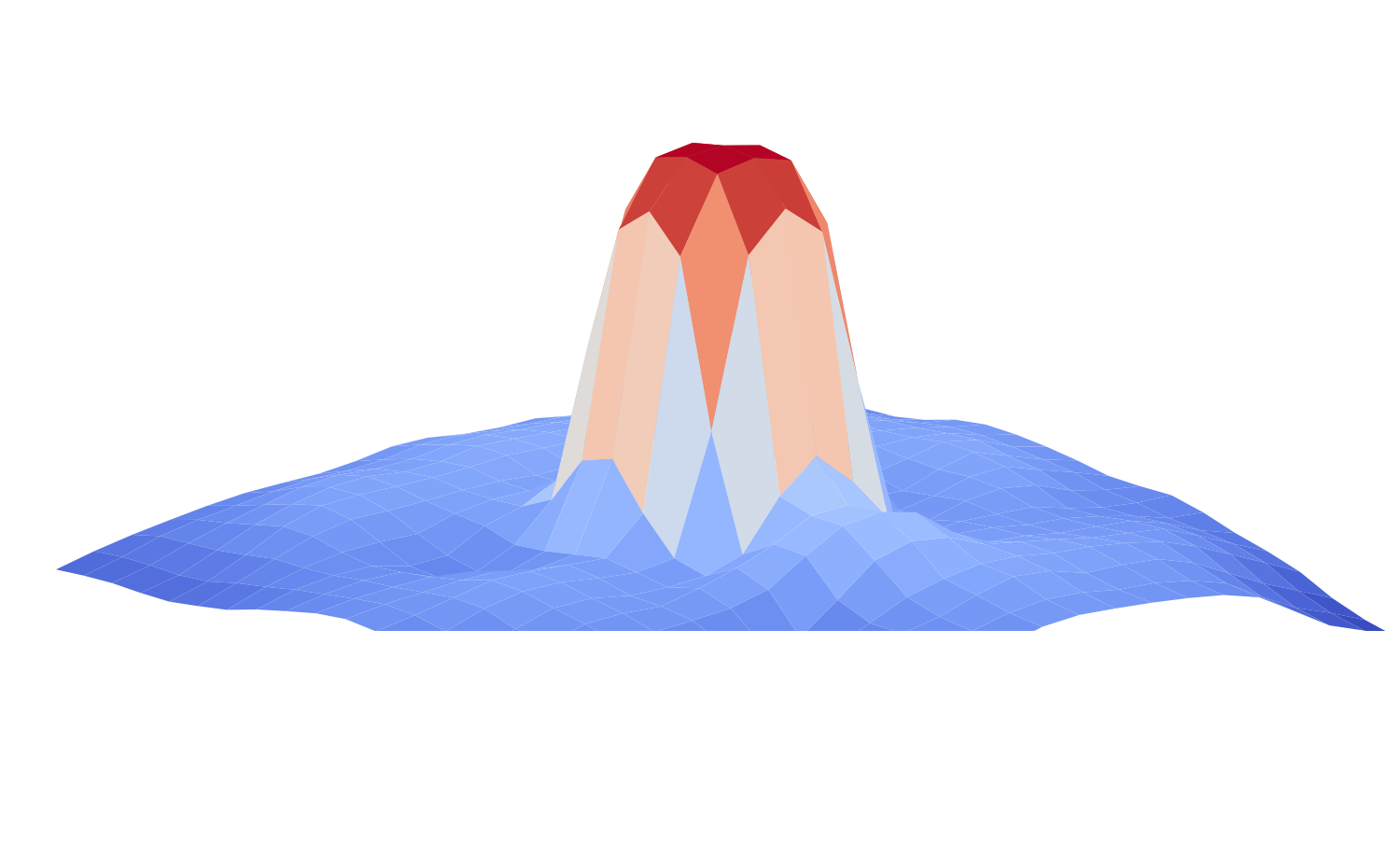}
   \footnotesize Original
\end{minipage}
\begin{minipage}{0.16\textwidth}
  \centering
  \includegraphics[trim= 5mm 15mm 5mm 5mm, clip, width=\linewidth , keepaspectratio ]{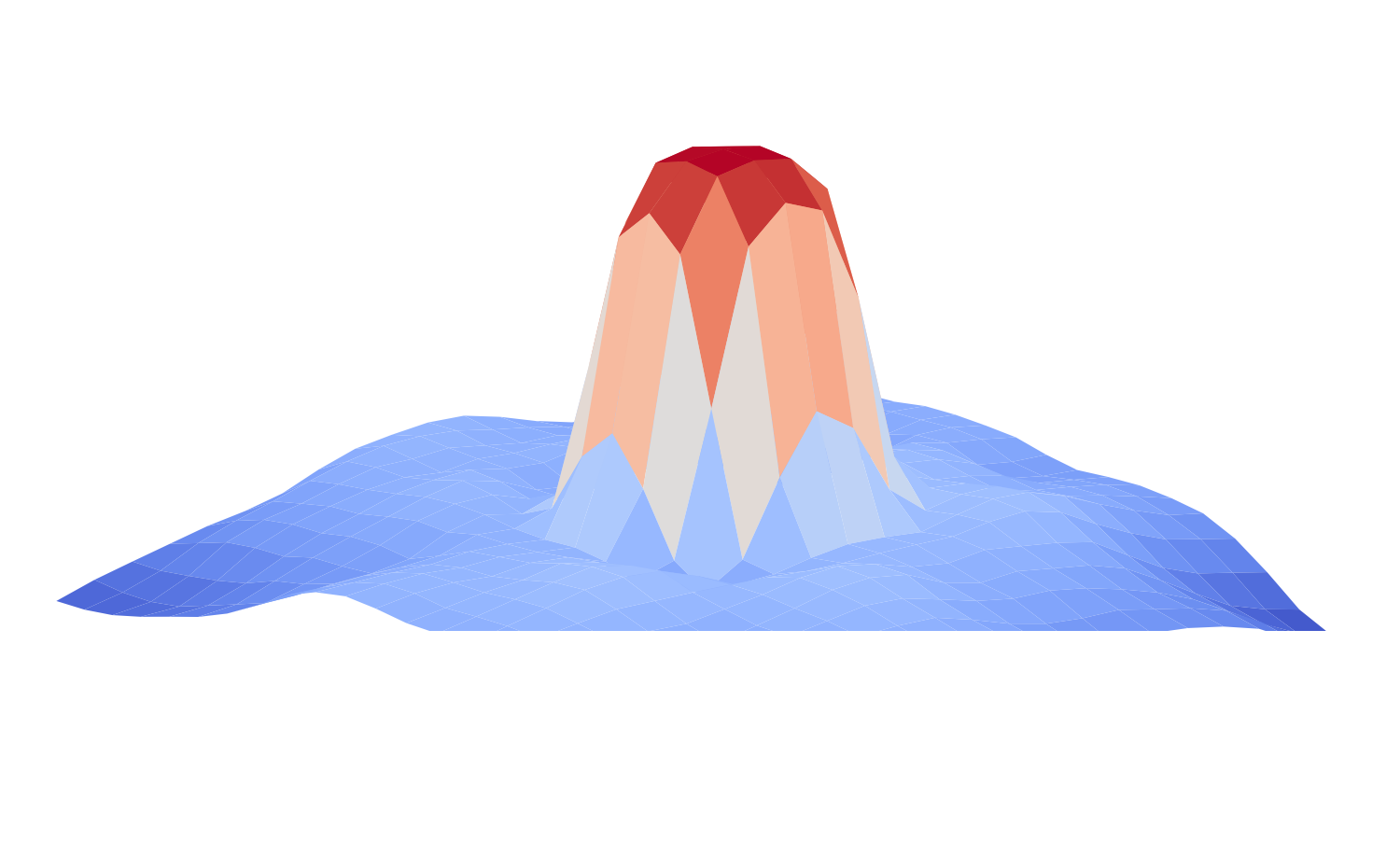}
    \footnotesize Class Label

\end{minipage}
\begin{minipage}{0.16\textwidth}
  \centering
  \includegraphics[trim= 5mm 15mm 5mm 5mm, clip, width=\linewidth , keepaspectratio]{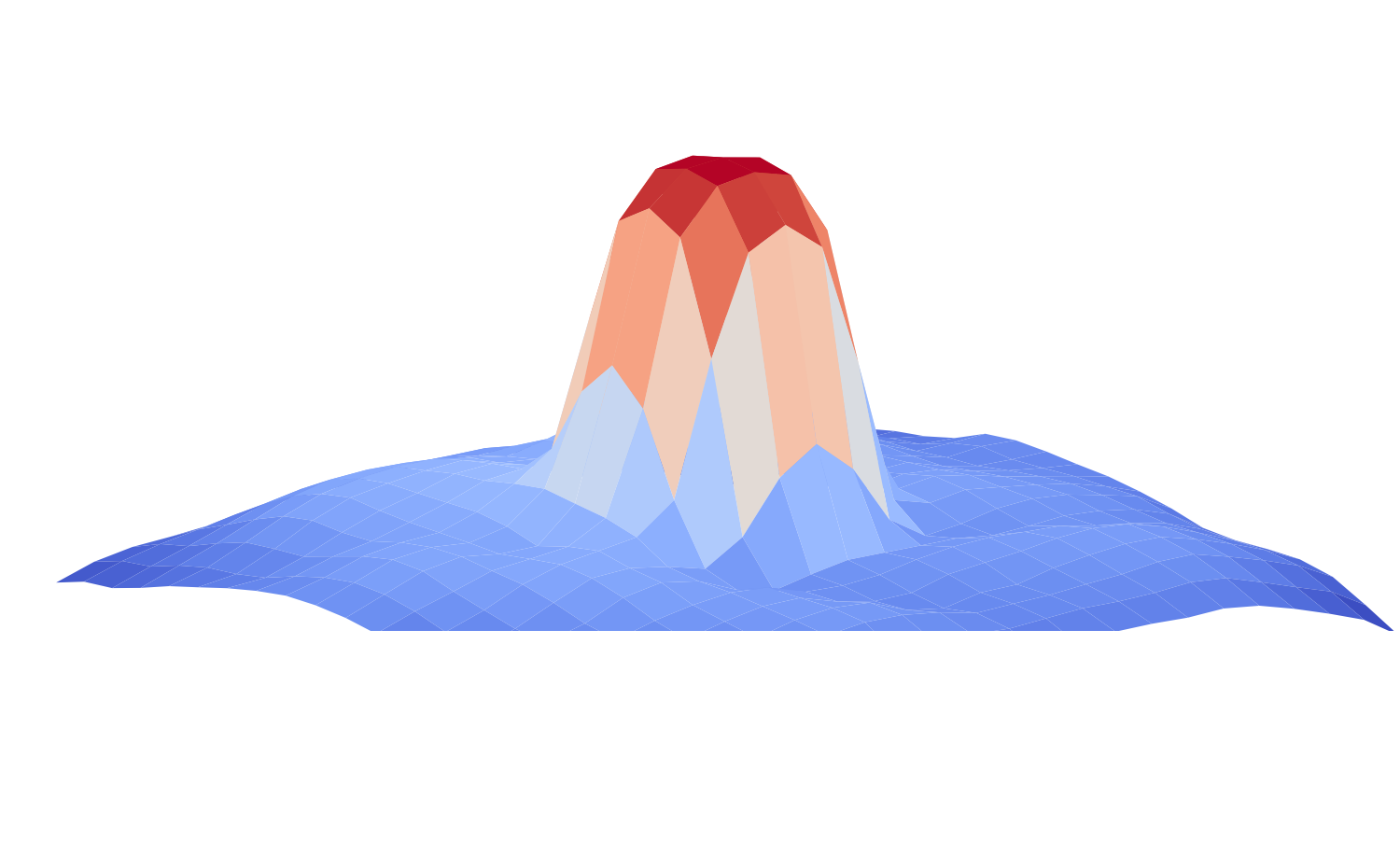}
     \footnotesize BLIP-2

\end{minipage}
\begin{minipage}{0.16\textwidth}
  \centering
  \includegraphics[trim= 5mm 15mm 5mm 5mm, clip, width=\linewidth , keepaspectratio]{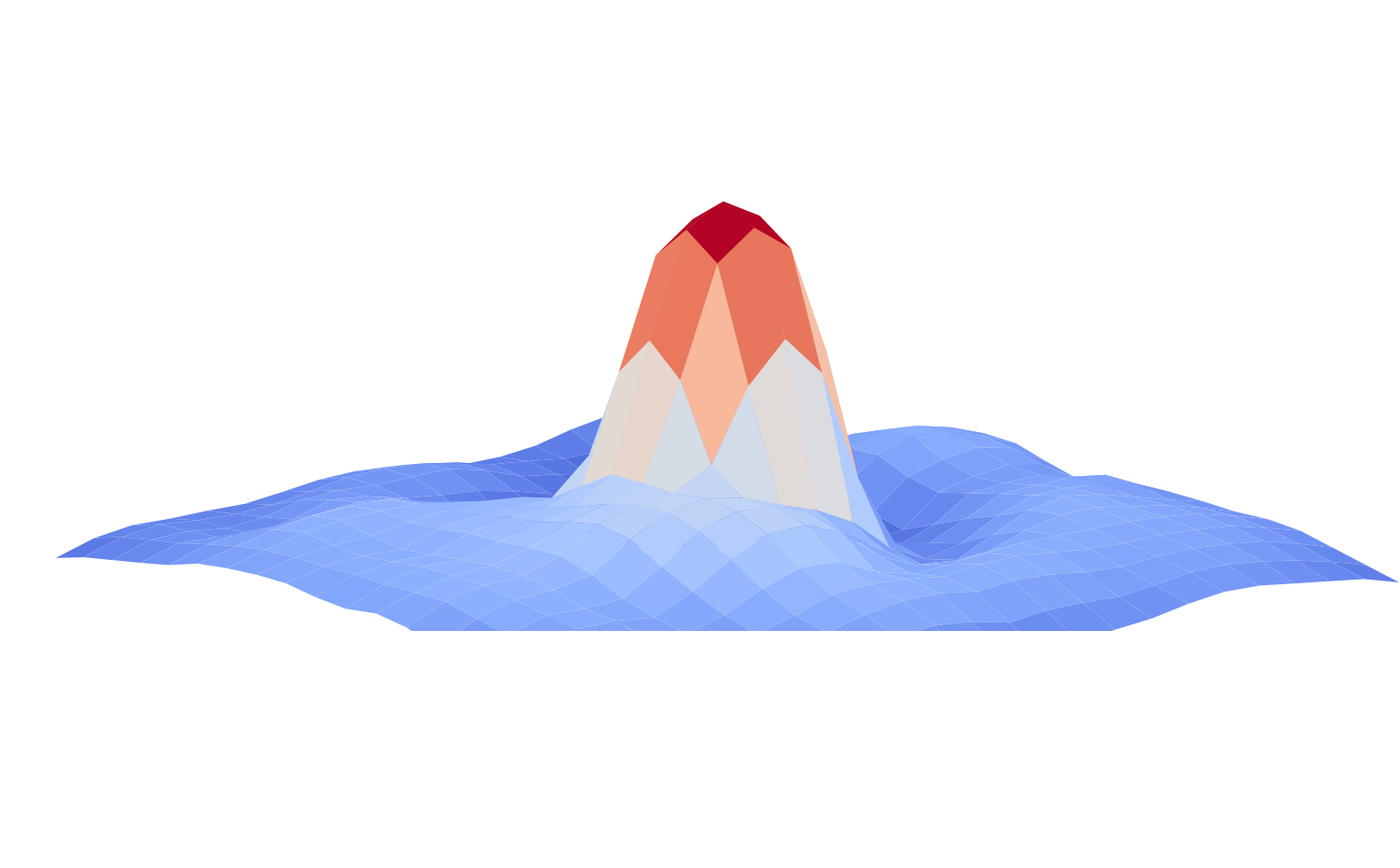}
     \footnotesize Color
\end{minipage}
\begin{minipage}{0.16\textwidth}
  \centering
  \includegraphics[trim= 5mm 15mm 5mm 5mm, clip, width=\linewidth , keepaspectratio]{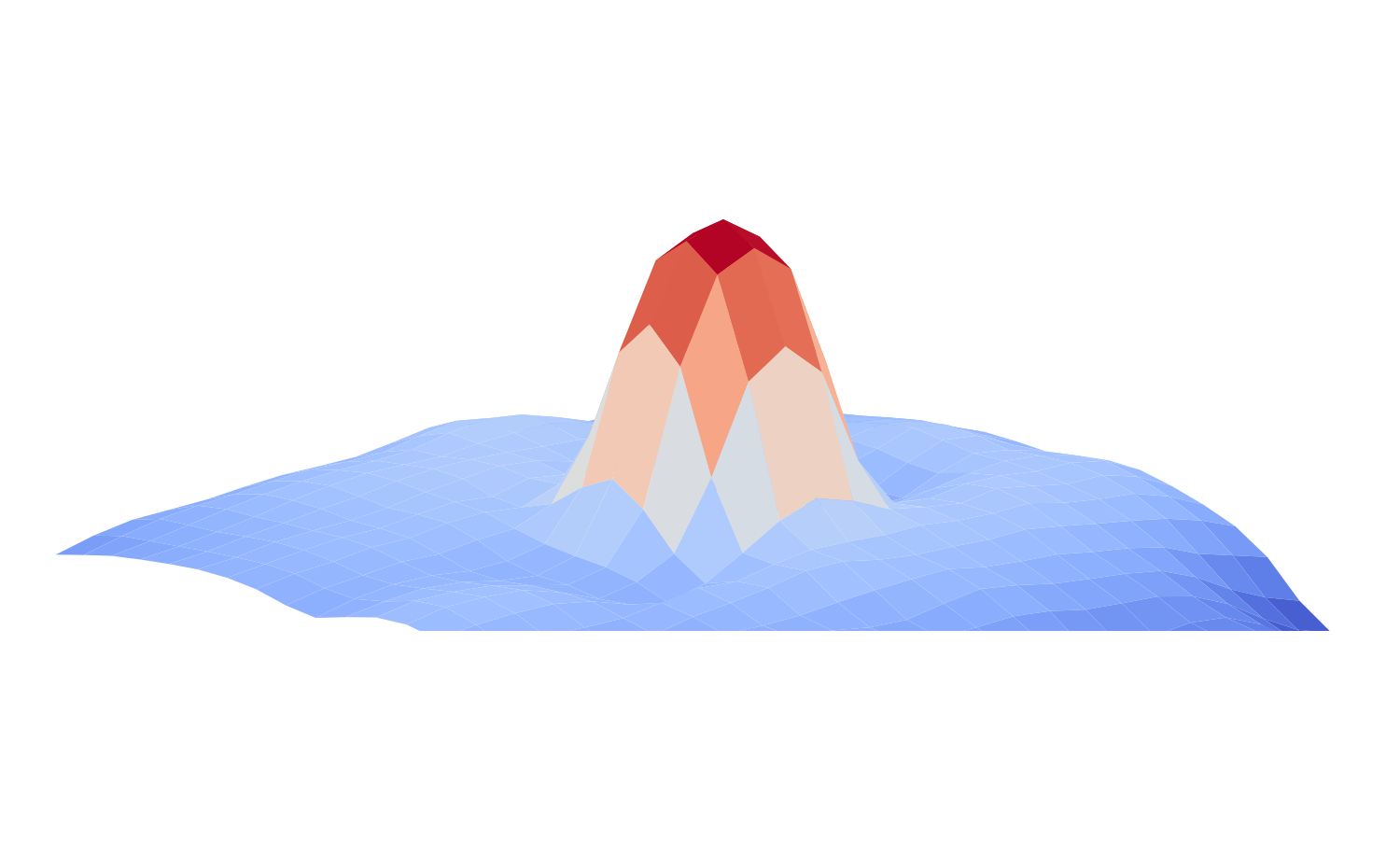}
     \footnotesize Texture

\end{minipage}
\begin{minipage}{0.16\textwidth}
  \centering
  \includegraphics[trim= 5mm 15mm 5mm 5mm, clip, width=\linewidth , keepaspectratio]{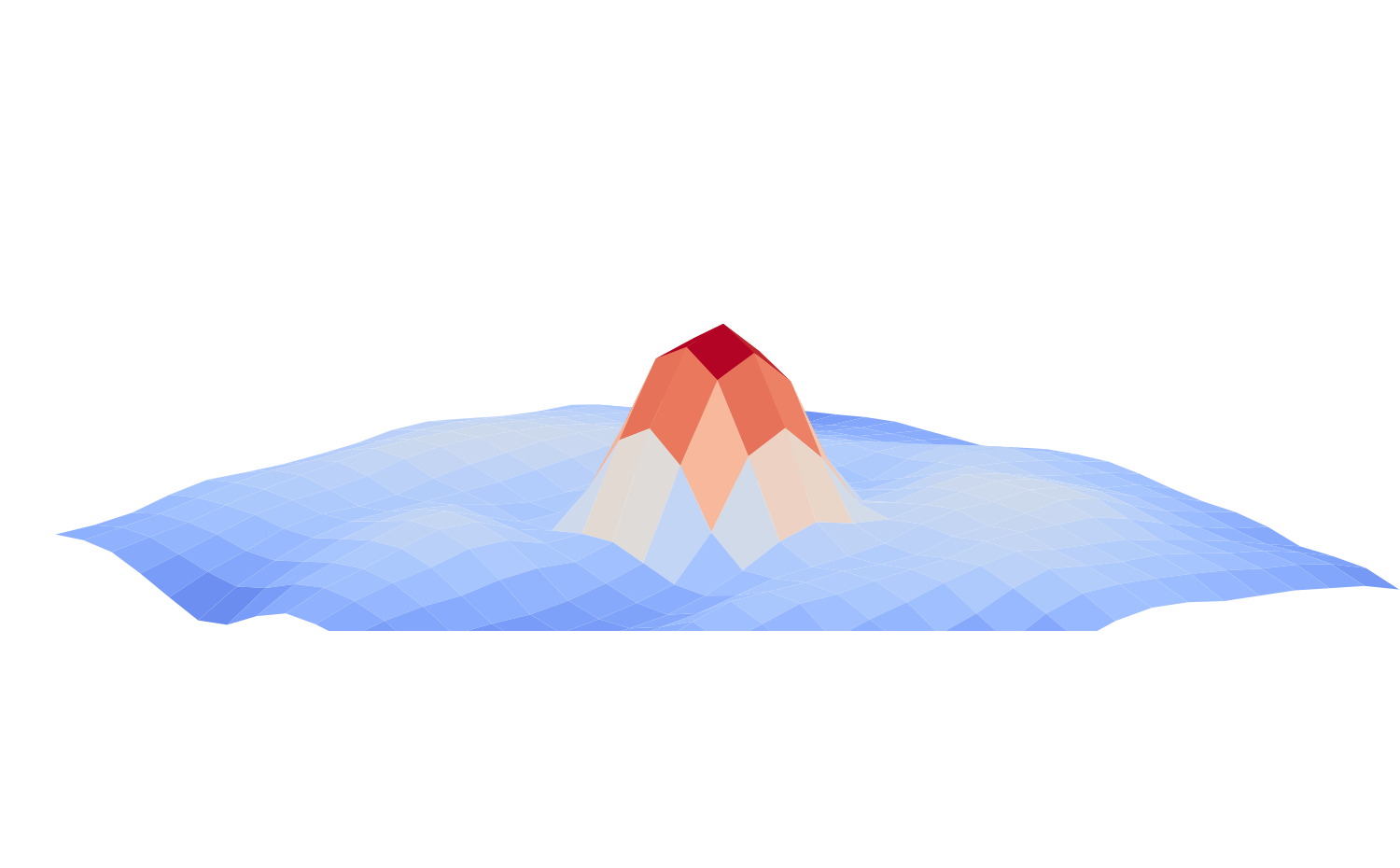}
     \footnotesize Adversarial
\end{minipage}

\end{minipage}
\hfill
  \caption{The loss surfaces \emph{(flipped)} of the ViT-S depicted on \texttt{ImageNet-B}. Significant distribution shifts result in narrow and shallow surfaces at convergence.}
  \label{fig:loss_surfaces}
\end{figure}

\begin{figure}[t]
\begin{minipage}{\textwidth}


\begin{minipage}{0.16\textwidth}
  \centering
  \includegraphics[ trim= 5mm 15mm 5mm 5mm, clip, width=\linewidth , keepaspectratio]{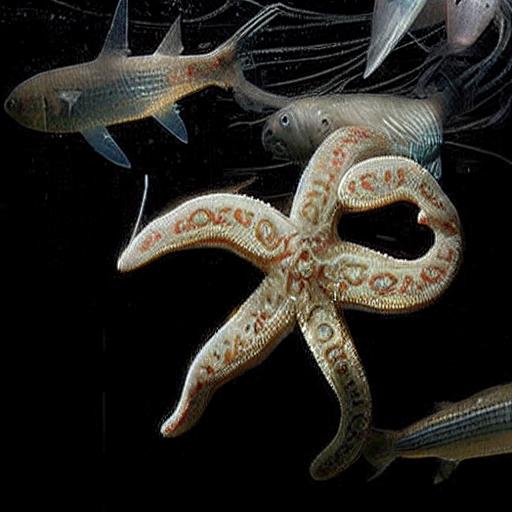}
   \footnotesize LANCE 
\end{minipage}
\begin{minipage}{0.16\textwidth}
  \centering
  \includegraphics[trim= 5mm 15mm 5mm 5mm, clip, width=\linewidth , keepaspectratio]{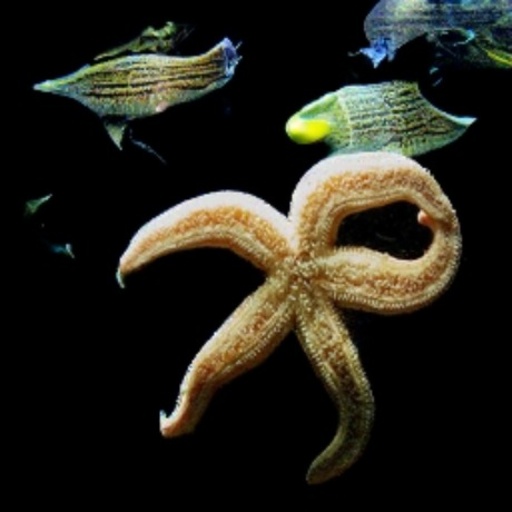}
    \footnotesize $\lambda=-20$ 

\end{minipage}
\begin{minipage}{0.16\textwidth}
  \centering
  \includegraphics[trim= 5mm 15mm 5mm 5mm, clip, width=\linewidth , keepaspectratio]{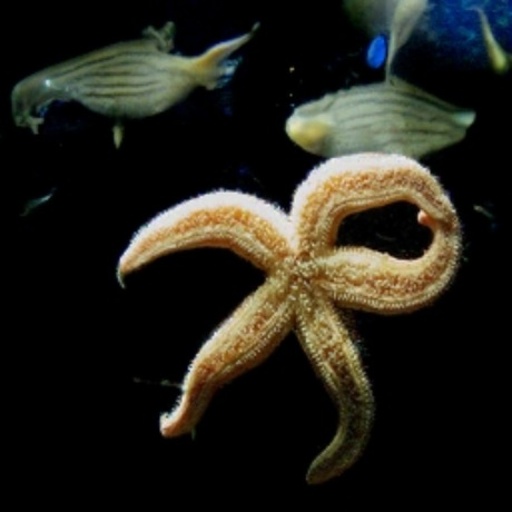}
     \footnotesize $\lambda=20$ 

\end{minipage}
\begin{minipage}{0.16\textwidth}
  \centering
  \includegraphics[trim= 5mm 15mm 5mm 5mm, clip, width=\linewidth , keepaspectratio]{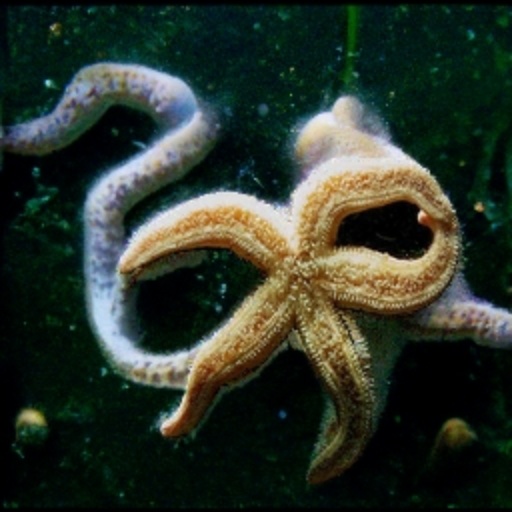}
 \footnotesize $\lambda_{adv}=20$
\end{minipage}
\begin{minipage}{0.16\textwidth}
  \centering
  \includegraphics[trim= 5mm 15mm 5mm 5mm, clip, width=\linewidth , keepaspectratio]{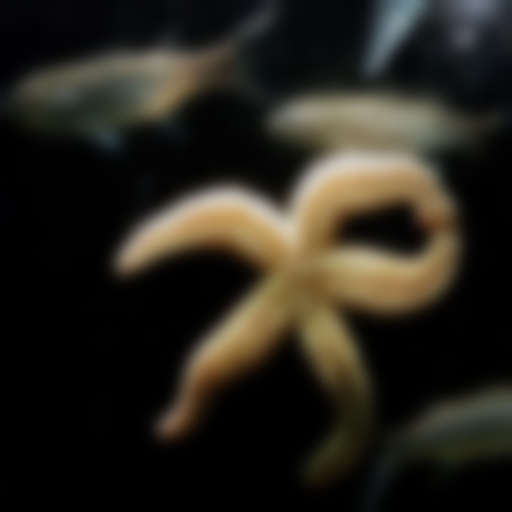}
     \footnotesize Blur

\end{minipage}
\begin{minipage}{0.16\textwidth}
  \centering
  \includegraphics[trim= 5mm 15mm 5mm 5mm, clip, width=\linewidth , keepaspectratio]{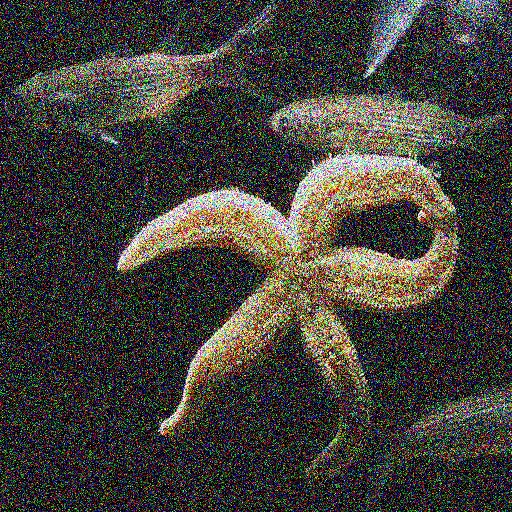}
     \footnotesize Noise
\end{minipage}

\begin{minipage}{0.16\textwidth}
  \centering
  \includegraphics[ trim= 5mm 15mm 5mm 5mm, clip, width=\linewidth , keepaspectratio]{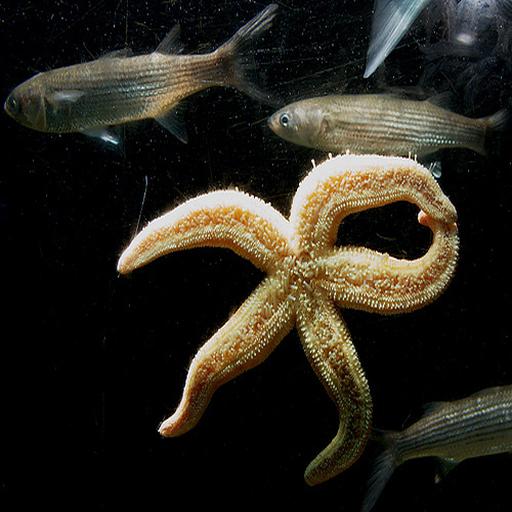}
   \footnotesize Original
\end{minipage}
\begin{minipage}{0.16\textwidth}
  \centering
  \includegraphics[trim= 5mm 15mm 5mm 5mm, clip, width=\linewidth , keepaspectratio]{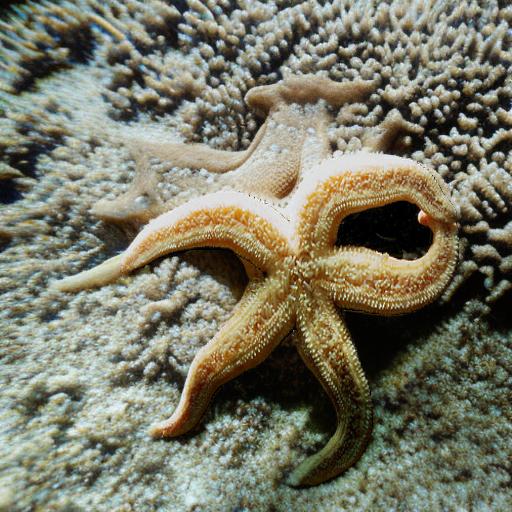}
    \footnotesize Class Label

\end{minipage}
\begin{minipage}{0.16\textwidth}
  \centering
  \includegraphics[trim= 5mm 15mm 5mm 5mm, clip, width=\linewidth , keepaspectratio]{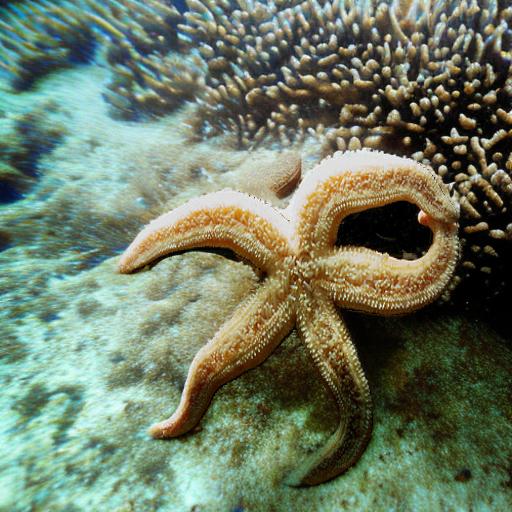}
     \footnotesize BLIP-2

\end{minipage}
\begin{minipage}{0.16\textwidth}
  \centering
  \includegraphics[trim= 5mm 15mm 5mm 5mm, clip, width=\linewidth , keepaspectratio]{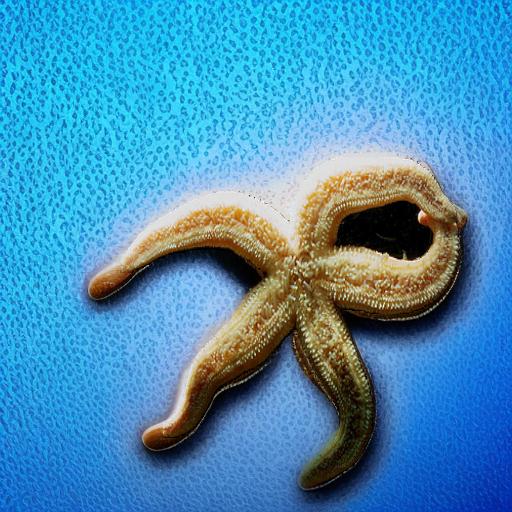}
     \footnotesize Color
\end{minipage}
\begin{minipage}{0.16\textwidth}
  \centering
  \includegraphics[trim= 5mm 15mm 5mm 5mm, clip, width=\linewidth , keepaspectratio]{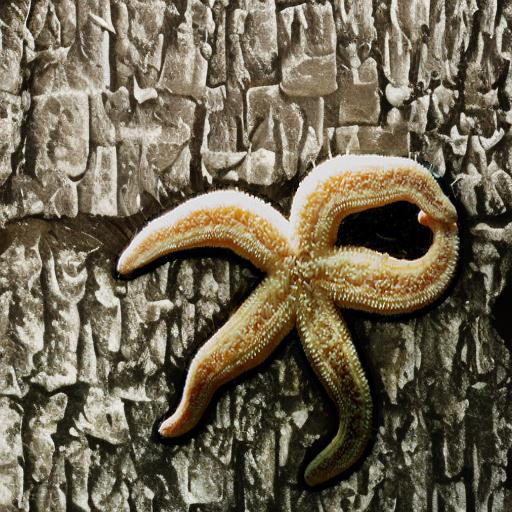}
     \footnotesize Texture

\end{minipage}
\begin{minipage}{0.16\textwidth}
  \centering
  \includegraphics[trim= 5mm 15mm 5mm 5mm, clip, width=\linewidth , keepaspectratio]{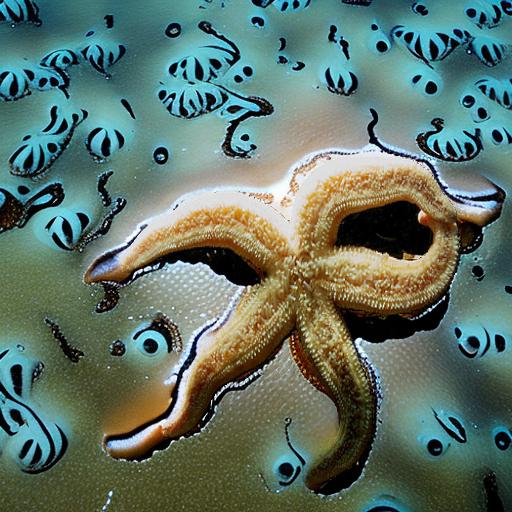}
     \footnotesize Adversarial
\end{minipage}

\end{minipage}
\hfill
  \caption{Qualitative comparison of our method \textit{(bottom row)} with previous related work \textit{(top row)}. Our method enables diversity and controlled background edits.}
  \label{fig:qualitative-comparison}
\end{figure}

\noindent \textbf{Natural ImageNet Training.} 
In Table \ref{tab:base_class_comparison}, we observe that background variations introduced by our method are more challenging for vision models, resulting in a performance drop of $13.5\%$ compared to ImageNet-E ($\lambda=20$) on natural background variations. When subjected to adversarial background changes, a substantial performance drop of $68.56\%$ is observed compared to ImageNet-E ($\lambda_{adv}=20$), highlighting the effectiveness of the unconstrained nature of our attack. Background variations by our method show a consistent decline in accuracy for both transformer-based and CNN models when exposed to diverse object-to-background changes. This decrease is especially noticeable in texture and color backgrounds. We find that as we moved from purely transformer-based architectures to convolution-based architectures, there is an overall improvement in accuracy across natural background changes. For instance,  the average accuracy across all backgrounds for ViT-T, Swin-T, and Res-50 on $\texttt{ImageNet-B}_{1000}$ is $76.95\%$, $89.22\%$ and $91.08\%$ respectively. Further, we observe that as the model capacity is increased across different model families, the robustness to background changes also increases. As is evident, the models are most vulnerable to adversarial background changes, resulting in a significant drop in average accuracy. Res-50 shows most drop on adversarial changes, which is expected as it serves as the discriminative model $\mathcal{F}_{\phi}$ (Eq. \ref{eq:adv_loss}) for generating adversarial examples.   In Figure \ref{fig:loss_surfaces}, we depict the loss surfaces of ViT-S and observe that these surfaces become narrower and shallower with more pronounced background variations, aligning with our results. We provide results on \texttt{ImageNet-B} and \texttt{COCO-DC} dataset in Appendix \ref{sec:prompt_evaluation} and \ref{sec:recent_models} with ablations across different background prompts. Visualizations are provided in Figure \ref{fig:qualitative-comparison} and Appendix \ref{sec:diversity}.

\begin{table}[t]
\fontsize{9pt}{9pt}\selectfont
\centering
\caption{Resilience evaluation of Zero-shot CLIP and EVA-CLIP  models on $\texttt{ImageNet-B}_{1000}$(Top-1 (\%) accuracy). Our natural object-to-background changes, including color and texture, perform favorably against state-of-the-art methods. We find that EVA-CLIP models show better performance across all background variations.}
\resizebox{1\linewidth}{!}{%
\begin{tabular}{lcccccccl}
\toprule
\multirow{2}{*}{Datasets} & 
\multicolumn{7}{c}{CLIP} & 
\\  \cmidrule(lr){2-8}
&  ViT-B/32 & ViT-B/16 & ViT-L/14 & Res50 & Res101 & Res50x4 & Res50x16 & \cellcolor{gray!20}Average \\
 \midrule
 Original  &  73.90 & 79.40  & 87.79& 70.69& 71.80 & 76.29 & 82.19 & \cellcolor{gray!20}77.43  \\
 \midrule

ImageNet-E ($\lambda$=-20) & 69.79 & 76.70  & 82.89 & 67.80 & 69.99&  72.70& 77.00& \cellcolor{gray!20}73.83\dec{3.60}\\
ImageNet-E ($\lambda$=20) & 67.97 & 76.16   & 82.12 & 67.37 & 39.89 & 72.62 &77.07 & \cellcolor{gray!20}73.31\dec{4.12} \\
ImageNet-E ($\lambda_{adv}=20$) & 62.82 & 70.50  & 77.57&  59.98& 65.85 & 67.07 &67.07 & \cellcolor{gray!20}68.23\dec{9.20}\\
LANCE  & 54.99 & 54.19  & 57.48&  58.05& 60.02 &60.39  & 73.37& \cellcolor{gray!20}59.78\dec{17.65}\\
 \midrule
 
%
Class label  & 78.49 & 83.69  & 88.79& 76.60& 77.00 & 82.09 & 84.50 & \cellcolor{gray!20}81.59\inc{4.16} \\
BLIP-2 Captions & 68.79 & 72.29 & 79.19 & 65.20 & 68.40 & 71.20 & 75.40 & \cellcolor{gray!20}71.49\dec{5.94 } \\
%
Color & 48.30 & 61.00 & 69.51 & 50.50 & 54.80 & 60.30 & 69.28 & \cellcolor{gray!20}\textbf{59.14}\dec{18.29} \\
%
%
Texture & 49.60 & 62.39  & 66.99 & 51.69 & 53.20 & 60.79 & 67.49 & \cellcolor{gray!20}\textbf{58.88}\dec{18.55}\\

Adversarial  & 25.5 & 34.89  & 48.19& 18.29& 24.40 & 30.29 & 48.49 & \cellcolor{gray!20}\textbf{32.87}\dec{46.25} \\

\midrule

\multirow{3}{*}{Datasets} & 

\multicolumn{7}{c}{EVA-CLIP} 
\\  \cmidrule(lr){2-8}

 & g/14 & g/14+ & B/16 & L/14 & L/14+ & E/14 & E/14+ & \cellcolor{gray!20}Average \\
 \midrule
 Original  & 88.80  & 92.69  & 89.19& 91.10& 91.99 & 93.80 & 94.60 & \cellcolor{gray!20}91.74  \\
 \midrule

ImageNet-E ($\lambda$=-20) & 84.74 & 88.98   & 85.55 & 89.19 & 88.78 & 92.02 & 91.81& \cellcolor{gray!20}88.72\dec{3.02}\\
ImageNet-E ($\lambda$=20) & 84.10 & 89.40  & 85.81 & 88.51 & 89.69& 92.69 & 92.50& \cellcolor{gray!20}88.95\dec{2.79}\\
ImageNet-E ($\lambda_{adv}=20$) & 79.69 &  85.45 & 80.20& 84.04 & 85.95 & 89.89 &89.59 & \cellcolor{gray!20}84.97\dec{6.77}\\
LANCE  & 70.25 & 77.40  & 73.26& 76.63 & 77.46 & 80.95 & 78.65& \cellcolor{gray!20}76.37\dec{15.37}\\
\midrule
Class label & 90.10 & 92.90& 88.61 & 91.31 & 91.90& 93.40& 93.41& \cellcolor{gray!20}91.66\dec{0.08} \\
BLIP-2 Caption & 80.31 & 84.29  & 82.10 & 82.50 & 84.80 & 86.90 & 86.90 & \cellcolor{gray!20}83.97\dec{7.77} \\

Color & 73.50 & 80.50 & 73.20 & 80.70 & 84.61 & 84.39 & 87.00 & \cellcolor{gray!20}80.55\dec{11.19} \\
 Texture & 75.30& 78.90&74.40 & 80.80& 82.10& 83.60& 85.60& \cellcolor{gray!20}80.10\dec{11.64}\\
Adversarial   & 55.59 & 62.49  & 48.70 & 65.39 & 73.59 & 70.29 & 73.29 & \cellcolor{gray!20}\textbf{64.19}\dec{27.55} \\
\bottomrule
\end{tabular}%
}
\label{tab:base_zs_comparison}
\end{table}

\noindent \textbf{Multimodal Training.}
In Table \ref{tab:base_zs_comparison}, we observe that compared to ImageNet-E ($\lambda=20$), our natural background variations lead to an average performance drop of $15.66\%$ and $8.85\%$ on CLIP and EVA-CLIP models. On comparing with ImageNet-E ($\lambda_{adv}=20$), our adversarial background variations lead to an average performance drop of $35.36\%$ and $20.78\%$ on CLIP and EVA-CLIP models. Similar to results mentioned in Table \ref{tab:base_class_comparison}, zero-shot robustness shows similar trend across different background changes. However, for background variations induced using class label information the performance increases in CLIP-based models. This reason could be the use of CLIP text encoder utilized for generating the textual embedding $e_{\mathcal{T}}$ for guiding the generation process of the diffusion model. 
On EVA-CLIP, which proposed changes to stabilize the training of CLIP models on large-scale datasets, we observe significant improvement in zero-shot performance across all background changes. In Appendix \ref{sec:prompt_evaluation}, we delve into the comparison between multimodal and unimodal models, offering detailed results on \texttt{ImageNet-B} dataset.

\begin{wrapfigure}[12]{r}{0.35\linewidth}
\centering
\vspace{-2em}
\includegraphics[width=1\linewidth]{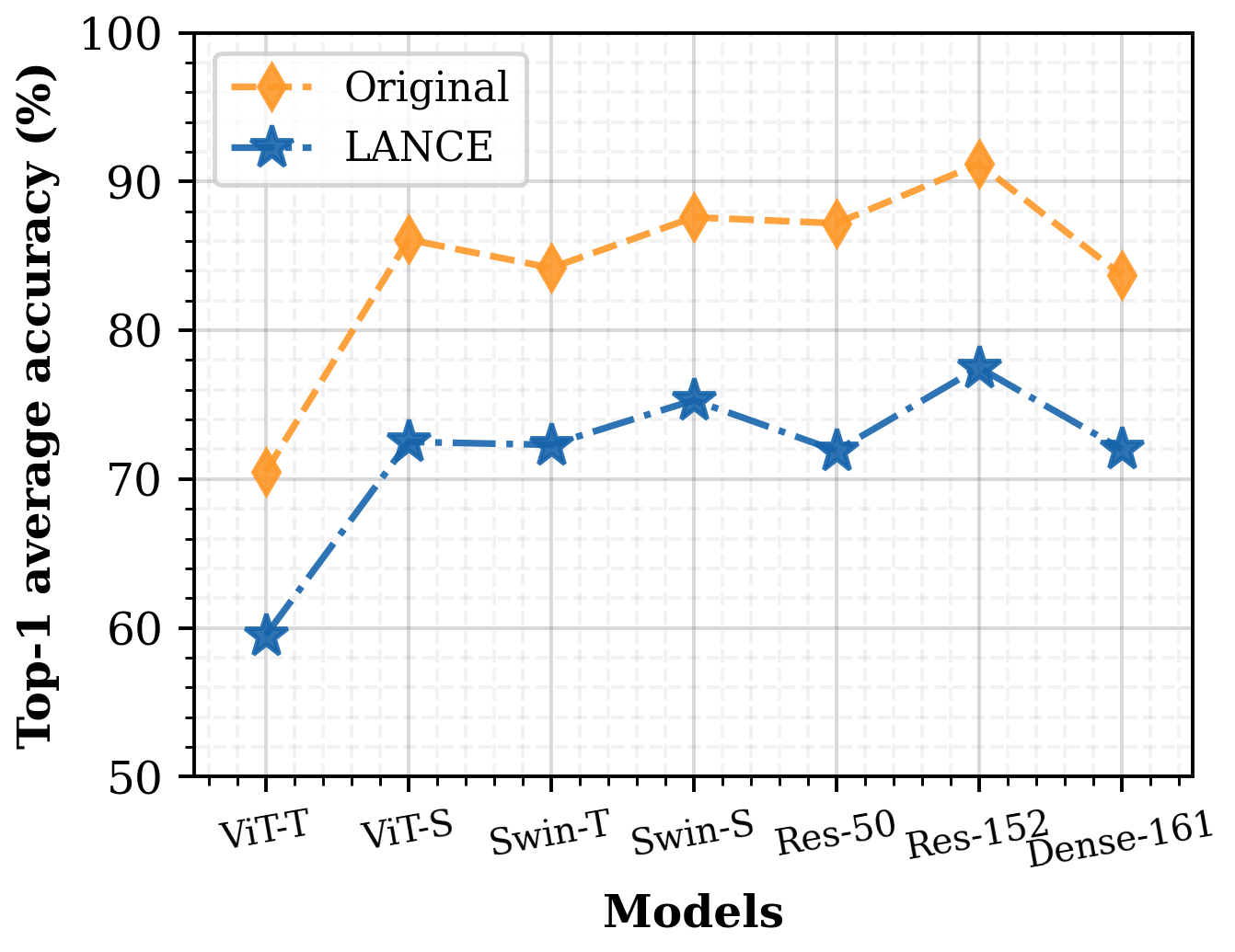}

\caption{Evaluating LANCE on $\texttt{ImageNet-B}_{1000}$ dataset with masked background.}
\label{fig:distortion}

\end{wrapfigure}

\noindent \textbf{Object Semantic distortion:} It's noteworthy to mention that in both Table \ref{tab:base_class_comparison} and \ref{tab:base_zs_comparison}, we observe a significant drop in performance of models across background changes induced by LANCE method. However, we discover that the drop in performance is not necessarily due to the induced background changes, rather than distorting the object semantics, making it unsuitable for evaluating object-to-background context. This observation is supported by Figure \ref{fig:distortion}, where we evaluate performance on original and LANCE generate images while masking the background. A significant performance drop is evident across all models, emphasizing the distortion of object semantics. We provide a detailed discussion with FID\cite{heusel2017gans} comparison and visualizations in Appendix \ref{sec:comparison_lance} and \ref{sec:qualitative comparison}.

\begin{figure}[t]
\begin{minipage}{\textwidth}

\centering

\begin{minipage}{0.49\textwidth}
  \centering
  \includegraphics[height=3.4cm, width=\linewidth , keepaspectratio]{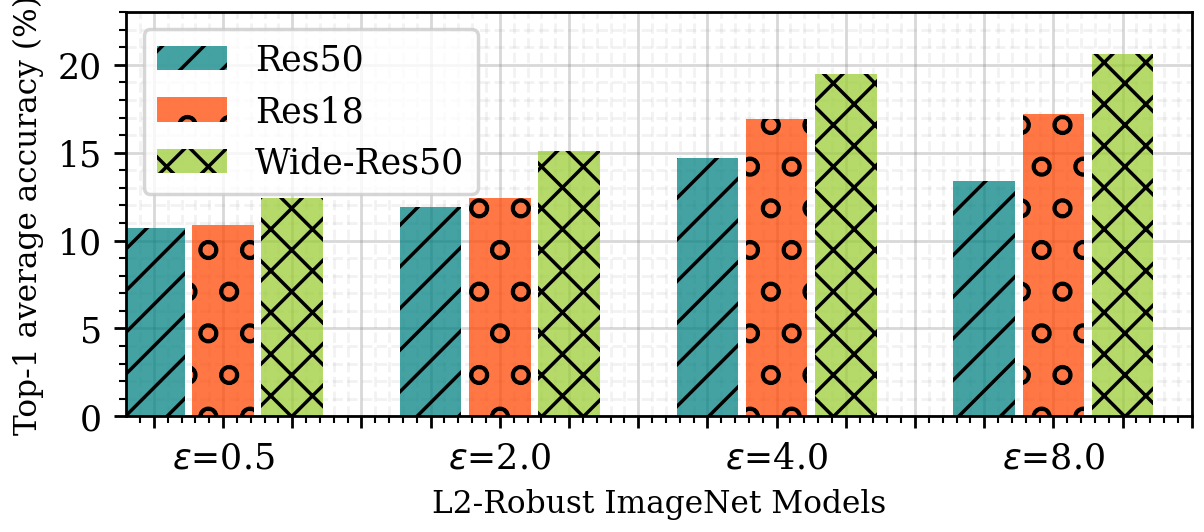}
\end{minipage}
\begin{minipage}{0.49\textwidth}
  \centering
  \includegraphics[height=3.4cm, width=\linewidth, keepaspectratio ]{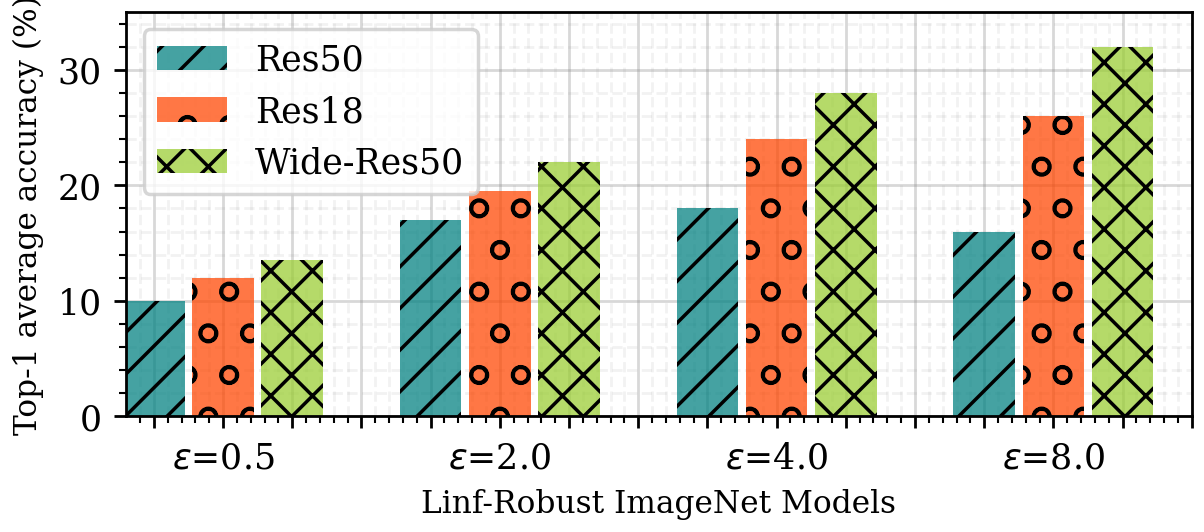}
\end{minipage}

  \begin{minipage}{0.16\textwidth}
  \centering
  \includegraphics[height=3.4cm, width=\linewidth , keepaspectratio]{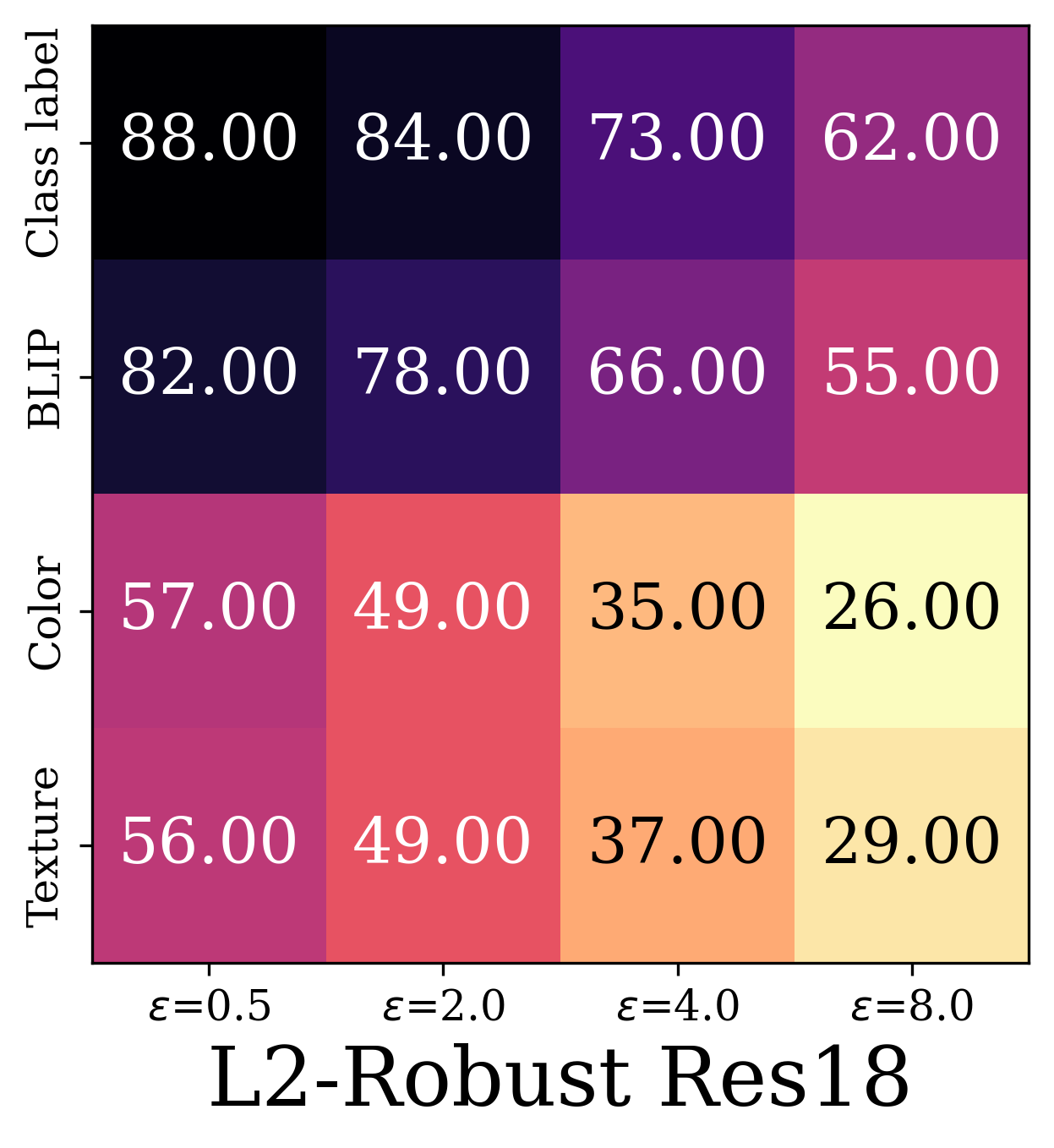}
\end{minipage}
  \begin{minipage}{0.16\textwidth}
  \centering
  \includegraphics[height=3.4cm, width=\linewidth , keepaspectratio]{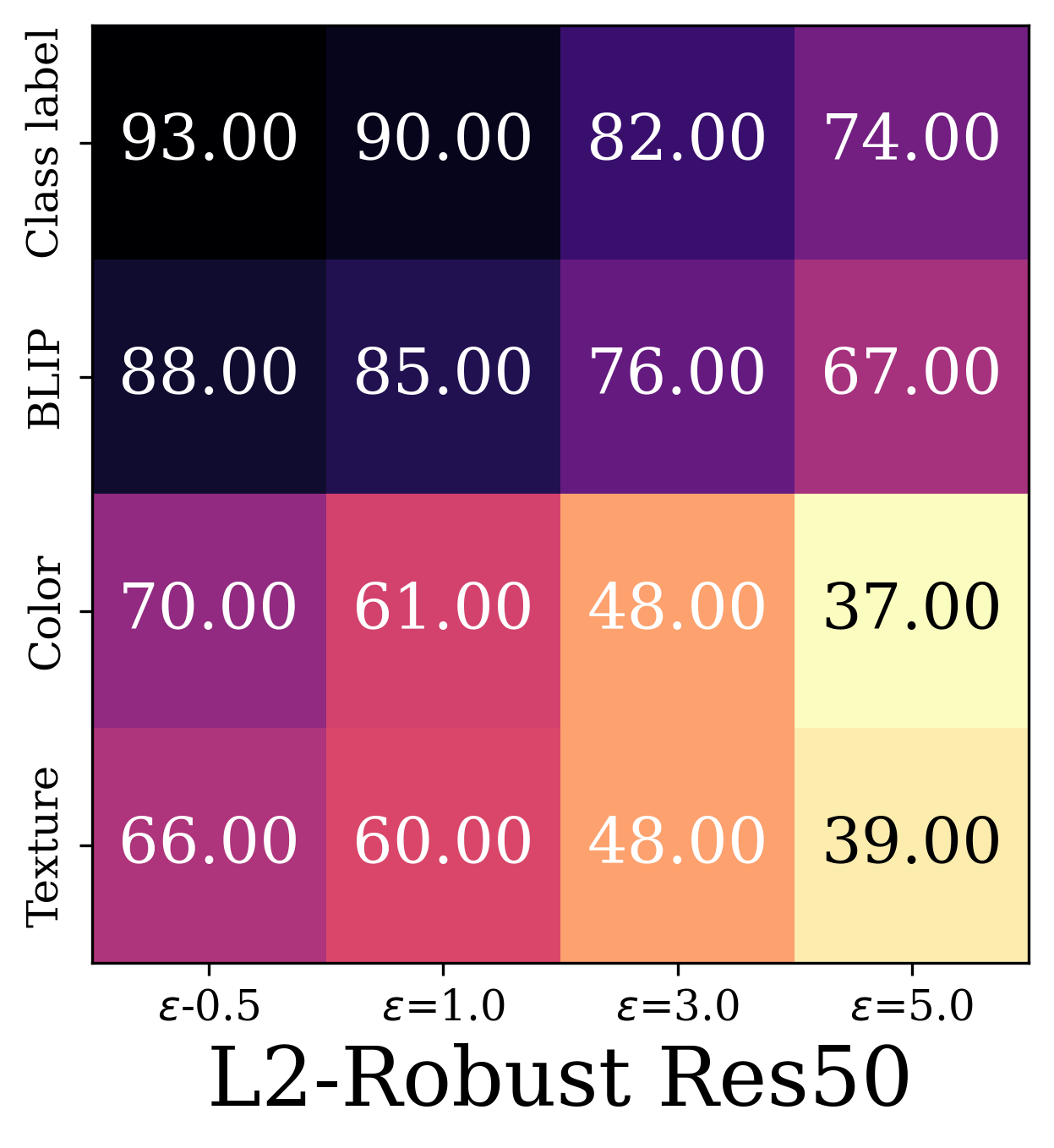}
\end{minipage}
\begin{minipage}{0.16\textwidth}
  \centering
  \includegraphics[height=3.4cm, width=\linewidth, keepaspectratio ]{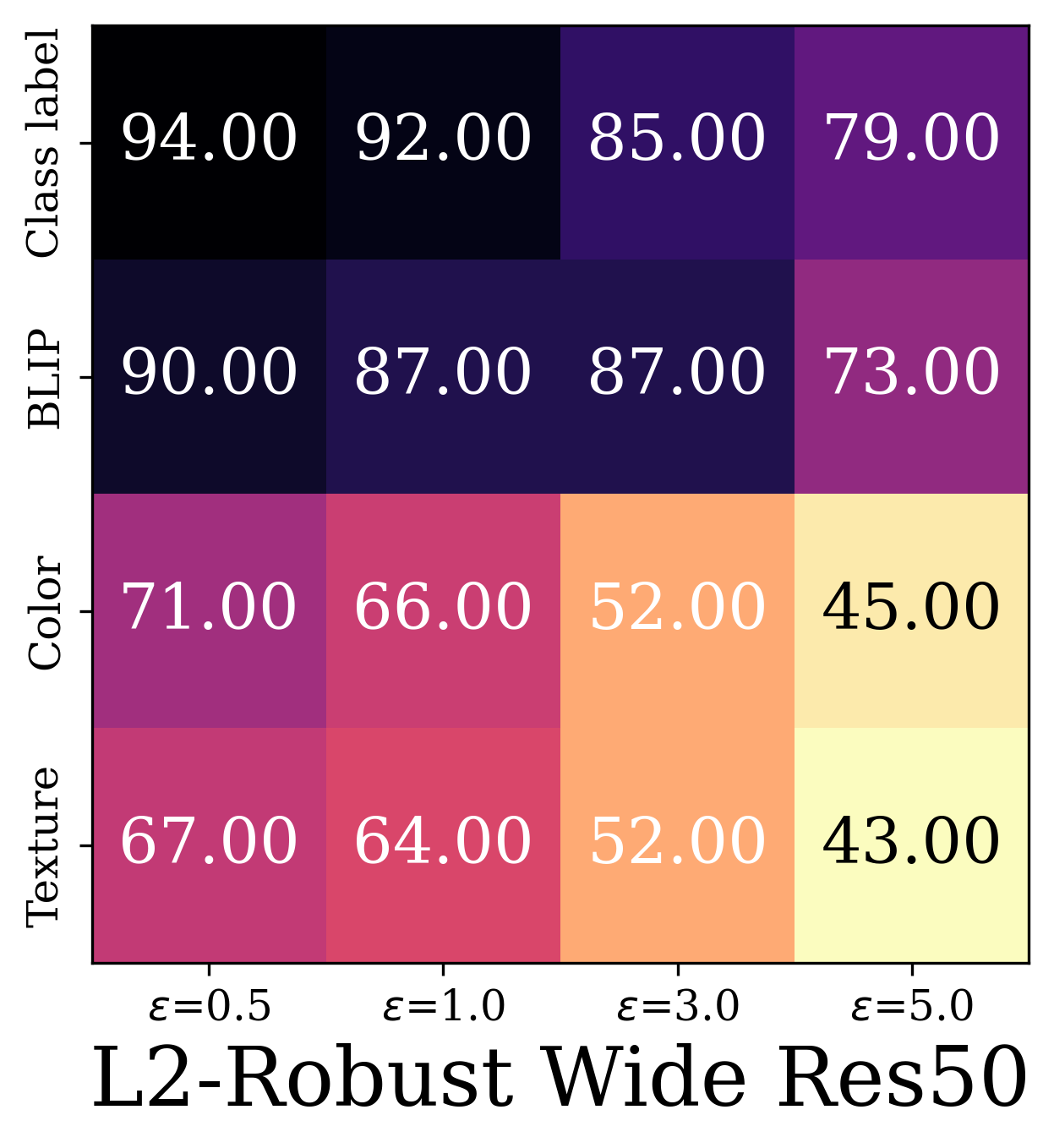}
\end{minipage}
\begin{minipage}{0.16\textwidth}
  \centering
  \includegraphics[height=3.4cm, width=\linewidth, keepaspectratio ]{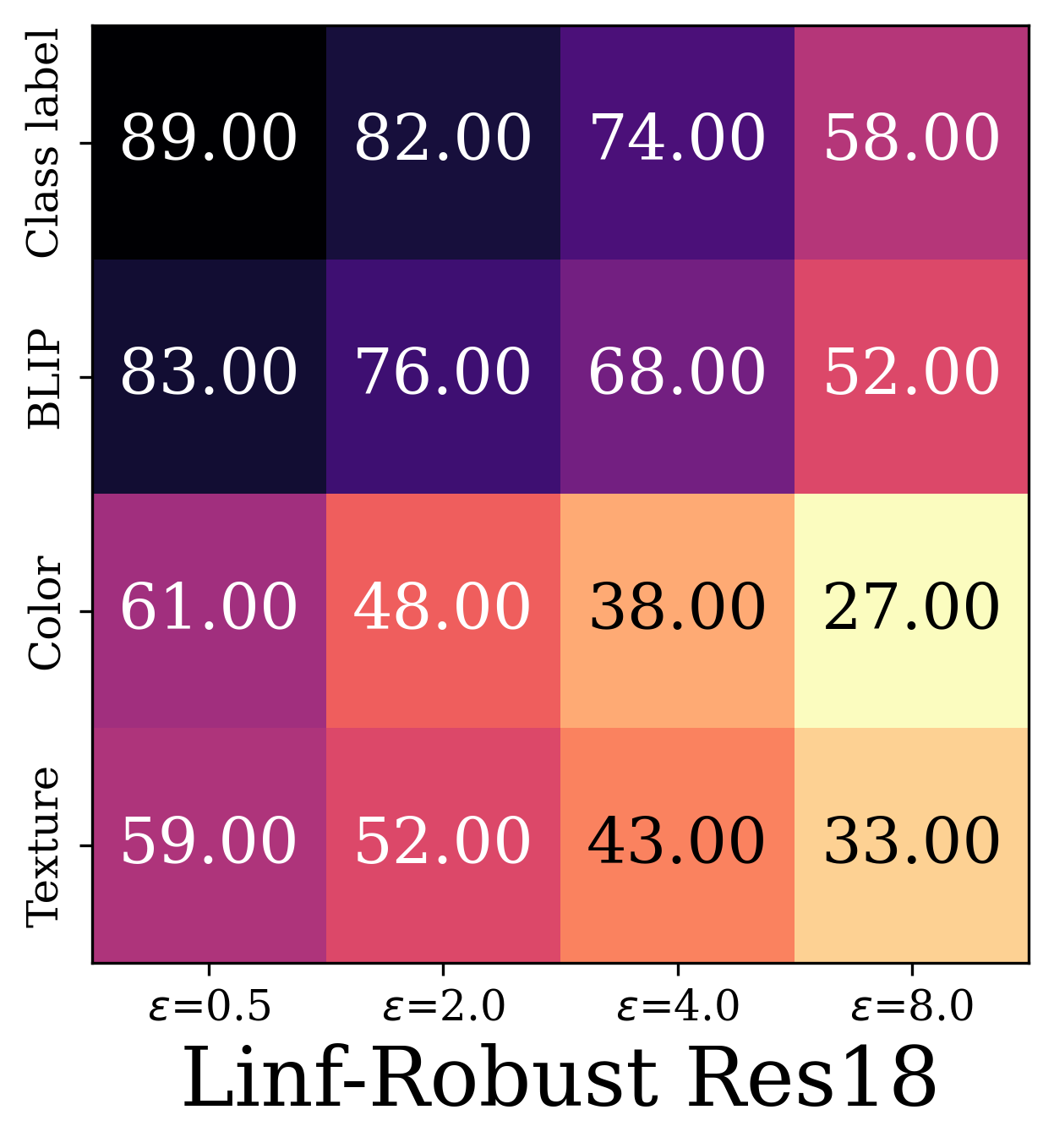}
\end{minipage}
\begin{minipage}{0.16\textwidth}
  \centering
  \includegraphics[height=3.4cm, width=\linewidth, keepaspectratio ]{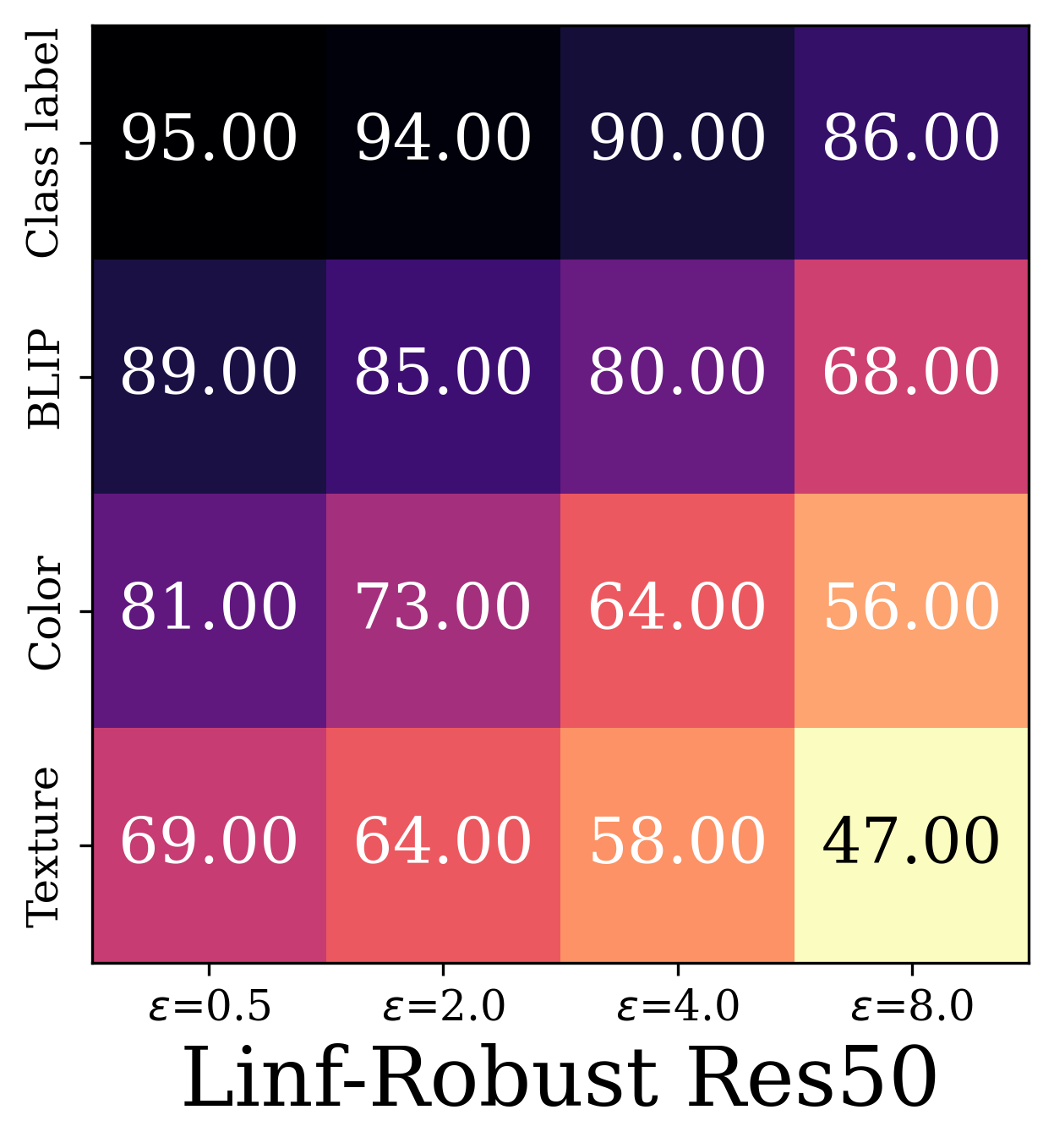}
\end{minipage}
\begin{minipage}{0.16\textwidth}
  \centering
  \includegraphics[height=3.4cm, width=\linewidth, keepaspectratio ]{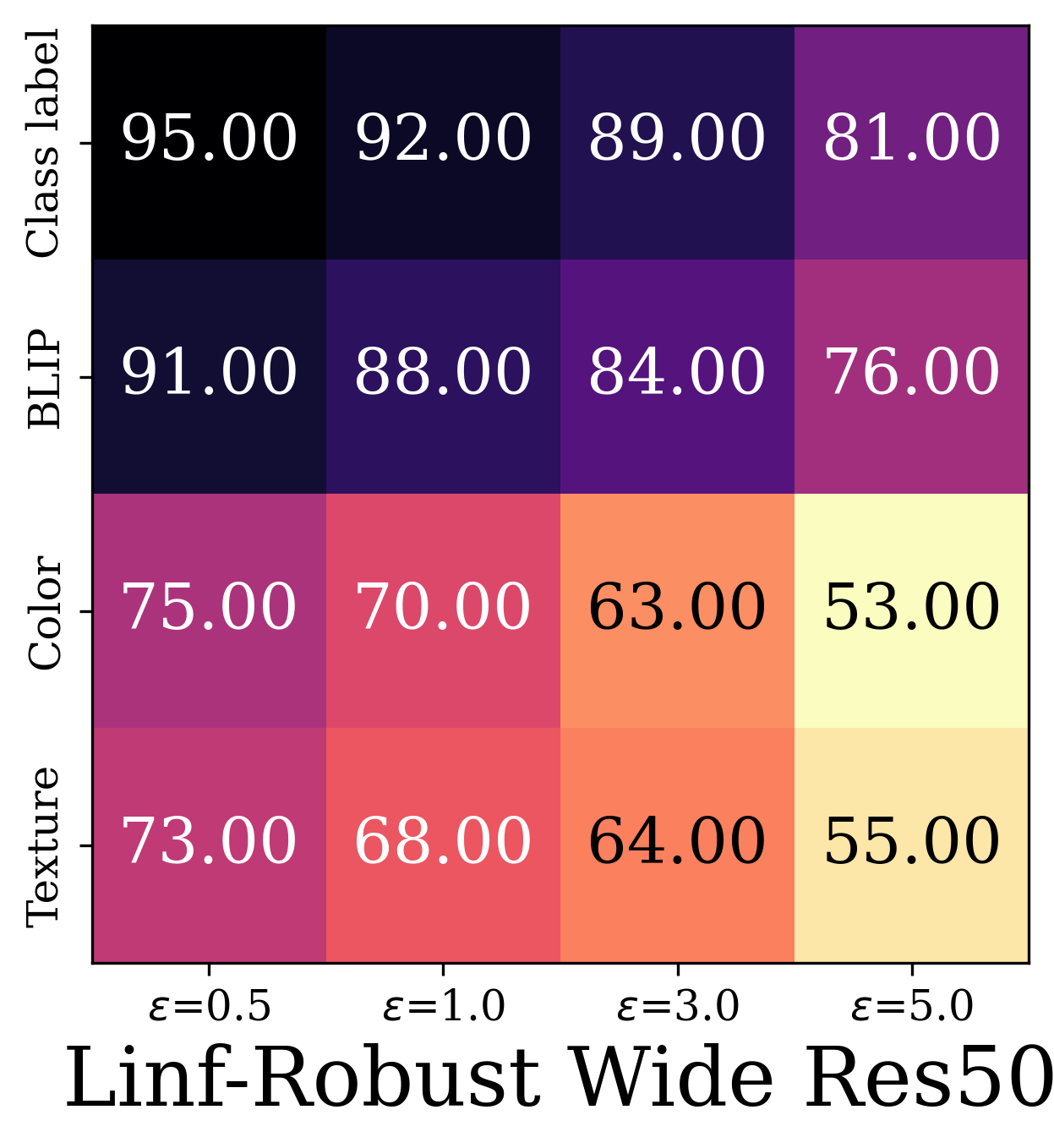}
\end{minipage}
\end{minipage}
  \caption{The top row plots the Top-1(\%) accuracy achieved by adversarially trained ResNet models on adversarial background changes on $\texttt{ImageNet-B}_{1000}$ and the bottom row indicates for the case of non-adversarial background changes on \texttt{ImageNet-B}. }
  \label{fig:adv_results}
\end{figure}

\subsection{Further Evaluations}

\noindent \textbf{Adversarial ImageNet Training.} 
As can be seen from Figure \ref{fig:adv_results}  \emph{(bottom row)}, our object-to-background compositional changes on $\texttt{ImageNet-B}$ lead to a significant decline in accuracy for adversarially trained models. This highlights the robustness of these models is limited to adversarial perturbations and does not transfer to different distribution shifts. In Figure \ref{fig:adv_results} \emph{(top row)}, when we evaluate these models on adversarial background changes on $\texttt{ImageNet-B}_{1000}$, the performance improves with an increase in adversarial robustness($\epsilon$) of the models. Furthermore, we also observe models with more capacity perform better, similar to results on natural training. For detailed results and comparison with baseline methods, refer to Appendix \ref{sec:comparison_adv_models}.

\noindent \textbf{Stylized ImageNet Training.} Despite the focus of Stylized ImageNet training \cite{geirhos2018imagenet} to encourage models to concentrate on the foreground of the scene by reducing background cues for prediction \cite{naseer2021intriguing}, our findings indicate that it is still susceptible to both natural and adversarial object-to-background variations (see Table \ref{tab:stylised}). Consequently, its applicability appears to be constrained to specific distribution shifts.

\begin{wrapfigure}[10]{r}{0.45\linewidth}
\centering
\includegraphics[width=1\linewidth]{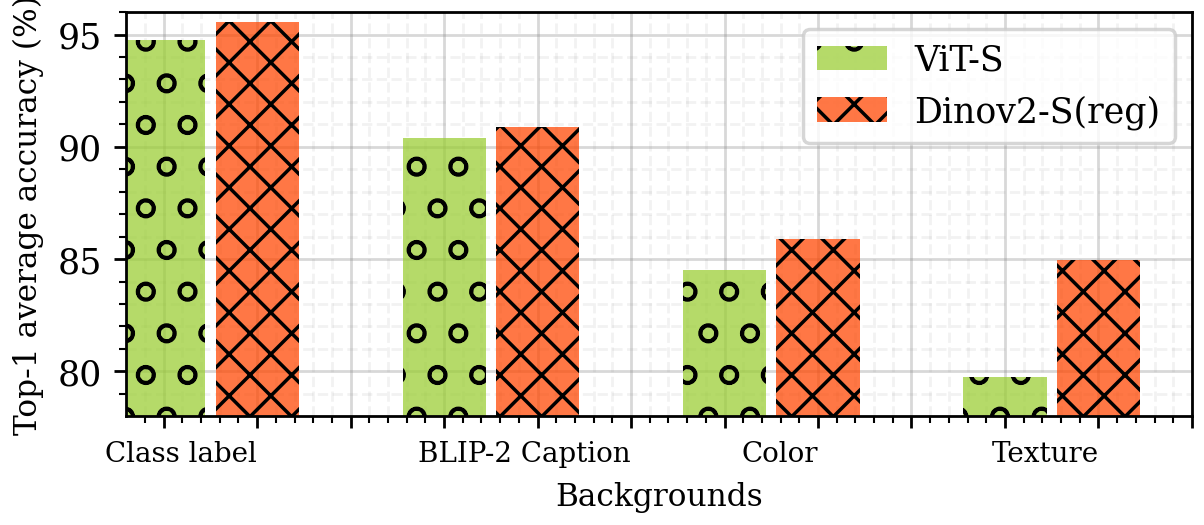}
\caption{Evaluating Dinov2 models on \texttt{ImageNet-B} background changes.}
\label{fig:dinov2}
\end{wrapfigure}

\noindent \textbf{Self-Supervised Training.} Improved performance is observed in Dinov2 models across object-to-background variations (see Figure \ref{fig:dinov2}). We hypothesize this improvement is acquired through training on extensive curated datasets and the utilization of additional learnable registers/tokens during training for refining the interpretability of attention maps. For more details, refer to Appendix \ref{sec: DINOv2}.

\begin{figure}[!t]
    \centering
    \includegraphics[height=3.8cm, width=\linewidth , keepaspectratio]{
    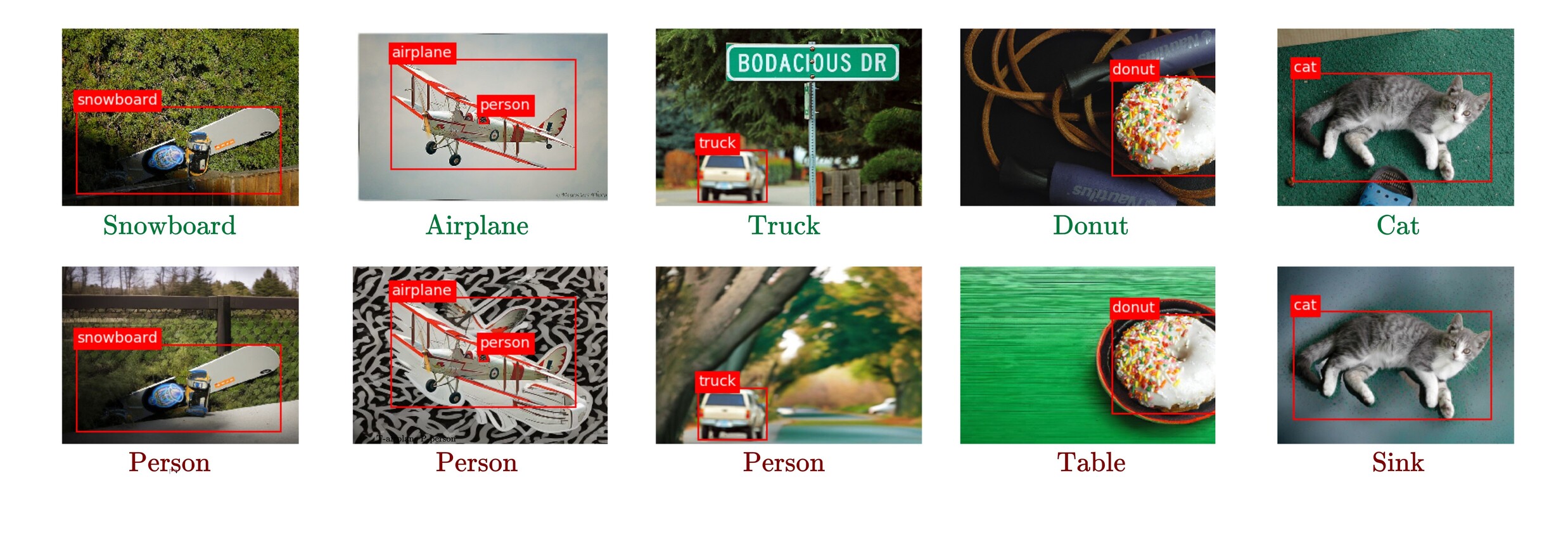
    }
    \caption{
Correct predictions by Mask-RCNN and Res-50  on the original image \emph{(top row)} and the  corresponding predictions on altered backgrounds \emph{(bottom row)}. }
    \label{detr}

\end{figure}

\noindent \textbf{Segmentation and Detection.}
We observe a consistent decrease in AP scores on object detection and instance segmentation tasks across background variations generated on \texttt{COCO-DC}(see Table \ref{tab:AP}). The adversarial background results in the lowest AP scores, but still remains at a reasonable level given that the adversarial examples are generated using a classification model, with limited cross-task transferability. Moreover, our qualitative observations suggest detection and segmentation models exhibit greater resilience to changes in the background compared to classifiers (see Figure \ref{detr} and Appendix \ref{sec:detection results}).

\noindent \textbf{Image Captioning.} Table \ref{tab:blip} shows the CLIP scores between captions from clean and generated images using the BLIP-2 model. Scores decrease with color, texture, and adversarial background changes (Appendix  \ref{sec:image captioning} for qualitative results).
\begin{table}[t]
\centering
\begin{minipage}{0.45\textwidth} 
\centering
\caption{\small Stylized Training Evaluation }
\label{tab:stylised}
\setlength{\tabcolsep}{2pt}
\scalebox{0.5}[0.5]{%
\begin{tabular}{llccl}
\toprule
\multirow{2}{*}{Datasets} & 
\multirow{2}{*}{Background} & 
\multicolumn{2}{c}{Stylized Trained models} \\
\cmidrule(lr){3-4}
& &  DeiT-S & DeiT-T & \cellcolor{gray!20}Average \\
\midrule
\multirow{5}{*}{\texttt{ImageNet-B}}& Original    & 91.22 & 87.21 & \cellcolor{gray!20} 89.21 \\
& Class label    & 89.35 & 85.35 & \cellcolor{gray!20} 87.35\dec{1.86} \\
& BLIP-2 Caption   & 84.01 & 79.19 & \cellcolor{gray!20} 81.60\dec{7.61} \\
& Color   & 66.57 & 57.54 & \cellcolor{gray!20} 62.05\dec{27.15} \\
& Texture  & 64.08 & 54.82 & \cellcolor{gray!20} 59.45\dec{29.76} \\
\cmidrule(lr){1-5}
\multirow{1}{*}{$\texttt{ImageNet-B}_{1000}$} & Original   &  89.60  & 85.90  &\cellcolor{gray!20} 87.75 \\
&Adversarial  & 15.90 &  10.80  &\cellcolor{gray!20} 13.35\dec{74.40} \\
\bottomrule
\end{tabular}%
}
\end{minipage}%
\hfill
\begin{minipage}{0.25\textwidth}
\caption{\small Image-to-Caption (BLIP-2) Evaluation}
\label{tab:blip}
\centering
\setlength{\tabcolsep}{2pt}
\scalebox{0.5}[0.5]{%
    \begin{tabular}{ll|c}
      \toprule
      Dataset & Background & CLIP Score \\
      \midrule
     \texttt{ImageNet-B} &Class Label         & 0.75                \\
      &BLIP-2 Caption        & 0.84                \\
      &Color               & 0.66                \\
      &Texture             & 0.67                \\
      \midrule
      $\texttt{ImageNet-B}_{1000}$&Adversarial         & 0.62                \\
      \bottomrule
    \end{tabular}
    }
  
  \end{minipage}
\hfill
  \begin{minipage}{0.25\textwidth}
\caption{\small Mask AP and Segment AP score on \texttt{COCO-DC}}
\label{tab:AP}
\centering
\setlength{\tabcolsep}{2pt}
\scalebox{0.5}[0.5]{%
\begin{tabular}{l|c|c}

      \toprule
      Background & Box AP & Segment AP\\
      \midrule
      Original            & 57.99            & 56.29               \\
      BLIP-2 Caption       & 47.40            & 44.75               \\
      Color               &  48.12           & 45.09               \\
      Texture             & 45.79            & 43.07               \\
      \bottomrule
      Adversarial         & 37.10            & 34.91                  \\
      \bottomrule
    \end{tabular}
    }
  \end{minipage}
\end{table}

\section{Conclusion}

In this study, we propose \textsc{ObjectCompose}, a method for generating object-to-background compositional changes. Our method addresses the limitations of current works, specifically distortion of object semantics and diversity in background changes. We accomplish this by utilizing the capabilities of image-to-text and image-to-segmentation foundational models to preserve the object semantics, while we optimize for diverse object-to-background compositional changes by modifying the textual prompts or optimizing the latents of the text-to-image model. \textsc{ObjectCompose} offers a complimentary evaluation protocol to the existing ones, for comprehensive evaluations across current vision-based models to reveal their vulnerability to background alterations. In Appendix \ref{sec:insights}, we elaborate on the initial insights gained from our work and discuss current limitations and future directions.

\newpage

\bibliographystyle{splncs04}
\bibliography{reference}

\newpage
\appendix

\section{Appendix}
\section*{Overview}
\textbf{}

{\hspace{0cm} \ref{sec:prompt}. List of Prompts}

{\hspace{0cm} \ref{sec:algorithm}. Algorithm}

{\hspace{0cm} \ref{sec:comparison_lance}. Object Distortion in LANCE}

{\hspace{0cm} \ref{sec:qualitative comparison}. Qualitative Comparison with Related Works}

{\hspace{0cm} \ref{sec:prompt_evaluation}. Ablation on Background Changes}

{\hspace{0cm} \ref{sec:comparison_adv_models}. Evaluation on Adversarially Trained models}

{\hspace{0cm} \ref{sec:recent_models}. Evaluation on Recent Vision models}

{\hspace{0cm} \ref{sec: DINOv2}. Evaluation on DINOv2 models}

{\hspace{0cm} \ref{sec:image captioning}. Vision Language model for Image Captioning}

{\hspace{0cm} \ref{sec:detection results}. Qualitative Results on Detection}

{\hspace{0cm} \ref{sec:FastSAM}. Effect of Background Change on Segmentation Models}

{\hspace{0cm} \ref{sec:feature space}. Exploring Feature Space of Vision Models }

{\hspace{0cm} \ref{sec:diversity}. Diversity and Diffusion Parameter Ablation}

{\hspace{0cm} \ref{sec:misclassified}. Misclassified Samples}

{\hspace{0cm} \ref{sec:external Factors}. Potential External Factors}

{\hspace{0cm} \ref{sec:Dataset}. Dataset Distribution and Comparison}

{\hspace{0cm} \ref{sec: rmasked}. Evaluation on Background/Foreground Images}

{\hspace{0cm} \ref{sec:insights}. Insights}

{\hspace{0cm} \ref{sec:calibration}. Calibration Metrics}

{\hspace{0cm} \ref{sec:adv_loss_ablation}. Ablation on Adversarial Loss}

{\hspace{0cm} \ref{sec:statements}. Reproducibility and Ethics Statement}

\subsection{List of prompts}
We provide the list of prompts that are used to guide the diffusion model to generate diverse background changes, encompassing different distribution shifts with respect to the original data distribution.
\label{sec:prompt}
\begin{table}[ht]
\caption{Prompts used to create background alterations}
\centering
\begin{tabular}{p{0.25\linewidth}p{0.65\linewidth}}
\hline
\textbf{Background} & \textbf{Prompts} \\
\hline
Class label & \textit{"This is a picture of a \textit{class name}"} \\
BLIP-2 Caption & Captions generated from BLIP-2 image to caption \\
\hline
Color\textsubscript{prompt-1} & \textit{"This is a picture of a vivid red background"} \\
Color\textsubscript{prompt-2} & \textit{"This is a picture of a vivid green background"} \\
Color\textsubscript{prompt-3} & \textit{"This is a picture of a vivid blue background"} \\
Color\textsubscript{prompt-4} & \textit{"This is a picture of a vivid colorful background"} \\
\hline
Texture\textsubscript{prompt-1} & \textit{"This is a picture of textures in the background"} \\
Texture\textsubscript{prompt-2} & \textit{"This is a picture of intricately textured background"} \\
Texture\textsubscript{prompt-3} & \textit{"This is a picture of colorful textured background"} \\
Texture\textsubscript{prompt-4} & \textit{"This is a photo of distorted textures in the background"} \\
\hline
Adversarial & Captions generated from BLIP-2 image to caption. \\
\hline
\end{tabular}
\label{prompt}
\end{table}

\newpage
\subsection{Algorithm}
\label{sec:algorithm}
We provide the algorithm (Algo. \ref{Attack-algo}) for our approach of generating adversarial backgrounds by optimizing the textual and visual conditioning of the diffusion model. We also tried to optimize only the conditional embeddings or the latent embeddings, but achieve better attack success rate by optimizing both. Note that for crafting adversarial examples on \texttt{COCO-DC} we use ImageNet trained ResNet-50 classifiers and our adversarial objective is to maximize the feature representation distance between clean and adversarial samples. Furthermore, for introducing desired non-adversarial background changes using the textual description $\mathcal{T^{'}}$, the optimization of the latent and embedding is not needed.

\begin{algorithm}[h]
\caption{Background Generation}
\label{Attack-algo}
\begin{algorithmic}[1]

\Require Conditioning module $\mathcal{C}$, Diffusion model $\epsilon_{\theta}$, Autoencoder $\mathcal{V}$, CLIP text encoder $\psi_{\texttt{CLIP}}$,
image $\mathcal{I}$, class label $\bm{y}$, classifier $\mathcal{F}_{\phi}$, denoising steps $T$, guidance scale $\lambda$, attack iterations $N$, and learning rate $\beta$ for AdamW optimizer $\mathcal{A}$.
\State Get the textual and visual conditioning from the image $\mathcal{I}$
\begin{equation*}
    \mathcal{C}(\mathcal{I}, \bm{y}) = \mathcal{T}_{\mathbf{B}}, \mathcal{M}
\end{equation*}
\State Modify $\mathcal{T}_{\mathbf{B}}$ to $\mathcal{T}$ for desired background change.
\State Map the mask $\mathcal{M}$ and image $\mathcal{I}$ to latent space: $i, m \gets \mathcal{V}_{\texttt{ENC}}(\mathcal{I},\mathcal{M})$ 
\State Get the embedding of the textual discription $\mathcal{T}$: $e_{\mathcal{T}} \gets \psi_{\texttt{CLIP}}(\mathcal{T})$
\State Randomly initialize the latent $z_{T}$
\State Get the denoised latent $z_t$ at time step $t$.
\For {$n\in [1, 2, \ldots N]$}
\For {$t\in [t, t+1, \ldots T]$}
\State $\hat{\epsilon}_{\theta}^{t}(z_{t},e_{\mathcal{T}},i, m) = \epsilon_{\theta}^{t}(z_{t},i, m) + \lambda \left( \epsilon_{\theta}^{t}(z_{t},e_{\mathcal{T}}, i, m) - \epsilon_{\theta}^{t}(z_{t},i, m) \right)$
\State From noise estimate $\hat{\epsilon}_{\theta}$ get $z_{t-1}$.
\EndFor
\State Project the latents to pixel space: $\mathcal{I}_{adv} \gets \mathcal{V}_{\texttt{DEC}}(z_0)$
\State Compute Adversarial Loss: \begin{equation*}
\mathcal{L}_{adv} = \mathcal{L}_{CE}(\mathcal{F}_{\phi}(\mathcal{I}_{adv}), \bm{y}) \end{equation*}
\State Update $z_t$ and $e_{\mathcal{T}}$ using $\mathcal{A}$ to maximize $\mathcal{L}_{adv}$:
\begin{equation*}
   		    z_t, e_\mathcal{T} \gets \mathcal{A}\left(\nabla_{z_t}\mathcal{L}_{adv},\nabla_{e_{\mathcal{T}}}\mathcal{L}_{adv} \right)
    		\end{equation*}
\EndFor

\State \hrulefill

\State Generate Adversarial image $\mathcal{I}_{adv}$ using updated $z_t$ and $e_{\mathcal{T}}$. 
\end{algorithmic}
\end{algorithm}

\newpage

\subsection{Object Distortion in LANCE}
\label{sec:comparison_lance}

\begin{figure}[h]
\begin{minipage}{\textwidth}
\centering
\begin{minipage}{\textwidth}
\centering
\Large
    \footnotesize Original
\end{minipage}
\begin{minipage}{0.135\textwidth}
  \centering
  \includegraphics[height=2.1cm, width=\linewidth, keepaspectratio]{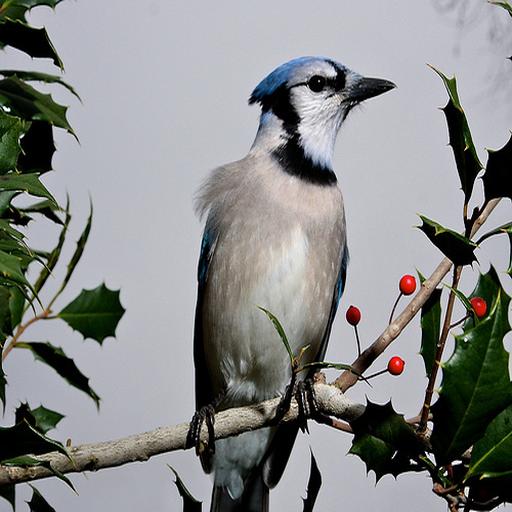}
\end{minipage}
\begin{minipage}{0.135\textwidth}
  \centering
  \includegraphics[height=2.1cm, width=\linewidth, keepaspectratio]{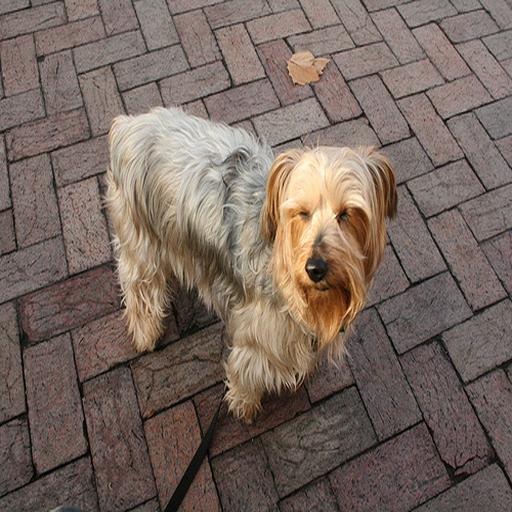}
\end{minipage}
\begin{minipage}{0.135\textwidth}
  \centering
  \includegraphics[height=2.1cm, width=\linewidth, keepaspectratio]{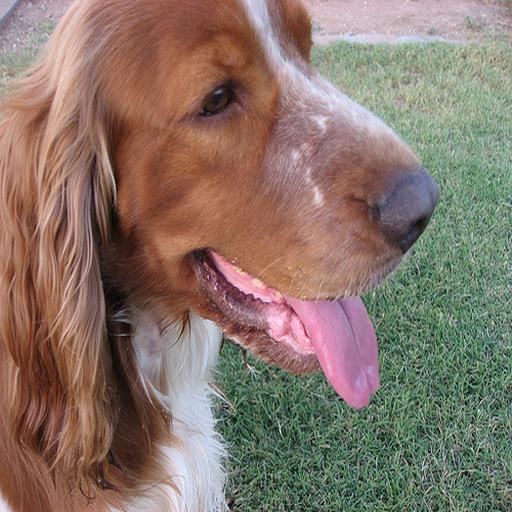}
\end{minipage}
\begin{minipage}{0.135\textwidth}
  \centering
  \includegraphics[height=2.1cm, width=\linewidth, keepaspectratio]{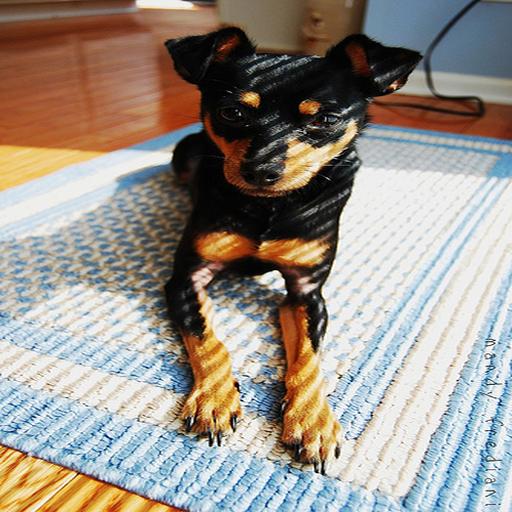}
\end{minipage}
\begin{minipage}{0.135\textwidth}
  \centering
  \includegraphics[height=2.1cm, width=\linewidth, keepaspectratio]{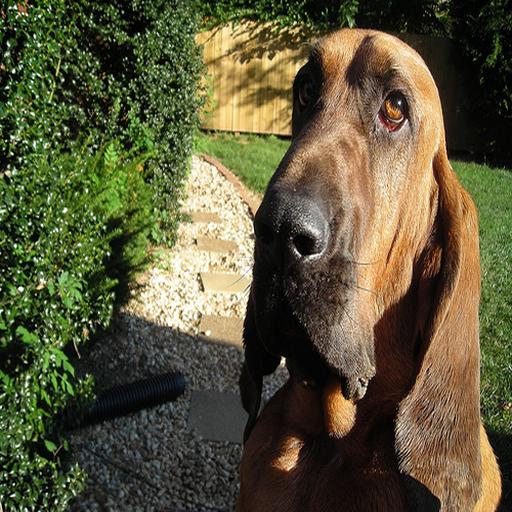}
\end{minipage}
\begin{minipage}{0.135\textwidth}
  \centering
  \includegraphics[height=2.1cm, width=\linewidth, keepaspectratio]{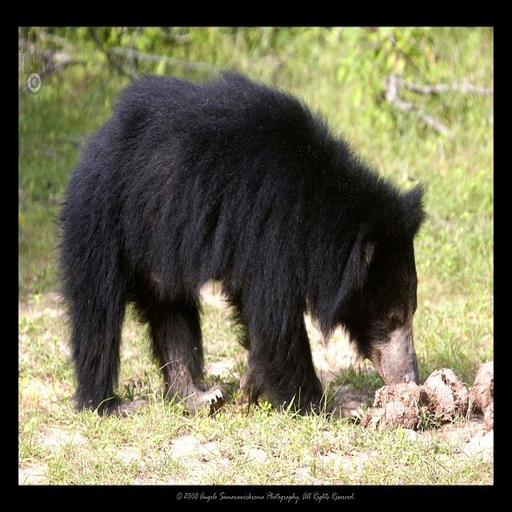}
\end{minipage}
\begin{minipage}{0.135\textwidth}
  \centering
  \includegraphics[height=2.1cm, width=\linewidth, keepaspectratio]{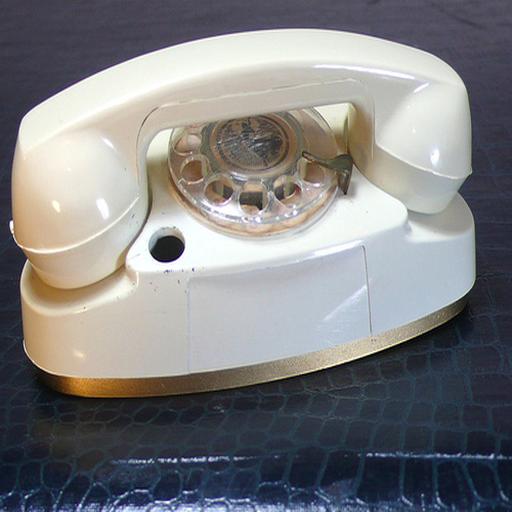}
\end{minipage}

\begin{minipage}{\textwidth}
\centering
\Large
    \footnotesize LANCE
\end{minipage}
\begin{minipage}{0.135\textwidth}
  \centering
  \includegraphics[height=2.1cm, width=\linewidth, keepaspectratio]{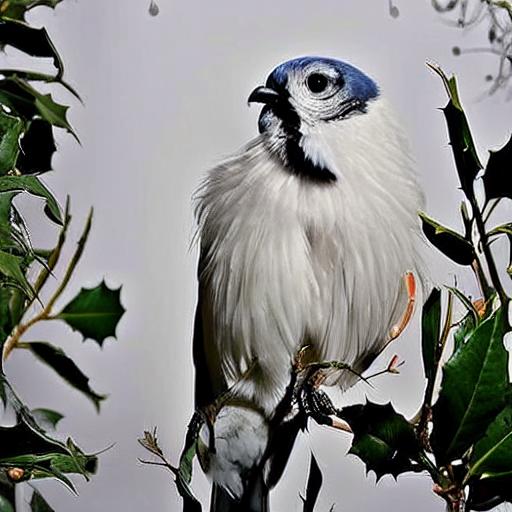}
\end{minipage}
\begin{minipage}{0.135\textwidth}
  \centering
  \includegraphics[height=2.1cm, width=\linewidth, keepaspectratio]{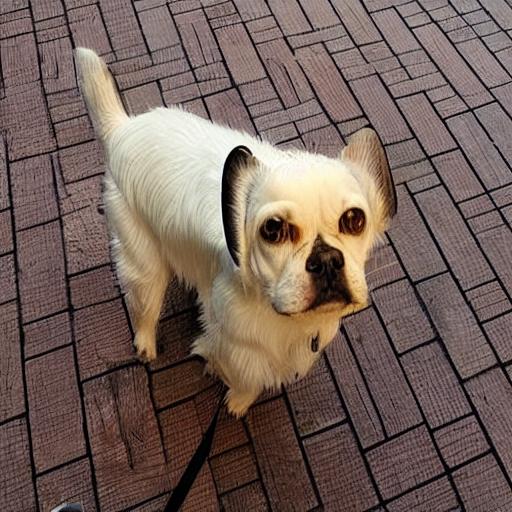}
\end{minipage}
\begin{minipage}{0.135\textwidth}
  \centering
  \includegraphics[height=2.1cm, width=\linewidth, keepaspectratio]{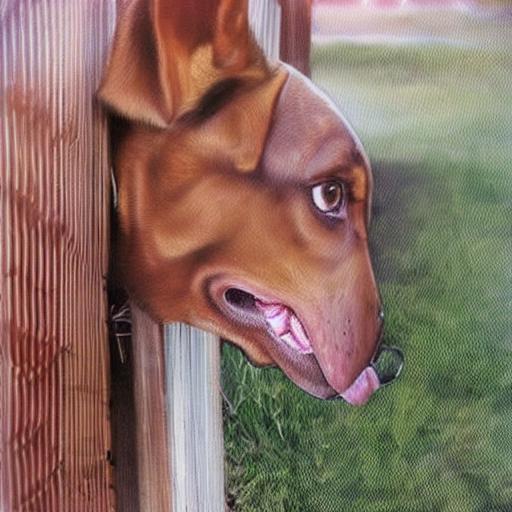}
\end{minipage}
\begin{minipage}{0.135\textwidth}
  \centering
  \includegraphics[height=2.1cm, width=\linewidth, keepaspectratio]{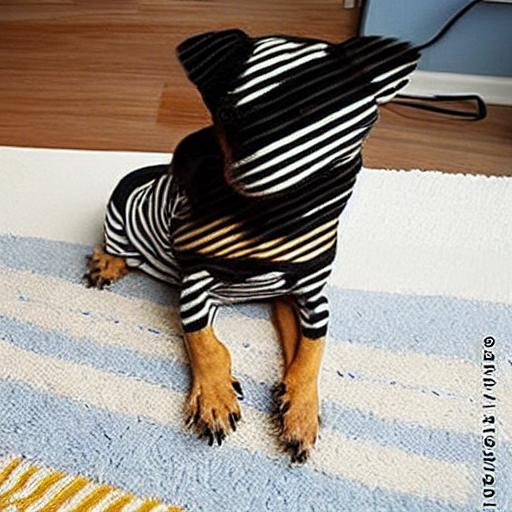}
\end{minipage}
\begin{minipage}{0.135\textwidth}
  \centering
  \includegraphics[height=2.1cm, width=\linewidth, keepaspectratio]{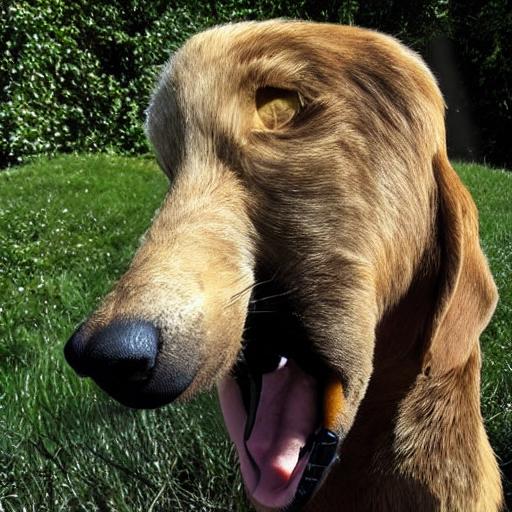}
\end{minipage}
\begin{minipage}{0.135\textwidth}
  \centering
  \includegraphics[height=2.1cm, width=\linewidth, keepaspectratio]{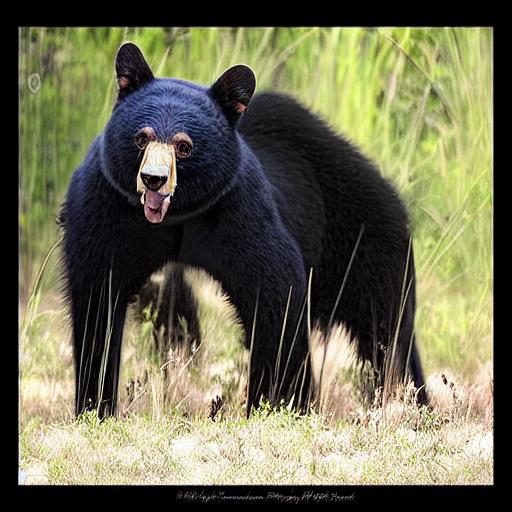}
\end{minipage}
\begin{minipage}{0.135\textwidth}
  \centering
  \includegraphics[height=2.1cm, width=\linewidth, keepaspectratio]{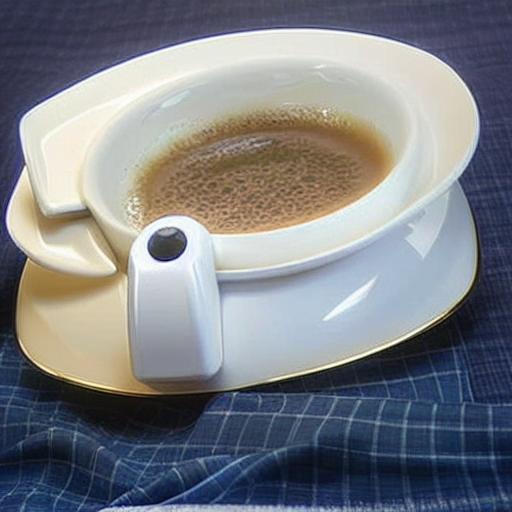}
\end{minipage}


\begin{minipage}{\textwidth}
\centering
\Large
    \footnotesize Ours
\end{minipage}
\begin{minipage}{0.135\textwidth}
  \centering
  \includegraphics[height=2.1cm, width=\linewidth, keepaspectratio]{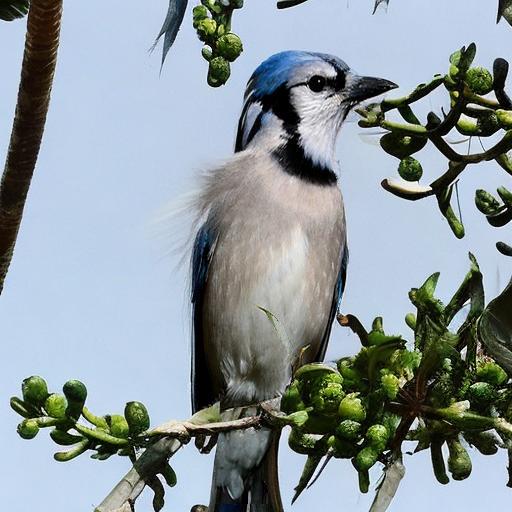}
\end{minipage}
\begin{minipage}{0.135\textwidth}
  \centering
  \includegraphics[height=2.1cm, width=\linewidth, keepaspectratio]{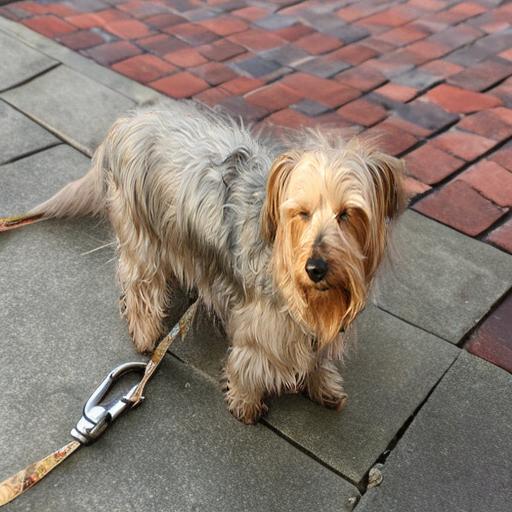}
\end{minipage}
\begin{minipage}{0.135\textwidth}
  \centering
  \includegraphics[height=2.1cm, width=\linewidth, keepaspectratio]{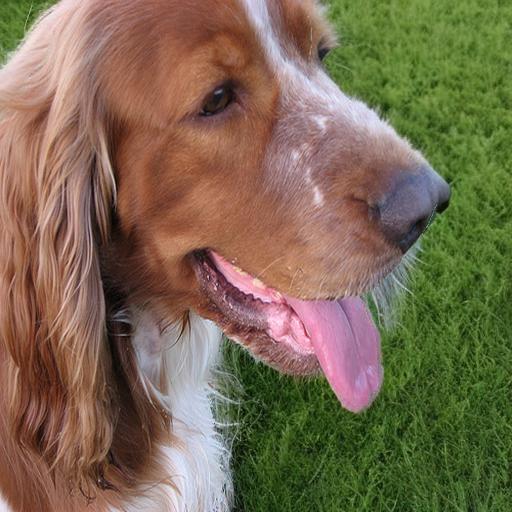}
\end{minipage}
\begin{minipage}{0.135\textwidth}
  \centering
  \includegraphics[height=2.1cm, width=\linewidth, keepaspectratio]{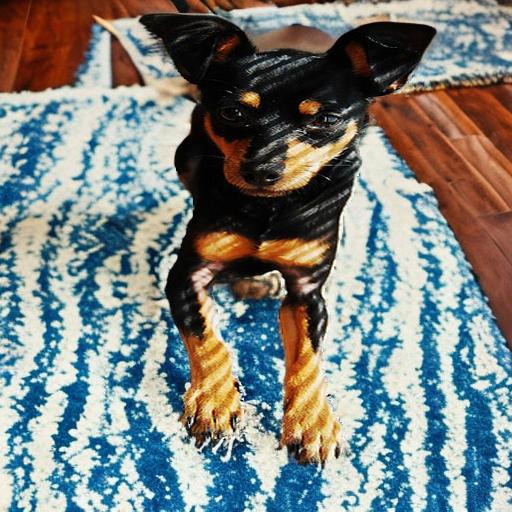}
\end{minipage}
\begin{minipage}{0.135\textwidth}
  \centering
  \includegraphics[height=2.1cm, width=\linewidth, keepaspectratio]{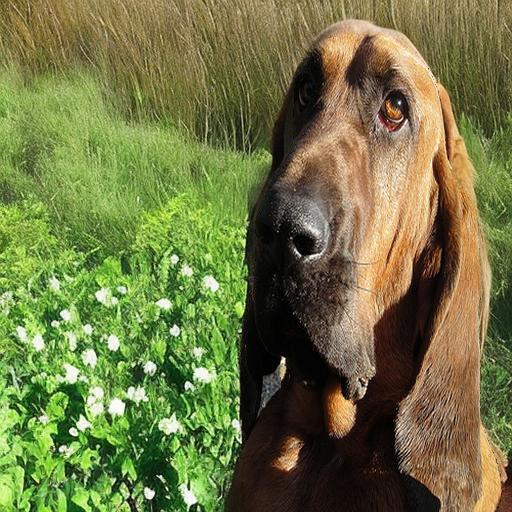}
\end{minipage}
\begin{minipage}{0.135\textwidth}
  \centering
  \includegraphics[height=2.1cm, width=\linewidth, keepaspectratio]{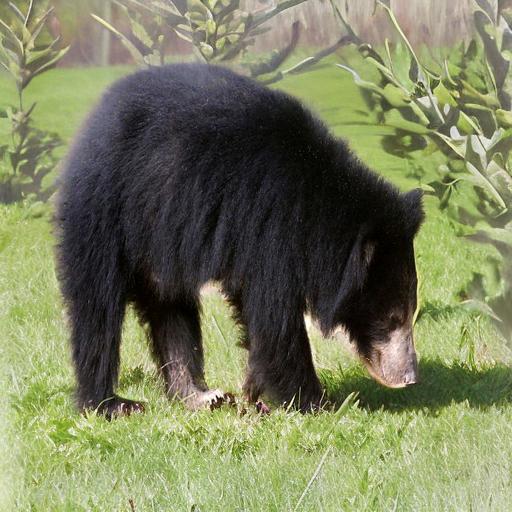}
\end{minipage}
\begin{minipage}{0.135\textwidth}
  \centering
  \includegraphics[height=2.1cm, width=\linewidth, keepaspectratio]{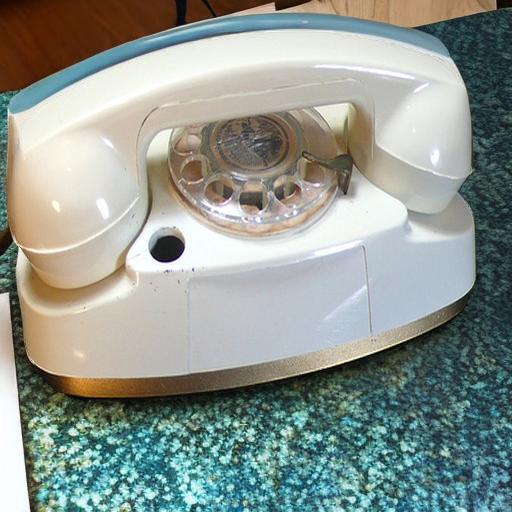}
\end{minipage}

\end{minipage}
\hfill
  \caption{Background Compositional changes on $\texttt{ImageNet-B}_{1000}$ dataset using LANCE and our method. LANCE fails to preserve object semantics, while our method exclusively edits the background.}
  \label{fig:lance}
\end{figure}

\begin{figure}[h]
\begin{minipage}{\textwidth}

\centering
\begin{minipage}{\textwidth}
\centering
\Large
    \footnotesize Original
\end{minipage}
\begin{minipage}{0.135\textwidth}
  \centering
  \includegraphics[height=2.1cm, width=\linewidth, keepaspectratio]{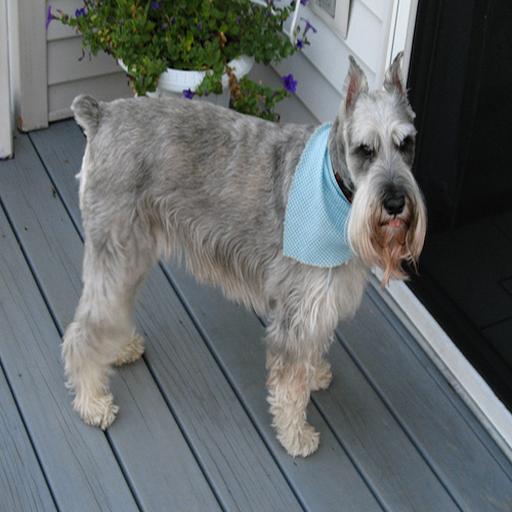}
\end{minipage}
\begin{minipage}{0.135\textwidth}
  \centering
  \includegraphics[height=2.1cm, width=\linewidth, keepaspectratio]{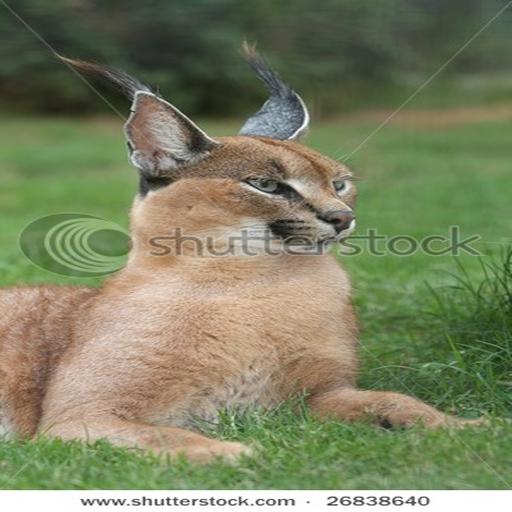}
\end{minipage}
\begin{minipage}{0.135\textwidth}
  \centering
  \includegraphics[height=2.1cm, width=\linewidth, keepaspectratio]{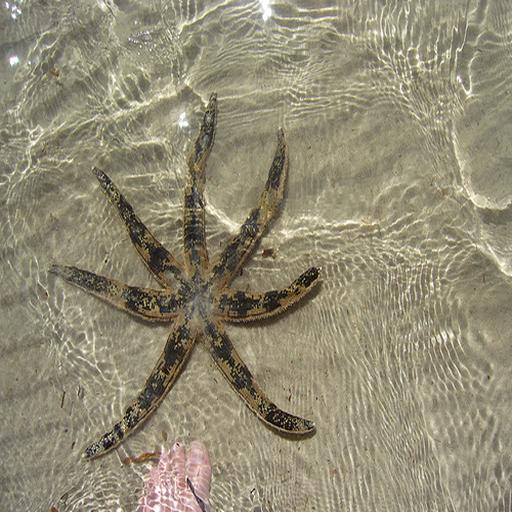}
\end{minipage}
\begin{minipage}{0.135\textwidth}
  \centering
  \includegraphics[height=2.1cm, width=\linewidth, keepaspectratio]{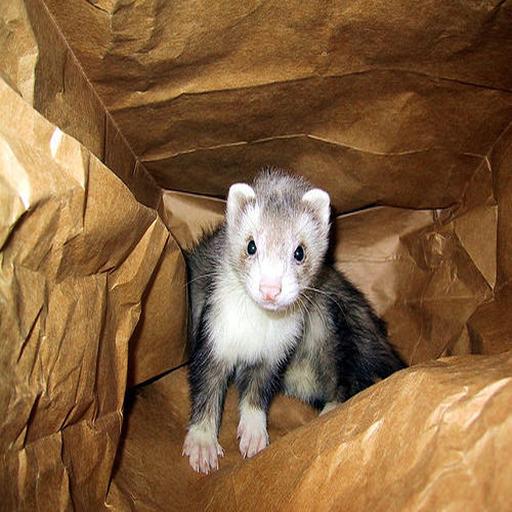}
\end{minipage}
\begin{minipage}{0.135\textwidth}
  \centering
  \includegraphics[height=2.1cm, width=\linewidth, keepaspectratio]{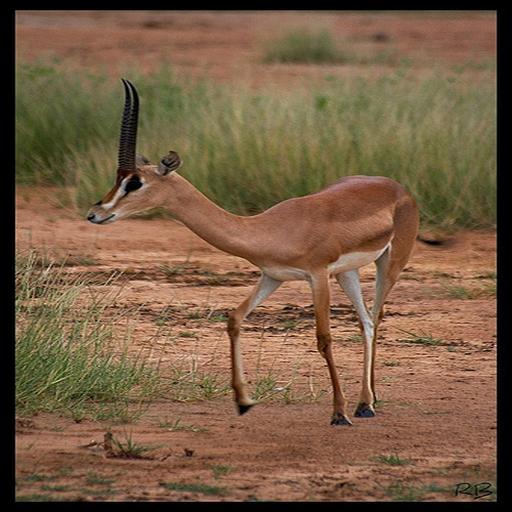}
\end{minipage}
\begin{minipage}{0.135\textwidth}
  \centering
  \includegraphics[height=2.1cm, width=\linewidth, keepaspectratio]{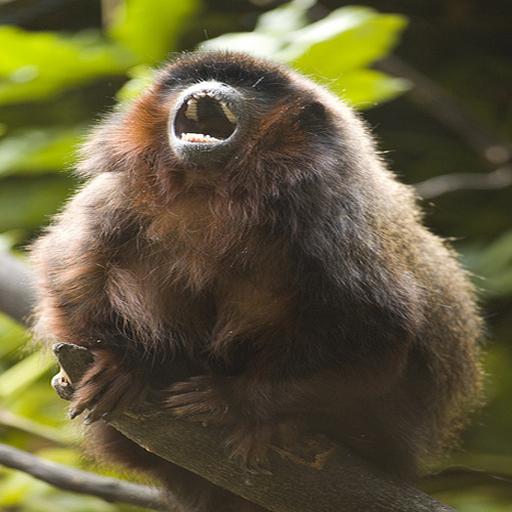}
\end{minipage}
\begin{minipage}{0.135\textwidth}
  \centering
  \includegraphics[height=2.1cm, width=\linewidth, keepaspectratio]{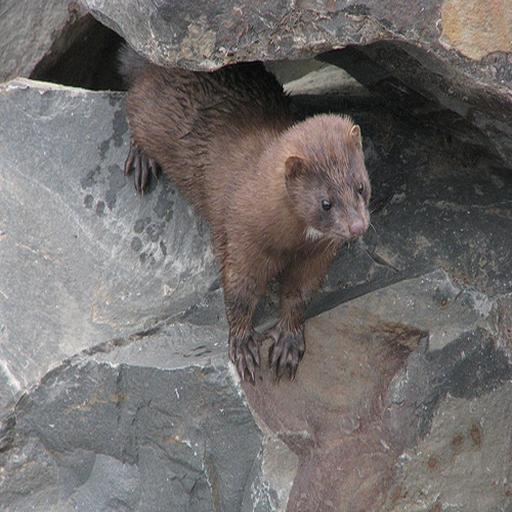}
\end{minipage}

\begin{minipage}{\textwidth}
\centering
\Large
    \footnotesize LANCE
\end{minipage}
\begin{minipage}{0.135\textwidth}
  \centering
  \includegraphics[height=2.1cm, width=\linewidth, keepaspectratio]{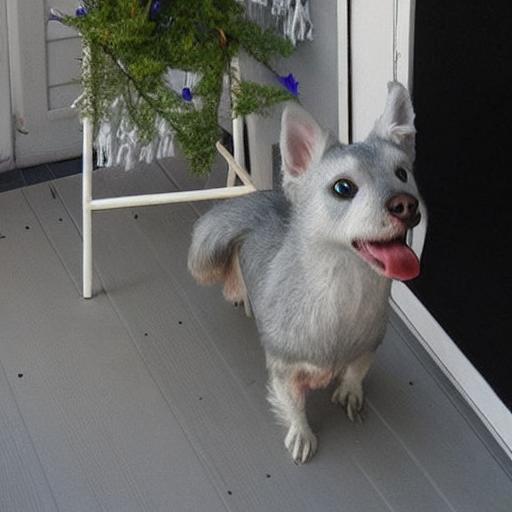}
\end{minipage}
\begin{minipage}{0.135\textwidth}
  \centering
  \includegraphics[height=2.1cm, width=\linewidth, keepaspectratio]{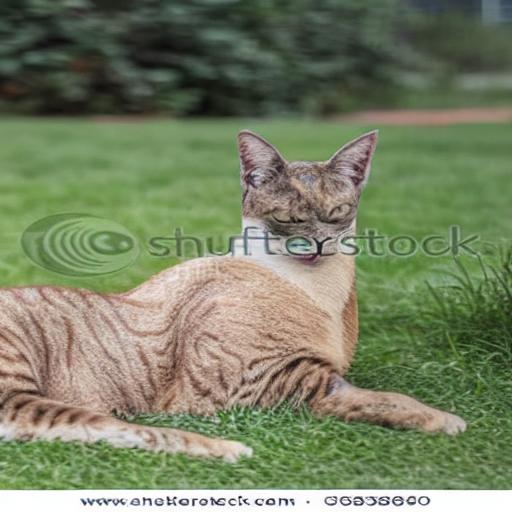}
\end{minipage}
\begin{minipage}{0.135\textwidth}
  \centering
  \includegraphics[height=2.1cm, width=\linewidth, keepaspectratio]{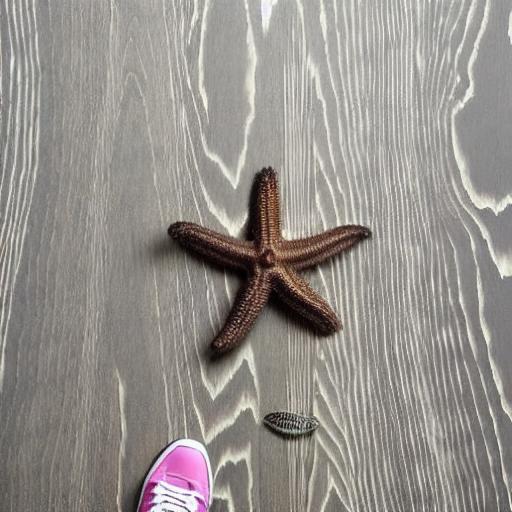}
\end{minipage}
\begin{minipage}{0.135\textwidth}
  \centering
  \includegraphics[height=2.1cm, width=\linewidth, keepaspectratio]{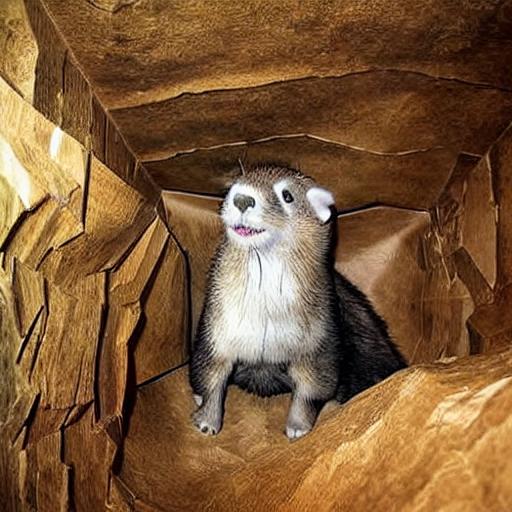}
\end{minipage}
\begin{minipage}{0.135\textwidth}
  \centering
  \includegraphics[height=2.1cm, width=\linewidth, keepaspectratio]{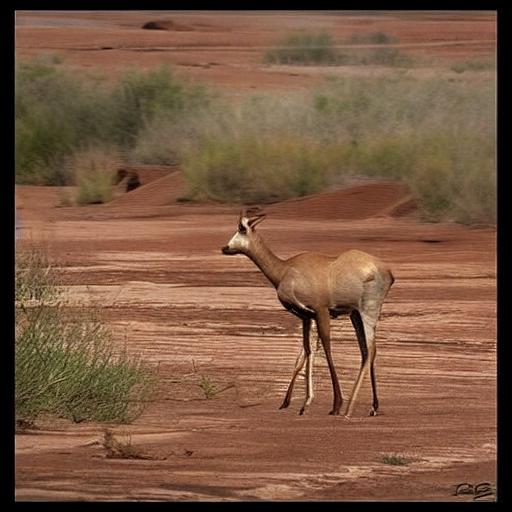}
\end{minipage}
\begin{minipage}{0.135\textwidth}
  \centering
  \includegraphics[height=2.1cm, width=\linewidth, keepaspectratio]{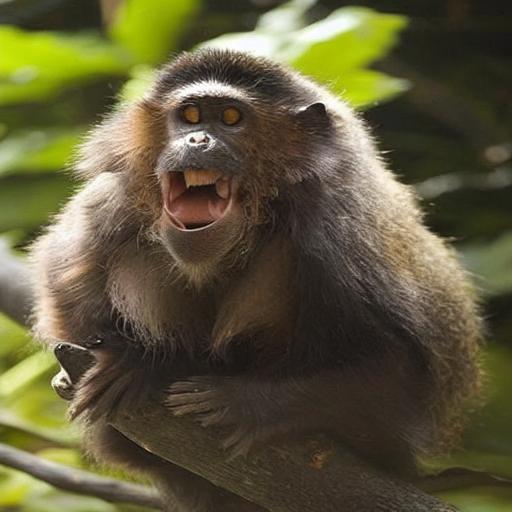}
\end{minipage}
\begin{minipage}{0.135\textwidth}
  \centering
  \includegraphics[height=2.1cm, width=\linewidth, keepaspectratio]{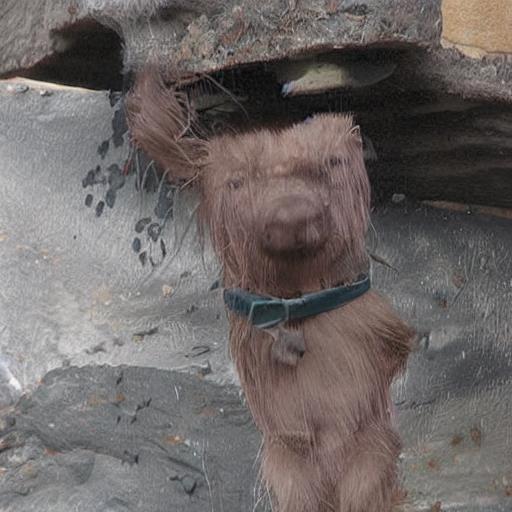}
\end{minipage}


\begin{minipage}{\textwidth}
\centering
\Large
    \footnotesize Ours
\end{minipage}
\begin{minipage}{0.135\textwidth}
  \centering
  \includegraphics[height=2.1cm, width=\linewidth, keepaspectratio]{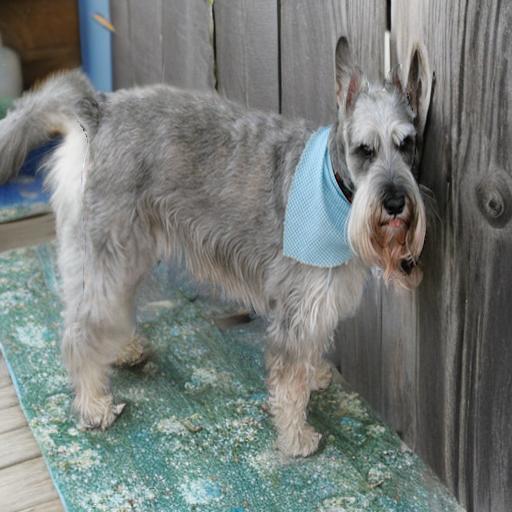}
\end{minipage}
\begin{minipage}{0.135\textwidth}
  \centering
  \includegraphics[height=2.1cm, width=\linewidth, keepaspectratio]{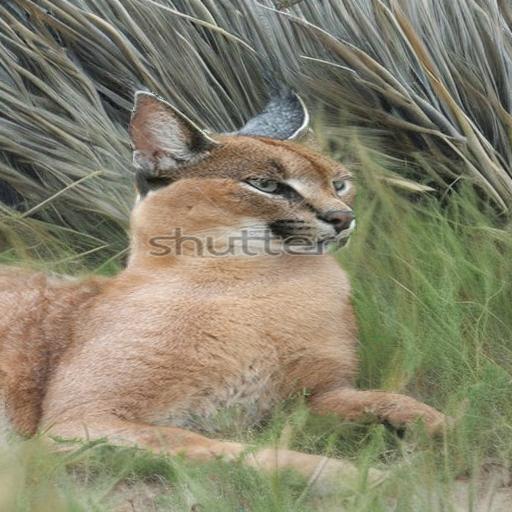}
\end{minipage}
\begin{minipage}{0.135\textwidth}
  \centering
  \includegraphics[height=2.1cm, width=\linewidth, keepaspectratio]{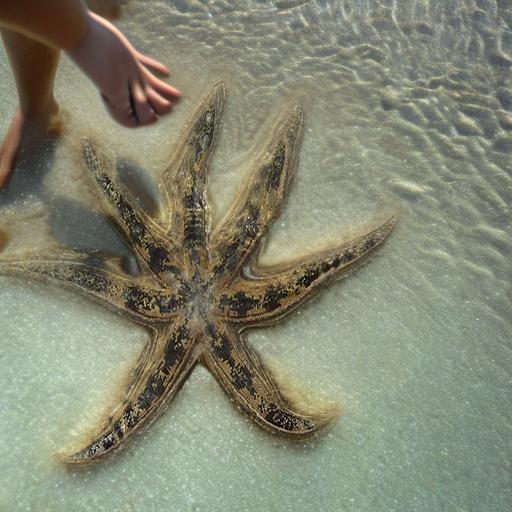}
\end{minipage}
\begin{minipage}{0.135\textwidth}
  \centering
  \includegraphics[height=2.1cm, width=\linewidth, keepaspectratio]{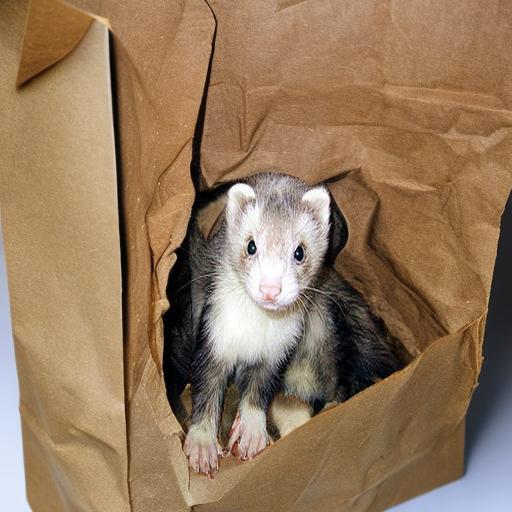}
\end{minipage}
\begin{minipage}{0.135\textwidth}
  \centering
  \includegraphics[height=2.1cm, width=\linewidth, keepaspectratio]{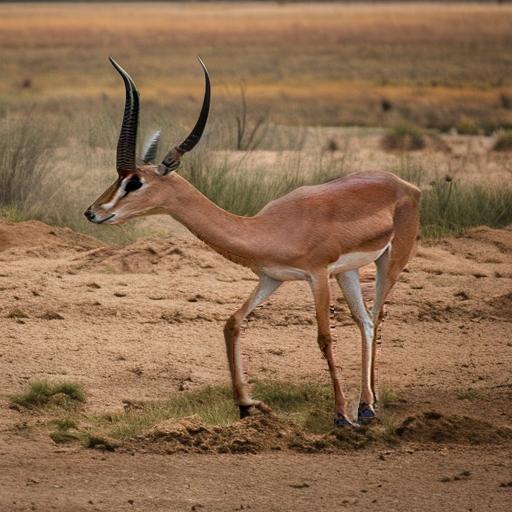}
\end{minipage}
\begin{minipage}{0.135\textwidth}
  \centering
  \includegraphics[height=2.1cm, width=\linewidth, keepaspectratio]{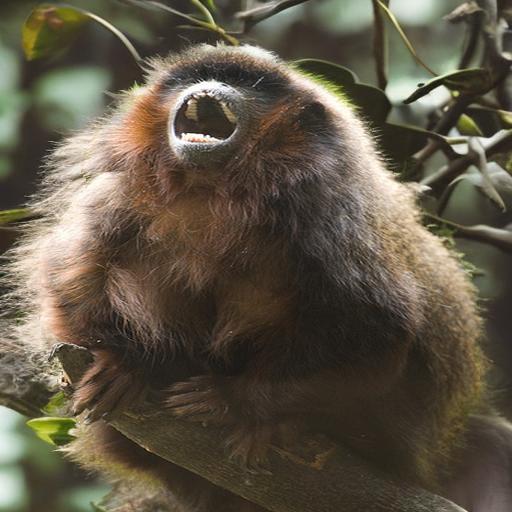}
\end{minipage}
\begin{minipage}{0.135\textwidth}
  \centering
  \includegraphics[height=2.1cm, width=\linewidth, keepaspectratio]{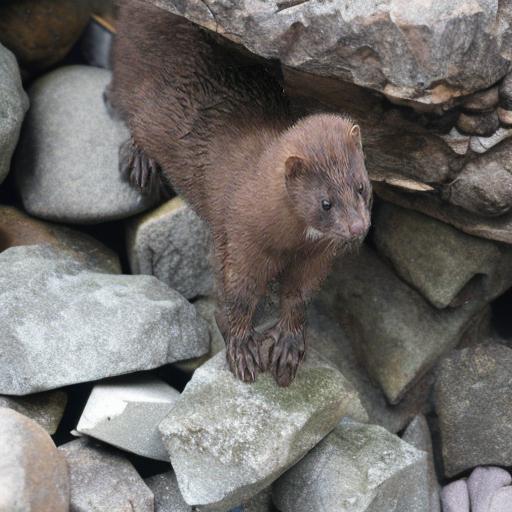}
\end{minipage}

\end{minipage}
\hfill
  \caption{Background Compositional changes on $\texttt{ImageNet-B}_{1000}$ dataset using LANCE and our method. LANCE fails to preserve object semantics, while our method exclusively edits the background.}
  \label{fig:lance-2}
\end{figure}

In \cite{prabhu2023lance}, LANCE method is proposed, which is closely relevant to our approach. LANCE leverages the capabilities of language models to create textual prompts, facilitating diverse image alterations using the prompt-to-prompt image editing method \cite{hertz2022prompt} and null-text inversion \cite{mokady2023null} for real image editing. However, this reliance on prompt-to-prompt editing imposes constraints, particularly limiting its ability to modify only specific words in the prompt. Such a limitation restricts the range of possible image transformations. Additionally, the global nature of their editing process poses challenges in preserving object semantics during these transformations. In contrast, our method employs both visual and textual conditioning, effectively preserving object semantics while generating varied background changes. This approach aligns well with our goal of studying object-to-background context. 
We use open-sourced code from LANCE to compare it against our approach both quantitatively and qualitatively. We use a subset of 1000 images, named $\texttt{ImageNet-B}_{1000}$, for comparison. We observe that our natural object-to-background changes including color and texture perform favorably against LANCE, while our adversarial object-to-background changes perform significantly better as shown in the Table \ref{tab:base_class_comparison}. Since LANCE relies on global-level image editing, it tends to alter the object semantics and distort the original object shape in contrast to our approach which naturally preserves the original object and alters the object-to-background composition only. This can be observed in qualitative examples provided in Figures \ref{fig:lance} and \ref{fig:lance-2}.  We further validate this effect by masking the background of original and LANCE-generated counterfactual images. As reported in Table \ref{tab:lance-masked}, when the background is masked in LANCE-generated counterfactual images, overall accuracy drops from $84.35\%$ to $71.57\%$. This drop in accuracy compared to original images with masked background, shows that the LANCE framework has distorted the original object semantics  during optimization. In contrast to this, our proposed approach allows us to study the correlation of object-to-background compositional changes without distorting the object semantics.

We calculate the FID score by comparing the background changes applied on our \texttt{ImageNet-B} dataset with the original ImageNet val. set (Tab. \ref{tab:fid}). Our background modifications such as \emph{Class Label}, \emph{BLIP Caption}, \& \emph{Color} achieve FID scores close to the original images, while our more complex background changes (\emph{Texture}, \emph{Adversarial}) show significant improvement over related works\cite{prabhu2023lance, zhang2024imagenet}.

\begin{table*}[h]
\fontsize{6pt}{6pt}\selectfont
\centering
\caption{Performance evaluation and comparison on $\texttt{ImageNet-B}_{1000}$ dataset. The drop in accuracy of LANCE dataset when the background is masked clearly highlight the image manipulation being done on the object of interest.}
\resizebox{1\linewidth}{!}{%
\begin{tabular}{lcccccccc}
\toprule

\multirow{2}{*}{Dataset} & 
\multicolumn{7}{c}{\textbf{Masked Background}}
\\  \cmidrule(lr){2-8}

& ViT-T & ViT-S & Swin-T & Swin-S & Res-50 &Res-152& Dense-161 & \cellcolor{gray!20} Average\\
 \midrule

Original & 70.5 &  86.1& 84.2 & 87.6 & 87.2 & 91.2 & 83.7&\cellcolor{gray!20} 84.35 \\
\cmidrule(lr){1-8}
LANCE & 59.5 & 72.5  & 72.3 & 75.3 & 71.9 & 77.5 & 72.0 & \cellcolor{gray!20} 71.57\\
\bottomrule

\end{tabular}%
}

\label{tab:lance-masked}
\end{table*}

\begin{table}[h]
\fontsize{12pt}{12pt}\selectfont
\centering
    \caption{FID comparison \emph{(lower is better)}.}
 \setlength{\tabcolsep}{12pt}
    \scalebox{0.55}[0.65]{
    \begin{tabular}{cccccccccc}
    \toprule
     LANCE\cite{prabhu2023lance} & BG\cite{zhang2024imagenet}  & Material\cite{zhang2024imagenet} & Texture\cite{zhang2024imagenet} & Ours: & Class & BLIP & Color  & Texture & Adv \\
    \midrule
88.51 & 68.99 & 120.18 & 132.28 &  & 35.05  & 30.98 & 31.65 & 45.11 & 67.57\\
    \bottomrule
    \end{tabular}}
         
    \label{tab:fid}
\end{table}

\newpage

\subsection{Qualitative Comparison with Related Works}
\label{sec:qualitative comparison}

\begin{figure}[h]
\begin{minipage}{\textwidth}

\centering

\begin{minipage}{0.16\textwidth}
  \centering
  \includegraphics[ trim= 5mm 15mm 5mm 5mm, clip, width=\linewidth , keepaspectratio]{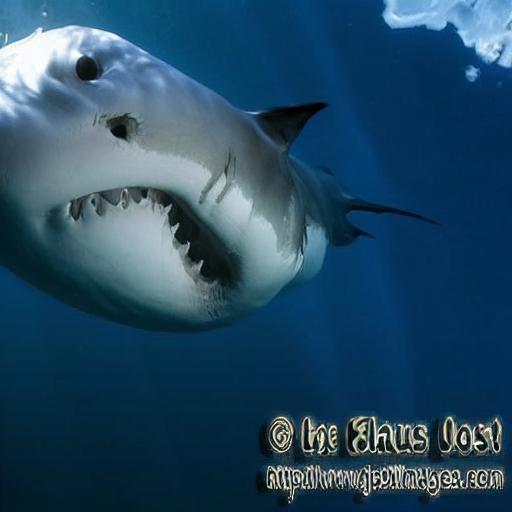}
   \footnotesize LANCE
\end{minipage}
\begin{minipage}{0.16\textwidth}
  \centering
  \includegraphics[trim= 5mm 15mm 5mm 5mm, clip, width=\linewidth , keepaspectratio ]{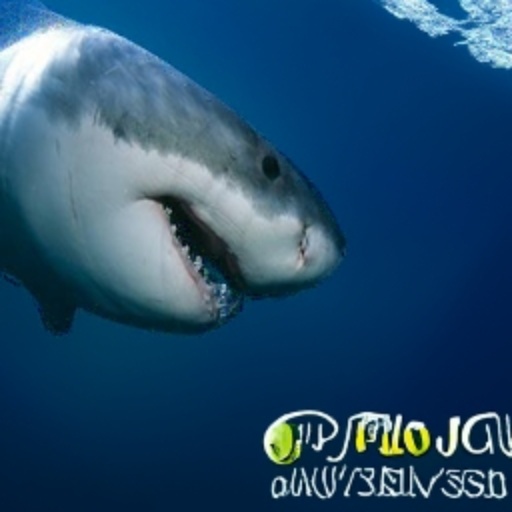}
    \footnotesize $\lambda=-20$ 

\end{minipage}
\begin{minipage}{0.16\textwidth}
  \centering
  \includegraphics[trim= 5mm 15mm 5mm 5mm, clip, width=\linewidth , keepaspectratio]{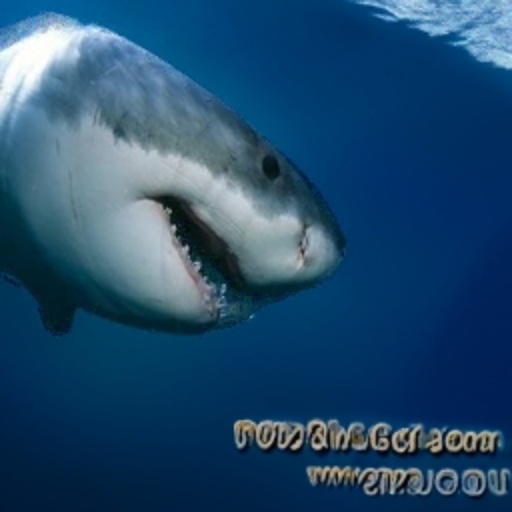}
     \footnotesize $\lambda=20$ 

\end{minipage}
\begin{minipage}{0.16\textwidth}
  \centering
  \includegraphics[trim= 5mm 15mm 5mm 5mm, clip, width=\linewidth , keepaspectratio]{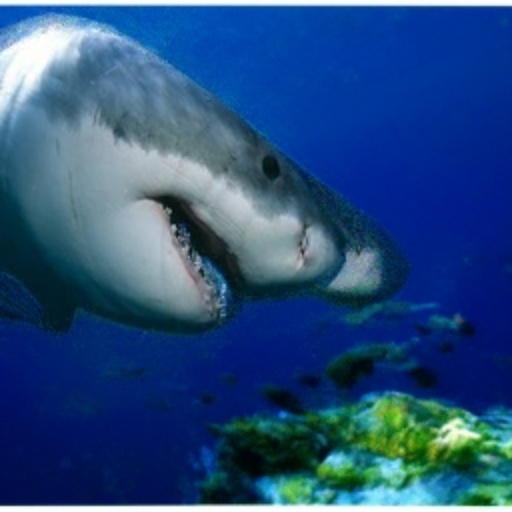}
 \footnotesize $\lambda_{adv}=20$
\end{minipage}
\begin{minipage}{0.16\textwidth}
  \centering
  \includegraphics[trim= 5mm 15mm 5mm 5mm, clip, width=\linewidth , keepaspectratio]{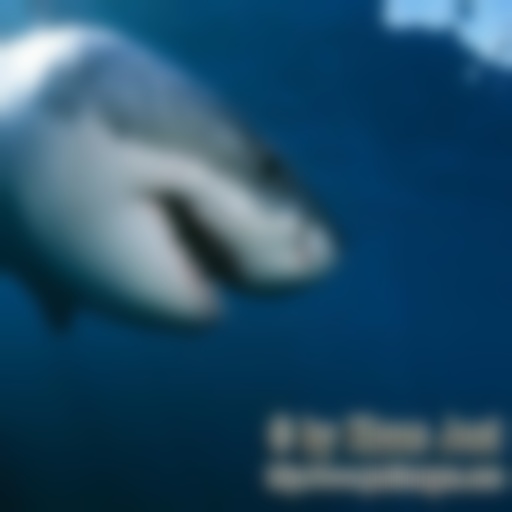}
     \footnotesize Blur

\end{minipage}
\begin{minipage}{0.16\textwidth}
  \centering
  \includegraphics[trim= 5mm 15mm 5mm 5mm, clip, width=\linewidth , keepaspectratio]{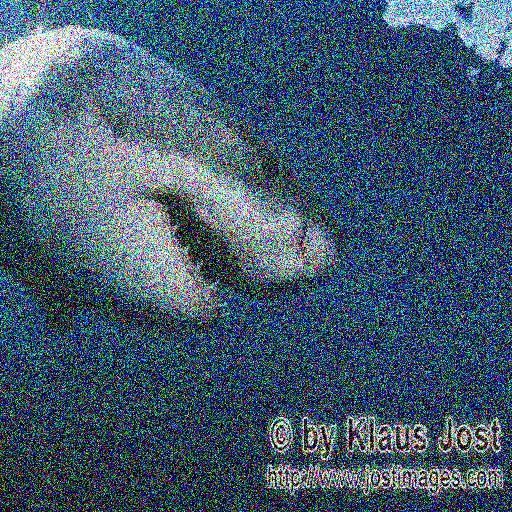}
     \footnotesize Noise
\end{minipage}

\begin{minipage}{0.16\textwidth}
  \centering
  \includegraphics[ trim= 5mm 15mm 5mm 5mm, clip, width=\linewidth , keepaspectratio]{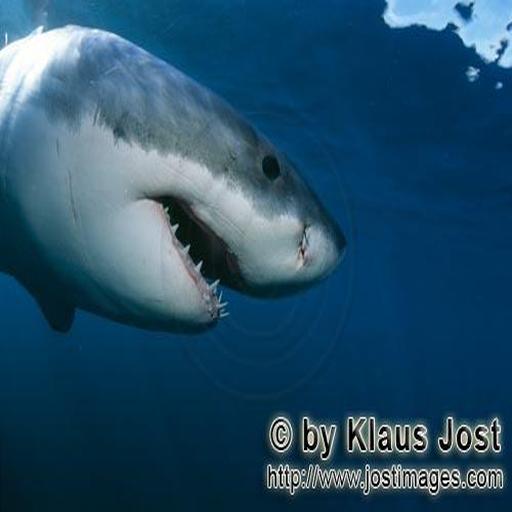}
   \footnotesize Original
\end{minipage}
\begin{minipage}{0.16\textwidth}
  \centering
  \includegraphics[trim= 5mm 15mm 5mm 5mm, clip, width=\linewidth , keepaspectratio ]{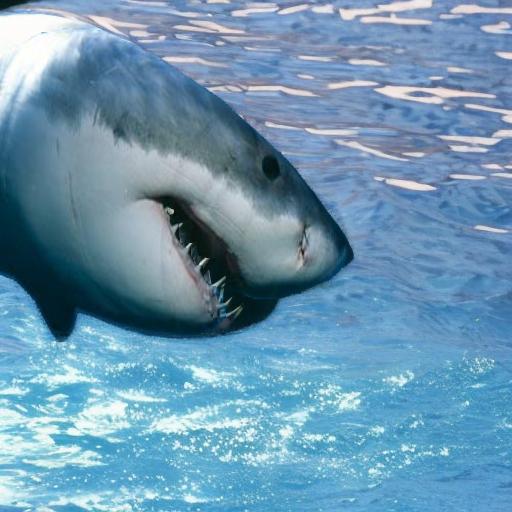}
    \footnotesize Class Label

\end{minipage}
\begin{minipage}{0.16\textwidth}
  \centering
  \includegraphics[trim= 5mm 15mm 5mm 5mm, clip, width=\linewidth , keepaspectratio]{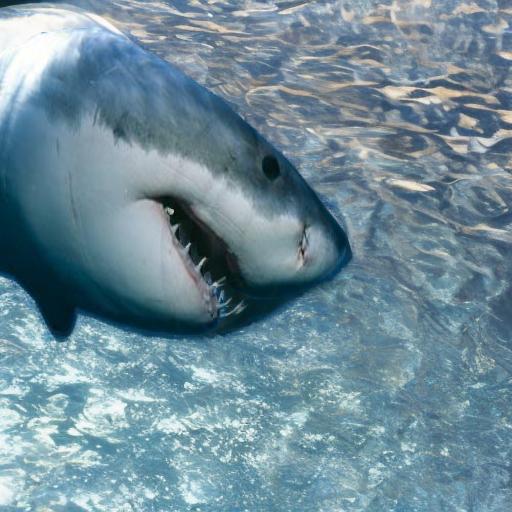}
     \footnotesize BLIP-2

\end{minipage}
\begin{minipage}{0.16\textwidth}
  \centering
  \includegraphics[trim= 5mm 15mm 5mm 5mm, clip, width=\linewidth , keepaspectratio]{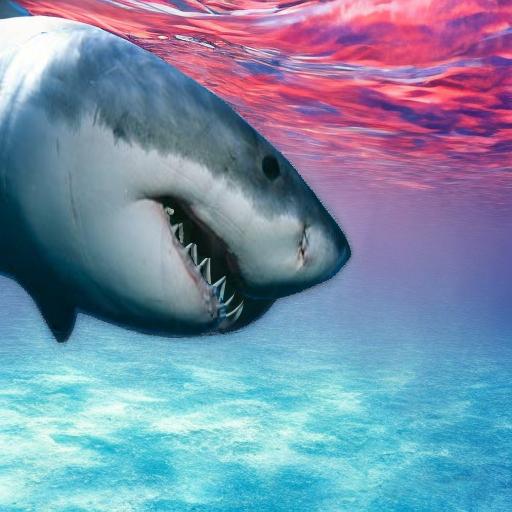}
     \footnotesize Color
\end{minipage}
\begin{minipage}{0.16\textwidth}
  \centering
  \includegraphics[trim= 5mm 15mm 5mm 5mm, clip, width=\linewidth , keepaspectratio]{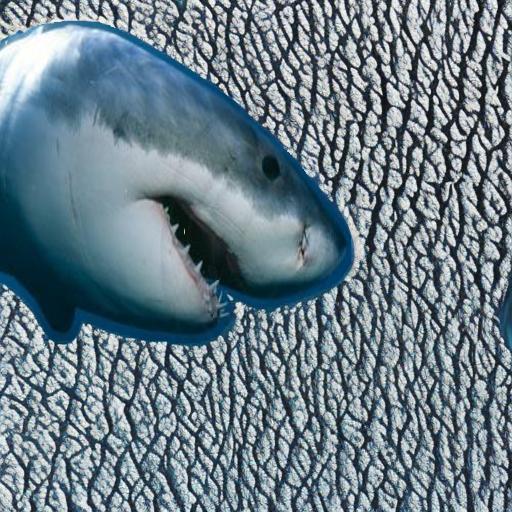}
     \footnotesize Texture

\end{minipage}
\begin{minipage}{0.16\textwidth}
  \centering
  \includegraphics[trim= 5mm 15mm 5mm 5mm, clip, width=\linewidth , keepaspectratio]{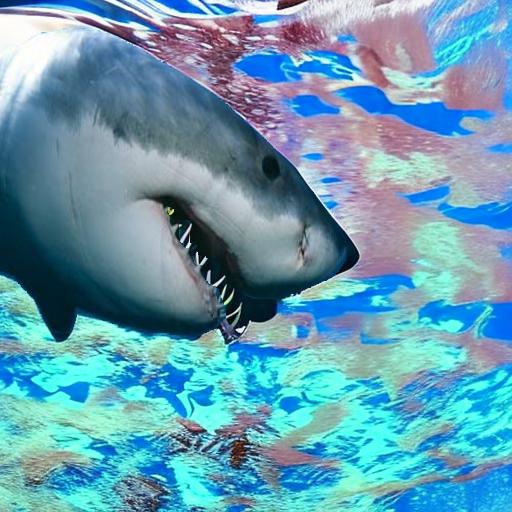}
     \footnotesize Adversarial
\end{minipage}

\end{minipage}
\hfill
  \caption{Qualitative comparison of our background changes \textit{(bottom row)} with previous related work \textit{(top row)}. Our method enables diversity and controlled background edits.}
  \label{fig:comparison-001}
\end{figure}

\FloatBarrier

\subsection{Ablation on Background Changes}
\label{sec:prompt_evaluation}
In this section, we report results on \texttt{ImageNet-B} and \texttt{COCO-DC} for uni-modal classifiers in Table \ref{tab:appendix_classification} and Table \ref{tab:appendix_clip_classification} reports zero-shot classification results on \texttt{ImageNet-B}. Furthermore, ablations across diverse color and texture prompts is provided in Table \ref{tab:appendix_zs_classification} and \ref{tab:appendix_coco-classification-ablation}.

\begin{table*}
\fontsize{7pt}{8pt}\selectfont
\centering
\caption{\small Resilience of Transformer and CNN models trained on ImageNet and COCO training sets against our proposed object-to-background context variations. We report top-1 (\%) accuracy. We observe that CNN-based models are relatively more robust than Transformers.}
\resizebox{1\linewidth}{!}{%
\begin{tabular}{llcccccccl}
\toprule
\multirow{3}{*}{Datasets} & \multirow{3}{*}{Background} & 
\multicolumn{4}{c}{Transformers} & 
\multicolumn{3}{c}{CNN}
\\  \cmidrule(lr){3-6}
\cmidrule(lr){7-9}
 & & ViT-T & ViT-S & Swin-T & Swin-S & Res-50 & Res-152 & Dense-161 & \cellcolor{gray!20} Average\\
\midrule
\multirow{6}{*}{\texttt{ImageNet-B}}& Original  & 96.04  & 98.18 & 98.65 & 98.84  &98.65 &99.27 & 98.09  & \cellcolor{gray!20} 98.25 \\
%
& Class label & 92.82 & 94.75 &96.18 & 96.55& 97.24& 97.56& 95.8&\cellcolor{gray!20}  95.84\dec{2.41} \\
& BLIP-2 Caption & 86.77& 90.41 & 92.71 & 93.60 & 94.46   & 95.35 & 91.62 & \cellcolor{gray!20} 92.13\dec{6.12}  \\
%
& Color & 70.64 & 84.52 & 86.84 & 88.84 & 89.44 & 92.89 & 83.19 & \cellcolor{gray!20} 85.19\dec{13.06} \\
%
%
& Texture & 68.24 & 79.73 & 81.09 & 84.41  & 83.21   & 87.66 &77.29 \ & \cellcolor{gray!20} 80.23\dec{18.02} \\
\midrule
\multirow{2}{*}{$\texttt{ImageNet-B}_{1000}$}& Original  &  95.01 & 97.50  & 97.90 & 98.30& 98.50 & 99.10 & 97.20 & \cellcolor{gray!20} 97.64  \\

& Adversarial & 18.40 & 32.10  & 25.00& 31.70& 2.00 & 28.00 & 14.40 & \cellcolor{gray!20} 21.65\dec{75.99}  \\

\midrule

\multirow{6}{*}{\texttt{COCO-DC}}& Original  & 82.96 & 86.24  & 88.55& 90.23& 88.55  & 89.08 & 86.77  & \cellcolor{gray!20} 87.21 \\
& BLIP-2 Caption & 82.69 & 84.73  & 86.24 & 86.95  & 88.46 & 86.69 & 85.01 & \cellcolor{gray!20} 
85.67\dec{1.54}\\
%
& Color & 55.54 & 61.04 & 70.09 & 72.13 & 74.97  & 75.10 & 66.19  & \cellcolor{gray!20} 66.66\dec{20.55} \\
%
%
& Texture & 52.52 & 58.82 & 68.05 & 70.09 & 70.71 & 74.77 & 63.79\  & \cellcolor{gray!20} 63.99\dec{23.22} \\
%
& Adversarial & 49.68 & 55.72 & 61.93 & 69.12 & 55.45 & 61.13 & 57.76& \cellcolor{gray!20} 58.68\dec{28.52} \\
\bottomrule
\end{tabular}%
}
\label{tab:appendix_classification}
\end{table*}

\begin{table}[H]
\fontsize{8pt}{8pt}\selectfont
\centering
\caption{Comparative Evaluation of Zero-shot CLIP and Eva CLIP Vision-Language Models on \texttt{ImageNet-B} and $\texttt{ImageNet-B}_{1000}$. Top-1(\%) accuracy is reported. We find that Eva CLIP models showed more robustness in all object-to-background variations.}
\resizebox{1\linewidth}{!}{%
\begin{tabular}{llcccccccl}
\toprule
\multirow{3}{*}{Datasets} & \multirow{3}{*}{Background} & 
\multicolumn{7}{c}{CLIP} & 
\\  \cmidrule(lr){3-9}
& & ViT-B/32 & ViT-B/16 & ViT-L/14 & Res50 & Res101 & Res50x4 & Res50x16 & \cellcolor{gray!20} Average \\
 \midrule
\multirow{6}{*}{\texttt{ImageNet-B}}& Original  &  75.56 & 81.56  & 88.61& 73.06& 73.95 & 77.87 & 83.25 & \cellcolor{gray!20} 79.12 \\
%
&Class label  & 80.83 & 84.41  & 89.41& 78.87& 79.33 & 81.94 & 85.67 & \cellcolor{gray!20} 82.92\inc{3.80} \\
&BLIP-2 Captions & 69.33 & 73.66  & 79.07 & 67.44 & 68.70 & 71.55 & 75.78 & \cellcolor{gray!20} 72.22\dec{6.90} \\
%
&Color & 53.02 & 63.08 & 71.42 & 53.53 & 55.87 & 60.05 & 71.28 & \cellcolor{gray!20} 61.18\dec{17.94} \\
%
%
&Texture & 51.01 & 62.25  & 69.08 & 51.35 & 53.46 & 61.10 & 70.33 & \cellcolor{gray!20} 59.79\dec{19.33}\\
\midrule
\multirow{2}{*}{$\texttt{ImageNet-B}_{1000}$}& Original  &  73.90 & 79.40  & 87.79& 70.69& 71.80 & 76.29 & 82.19 & \cellcolor{gray!20} 77.43 \\
&Adversarial  & 25.5 & 34.89  & 48.19& 18.29& 24.40 & 30.29 & 48.49 & \cellcolor{gray!20} 32.87\dec{46.25} \\

\midrule

\multirow{3}{*}{Datasets} & \multirow{3}{*}{Background} & 

\multicolumn{7}{c}{EVA-CLIP} 
\\  \cmidrule(lr){3-9}

& & g/14 & g/14+ & B/16 & L/14 & L/14+ & E/14 & E/14+ & \cellcolor{gray!20} Average \\
 \midrule
\multirow{6}{*}{\texttt{ImageNet-B}}& Original  & 90.80&  93.71 & 90.24  & 93.71 & 93.69 & 95.38 & 95.84 & \cellcolor{gray!20} 93.34  \\
%
%
&Class label & 90.48 & 93.53& 90.20 &93.47& 93.49&94.78& 95.18& \cellcolor{gray!20} 93.02\dec{0.32} \\
&BLIP-2 Caption & 80.56 & 85.23  & 81.88 & 85.28 & 86.24 & 88.13 & 88.68 & \cellcolor{gray!20} 85.14\dec{8.20} \\

&Color & 77.25 & 83.96 & 76.24 & 83.63 & 85.79 & 88.70 & 88.33 & \cellcolor{gray!20} 83.41\dec{9.93} \\
& Texture & 75.93 & 82.76  & 74.44  & 82.56 & 86.35 & 87.84 & 88.44 & \cellcolor{gray!20}  82.62\dec{10.72}\\
\midrule
\multirow{2}{*}{$\texttt{ImageNet-B}_{1000}$}& Original   & 88.80  & 92.69  & 89.19& 91.10& 91.99 & 93.80 & 94.60 & \cellcolor{gray!20} 91.74  \\
& Adversarial   & 55.59 & 62.49  & 48.70 & 65.39 & 73.59 & 70.29 & 73.29 & \cellcolor{gray!20} 64.19\dec{27.55} \\
\bottomrule
\end{tabular}%
}
\label{tab:appendix_clip_classification}
\end{table}

\begin{table*}
\fontsize{10pt}{10pt}\selectfont
\centering
\caption{Performance evaluation of naturally trained classifiers and zero-shot CLIP models on \texttt{ImageNet-B}. The text prompts used for color and texture changes are provided in Table \ref{prompt}.}
\resizebox{1\linewidth}{!}{%
\begin{tabular}{lcccccccc}
\toprule
\multirow{2}{*}{\textbf{Background}} & 
\multicolumn{8}{c}{\textbf{Naturally Trained Models}}
\\  \cmidrule(lr){2-9}
\multirow{2}{*}{} & 
\multicolumn{4}{c}{\textbf{ViT}} & 
\multicolumn{3}{c}{\textbf{CNN}}
\\  \cmidrule(lr){2-5}
\cmidrule(lr){6-8}
& ViT-T & ViT-S & Swin-T & Swin-S & ResNet50 & ResNet152 & DenseNet & \cellcolor{gray!20} Average\\
 \midrule
Clean  & 96.04  & 98.18 & 98.65 & 98.84  &98.65 &99.27 & 98.09  & \cellcolor{gray!20} 98.25 \\
\cmidrule(lr){1-9}
$\text{Color}_{\text{prompt-1}}$ & 76.58& 86.43& 88.92 & 91.23 & 91.08  & 93.79 & 86.05  & \cellcolor{gray!20} 87.72 \\
$\text{Color}_{\text{prompt-2}}$ & 77.09& 87.57  & 89.33 & 90.99 & 90.62  & 93.40 & 86.61 & \cellcolor{gray!20} 87.94 \\
$\text{Color}_{\text{prompt-3}}$ & 76.80 & 86.97 & 88.74 & 90.99& 90.62  & 93.18 & 87.41  & \cellcolor{gray!20} 87.82 \\
$\text{Color}_{\text{prompt-4}}$ & 70.64 & 84.52 & 86.84 & 88.84 & 89.44 & 92.89 & 83.19 & \cellcolor{gray!20} 85.19 \\
\cmidrule(lr){1-9}
$\text{Texture}_{\text{prompt-1}}$ & 79.07 & 87.92 & 90.17 & 91.68 &  91.18  & 94.42 & 88.28 \ & \cellcolor{gray!20} 88.96\\
$\text{Texture}_{\text{prompt-2}}$ & 75.29& 85.84 & 87.74 & 90.32 & 89.01    & 93.04 & 84.77 \ & \cellcolor{gray!20} 86.57\\
$\text{Texture}_{\text{prompt-3}}$ & 67.97 & 82.54 & 86.17 & 87.99 & 87.99    & 91.28 & 82.99 \ & \cellcolor{gray!20} 83.85\\
$\text{Texture}_{\text{prompt-4}}$ & 68.24 & 79.73 & 81.09 & 84.41  & 83.21   & 87.66 &77.29 \ & \cellcolor{gray!20} 80.23\\

\midrule
\midrule

\multirow{2}{*}{\textbf{Background}} & 
\multicolumn{8}{c}{\textbf{CLIP Models}}
\\  \cmidrule(lr){2-9}
\multirow{2}{*}{} & 
\multicolumn{3}{c}{\textbf{ViT}} & 
\multicolumn{4}{c}{\textbf{CNN}}
\\  \cmidrule(lr){2-4}
\cmidrule(lr){5-8}
& ViT-B/32 & ViT-B/16 & ViT-L/14 & ResNet50 & ResNet101 & ResNet50x4 & ResNet50x16 & \cellcolor{gray!20} Average\\
 \midrule
Clean  &  75.56 & 81.56  & 88.61& 73.06& 73.95 & 77.87 & 83.25 & \cellcolor{gray!20} 79.12 \\
\cmidrule(lr){1-9}
$\text{Color}_{\text{prompt-1}}$ & 58.32 & 65.54 & 72.75 & 57.43 & 60.92 & 65.97 & 73.04 & \cellcolor{gray!20} 64.49 \\
$\text{Color}_{\text{prompt-2}}$ & 57.91 & 67.28 & 74.44 & 58.67 & 60.12 & 65.9 & 74.13 & \cellcolor{gray!20} 65.49 \\
$\text{Color}_{\text{prompt-3}}$ & 57.27 & 66.77 & 74.07 & 57.89 & 59.03 & 66.10 & 74.06 & \cellcolor{gray!20} 65.03 \\
$\text{Color}_{\text{prompt-4}}$ & 53.02 & 63.08 & 71.42 & 53.53 & 55.87 & 60.05 & 71.28 & \cellcolor{gray!20} 61.18 \\
\cmidrule(lr){1-9}
$\text{Texture}_{\text{prompt-1}}$ & 59.05 &  68.50  & 75.67 & 60.38 & 61.78 & 66.99 & 74.31\ & \cellcolor{gray!20} 66.67\\
$\text{Texture}_{\text{prompt-2}}$ & 58.60 & 68.01 & 74.40 & 58.29 & 59.56 & 66.34 & 74.67\ & \cellcolor{gray!20} 65.69\\
$\text{Texture}_{\text{prompt-3}}$ & 52.89 & 64.30  & 68.70 & 53.29 & 55.35 & 61.58 & 69.35 & \cellcolor{gray!20} 60.78\\
$\text{Texture}_{\text{prompt-4}}$ & 51.01 & 62.25  & 69.08 & 51.35 & 53.46 & 61.10 & 70.33 & \cellcolor{gray!20} 59.79\\
\bottomrule
\end{tabular}%
}
\label{tab:appendix_zs_classification}
\end{table*}

\begin{table*}
\fontsize{7.5pt}{7.5pt}\selectfont
\centering
\caption{Performance evaluation of naturally trained classifiers on \texttt{COCO-DC} dataset. The text prompts used for color and texture changes are provided in Table \ref{prompt}.}
\begin{tabular}{lccccccc}
\toprule

\multirow{2}{*}{Background} & 
\multicolumn{4}{c}{\textbf{ViT}} & 
\multicolumn{2}{c}{\textbf{CNN}}
\\  \cmidrule(lr){2-5}
\cmidrule(lr){6-7}
& ViT-T & ViT-S & Swin-T & Swin-S & ResNet50 & Dense-161 & \cellcolor{gray!20} Average\\
 \midrule
Clean & 82.96 & 86.24  & 88.55& 90.23& 88.55  & 86.77 & \cellcolor{gray!20} 87.21 \\

 \midrule
$\text{Color}_{\text{prompt-1}}$ & 61.66 & 65.92 & 73.38 & 73.73 & 75.86  & 71.6 & \cellcolor{gray!20} 70.35 \\

$\text{Color}_{\text{prompt-2}}$ & 64.86 & 70.01 & 76.84 & 77.10 & 77.81 & 75.06 & \cellcolor{gray!20} 73.61 \\

$\text{Color}_{\text{prompt-3}}$ & 62.64 & 67.52 & 73.29 & 74.09 & 77.28  & 73.64 & \cellcolor{gray!20} 71.41 \\

$\text{Color}_{\text{prompt-4}}$ & 55.54 & 61.04 & 70.09 & 72.13 & 74.97  & 66.19 & \cellcolor{gray!20} 66.66 \\

\cmidrule(lr){1-8}
$\text{Texture}_{\text{prompt-1}}$ & 67.96 & 70.36   & 75.42 & 78.70 & 79.94 & 73.55\ & \cellcolor{gray!20} 74.32\\

$\text{Texture}_{\text{prompt-2}}$ & 63.97 & 69.56 & 74.62 & 77.72 & 78.97 & 75.15\ & \cellcolor{gray!20} 73.33\\

$\text{Texture}_{\text{prompt-3}}$ & 52.52 & 58.82 & 68.05 & 70.09 & 70.71 & 63.79\ & \cellcolor{gray!20} 63.99\\

$\text{Texture}_{\text{prompt-4}}$ & 56.16 & 61.57 & 66.72 & 70.18 & 69.56 & 67.25\ & \cellcolor{gray!20} 65.24\\

\bottomrule
\end{tabular}
\label{tab:appendix_coco-classification-ablation}
\end{table*}

\newpage

\subsection{Evaluation on Adversarially Trained models}
\label{sec:comparison_adv_models}
In this section, we evaluate adversarially trained Res-18, Res-50, and Wide-Res-50 models across background changes induced by our methods and baseline methods (See Table \ref{tab:adv-model_comparison-2}, \ref{tab:adv-model_comparison-1}, and \ref{tab:adv-model_comparison-3}).

\begin{table*}[h]
\fontsize{5pt}{5pt}\selectfont
\centering
\caption{
Performance evaluation and comparison of our dataset with state of the art methods on adversarially trained Res-18 models. The images are generated on $\texttt{ImageNet-B}_{1000}$ dataset. We report top-1 average accuracy of models trained on various adversarial budget.
\small }
\resizebox{1\linewidth}{!}{%
\begin{tabular}{lccccccccc}
\toprule

\multirow{2}{*}{Datasets} & 
\multicolumn{4}{c}{$\ell_\infty$} & 
\multicolumn{4}{c}{$\ell_2$}& 
\\  \cmidrule(lr){2-5}
\cmidrule(lr){6-9}

& $\varepsilon$=0.5 & $\varepsilon$=2.0 &  $\varepsilon$=4.0 & $\varepsilon$=8.0 & $\varepsilon$=0.5 & $\varepsilon$=2.0 & $\varepsilon$=4.0 & $\varepsilon$=8.0 \\
 \midrule
Original & 88.00 & 78.30  & 69.70& 54.40& 84.60  & 81.40 & 68.80& 57.50& \\
\midrule
 ImageNet-E ($\lambda$=-20) & 84.50 & 77.01   & 69.10 & 54.80 & 83.40 & 80.00 & 68.80 & 59.00\\
 ImageNet-E ($\lambda$=20) & 81.41 & 74.94   & 66.36 & 52.52 & 81.61 & 75.75 & 65.65 & 55.05\\
 ImageNet-E ($\lambda_{adv}=20$) & 75.15 &  66.16  & 56.36 & 45.35 & 72.82 & 66.66& 55.75& 45.05\\
 LANCE  & 76.37 &  66.78  & 59.13 & 45.17 & 74.99 & 73.19 & 61.60& 48.68\\

 \midrule
Class label & 87.10 & 79.90 & 69.40 & 57.30 & 85.00  & 79.90 & 70.90& 57.80\\
BLIP-2 Caption & 80.90 & 73.10 & 67.10 & 51.00 &79.50& 75.30 &63.10&53.40 \\
\rowcolor{gray!20} Color & 56.90 & 45.80 & 35.40 & 25.70 & 53.20  & 46.80 & 32.80&23.40 \\
\rowcolor{gray!20} Texture & 59.20 & 47.10& 38.60 & 28.70 & 54.30  & 48.10 & 35.50&26.20 \\

\rowcolor{gray!20} Adversarial & 12.10 & 19.80& 24.60 & 26.80 & 10.90  & 12.40 & 16.90&17.20 \\

\bottomrule
\end{tabular}%
}
\label{tab:adv-model_comparison-2}
\end{table*}

\begin{table*}[h]
\fontsize{5pt}{5pt}\selectfont
\centering
\caption{
Performance evaluation and comparison of our dataset with state of the art methods on adversarially trained Res-50 models. The images are generated on $\texttt{ImageNet-B}_{1000}$ dataset. We report top-1 average accuracy of models trained on various adversarial budget.
\small }
\resizebox{1\linewidth}{!}{%
\begin{tabular}{lccccccccc}
\toprule

\multirow{2}{*}{Datasets} & 
\multicolumn{4}{c}{$\ell_\infty$} & 
\multicolumn{4}{c}{$\ell_2$}& 
\\  \cmidrule(lr){2-5}
\cmidrule(lr){6-9}

& $\varepsilon$=0.5 & $\varepsilon$=2.0 &  $\varepsilon$=4.0 & $\varepsilon$=8.0 & $\varepsilon$=0.5 & $\varepsilon$=2.0 & $\varepsilon$=4.0 & $\varepsilon$=8.0 \\
 \midrule
Original & 95.20 & 89.30  & 83.20& 72.40& 94.30  & 91.10 & 80.90& 70.80&  \\
\midrule
 ImageNet-E ($\lambda$=-20) & 93.10 & 89.20   & 82.00 & 70.70 & 91.70 & 88.50 & 79.60& 69.10\\
 ImageNet-E ($\lambda$=20) & 92.52 & 86.36   & 80.60 & 67.97 & 90.40 & 86.96 & 76.26 & 68.08\\
 ImageNet-E ($\lambda_{adv}=20$) & 84.44 &  78.78  & 71.71 & 58.58 & 80.50 & 76.76& 65.75& 56.86\\
 LANCE  & 81.94 & 78.96   & 70.11 & 59.72 & 83.52 & 80.46 & 69.83& 61.32\\

 \midrule
Class label & 92.40 & 88.50 & 82.70 & 72.50 & 90.70  & 88.60 & 80.20& 70.50\\
BLIP-2 Caption & 87.90 & 83.70 & 79.00 & 67.90 & 86.60 & 84.60 & 73.70& 65.70\\
\rowcolor{gray!20} Color & 70.80 & 60.30 & 53.20 & 39.50 & 67.20  & 58.50 &44.40 &34.20\\
\rowcolor{gray!20} Texture & 69.70 & 61.00 & 54.60 & 43.40 & 64.90  & 59.70 &48.00 & 37.70 \\

\rowcolor{gray!20} Adversarial & 10.80 & 17.10 & 18.10 & 16.60 & 10.70  & 11.90 &14.70& 13.40 \\

\bottomrule
\end{tabular}%
}
\label{tab:adv-model_comparison-1}
\end{table*}

\begin{table*}[t]
\fontsize{5pt}{5pt}\selectfont
\centering
\caption{
Performance evaluation and comparison of our dataset with state of the art methods on adversarially trained Wide Res-50 models. The images are generated on $\texttt{ImageNet-B}_{1000}$ dataset. We report top-1 average accuracy of models trained on various adversarial budget.
\small }
\resizebox{1\linewidth}{!}{%
\begin{tabular}{lccccccccc}
\toprule

\multirow{2}{*}{Datasets} & 
\multicolumn{4}{c}{$\ell_\infty$} & 
\multicolumn{4}{c}{$\ell_2$}& 
\\  \cmidrule(lr){2-5}
\cmidrule(lr){6-9}

& $\varepsilon$=0.5 & $\varepsilon$=2.0 &  $\varepsilon$=4.0 & $\varepsilon$=8.0 & $\varepsilon$=0.5 & $\varepsilon$=2.0 & $\varepsilon$=4.0 & $\varepsilon$=8.0 \\
 \midrule
Original & 96.20 & 92.60  & 89.10& 78.70& 95.60  & 94.00 & 87.00& 78.30&  \\
\midrule
 ImageNet-E ($\lambda$=-20) & 94.10 & 91.10   & 86.60 & 76.40 & 93.20 & 91.60 & 84.20 & 76.90 \\
 ImageNet-E ($\lambda$=20) & 92.82 & 89.29   & 83.53 & 75.15 & 91.21 & 88.68 & 81.11 & 73.63 \\
 ImageNet-E ($\lambda_{adv}=20$) & 87.17 &  76.56 & 66.16 & 82.92 & 82.22 & 82.00& 71.41& 62.42\\
 LANCE  & 84.18 & 81.34   & 77.07 & 64.51 & 83.48 & 83.09 & 77.40& 66.09\\

 \midrule
Class label & 93.50 & 90.60 & 87.30 & 78.80 & 92.10  & 90.80 &83.70 & 75.50\\
BLIP-2 Caption & 90.20 & 86.20 & 82.80 & 74.30 & 88.90 & 86.70 &80.00 &69.20 \\
\rowcolor{gray!20} Color &72.2 &66.60 & 60.70 & 51.20 & 68.10 & 65.10  & 51.40 &40.90  \\
\rowcolor{gray!20} Texture & 73.80& 66.70 & 61.70 & 53.60 & 67.20  & 64.60 & 52.00&44.20 \\
\rowcolor{gray!20} Adversarial & 13.90& 22.90 & 28.00 & 32.00 & 12.40  & 15.10 & 19.50&20.60 \\

\bottomrule
\end{tabular}%
}
\label{tab:adv-model_comparison-3}
\end{table*}

\FloatBarrier

\newpage
\subsection{Evaluation on Recent Vision Models}
\label{sec:recent_models}

We have conducted experiments on recent transformer CNN based models like DeiT \cite{touvron2021training} and ConvNeXt \cite{liu2022convnet}, and their results are presented in Table \ref{tab:sota-models}. We observe a consistent trend in model performance on our dataset, revealing that even the modern vision models are vulnerable to background changes.

\begin{table*}[h]
\fontsize{11pt}{11pt}\selectfont
\centering
\caption{Performance evaluation on naturally trained classifiers on \texttt{ImageNet-B} and $\texttt{ImageNet-B}_{1000}$ dataset. All models exhibit a marked decrease in accuracy when the background is modified, highlighting their sensitivity to changes in the environment. The decline in performance is minimal with class label backgrounds but more pronounced with texture and color alterations. The most significant accuracy drop occurs under adversarial conditions, underscoring the substantial challenge posed by such backgrounds to the classifiers.}
\resizebox{1\linewidth}{!}{%
\begin{tabular}{lllcccccccc}
\toprule

\multirow{2}{*}{Datasets}&
\multirow{2}{*}{Background} & 
\multicolumn{3}{c}{\textbf{Transformers}} & 
\multicolumn{5}{c}{\textbf{CNN}}
\\  \cmidrule(lr){3-5}
\cmidrule(lr){7-9}
& & DeiT-T & DeiT-S & DeiT-B  &\cellcolor{gray!20} Average & ConvNeXt-T & ConvNeXt-B & ConvNeXt-L & \cellcolor{gray!20} Average \\
 \midrule
\multirow{5}{*}{\texttt{ImageNet-B}} &Original & 96.36 & 99.27 & 99.41 &\cellcolor{gray!20} 98.34 & 99.07 & 99.21 & 99.40 & \cellcolor{gray!20} 99.22 \\
&$\text{Class label}$ & 94.18 & 96.85&  97.74  &\cellcolor{gray!20} 96.25 & 97.60 & 97.51 & 97.51 & \cellcolor{gray!20} 97.54\\

&$\text{BLIP-2 Caption}$ & 89.33  & 94.29& 95.07 &\cellcolor{gray!20} 92.89 & 94.64 & 94.82 & 95.47 & \cellcolor{gray!20}  94.97\\
&$\text{Color}$ & 80.96 & 89.48&91.11  & \cellcolor{gray!20} 87.13& 92.11 & 93.58 & 93.58 & \cellcolor{gray!20} 93.09  \\
&$\text{Texture}$ & 74.15 & 84.01&86.75 &\cellcolor{gray!20} 81.63 & 88.50 & 89.50 & 91.13 & \cellcolor{gray!20} 89.71 \\

\midrule

\multirow{1}{*}{$\texttt{ImageNet-B}_{1000}$} &Original & 95.44 & 99.10 & 99.10 &\cellcolor{gray!20}97.88 & 99.00 & 99.00  & 92.92 & \cellcolor{gray!20} 96.97 \\
&$\text{Adversarial}$ & 20.40 & 29.62&  34.81  &\cellcolor{gray!20} 28.27 & 32.88 & 42.52 & 48.60 & \cellcolor{gray!20} 41.33 \\

\bottomrule
\end{tabular}%
}
\label{tab:sota-models}
\end{table*}

\FloatBarrier
\newpage
\subsection{Evaluation on DINOv2 models}
\label{sec: DINOv2}

Our findings underscore the necessity of training vision models to prioritize discriminative and salient features, thereby diminishing their dependence on background cues. Recent advancements, such as the approaches by \cite{sitawarin2022part} employing a segmentation backbone for classification to improve adversarial robustness and by \cite{darcet2023vision} using additional learnable tokens known as \emph{registers} for interpretable attention maps, resonate with this perspective.  Our preliminary experiments with the DINOv2 models \cite{oquab2023dinov2}, as presented in Table \ref{tab:dinov2_reg}, corroborate this hypothesis. Across all the experiments, models with registers \emph{(learnable tokens)} provide more robustness to background changes, with significant improvement seen in the adversarial background changes.

\begin{table*}[h]
\fontsize{5pt}{5pt}\selectfont
\centering
\caption{Performance comparison of classifiers that are trained different on \texttt{ImageNet-B} dataset. The DINOv2 model with registers generally shows higher robustness to background changes, particularly in the presence of color, texture and adversarial backgrounds. This suggests that the incorporation of registers in DINOv2 enhances its ability to maintain performance despite challenging background alterations.}
\resizebox{1\linewidth}{!}{%
\begin{tabular}{lllcccccccc}
\toprule
\multirow{2}{*}{Dataset} &
\multirow{2}{*}{Background} &  
\multicolumn{3}{c}{\textbf{Dinov2}} &
\multicolumn{1}{c}{\textbf{}} &
\multicolumn{3}{c}{\textbf{Dinov2$_{\text{registers}}$}}
\\  \cmidrule(lr){3-5}
\cmidrule(lr){7-9}

 & & ViT-S & ViT-B & ViT-L & \cellcolor{gray!20}Average &  ViT-S & ViT-B & ViT-L & \cellcolor{gray!20} Average\\
 \midrule
\multirow{5}{*}{\texttt{ImageNet-B}} & Original &  96.78& 97.18 &  98.58 &\cellcolor{gray!20} 97.51 & 97.71 & 98.05 &99.14&\cellcolor{gray!20} 98.30 \\

&Class label  & 94.62& 96.02   & 97.18 &\cellcolor{gray!20}95.94 &95.55&96.44&97.94& \cellcolor{gray!20} 96.64 \\
&BLIP-2 Caption & 89.22& 91.73  & 94.33 &\cellcolor{gray!20} 91.76 &90.86&92.10&95.02& \cellcolor{gray!20} 92.66 \\
&Color  &  83.85 & 89.68  & 93.31 &\cellcolor{gray!20} 88.94 &85.88& 91.15&94.64&\cellcolor{gray!20} 90.55 \\
&Texture  & 83.63& 89.08   & 92.44 &\cellcolor{gray!20} 88.38&84.98&91.03 &93.97&\cellcolor{gray!20} 89.99 \\
\midrule
\multirow{1}{*}{$\texttt{ImageNet-B}_{1000}$} & Original  & 95.12& 96.50  & 98.10 &\cellcolor{gray!20} 96.57&97.91&97.80& 99.00& \cellcolor{gray!20} 98.23\\
&Adversarial  & 54.31& 71.62  & 80.87 &\cellcolor{gray!20} 68.93 &58.30&76.21&84.50& \cellcolor{gray!20} 73.00\\

\bottomrule
\end{tabular}%
}
\label{tab:dinov2_reg}
\end{table*}

\FloatBarrier

\newpage
\subsection{Vision Language model for Image Captioning}

\label{sec:image captioning}
\begin{figure}[H]
    \centering
    \includegraphics[width=\textwidth]{
    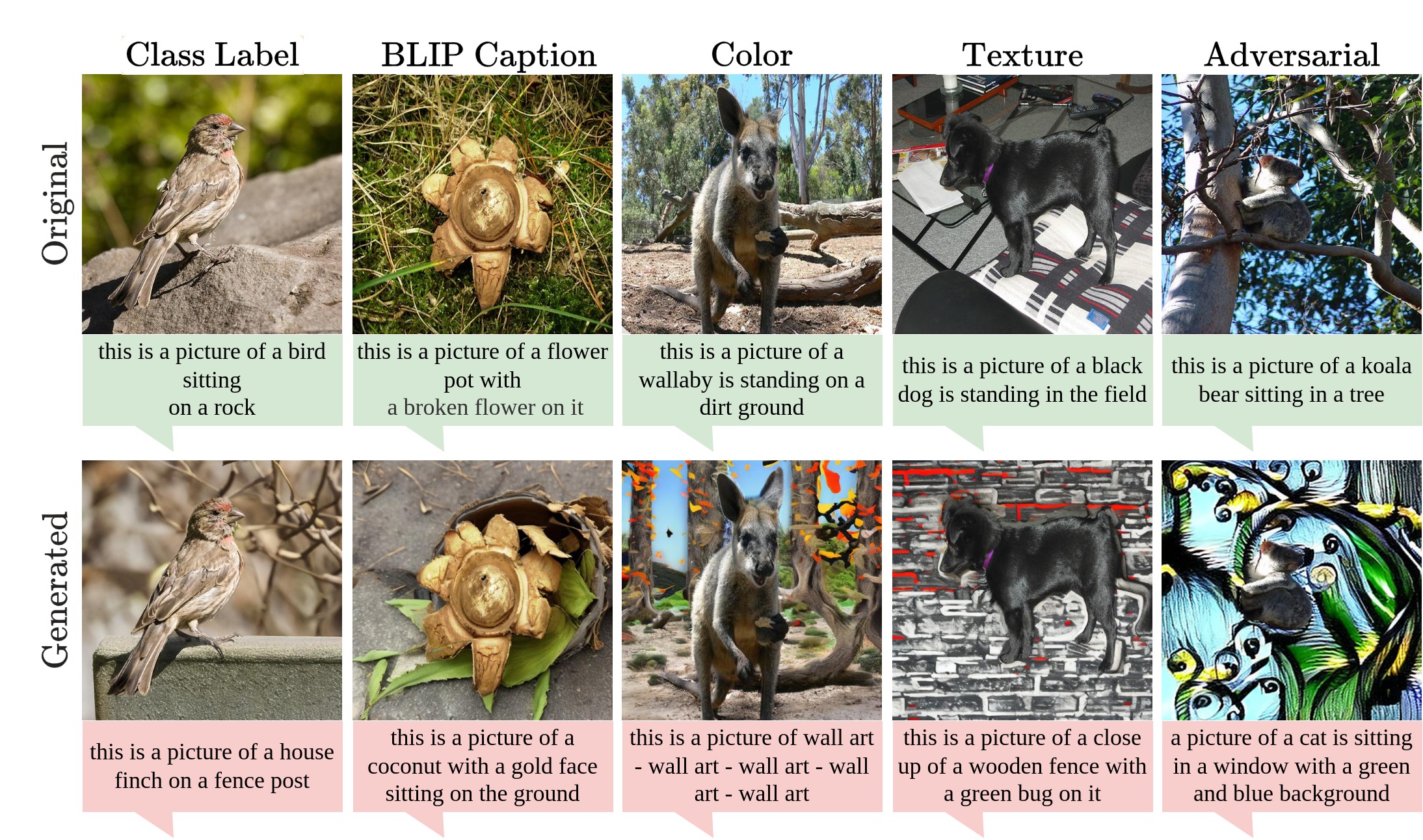
    }
    \caption{A visual comparison of BLIP-2 captions on clean and generated datasets. The top row shows captions on clean images, while the bottom row displays captions on generated images. As background complexity increases, BLIP-2 fails to accurately represent the true class in the image.}
    \label{clip-score}
\end{figure}

\begin{figure}[H]
    \centering
    \includegraphics[width=\linewidth]{
    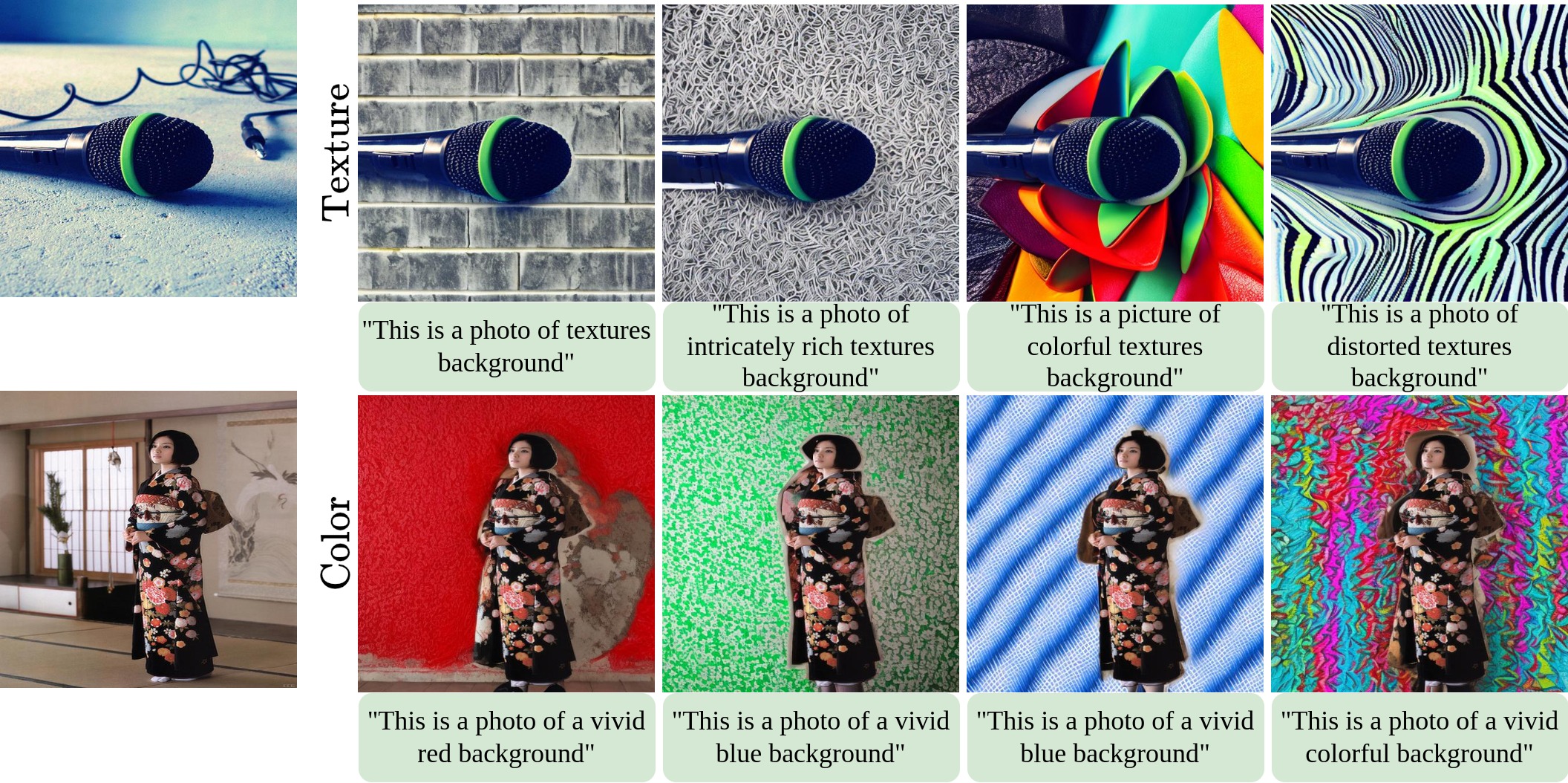
    }
    \caption{The figure illustrates the introduction of background variations achieved through a diverse set of texture and color text prompts}
    \label{main-color}
\end{figure}

\newpage

\subsection{Qualitative Results on Detection}
\label{sec:detection results}

\begin{figure*}[h!]
\begin{minipage}{\textwidth}

\centering


\begin{minipage}{0.19\textwidth}
  \centering
  \includegraphics[height=2.4cm, width=\linewidth , keepaspectratio]
  {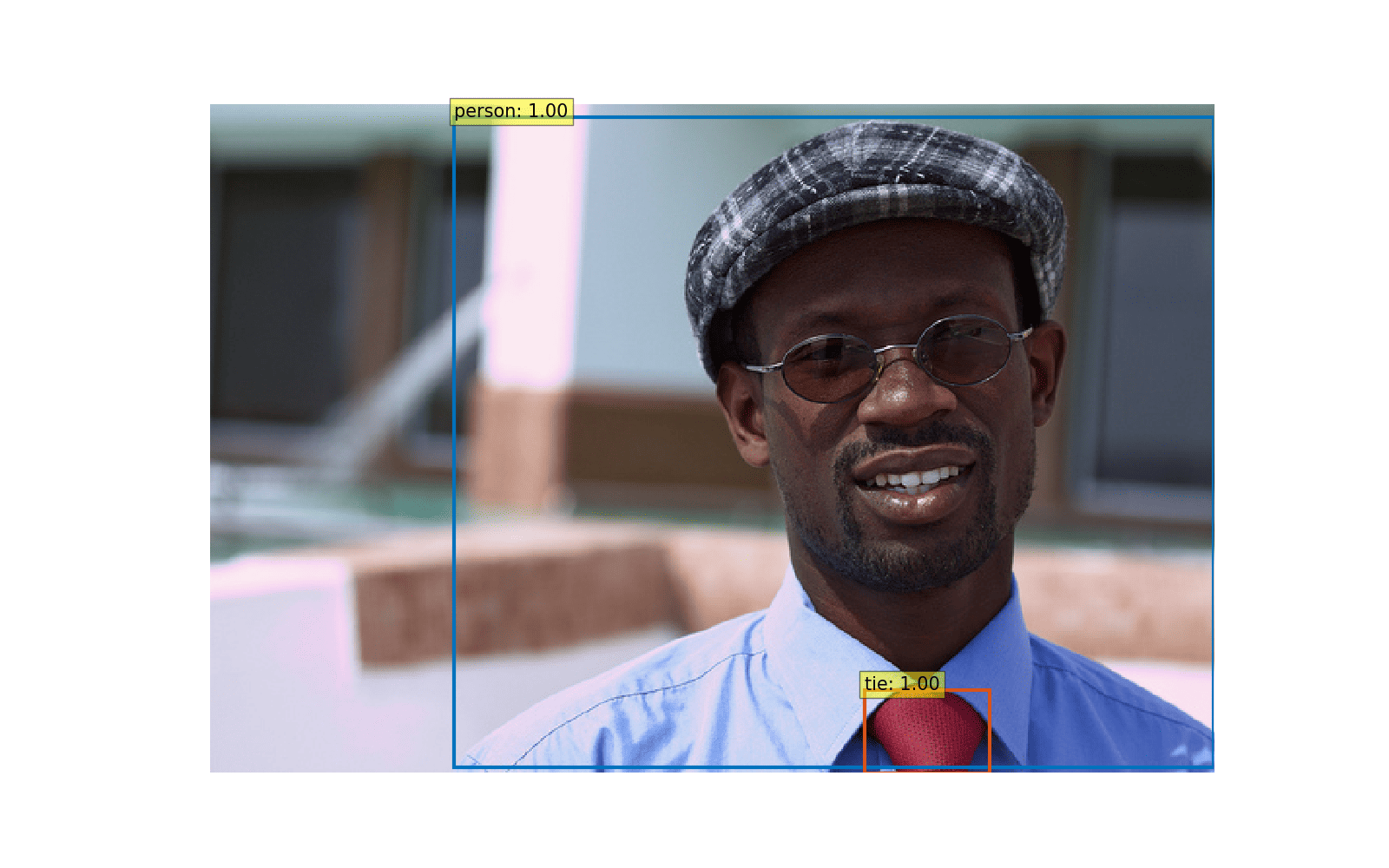}
\end{minipage}
\begin{minipage}{0.19\textwidth}
  \centering
  \includegraphics[height=3.2cm, width=\linewidth, keepaspectratio ]{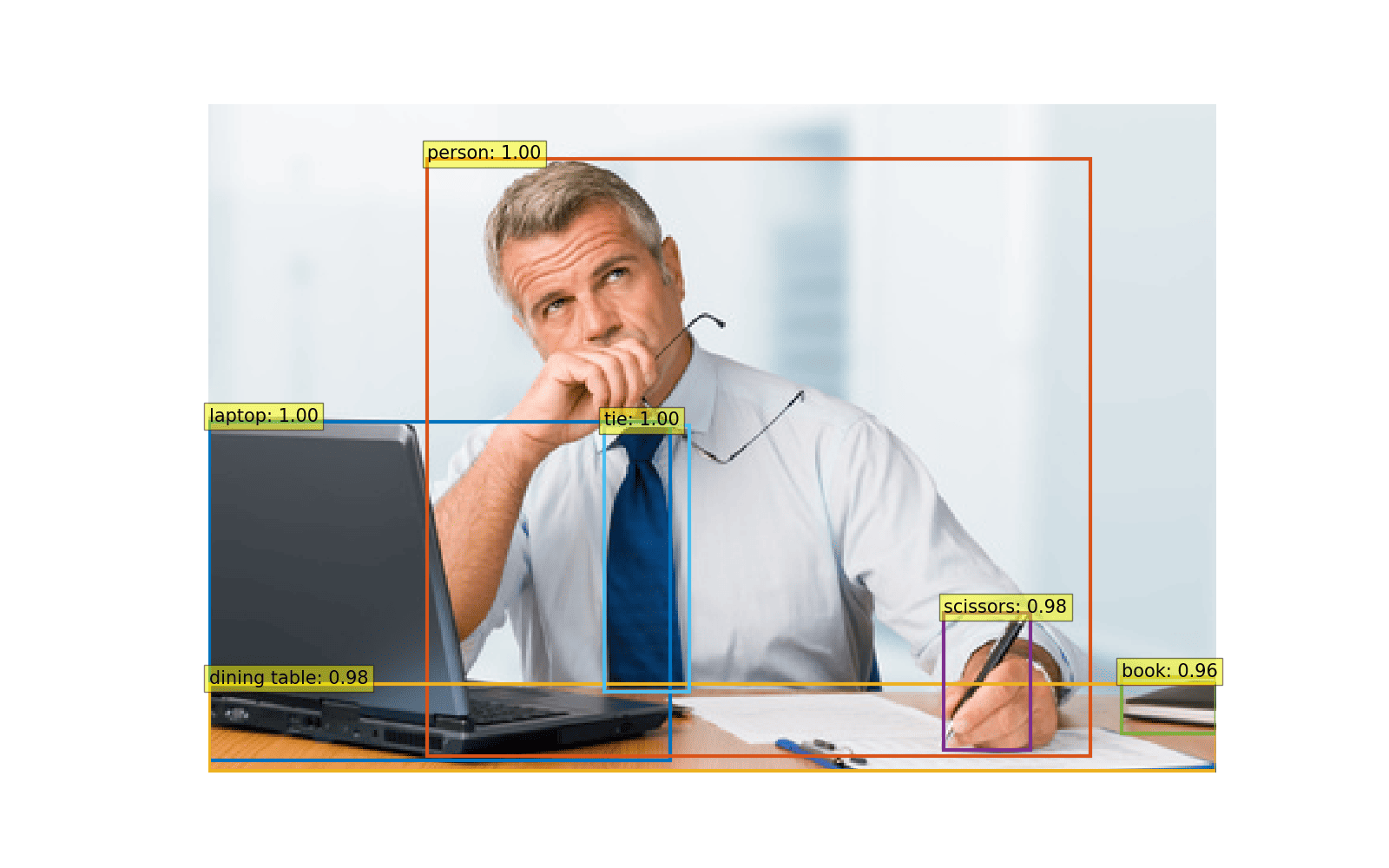}
\end{minipage}
\begin{minipage}{0.19\textwidth}
  \centering
  \includegraphics[height=3.2cm, width=\linewidth, keepaspectratio]{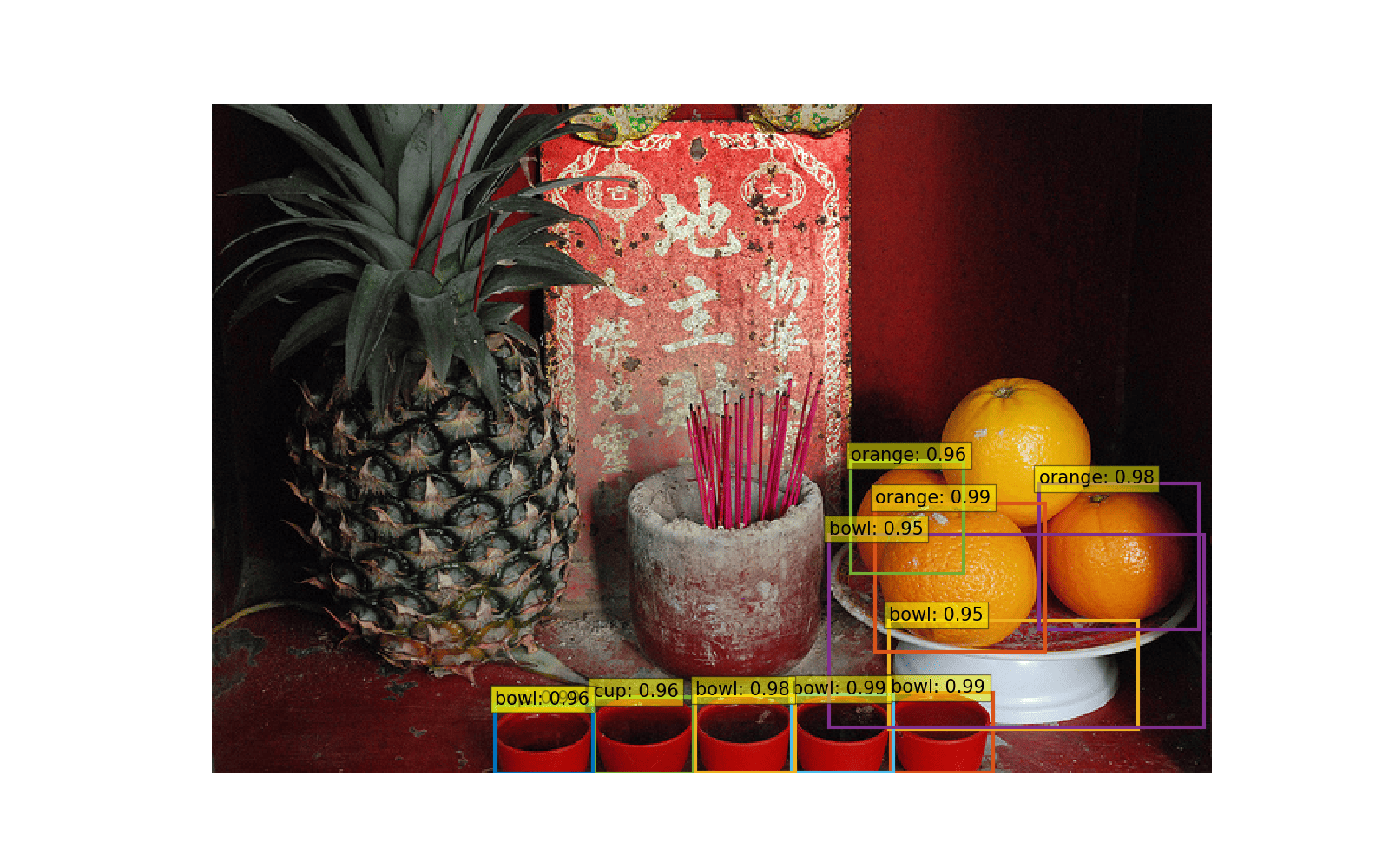}
\end{minipage}
\begin{minipage}{0.19\textwidth}
  \centering
  \includegraphics[height=3.2cm, width=\linewidth, keepaspectratio]{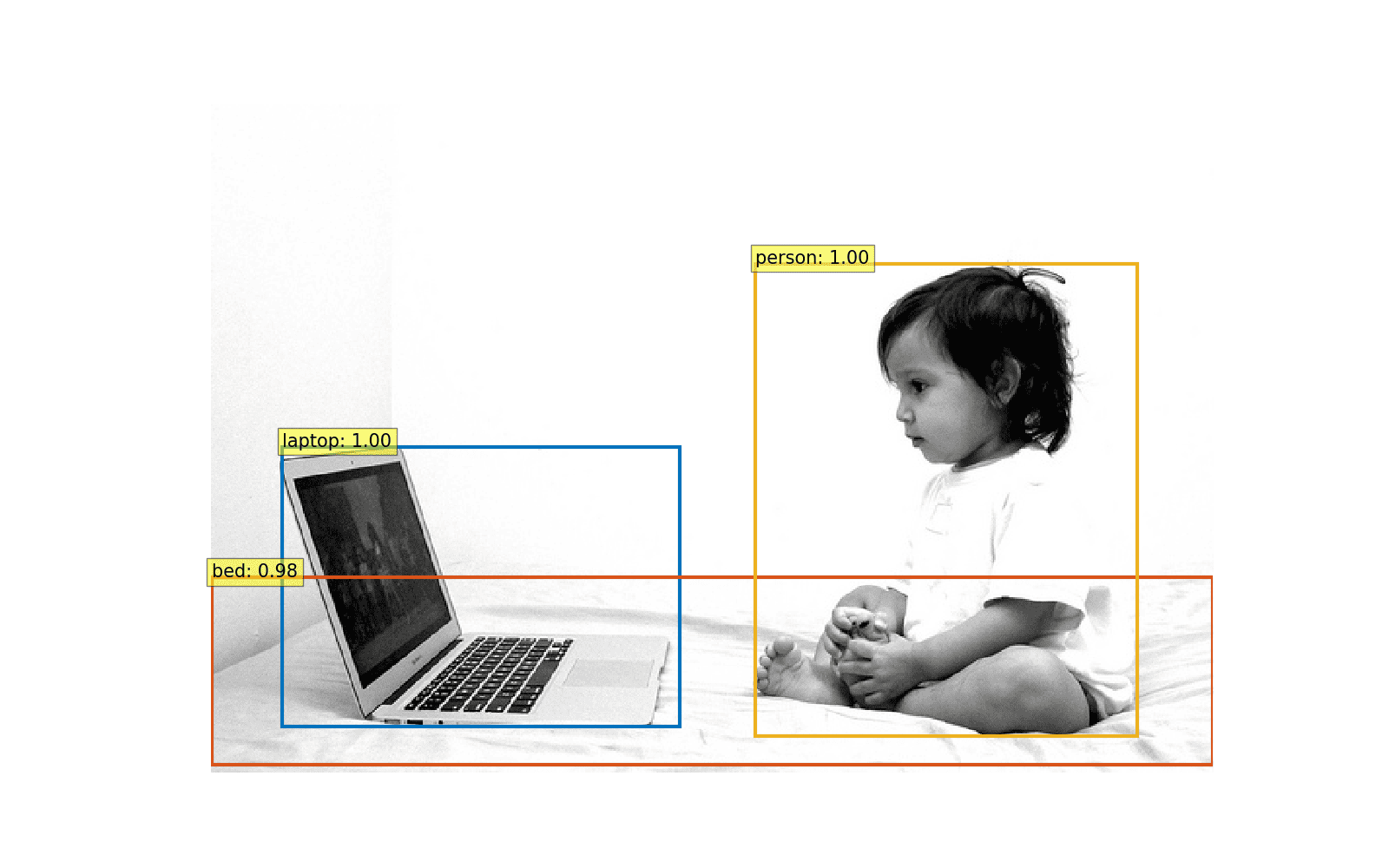}
\end{minipage}
\begin{minipage}{0.19\textwidth}
  \centering
  \includegraphics[height=3.2cm, width=\linewidth, keepaspectratio]{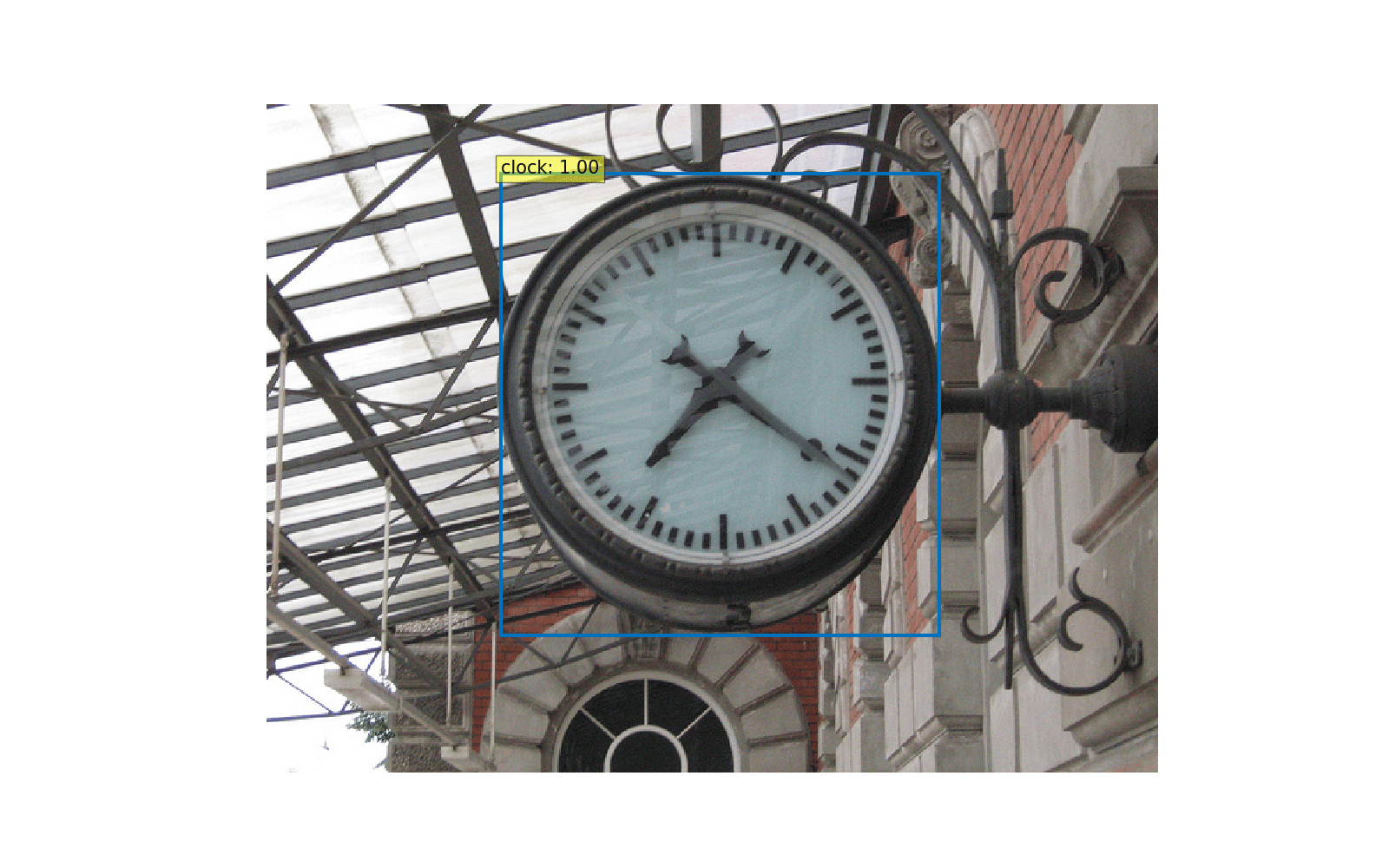}
\end{minipage}
\begin{minipage}{0.19\textwidth}
  \centering
  \includegraphics[height=3.2cm, width=\linewidth , keepaspectratio]{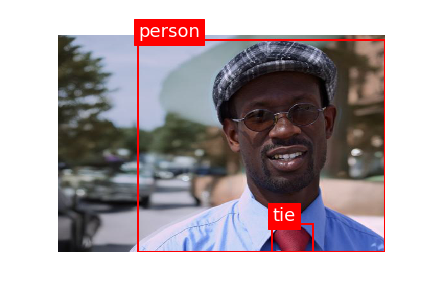}
\end{minipage}
\begin{minipage}{0.19\textwidth}
  \centering
  \includegraphics[height=3.2cm, width=\linewidth, keepaspectratio ]{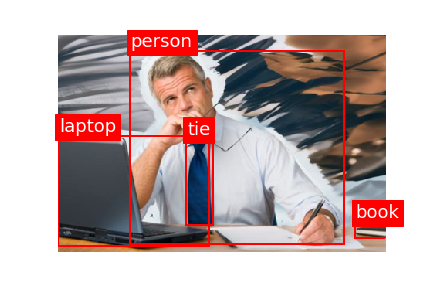}
\end{minipage}
\begin{minipage}{0.19\textwidth}
  \centering
  \includegraphics[height=3.2cm, width=\linewidth, keepaspectratio]{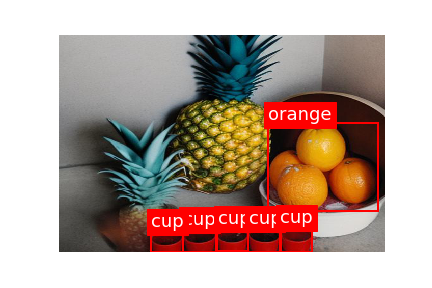}
\end{minipage}
\begin{minipage}{0.19\textwidth}
  \centering
  \includegraphics[height=3.2cm, width=\linewidth, keepaspectratio]{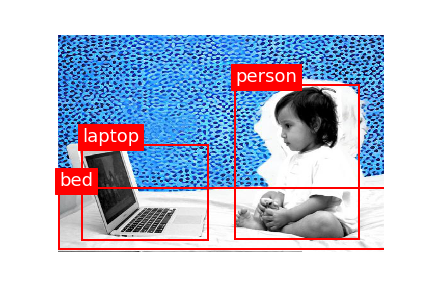}
\end{minipage}
\begin{minipage}{0.19\textwidth}
  \centering
  \includegraphics[height=3.2cm, width=\linewidth, keepaspectratio]{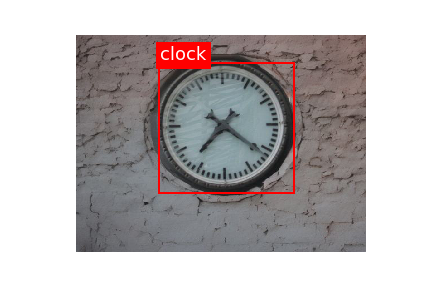}
\end{minipage}

\begin{minipage}{0.19\textwidth}
  \centering
  \includegraphics[height=2.4cm, width=\linewidth , keepaspectratio]{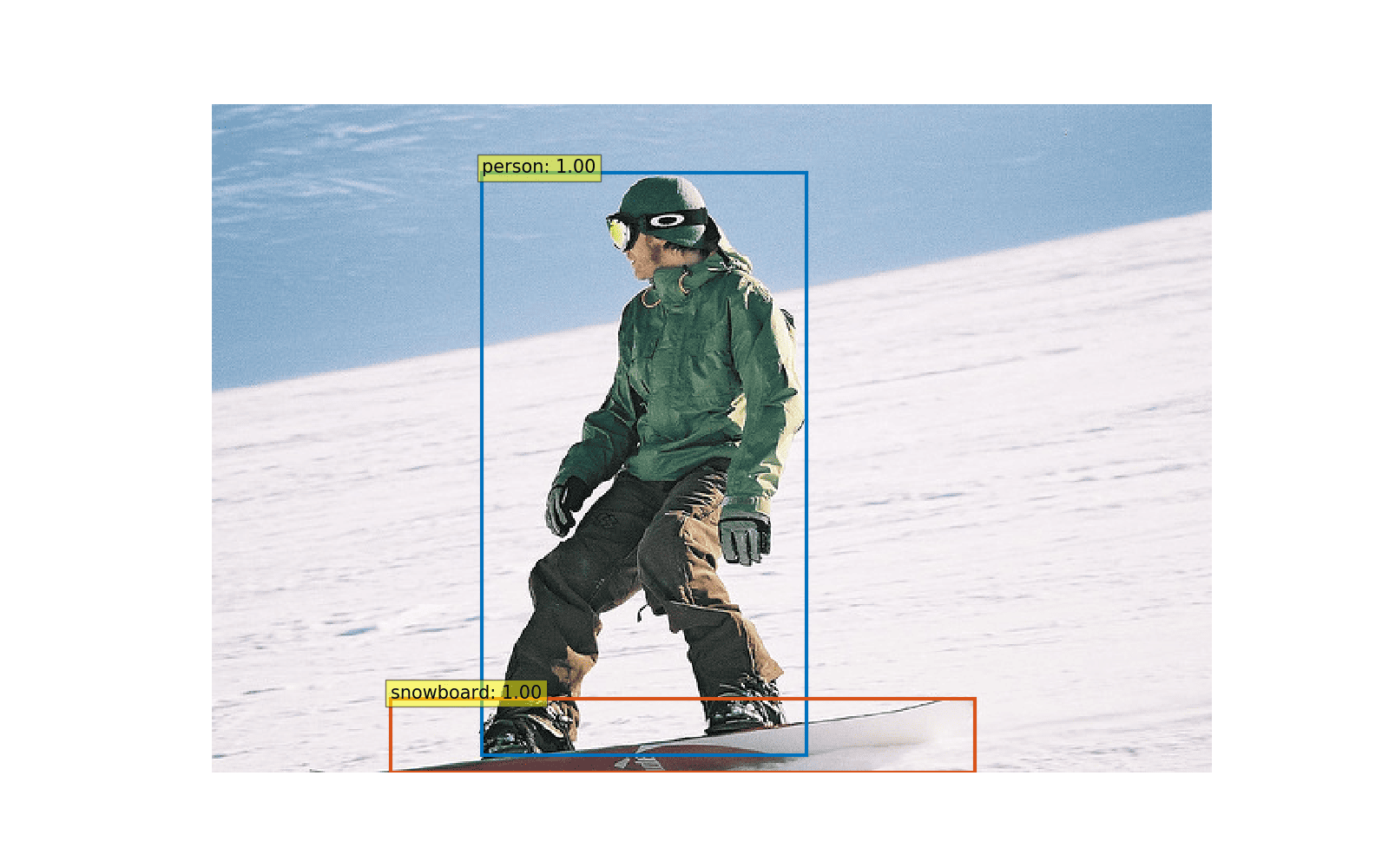}
\end{minipage}
\begin{minipage}{0.19\textwidth}
  \centering
  \includegraphics[height=3.2cm, width=\linewidth, keepaspectratio ]{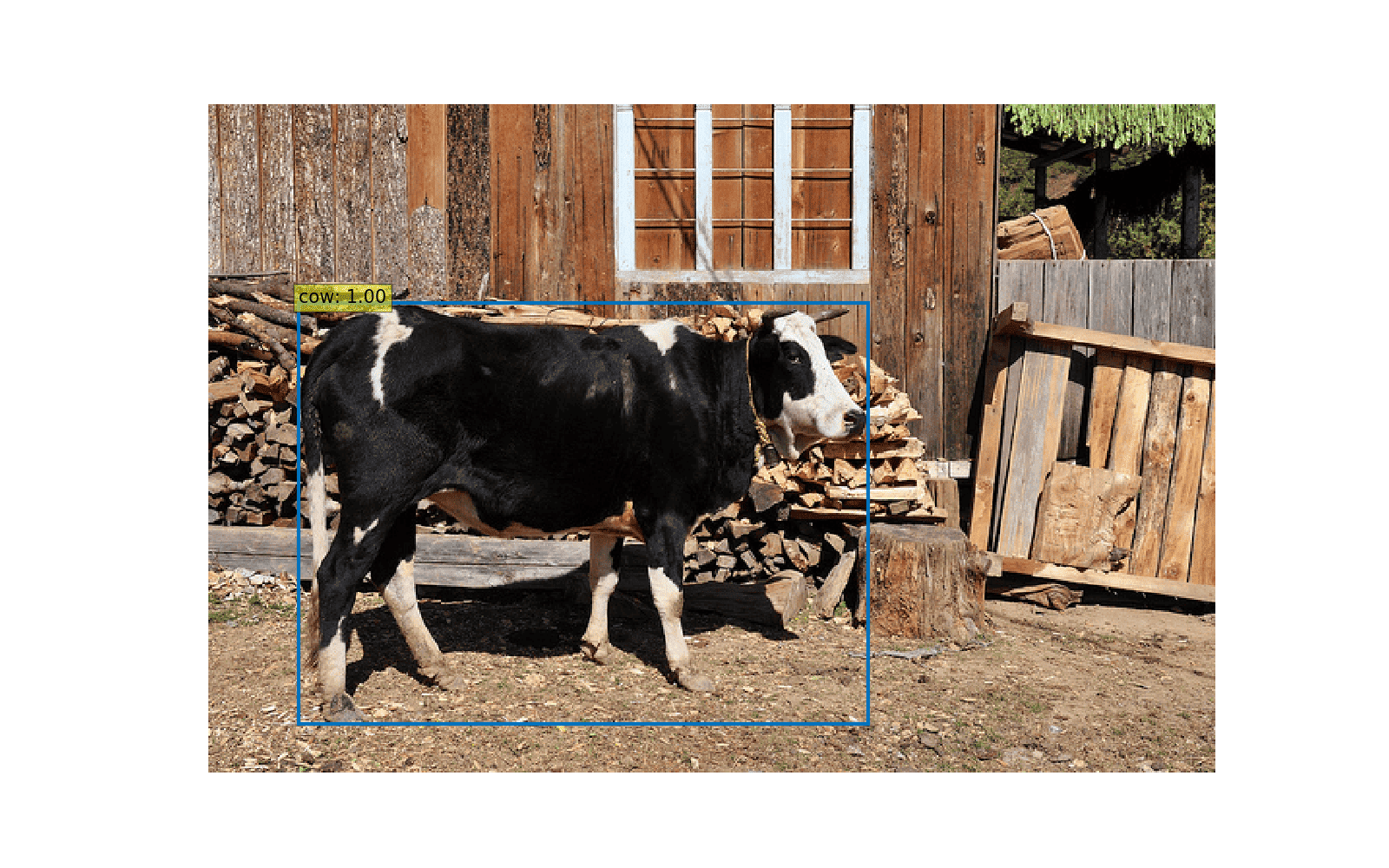}
\end{minipage}
\begin{minipage}{0.19\textwidth}
  \centering
  \includegraphics[height=3.2cm, width=\linewidth, keepaspectratio]{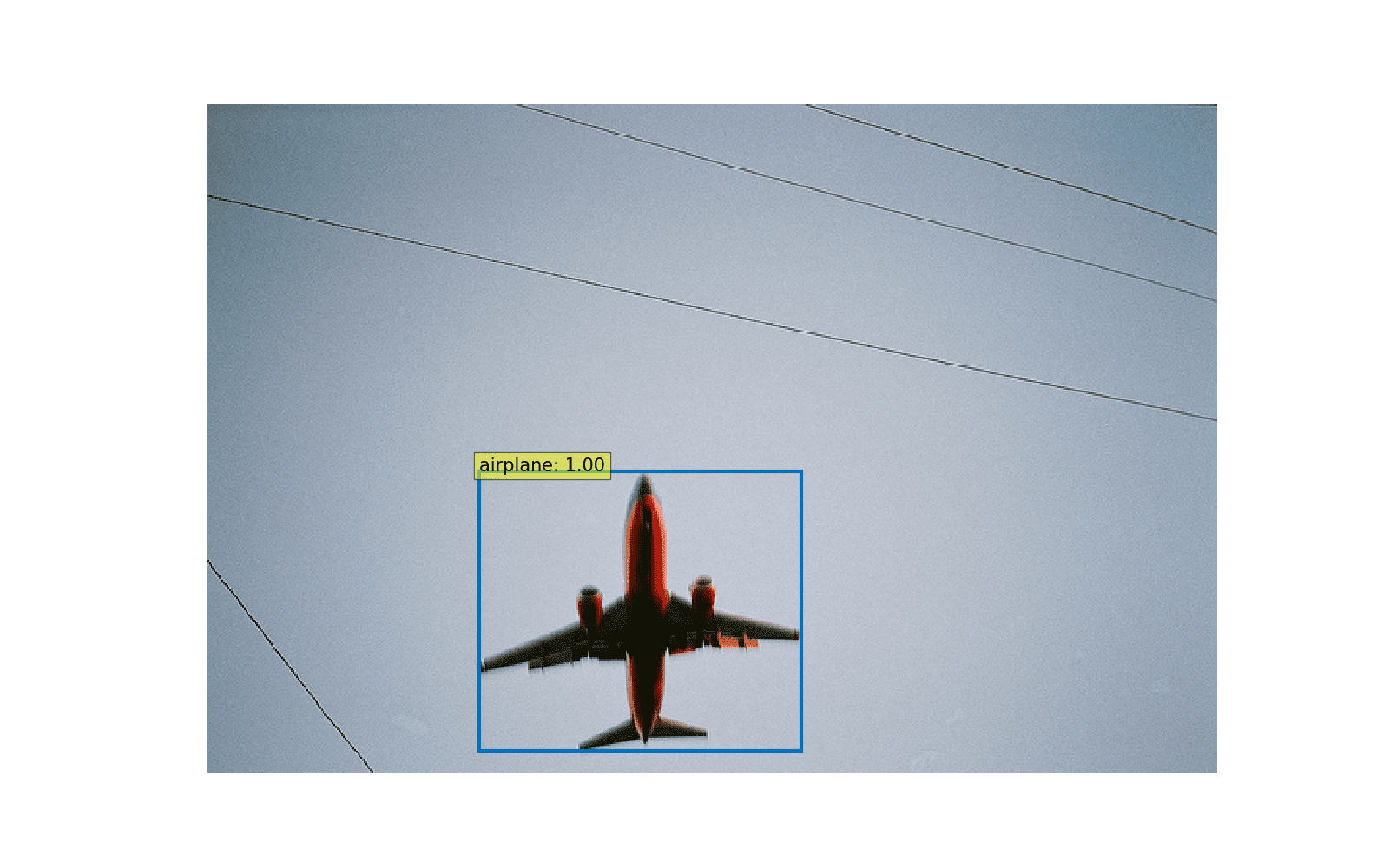}
\end{minipage}
\begin{minipage}{0.19\textwidth}
  \centering
  \includegraphics[height=3.2cm, width=\linewidth, keepaspectratio]{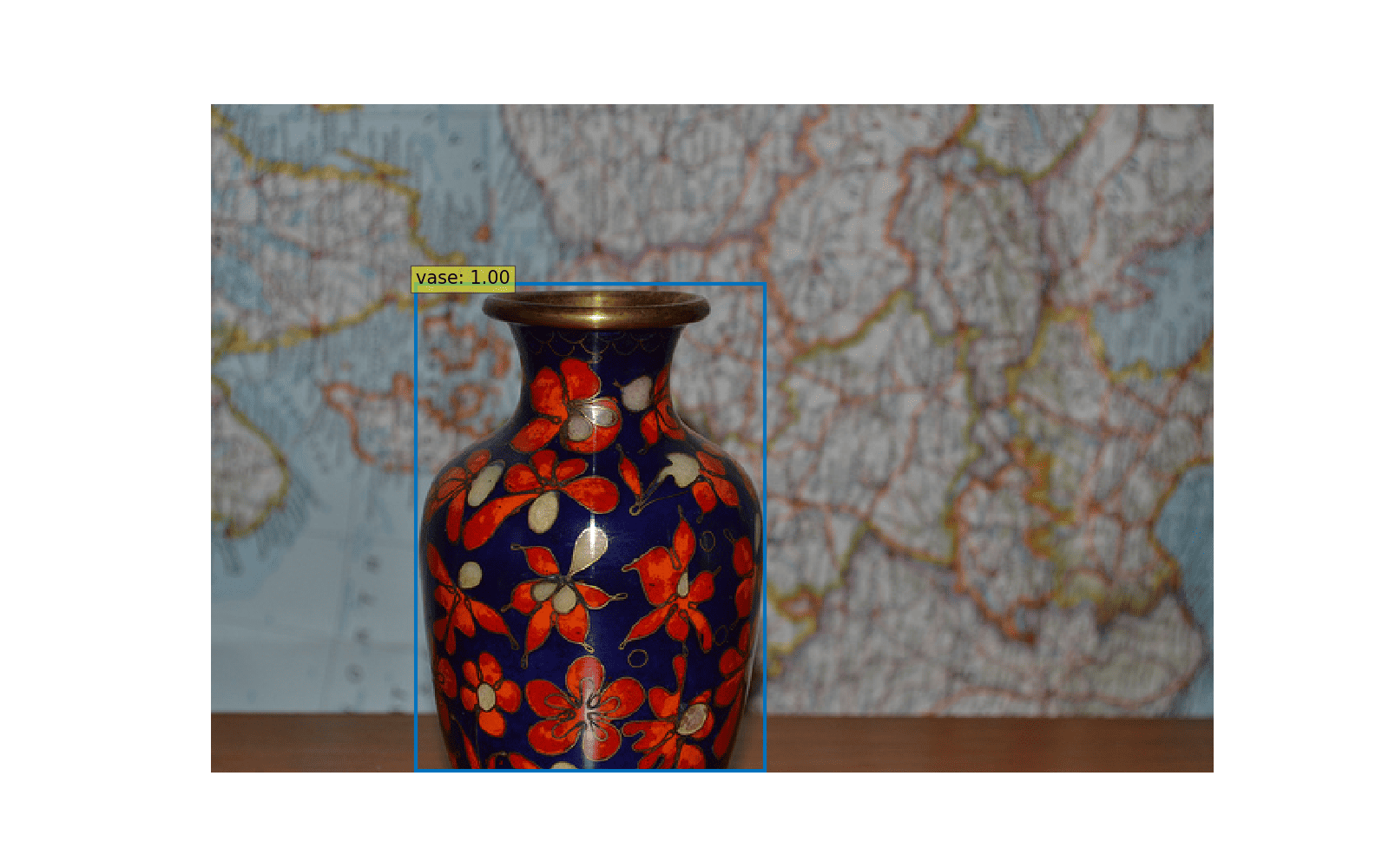}
\end{minipage}
\begin{minipage}{0.19\textwidth}
  \centering
  \includegraphics[height=3.2cm, width=\linewidth, keepaspectratio]{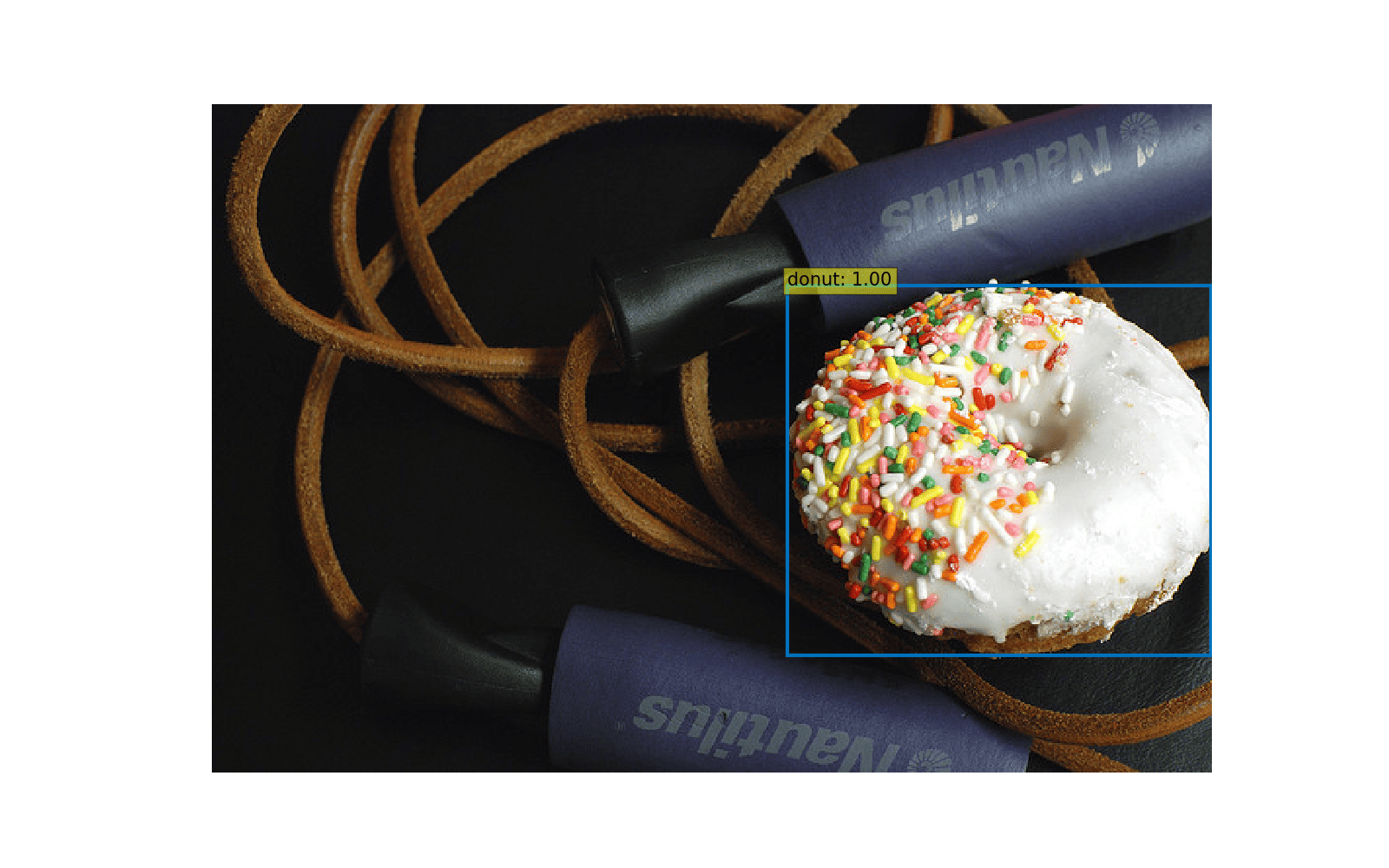}
\end{minipage}

\begin{minipage}{0.19\textwidth}
  \centering
  \includegraphics[height=3.2cm, width=\linewidth , keepaspectratio]{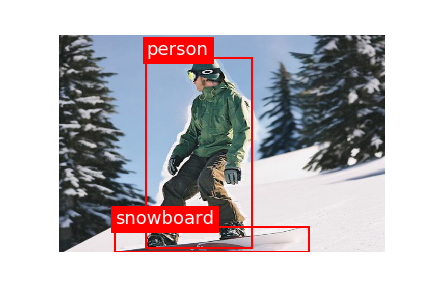}
\end{minipage}
\begin{minipage}{0.19\textwidth}
  \centering
  \includegraphics[height=3.2cm, width=\linewidth, keepaspectratio ]{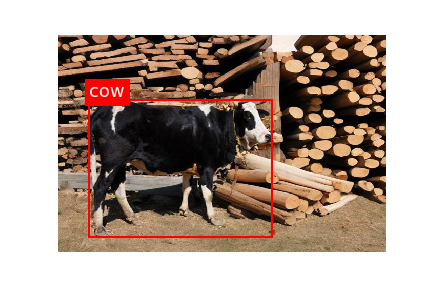}
\end{minipage}
\begin{minipage}{0.19\textwidth}
  \centering
  \includegraphics[height=3.2cm, width=\linewidth, keepaspectratio]{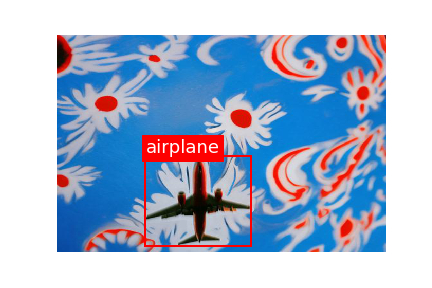}
\end{minipage}
\begin{minipage}{0.19\textwidth}
  \centering
  \includegraphics[height=3.2cm, width=\linewidth, keepaspectratio]{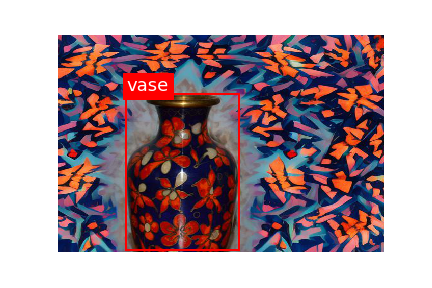}
\end{minipage}
\begin{minipage}{0.19\textwidth}
  \centering
  \includegraphics[height=3.2cm, width=\linewidth, keepaspectratio]{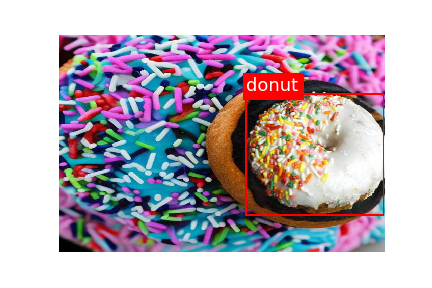}
\end{minipage}

\begin{minipage}{0.19\textwidth}
  \centering
  \includegraphics[height=2.4cm, width=\linewidth , keepaspectratio]{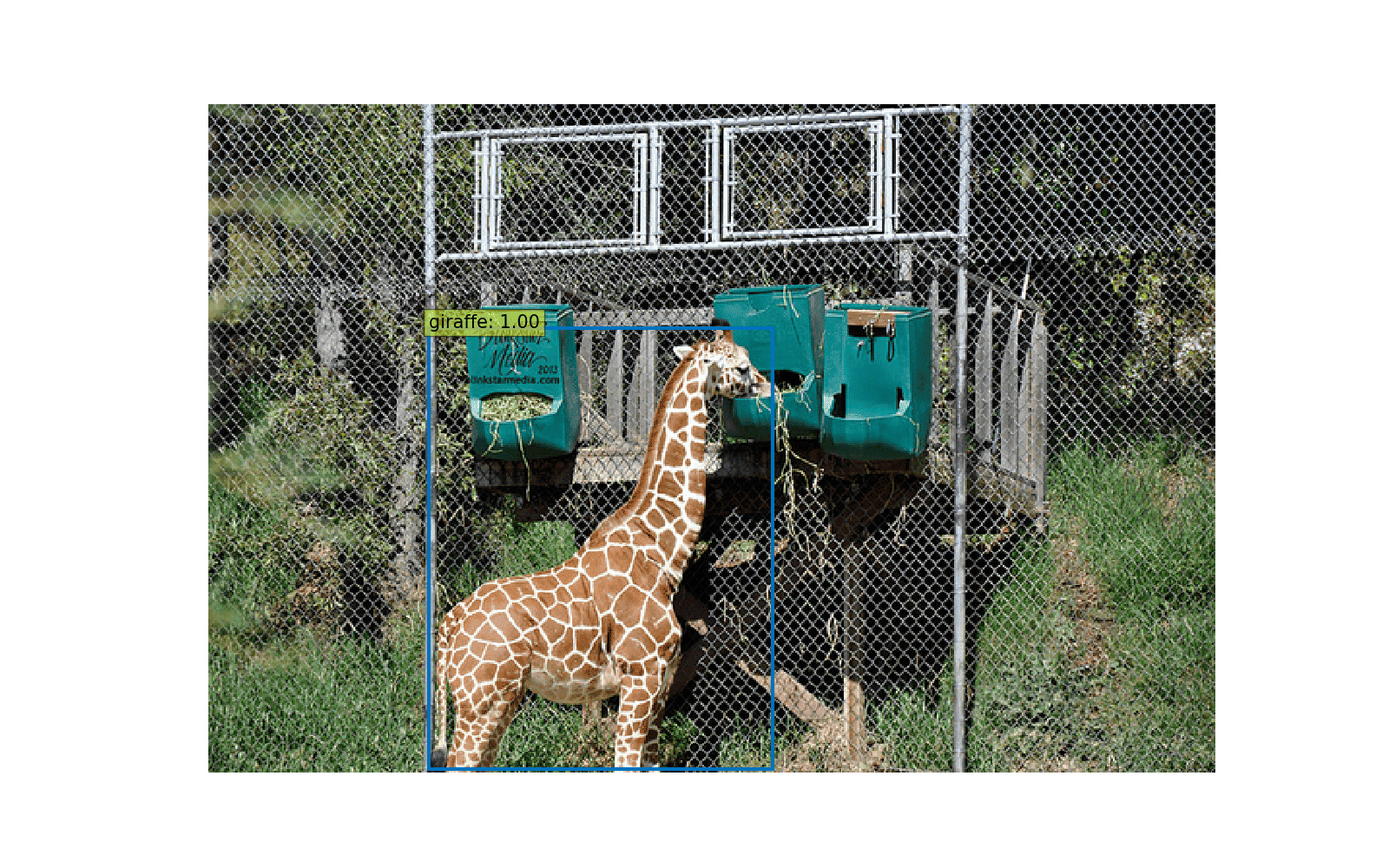}
\end{minipage}
\begin{minipage}{0.19\textwidth}
  \centering
  \includegraphics[height=3.2cm, width=\linewidth, keepaspectratio ]{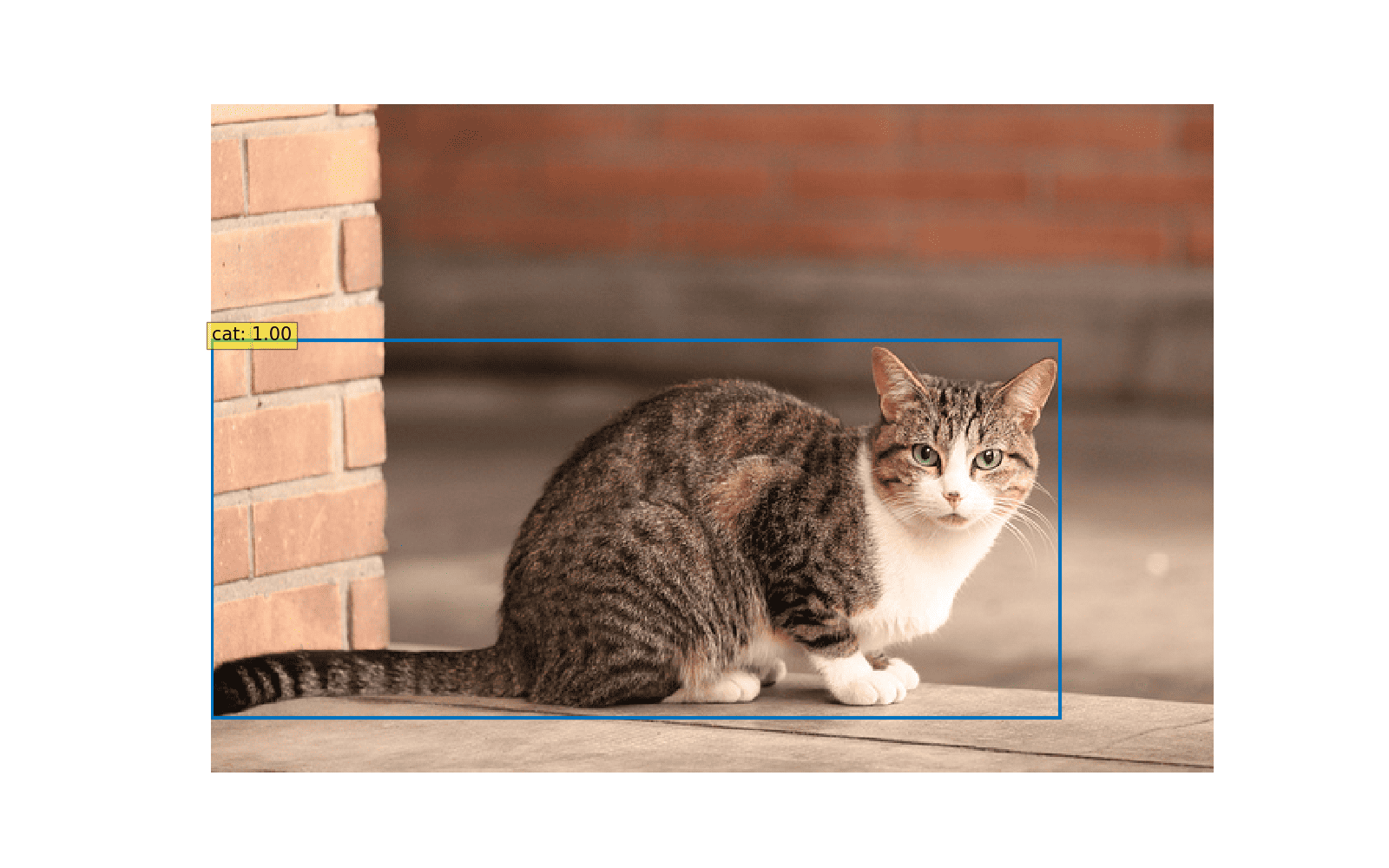}
\end{minipage}
\begin{minipage}{0.19\textwidth}
  \centering
  \includegraphics[height=3.2cm, width=\linewidth, keepaspectratio]{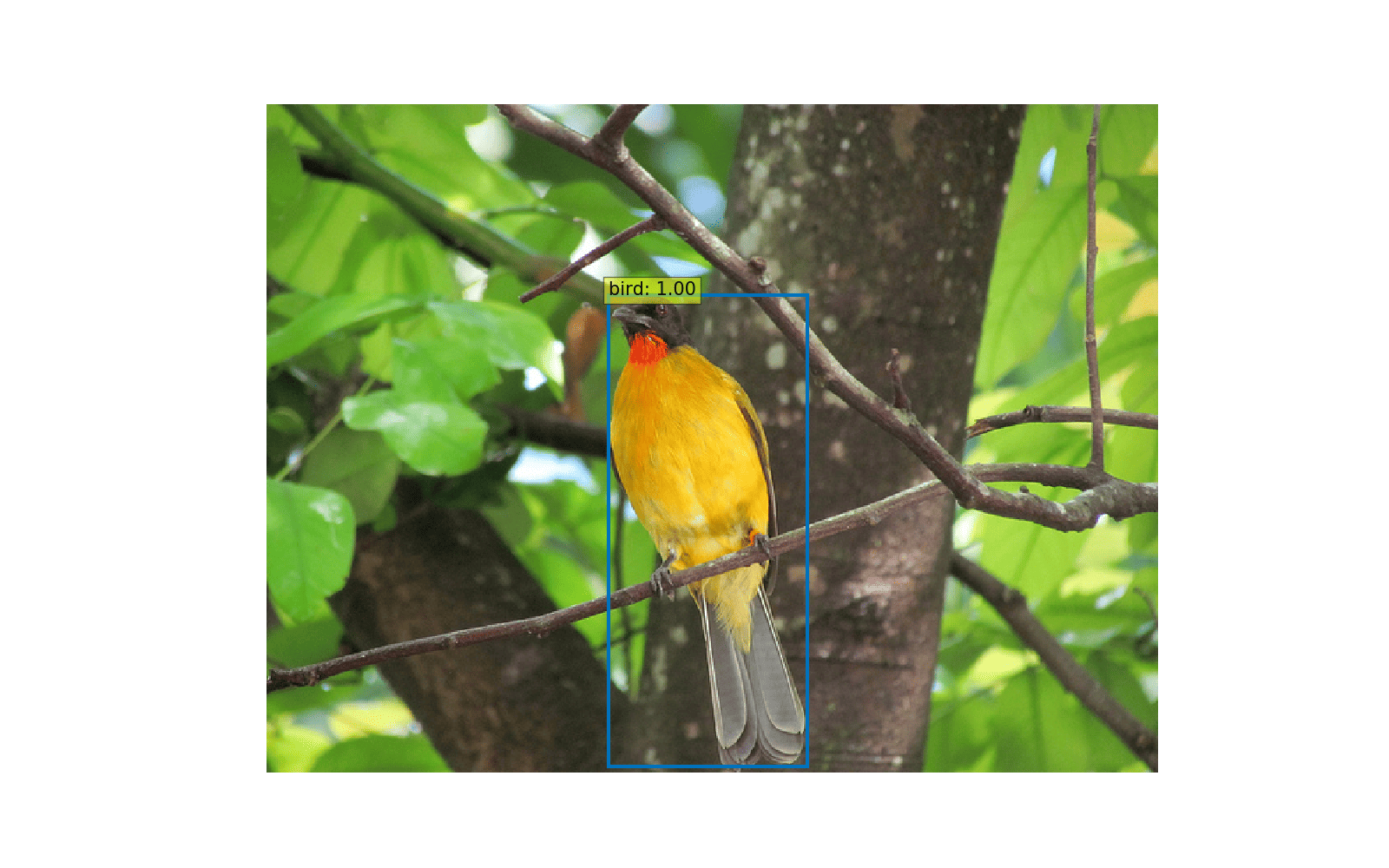}
\end{minipage}
\begin{minipage}{0.19\textwidth}
  \centering
  \includegraphics[height=3.2cm, width=\linewidth, keepaspectratio]{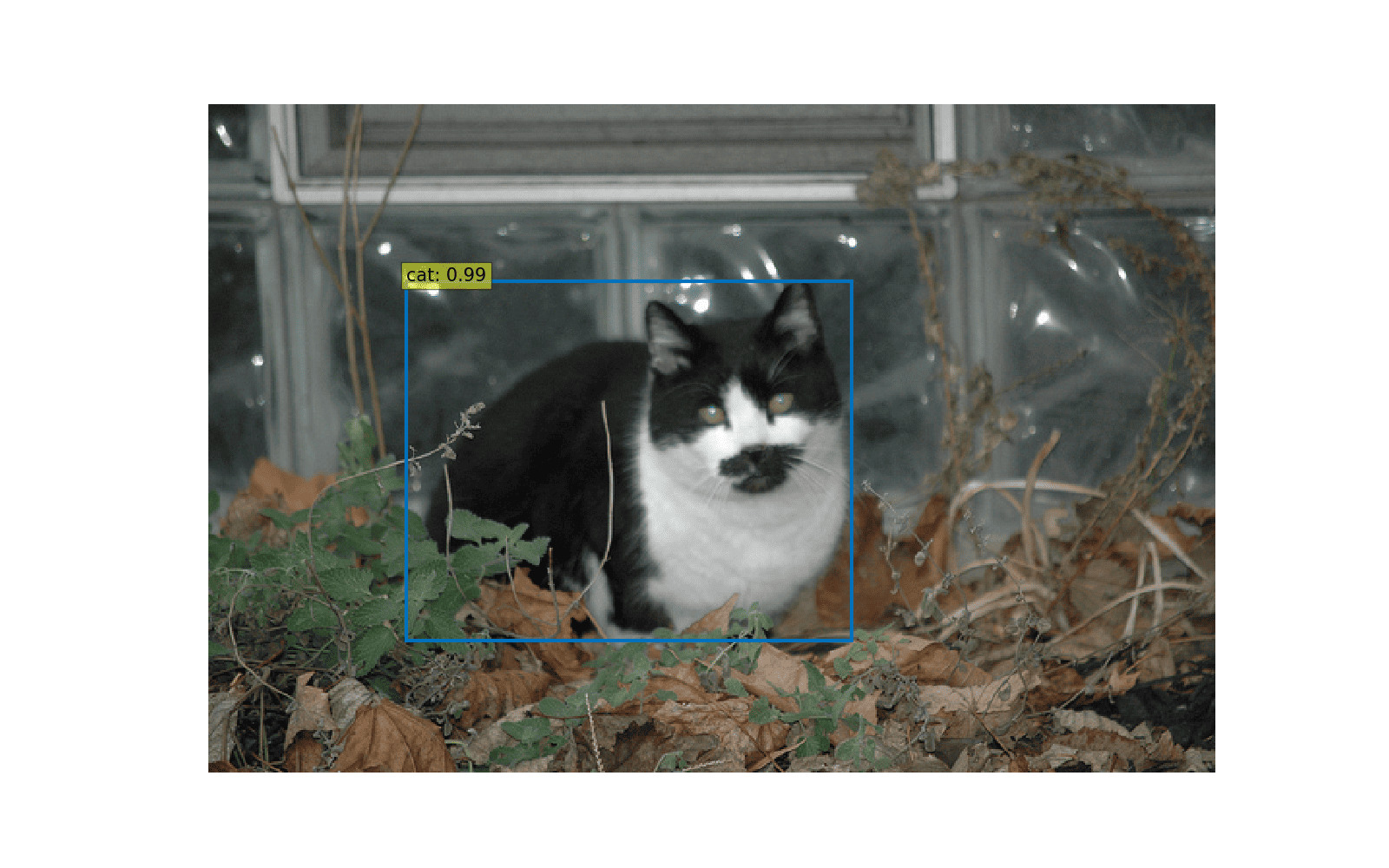}
\end{minipage}
\begin{minipage}{0.19\textwidth}
  \centering
  \includegraphics[height=3.2cm, width=\linewidth, keepaspectratio]{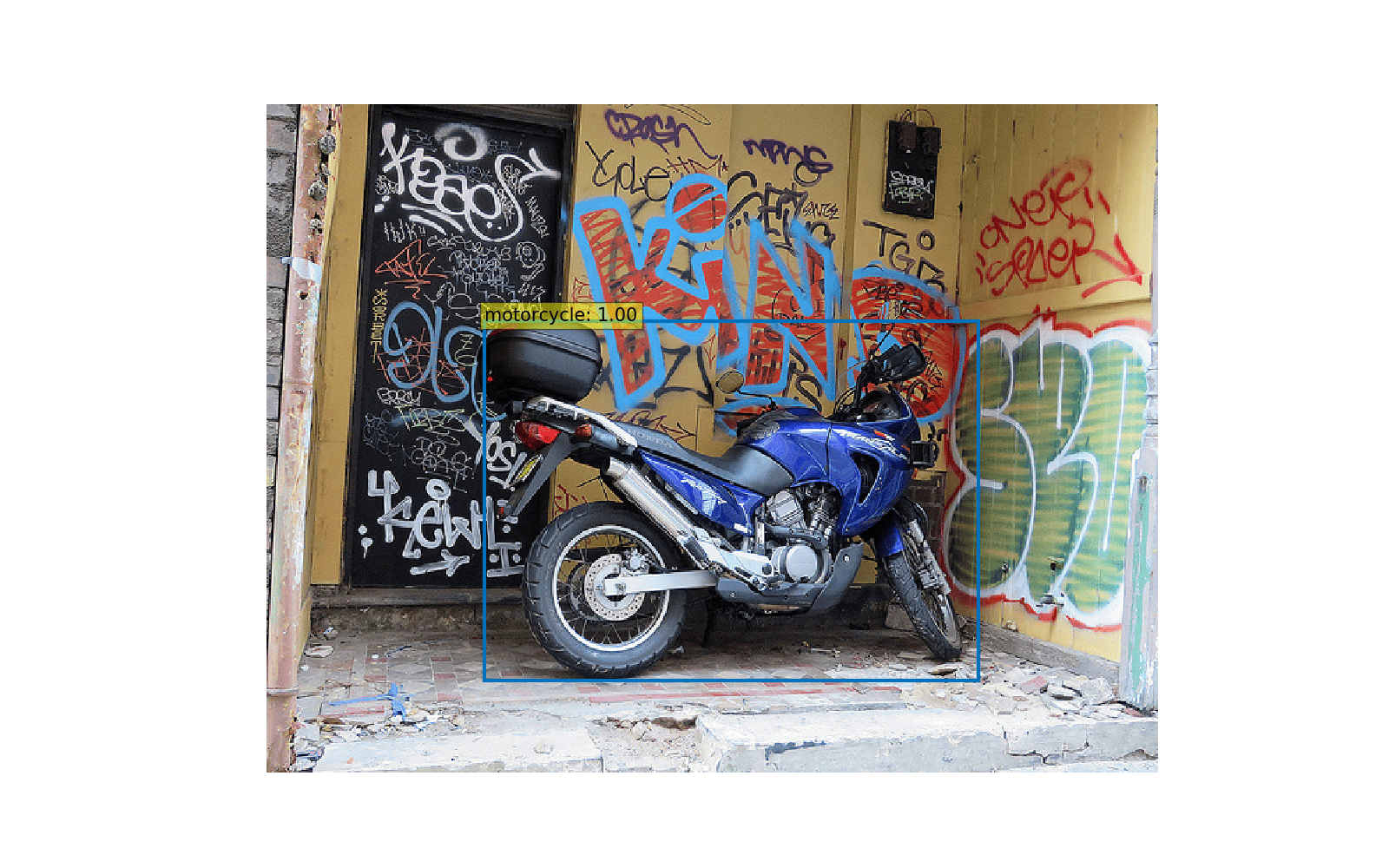}
\end{minipage}

\begin{minipage}{0.19\textwidth}
  \centering
  \includegraphics[height=3.2cm, width=\linewidth , keepaspectratio]{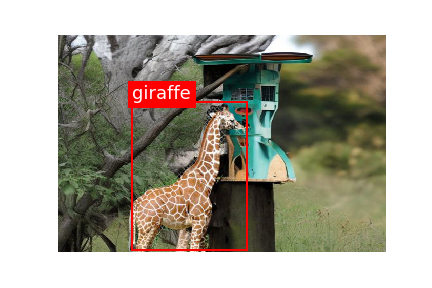}
\end{minipage}
\begin{minipage}{0.19\textwidth}
  \centering
  \includegraphics[height=3.2cm, width=\linewidth, keepaspectratio ]{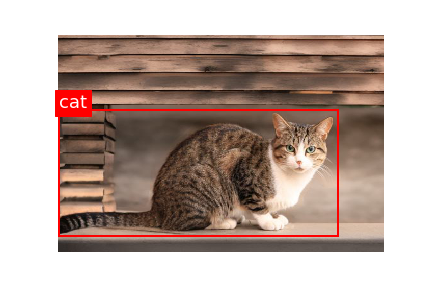}
\end{minipage}
\begin{minipage}{0.19\textwidth}
  \centering
  \includegraphics[height=3.2cm, width=\linewidth, keepaspectratio]{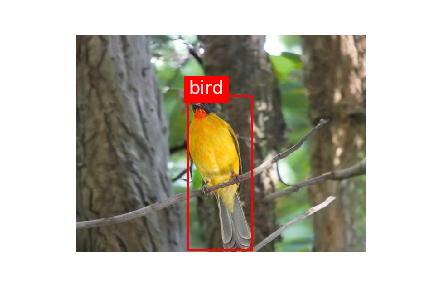}
\end{minipage}
\begin{minipage}{0.19\textwidth}
  \centering
  \includegraphics[height=3.2cm, width=\linewidth, keepaspectratio]{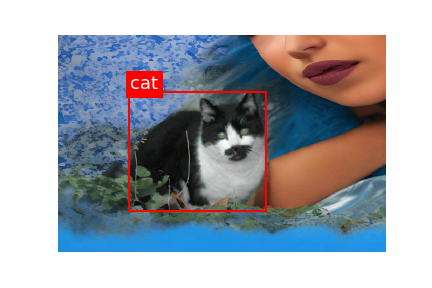}
\end{minipage}
\begin{minipage}{0.19\textwidth}
  \centering
  \includegraphics[height=3.2cm, width=\linewidth, keepaspectratio]{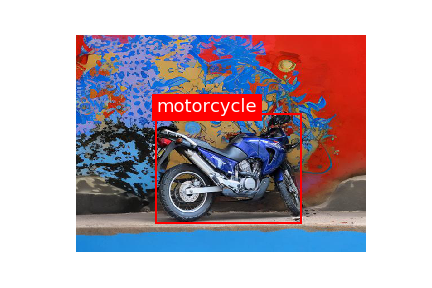}
\end{minipage}

\begin{minipage}{0.19\textwidth}
  \centering
  \includegraphics[height=2.4cm, width=\linewidth , keepaspectratio]{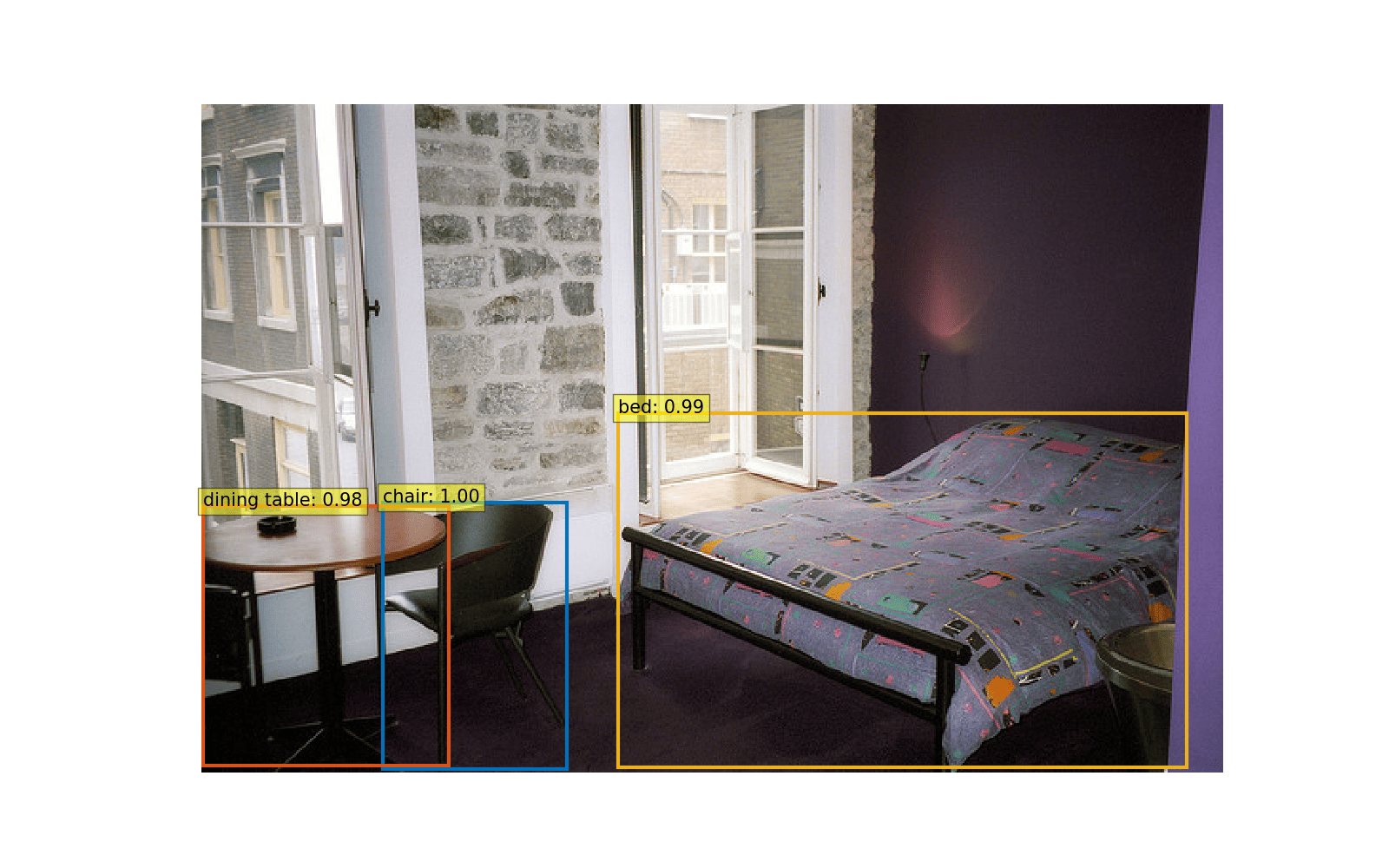}
\end{minipage}
\begin{minipage}{0.19\textwidth}
  \centering
  \includegraphics[height=3.2cm, width=\linewidth, keepaspectratio ]{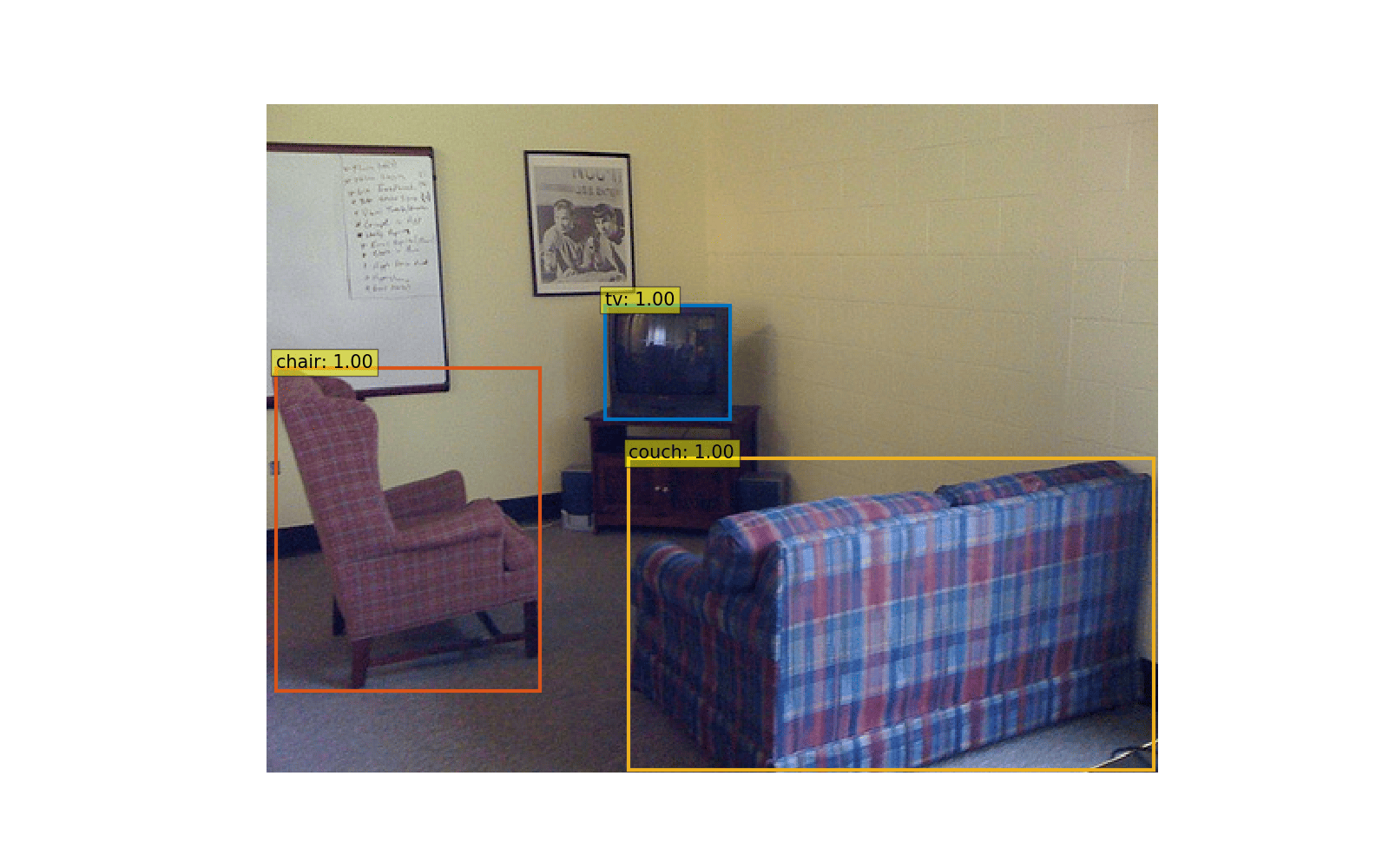}
\end{minipage}
\begin{minipage}{0.19\textwidth}
  \centering
  \includegraphics[height=3.2cm, width=\linewidth, keepaspectratio]{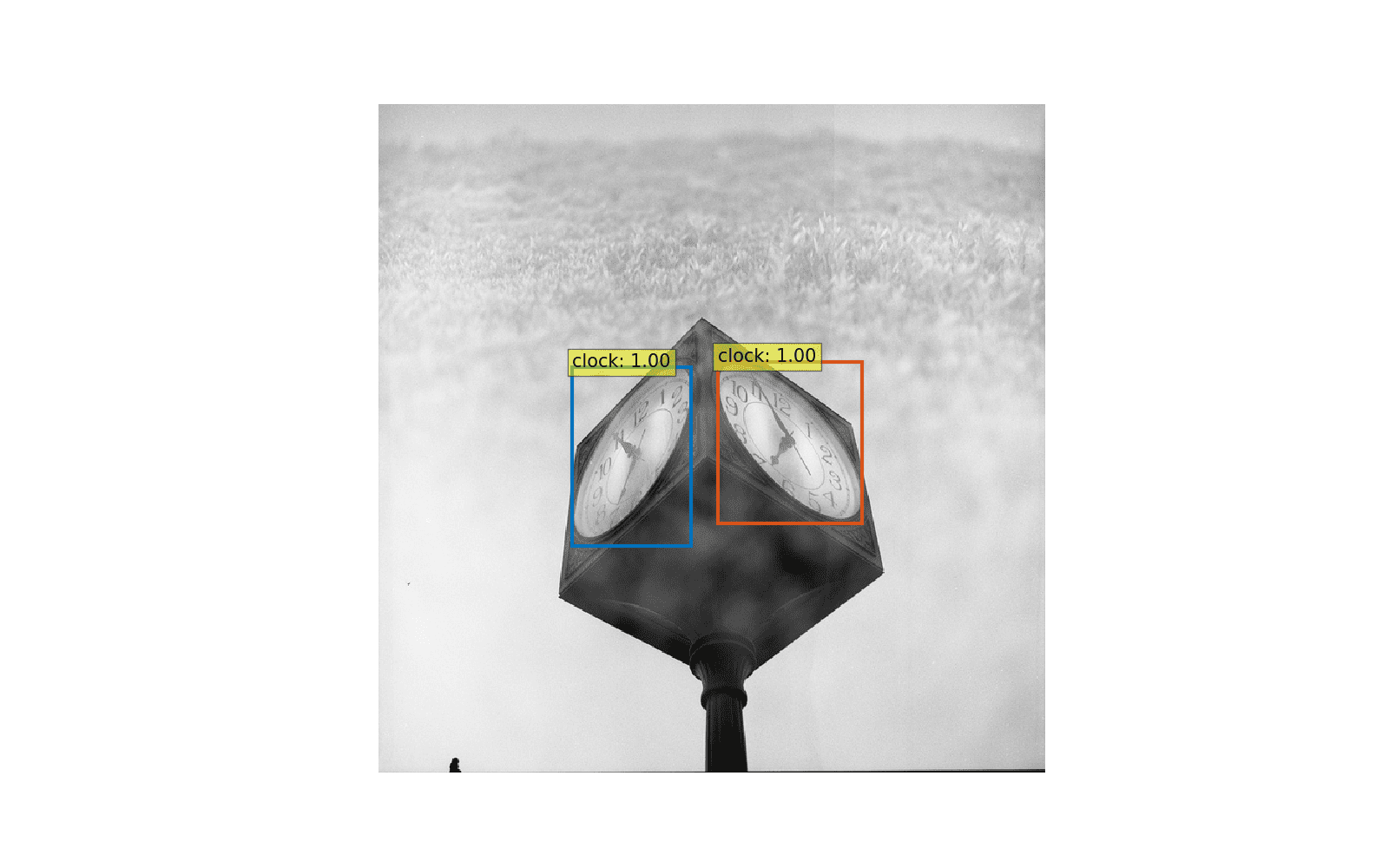}
\end{minipage}
\begin{minipage}{0.19\textwidth}
  \centering
  \includegraphics[height=3.2cm, width=\linewidth, keepaspectratio]{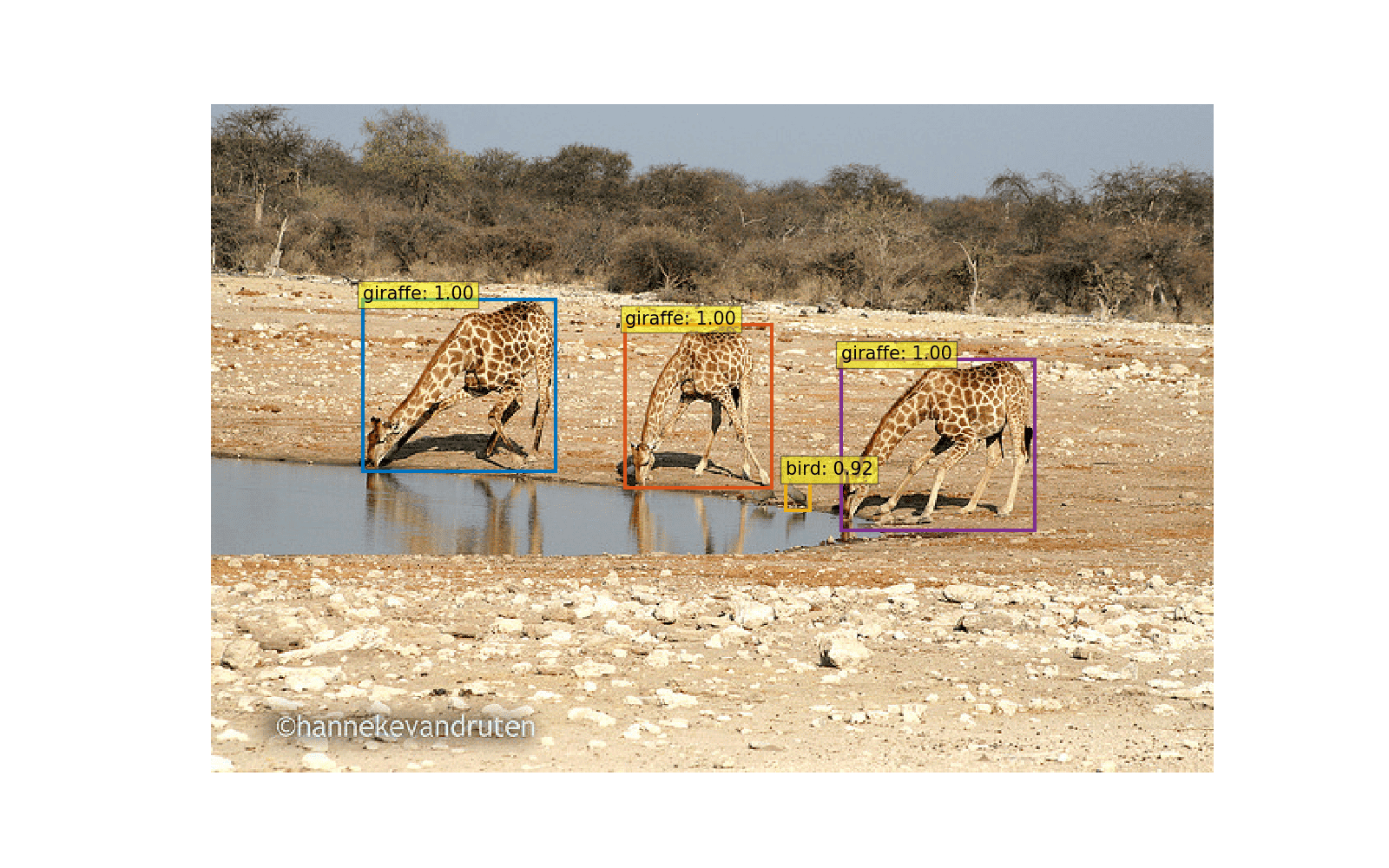}
\end{minipage}
\begin{minipage}{0.19\textwidth}
  \centering
  \includegraphics[height=3.2cm, width=\linewidth, keepaspectratio]{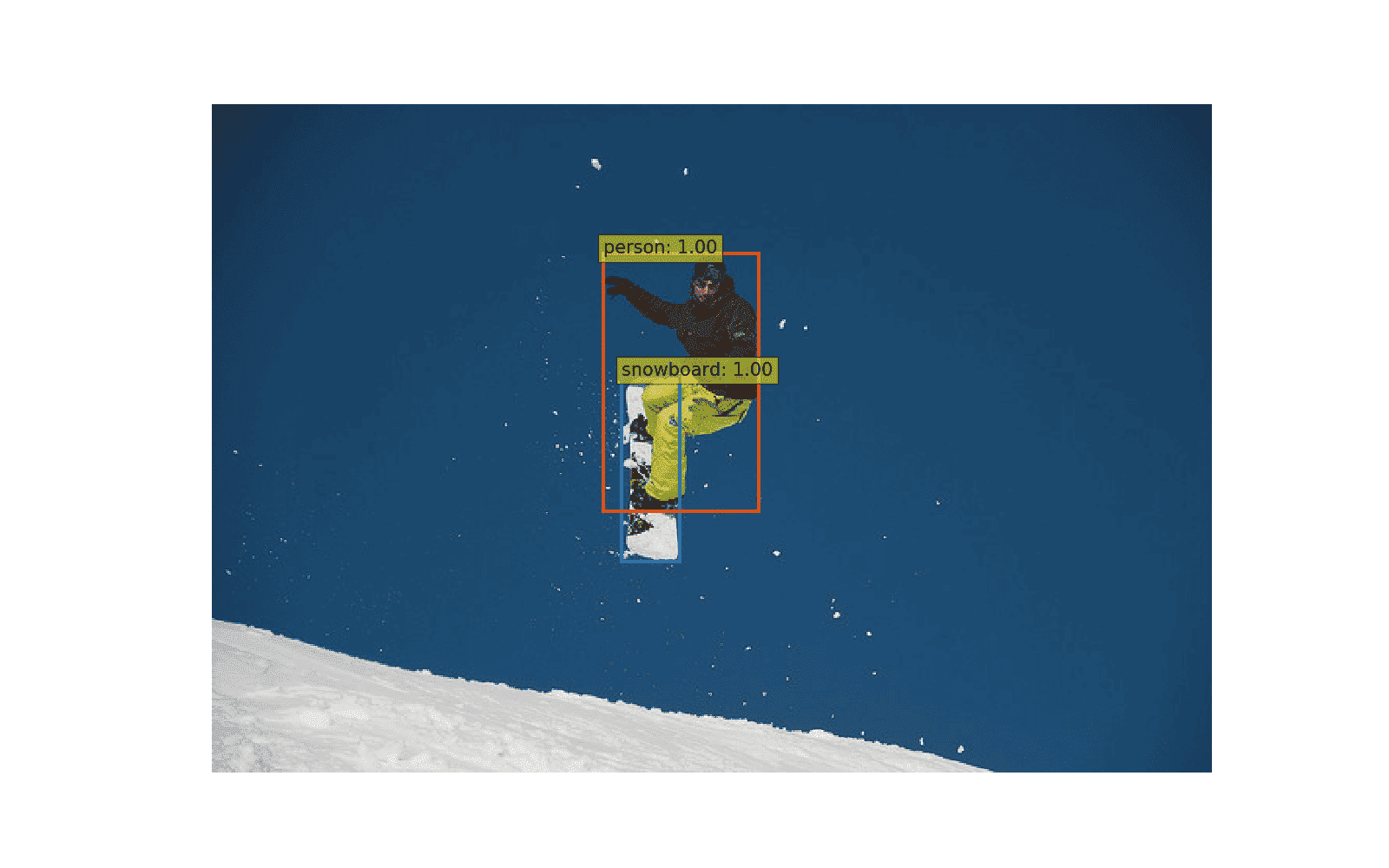}
\end{minipage}

\begin{minipage}{0.19\textwidth}
  \centering
  \includegraphics[height=3.2cm, width=\linewidth , keepaspectratio]{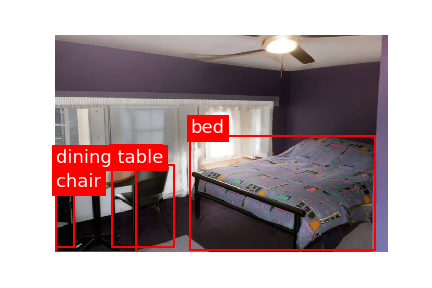}
\end{minipage}
\begin{minipage}{0.19\textwidth}
  \centering
  \includegraphics[height=3.2cm, width=\linewidth, keepaspectratio ]{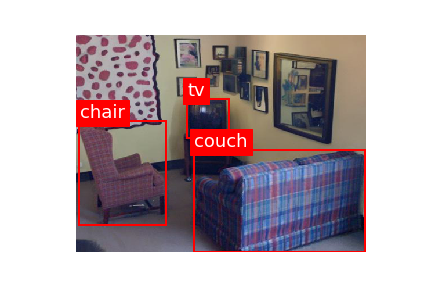}
\end{minipage}
\begin{minipage}{0.19\textwidth}
  \centering
  \includegraphics[height=3.2cm, width=\linewidth, keepaspectratio]{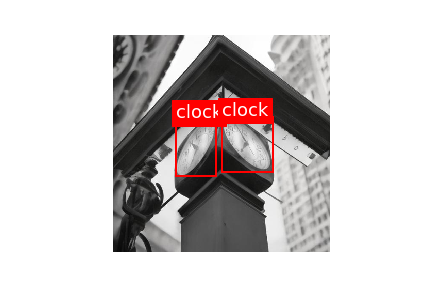}
\end{minipage}
\begin{minipage}{0.19\textwidth}
  \centering
  \includegraphics[height=3.2cm, width=\linewidth, keepaspectratio]{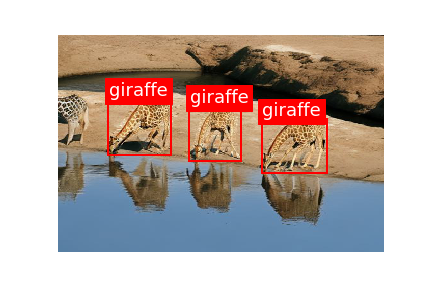}
\end{minipage}
\begin{minipage}{0.19\textwidth}
  \centering
  \includegraphics[height=3.2cm, width=\linewidth, keepaspectratio]{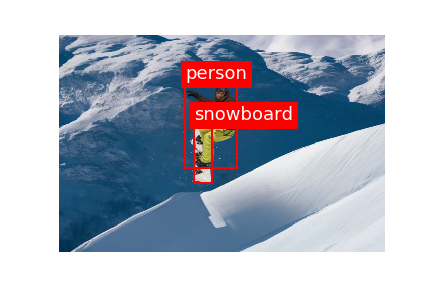}
\end{minipage}

\begin{minipage}{0.19\textwidth}
  \centering
  \includegraphics[height=2.4cm, width=\linewidth , keepaspectratio]{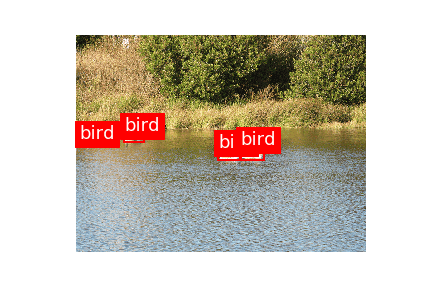}
\end{minipage}
\begin{minipage}{0.19\textwidth}
  \centering
  \includegraphics[height=3.2cm, width=\linewidth, keepaspectratio ]{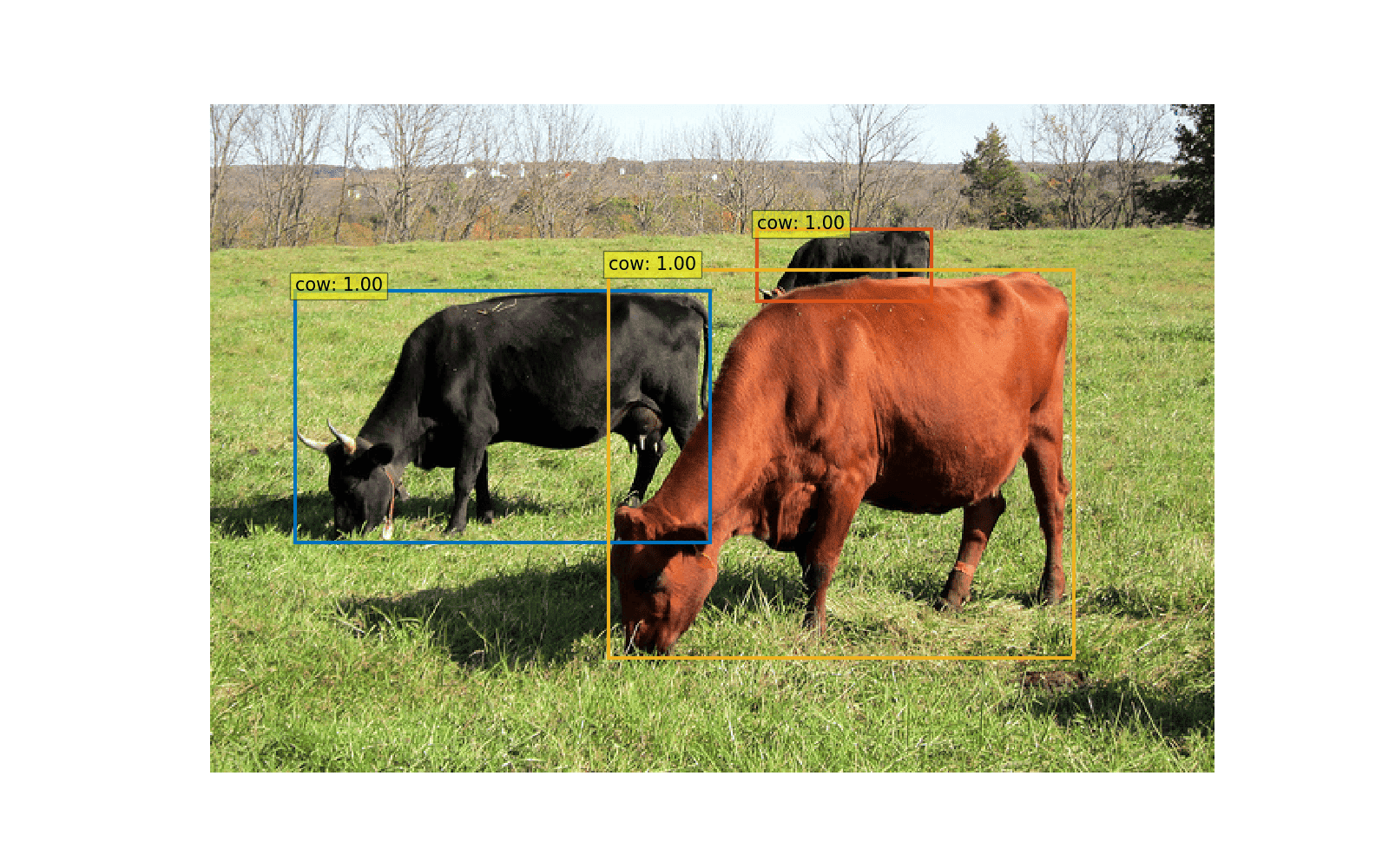}
\end{minipage}
\begin{minipage}{0.19\textwidth}
  \centering
  \includegraphics[height=3.2cm, width=\linewidth, keepaspectratio]{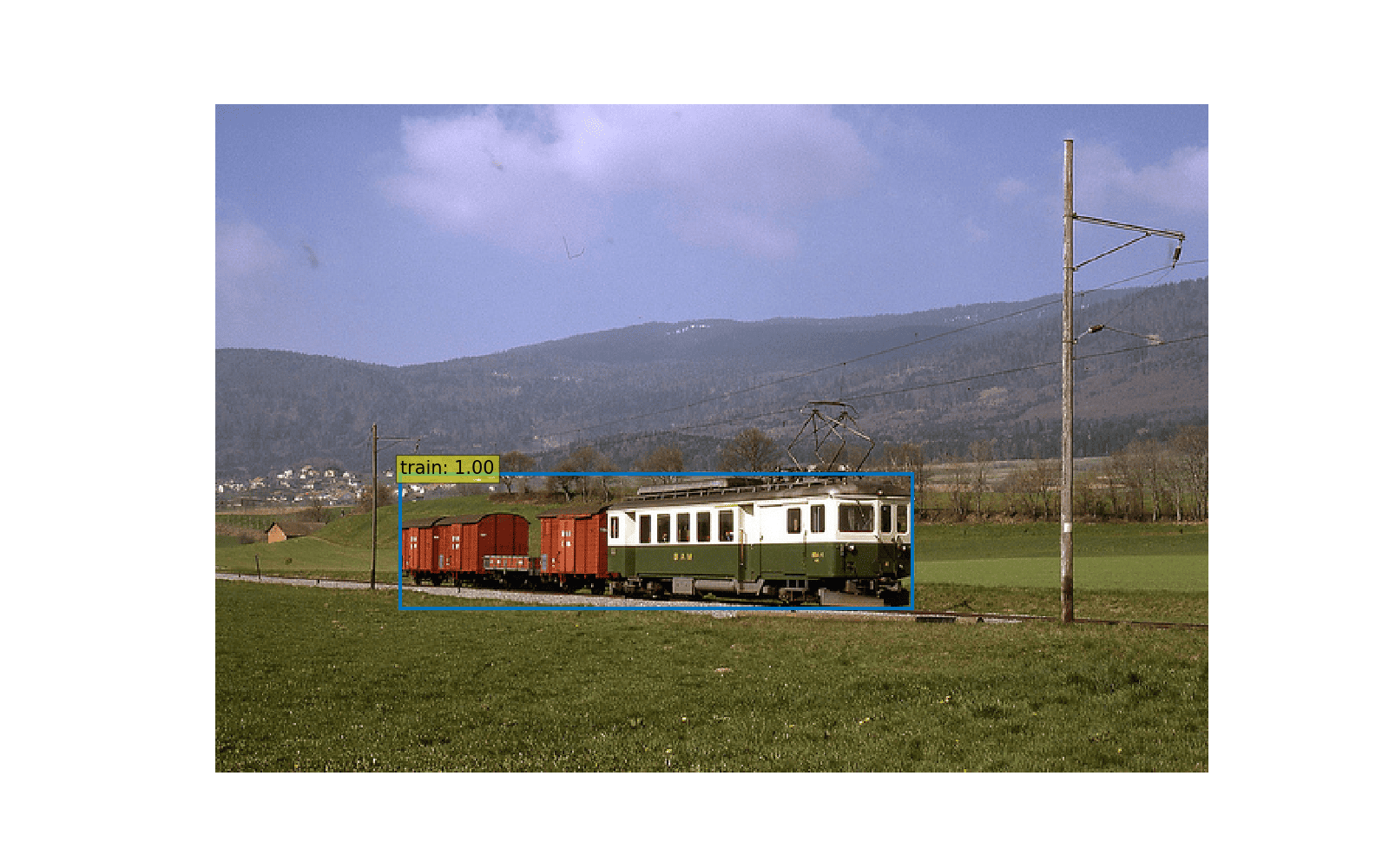}
\end{minipage}
\begin{minipage}{0.19\textwidth}
  \centering
  \includegraphics[height=3.2cm, width=\linewidth, keepaspectratio]{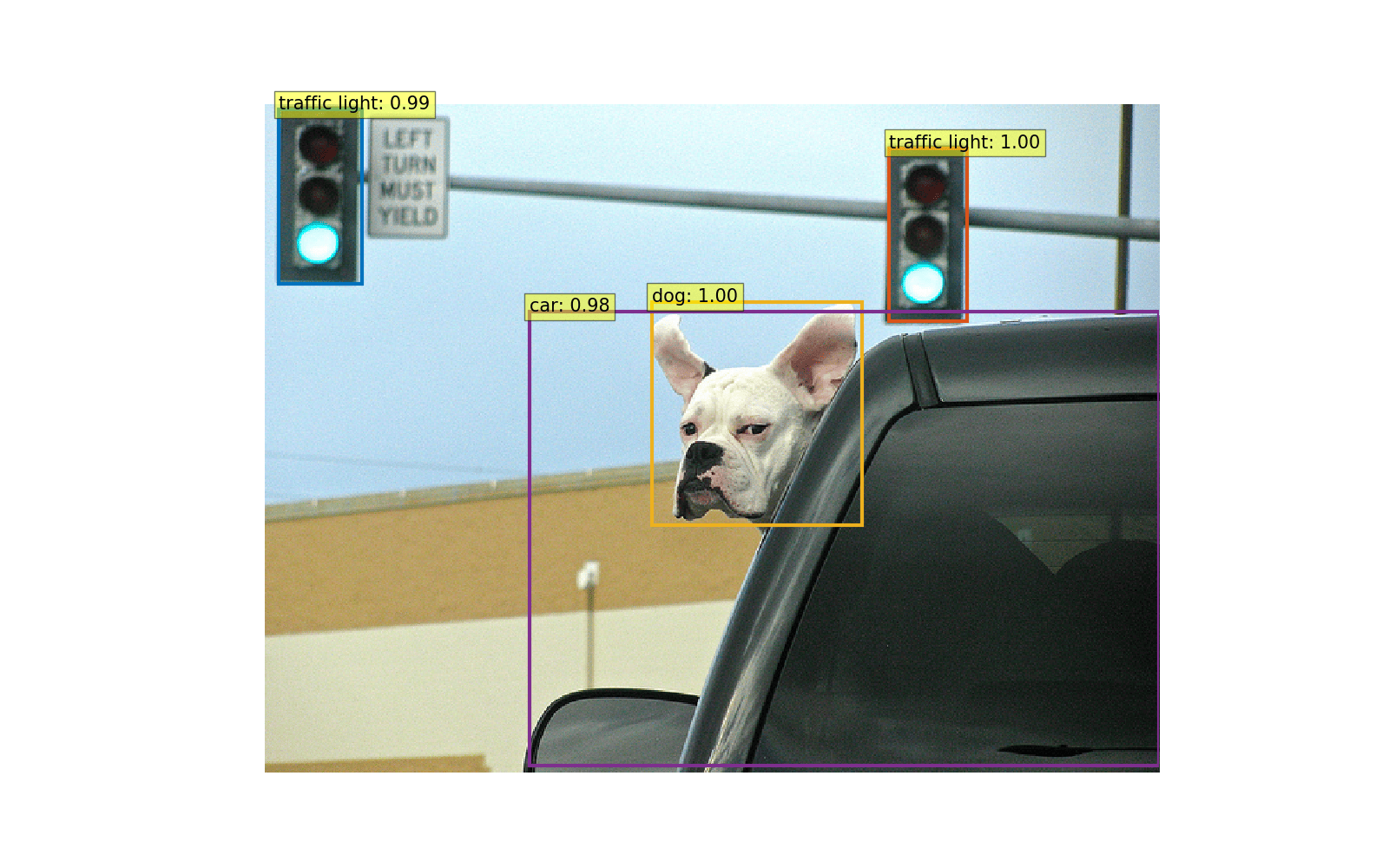}
\end{minipage}
\begin{minipage}{0.19\textwidth}
  \centering
  \includegraphics[height=3.2cm, width=\linewidth, keepaspectratio]{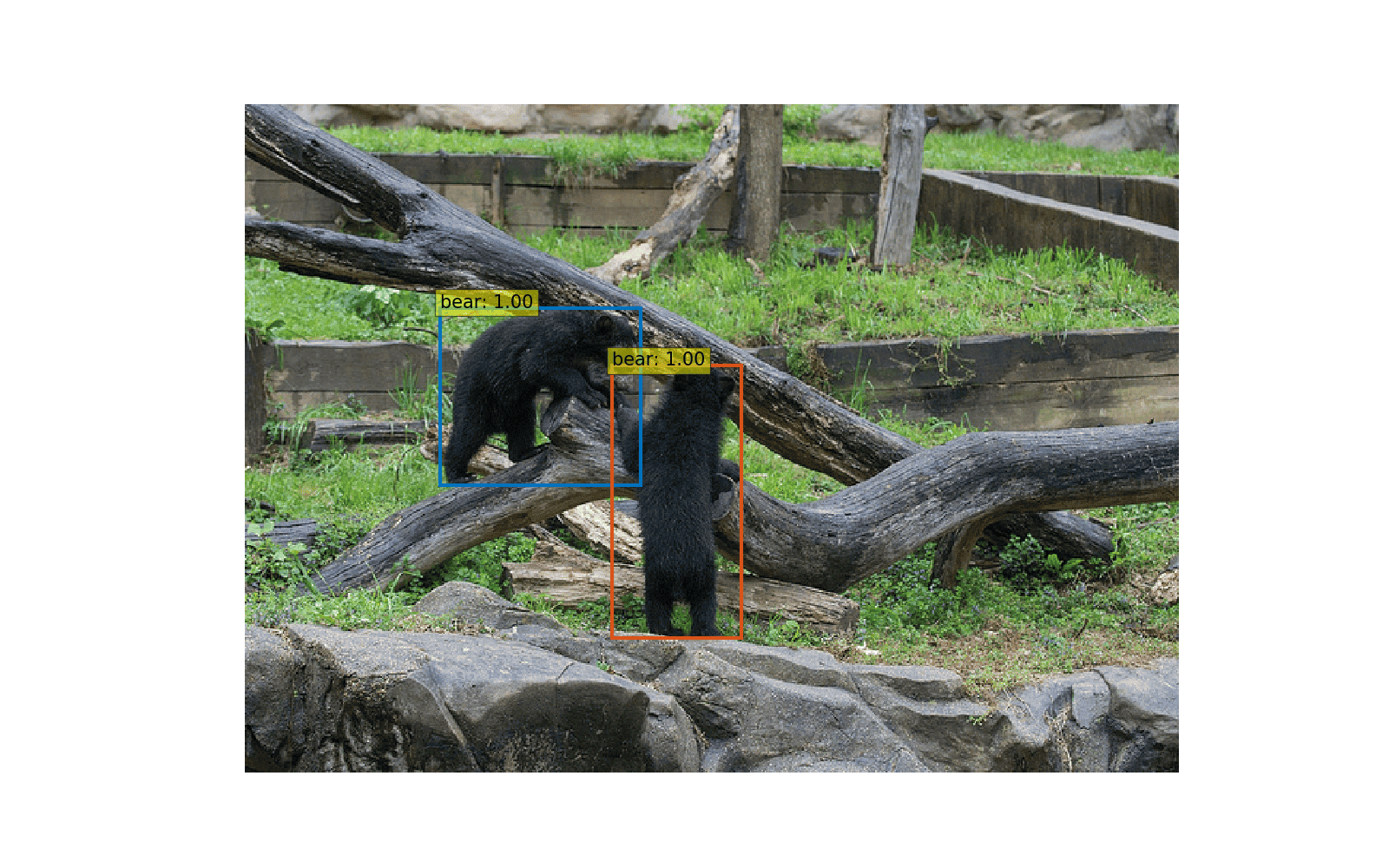}
\end{minipage}

\begin{minipage}{0.19\textwidth}
  \centering
  \includegraphics[height=3.2cm, width=\linewidth , keepaspectratio]{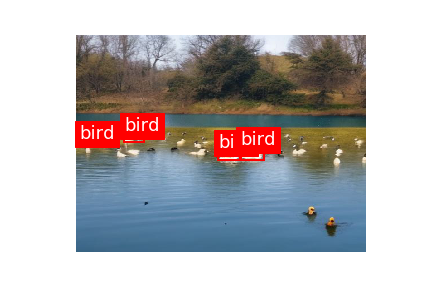}
\end{minipage}
\begin{minipage}{0.19\textwidth}
  \centering
  \includegraphics[height=3.2cm, width=\linewidth, keepaspectratio ]{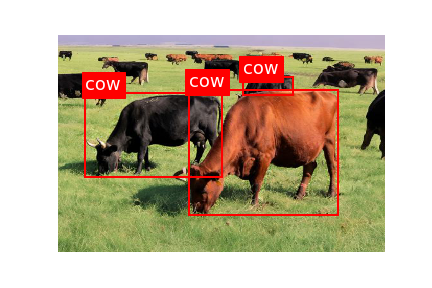}
\end{minipage}
\begin{minipage}{0.19\textwidth}
  \centering
  \includegraphics[height=3.2cm, width=\linewidth, keepaspectratio]{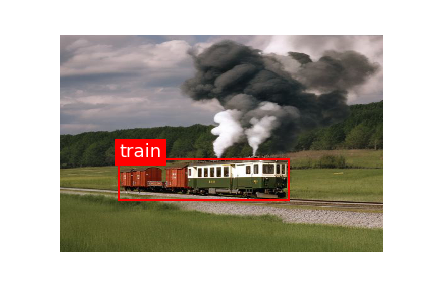}
\end{minipage}
\begin{minipage}{0.19\textwidth}
  \centering
  \includegraphics[height=3.2cm, width=\linewidth, keepaspectratio]{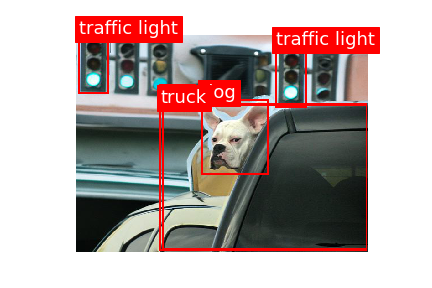}
\end{minipage}
\begin{minipage}{0.19\textwidth}
  \centering
  \includegraphics[height=3.2cm, width=\linewidth, keepaspectratio]{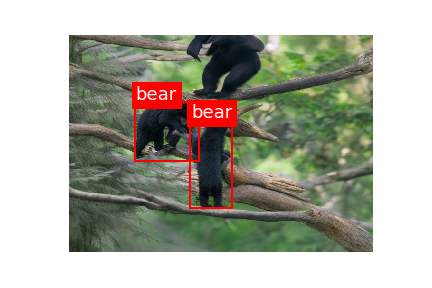}
\end{minipage}

\end{minipage}
\hfill
  \caption{ We use diverse prompts to capture the diverse background shifts on samples from \texttt{COCO-DC}. The figure illustrate a comparison of prediction of Mask-RCNN on both clean and generated samples on \texttt{COCO-DC}.  Each two adjacent rows represents the prediction of Mask-RCNN on clean \textit{(top)} and  generated images  \textit{(bottom)}. }
  \label{fig:diversity_detection}
\end{figure*}

\newpage

\subsection{Effect of Background Change on Segmentation Models}
\label{sec:FastSAM}
\begin{figure}[h!]
    \centering
    \includegraphics[width=\textwidth, height=0.35\textheight, keepaspectratio]{
    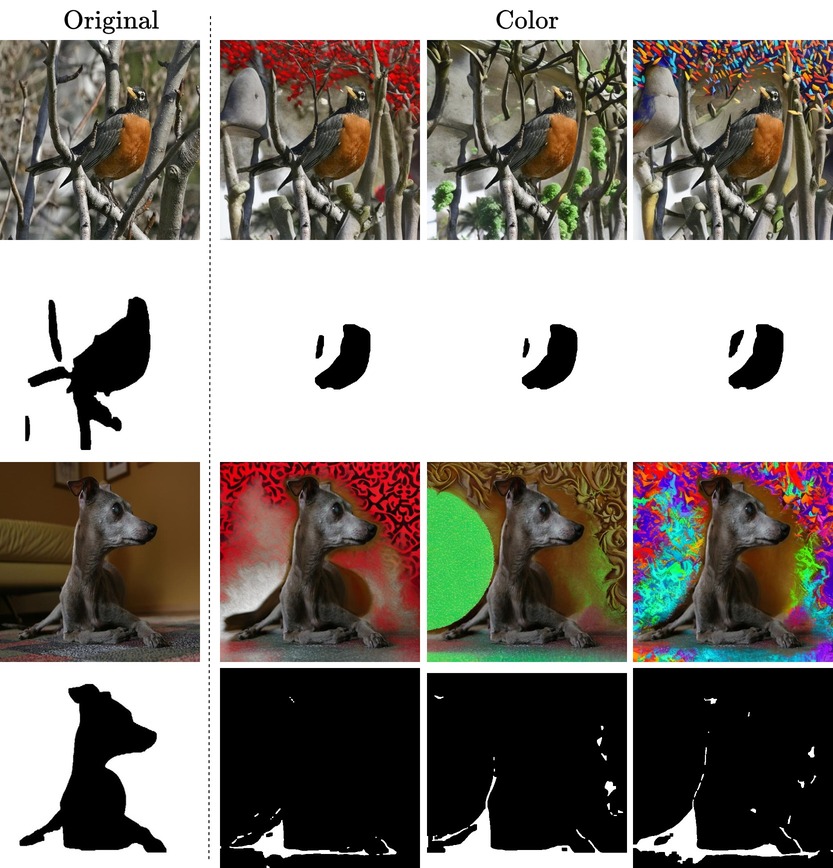
    }
    \caption{Instances illustrating FastSAM model's failure to accurately segment masks for the background color changes on \texttt{ImageNet-B} samples..}
    \label{sam_color-appendix}
    
\end{figure}

\begin{figure}[ht]
    \centering
    \includegraphics[width=\textwidth, height=0.35\textheight, keepaspectratio]{
    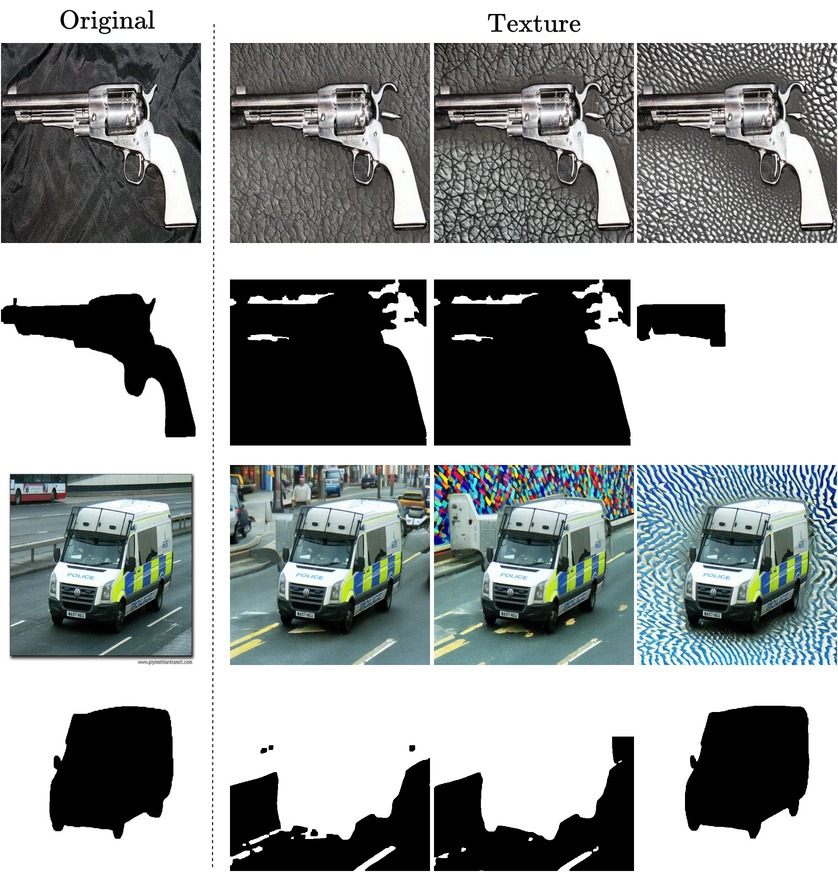
    }
    \caption{Instances illustrating FastSAM model's failure to accurately segment masks for the background texture changes on \texttt{ImageNet-B} samples.}
    \label{sam_texture-appendix}
    
\end{figure}

\begin{figure}[h!]
    \centering
    \includegraphics[width=\textwidth, height=0.4\textheight, keepaspectratio]{
    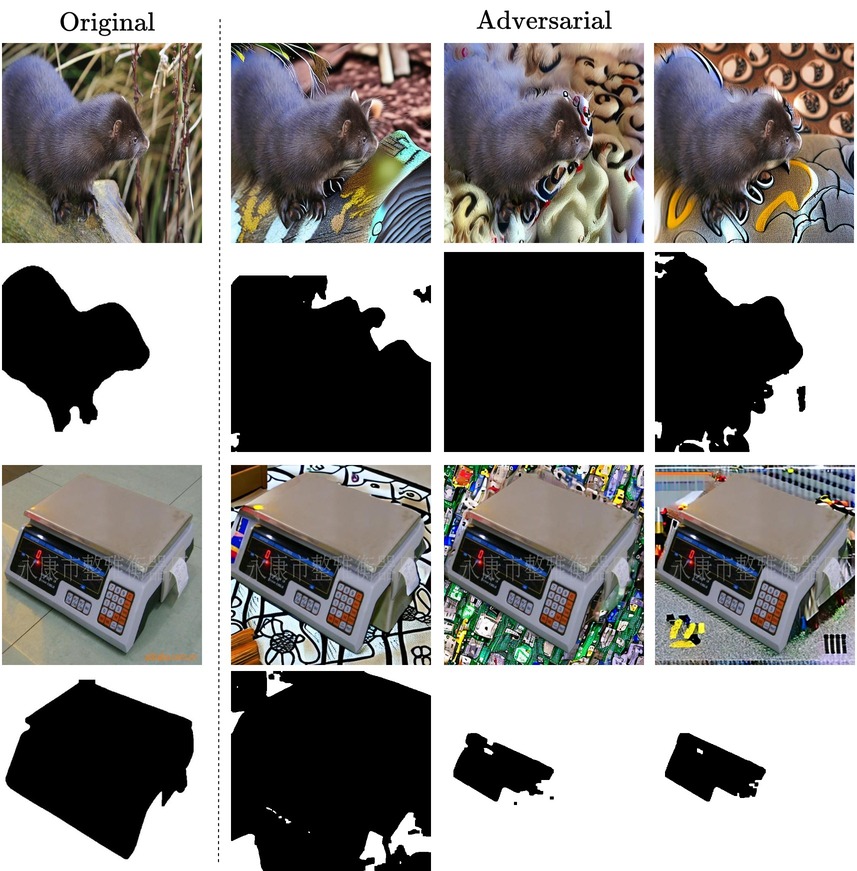
    }
    \caption{Instances illustrating FastSAM model's failure to accurately segment masks for the adversarial background changes on $\texttt{ImageNet-B}_{1000}$ samples.}
    \label{sam_adv-appendix}
    
\end{figure}

Figure \ref{sam_color-appendix}, \ref{sam_texture-appendix}, and \ref{sam_adv-appendix} provide failure cases of FastSAM to correctly segment the object in the images where background has been changed in terms of color, texture, and adversarial, respectively. Since we obtain the object masks for \texttt{ImageNet-B} using FastSAM, we compare those masks using IoU with the ones obtained by FastSAM on the generated dataset (see Table \ref{tab:sam}).

\begin{table}[h]
\begin{minipage}{0.60\textwidth}
  \centering
    \caption{\small IoU distribution of FastSAM. Percentage of images within an IoU range is reported.}
        \label{tab:sam}
    \scalebox{0.80}[0.80]{
    \begin{tabular}{lccccc}
      \toprule
      Background & 0.0-0.2 & 0.2-0.4 &0.4-0.6&0.6-0.8&0.8-1.0 \\
      \midrule
      Class Label         &  8.10   & 5.93& 8.02&13.03&       64.92               \\
      BLIP-2 Caption        & 5.70        & 4.81 & 6.92 &13.01 &69.56               \\
      Color               & 1.65 &1.39 &2.31 &4.99           & 89.65               \\
      Texture             & 2.11            & 1.02 &1.78&4.07&91.02       \\
      \midrule
      Adversarial         & 4.87 &2.91 &4.32 &10.63   & 77.27                  \\
      \midrule
    \end{tabular}
    }
    \end{minipage}%
\hfill
\begin{minipage}{0.35\textwidth}
\caption{\small DETR Object detection evaluation on \texttt{COCO-DC}}
\label{tab:obj}
\scalebox{0.8}[0.8]{
    \begin{tabular}{l|c|c}
      \toprule
      Background & Box AP & Recall AR \\
      \midrule
      Original            & 0.65            & 0.81               \\
      BLIP-2 Caption       & 0.53            & 0.76               \\
      Color               & 0.52            & 0.73               \\
      Texture             & 0.52            & 0.71               \\
      \bottomrule
      Adversarial         & 0.42            & 0.62                  \\
      \midrule
    \end{tabular}}
  \end{minipage}
\end{table}

\newpage
\FloatBarrier
\newpage

\subsection{Exploring Feature Space of Vision Models}
\label{sec:feature space}
\begin{figure}[h]
    \centering
    \includegraphics[width=\textwidth]{
    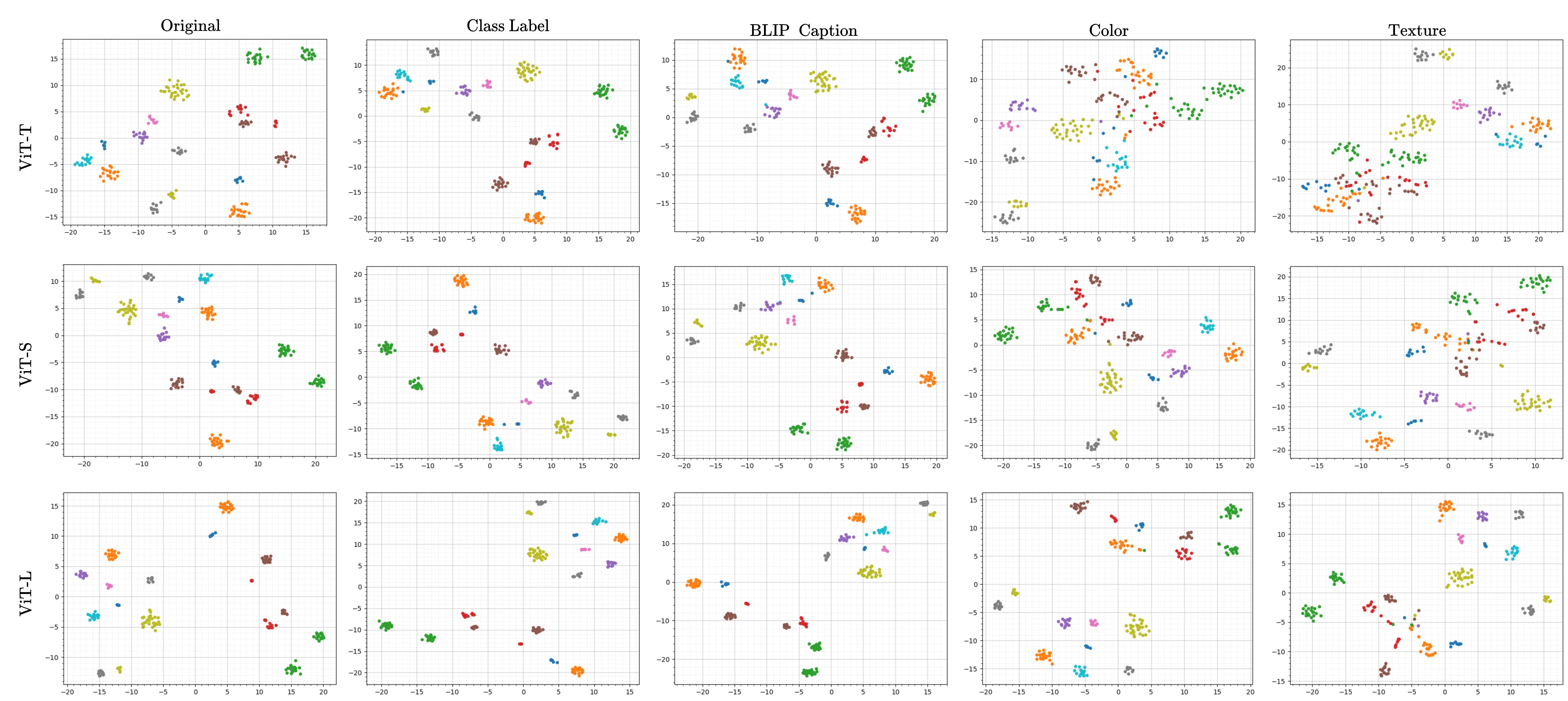
    }
    \caption{t-SNE visualization of classifier models on \texttt{ImageNet-B} dataset.}
    \label{tsne-classifier}

\end{figure}

\begin{figure}[h]
    \centering
    \includegraphics[width=\textwidth]{
    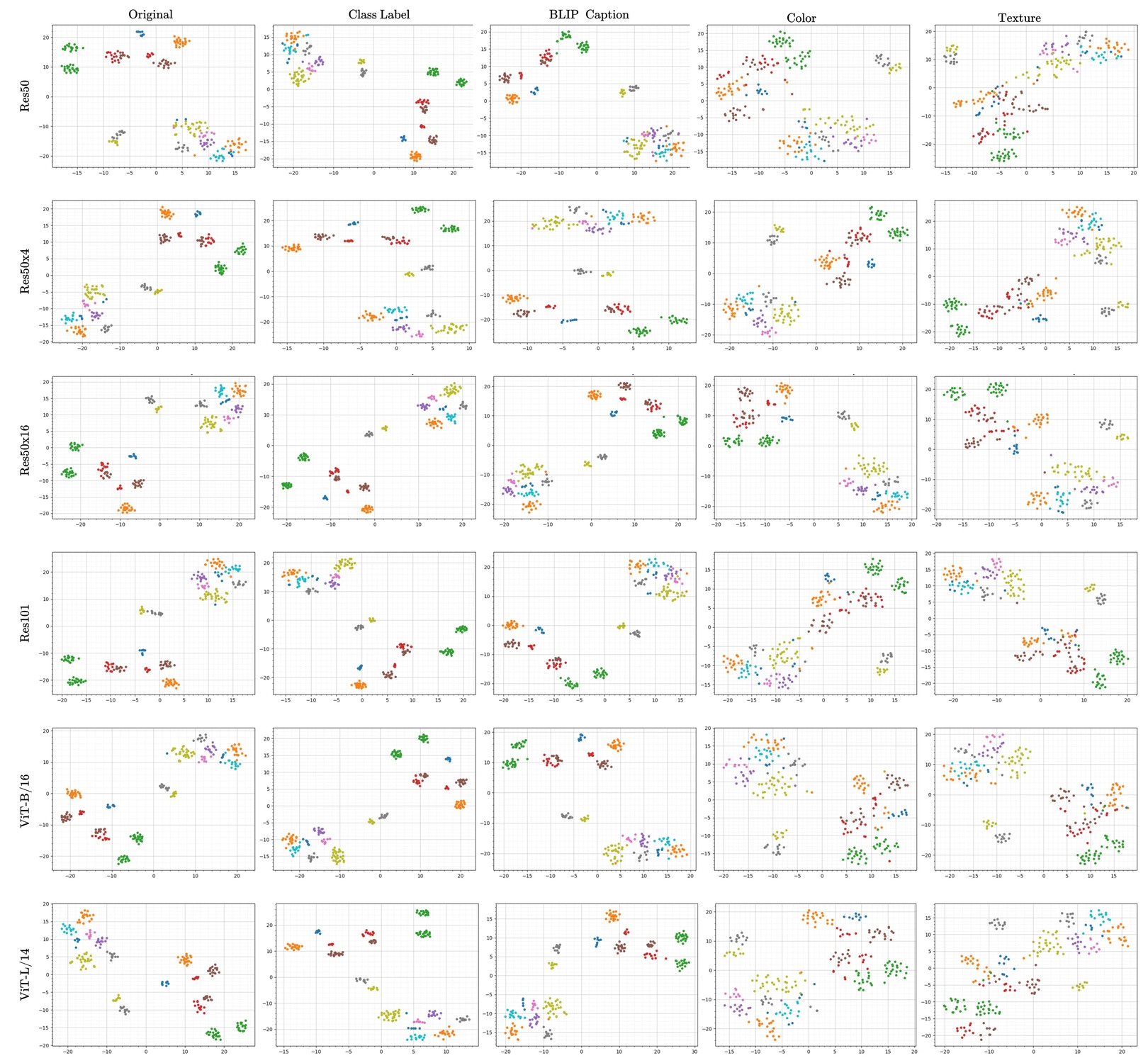
    }
    \caption{t-SNE visualization of CLIP Vision Encoder features on \texttt{ImageNet-B} dataset.}
    \label{tsne-vision}

\end{figure}

\begin{figure}[ht]
    \centering
   \includegraphics[width=\textwidth, height=0.35\textheight, keepaspectratio]{
    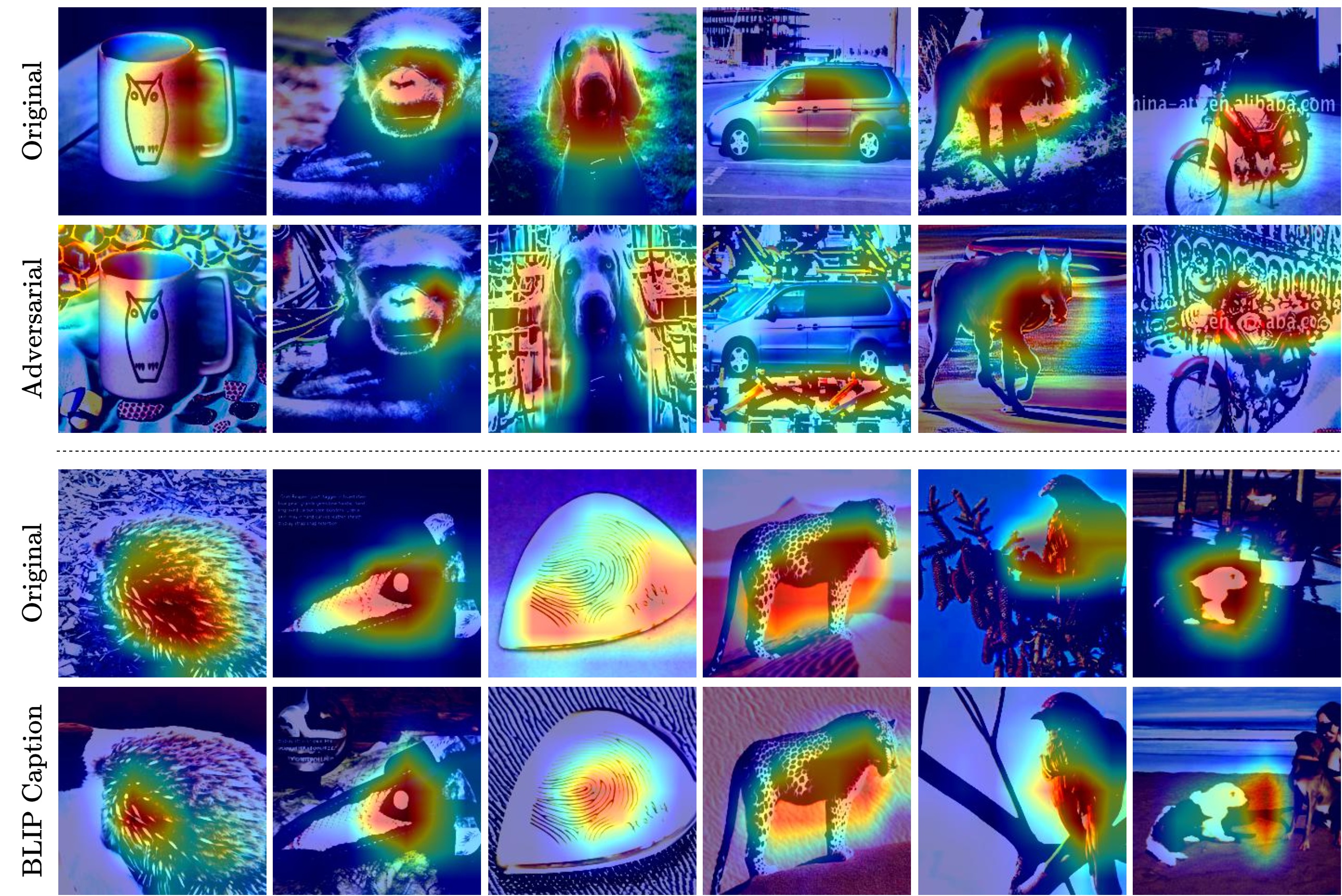
    }
    \caption{GradCAM \cite{Selvaraju_2019} visualization of adversarial and BLIP-2 background examples. The activation maps were generated on ImageNet pre-trained Res-50 model.}
    \label{grad-cam} 
\end{figure}

\begin{figure}[h!]
    \centering
    \includegraphics[width=\textwidth, height=0.35\textheight, keepaspectratio]{
    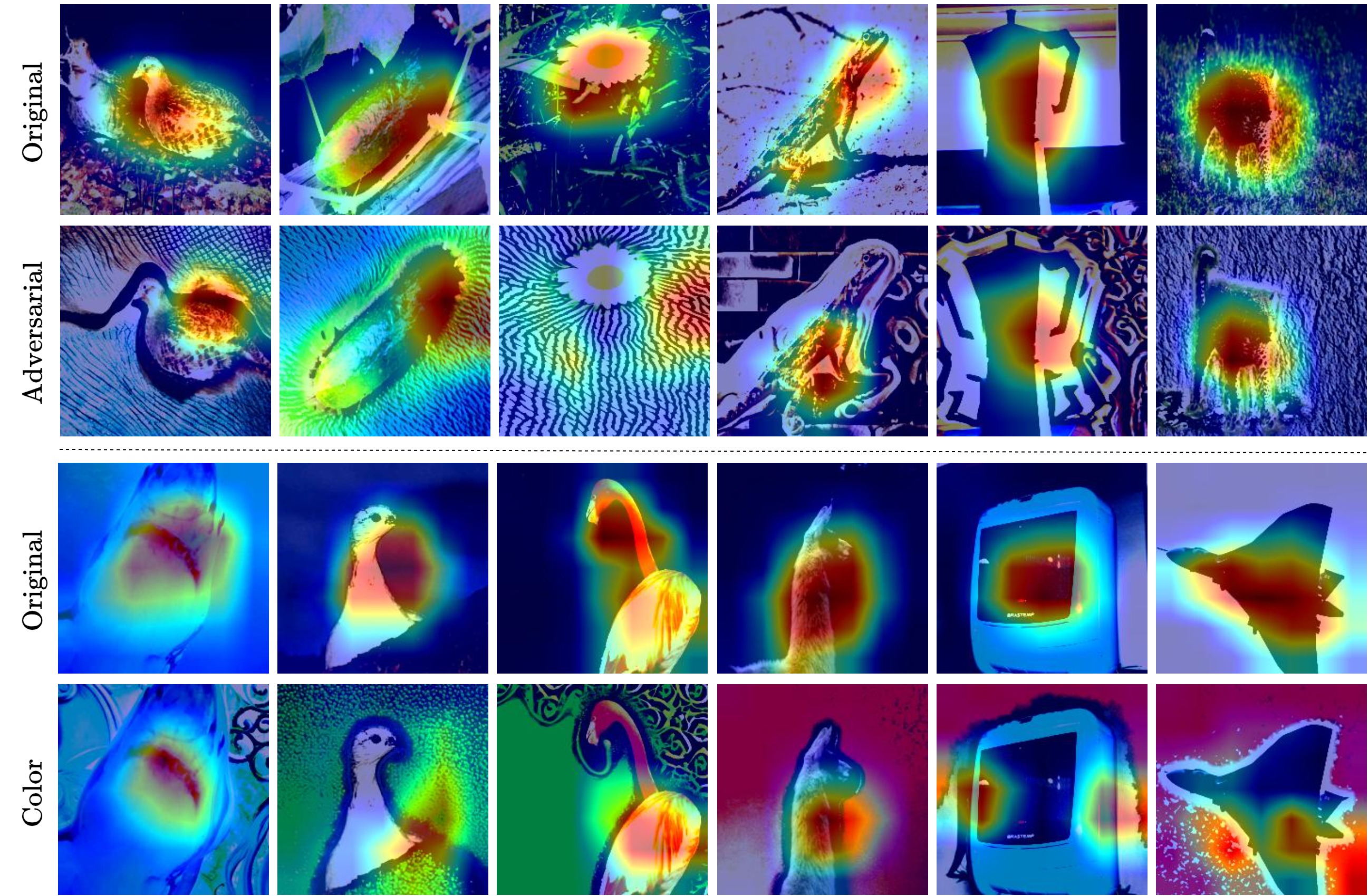}
    \caption{GradCAM \cite{Selvaraju_2019} visualization of  texture and color background changes. The activation maps were generated on ImageNet pre-trained Res-50 model.}
    \label{grad-cam-2}
\end{figure}

In Figure \ref{tsne-classifier} and \ref{tsne-vision}, we explore the visual feature space of vision and vision language model using t-SNE visualizations. We observe that as the background changes deviate further from the original background, a noticeable shift occurs in the feature space. The distinct separation or clustering of features belonging to the same class appears to decrease. This observation suggests a significant correlation between the model's decision-making process and the alterations in the background. Furthermore, we also show the GradCAM \cite{Selvaraju_2019} on generated background changes. We observe that diverse background changes significantly shift the attention of the model as can be seen from Figure \ref{grad-cam} and \ref{grad-cam-2}.

\newpage

\newpage

\newpage

\FloatBarrier
\newpage
\subsection{Diversity and Diffusion parameter ablation}
\label{sec:diversity}
\begin{figure}[h]
    \centering
    \includegraphics[width=\textwidth]{
    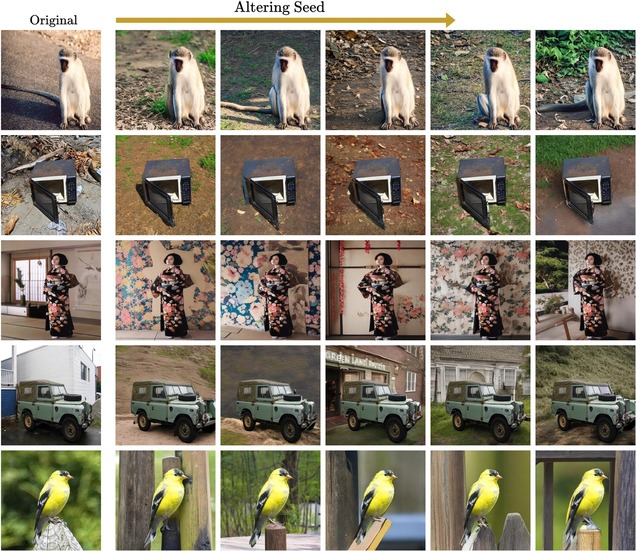
    }
    \caption{In this figure, examples are generated using BLIP-2 captions by altering the seed from left to right in the row. This highlights the high diversity achievable with the diffusion model when employing different starting noise latents.}
    \label{seed-appendix}
    
\end{figure}

\begin{figure}
    \centering
    \includegraphics[width=\textwidth]{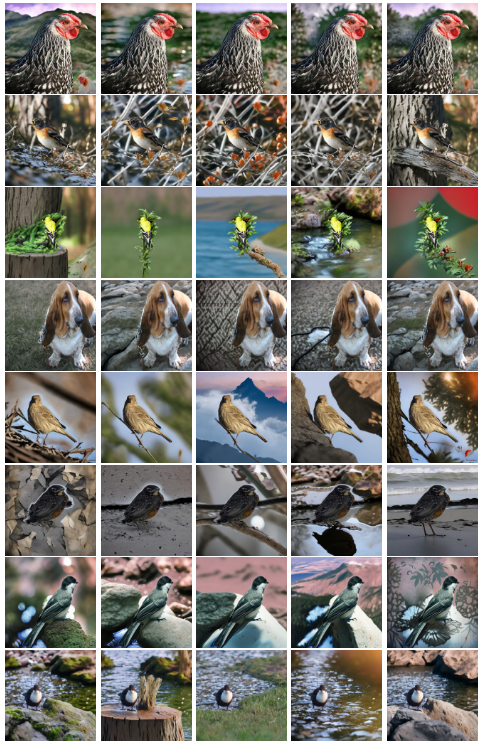}
    \caption{Using diverse prompts to capture for diverse background shifts on samples from \texttt{ImageNet-B}.}
    \label{fig:diversity}
\end{figure}

\begin{figure}
    \centering
    \includegraphics[width=\textwidth]{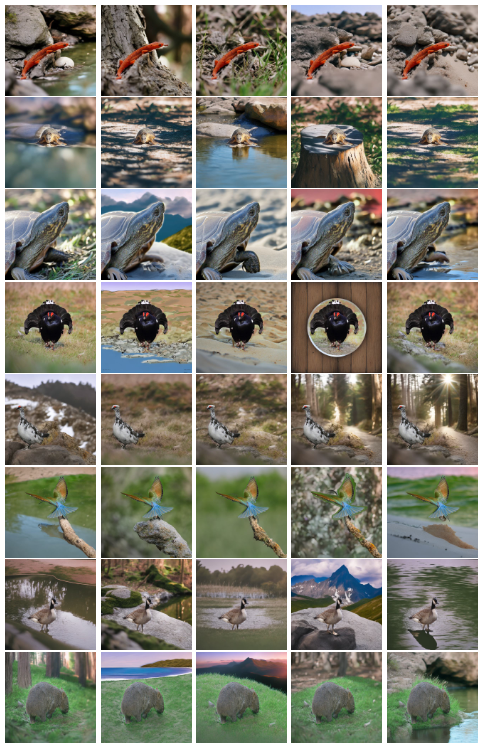}
    \caption{Using diverse prompts to capture for diverse background shifts on samples from \texttt{ImageNet-B}.}
    \label{fig:diversity_2}
\end{figure}

\begin{figure}[h]
    \centering
    \includegraphics[width=\textwidth]{
    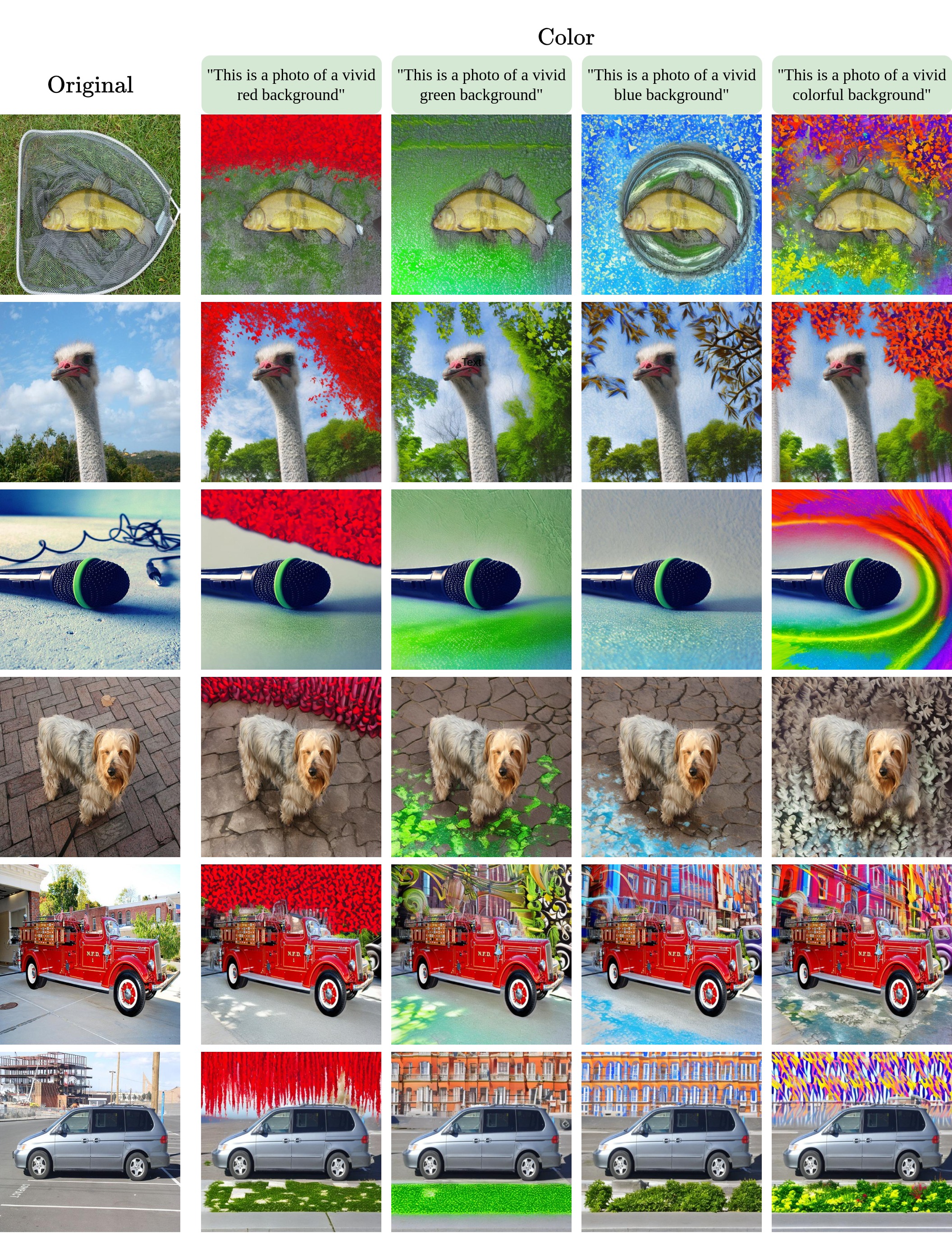
    }
    \caption{Images generated through diverse color prompts on \texttt{ImageNet-B}.}
    \label{color-appendix}
    
\end{figure}

\begin{figure}[h]
    \centering
    \includegraphics[width=\textwidth]{
    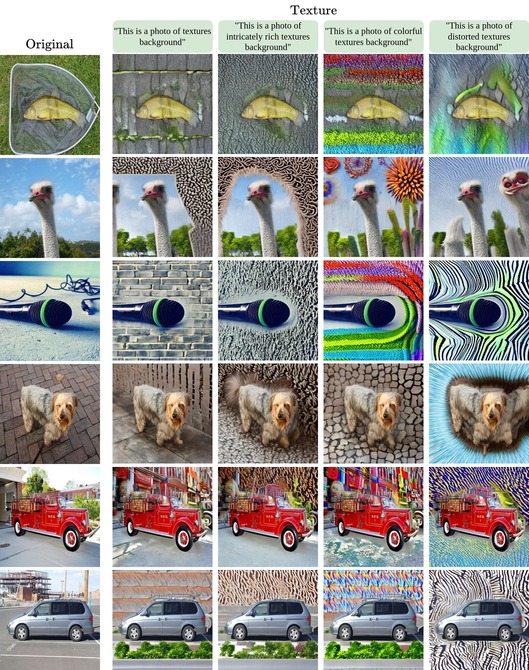
    }
    \caption{Images generated through diverse texture prompts on \texttt{ImageNet-B}.}
    \label{texture-appendix}
    
\end{figure}

\begin{figure}[h]
    \centering
    \includegraphics[width=\textwidth]{
    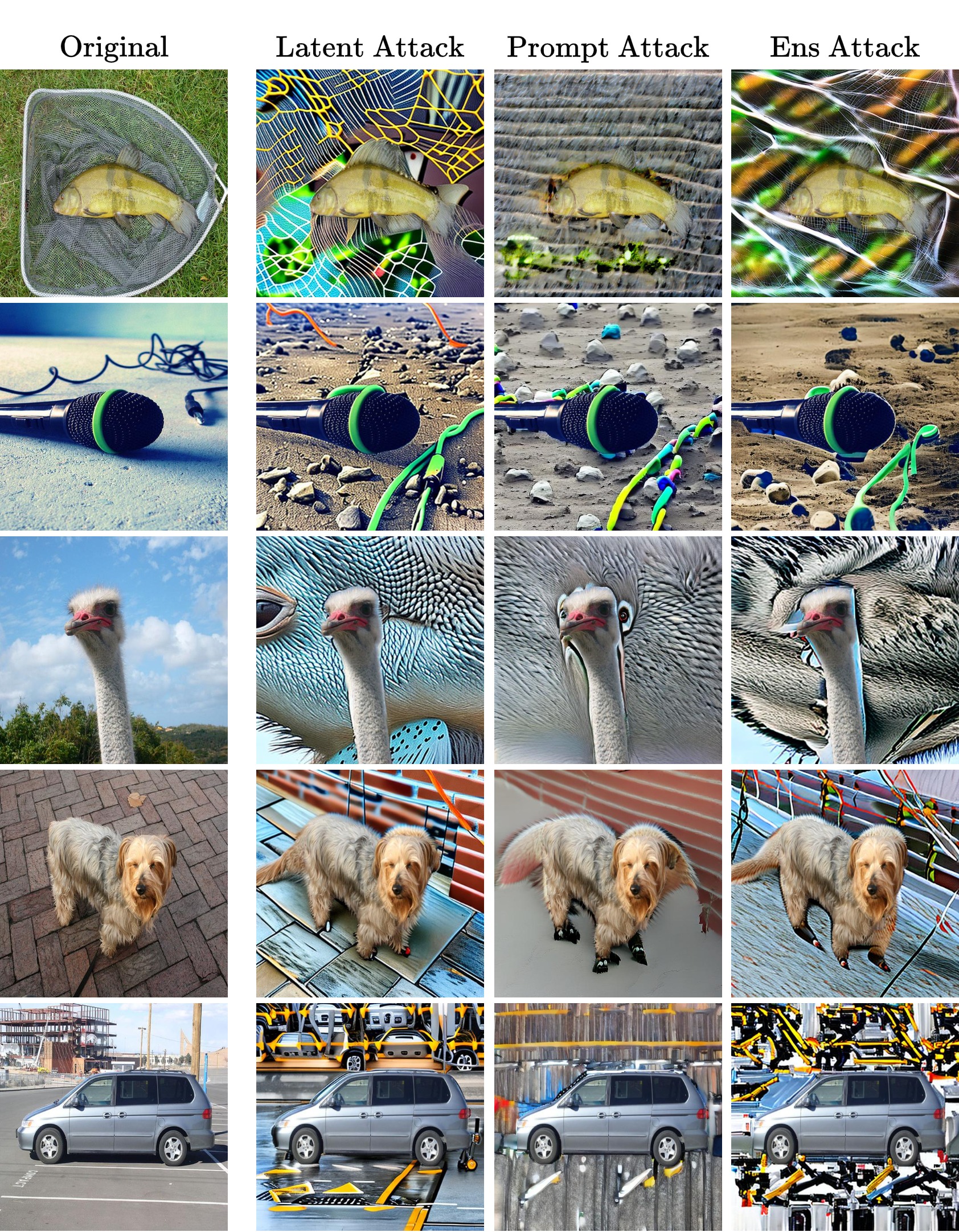
    }
    \caption{Images generated under various attack scenarios on $\texttt{ImageNet-B}_{1000}$. Here we show the visualization for latent, prompt, and ensemble attack that are generated by optimizing latent, text prompt embeddings, and both respectively.}
    \label{attack-appendix}
    
\end{figure}

\begin{figure}[h]
    \centering
    \includegraphics[width=\textwidth]{
    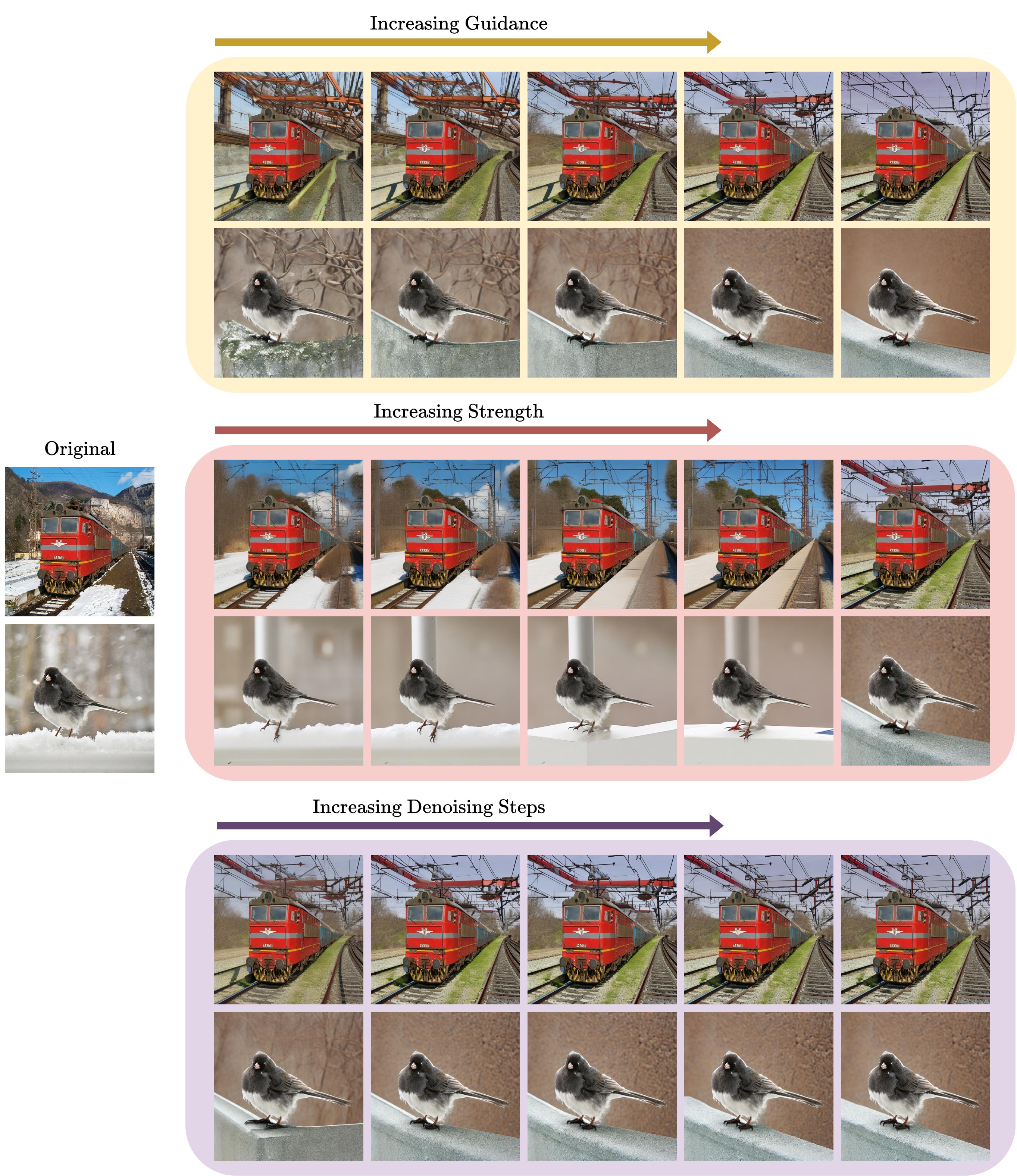
    }
    \caption{Visualization on samples taken from \texttt{ImageNet-B}. Varying parameters like guidance, strength, and denoising steps while using BLIP-2 caption as the prompt. Increasing guidance leads to more fine-detailed background changes. Additionally, greater strength correlated with more pronounced alterations from the original background. And, augmenting diffusion steps improves image quality.}
    \label{blip-abla}
\end{figure}

In this section, we qualitatively analyze the diversity in visual results of the diffusion model. In Figure \ref{seed-appendix}, we show that keeping textual and visual guidance fixed, the diffusion model is still able to generate diverse changes with similar background semantics at different seeds for the noise $z_T$. Furthermore, we explore the diversity in generating realistic background changes across an original image by using diverse class agnostic textual prompts, capturing different realistic backgrounds. Figure \ref{fig:diversity} and \ref{fig:diversity_2} show some of the qualitative results obtained on \texttt{ImageNet-B} samples using prompts generated from ChatGPT). 
Furthermore, we show the visual examples of color, texture, and adversarial attack on \texttt{ImageNet-B} dataset in Figure \ref{color-appendix}, \ref{texture-appendix}, and \ref{attack-appendix}. We also provide a visualization in Figure \ref{blip-abla} showing the effect of changing diffusion model parameters.

\newpage

\newpage

\newpage

\newpage

\FloatBarrier

\newpage

\FloatBarrier

\newpage

\subsection{Misclassified Samples}
\label{sec:misclassified}
\begin{figure}[h]
    \centering
    \includegraphics[width=\textwidth]{
    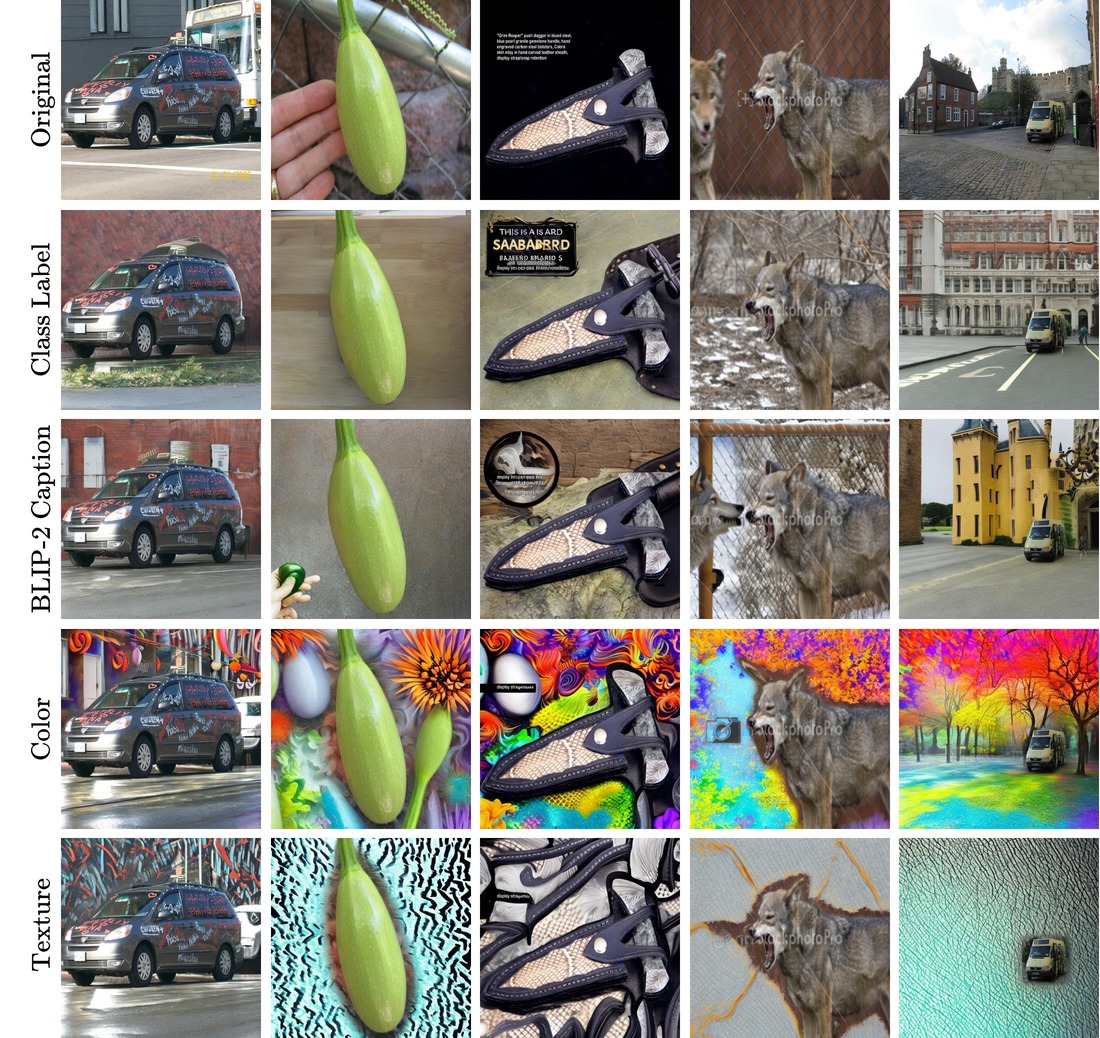
    }
    \caption{Images  misclassified by Res-50 across different background changes}
    \label{missclassified}
\end{figure}

\begin{figure*}[h]
\begin{minipage}{\textwidth}

\centering

\begin{minipage}{0.19\textwidth}
  \centering
  \includegraphics[height=2.4cm, width=\linewidth , keepaspectratio]{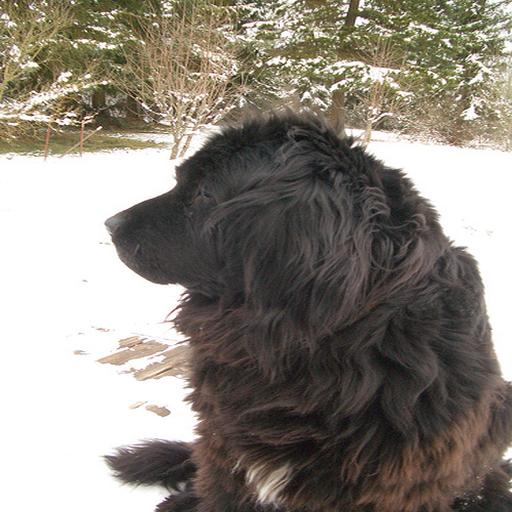}
\end{minipage}
\begin{minipage}{0.19\textwidth}
  \centering
  \includegraphics[height=2.4cm, width=\linewidth, keepaspectratio ]{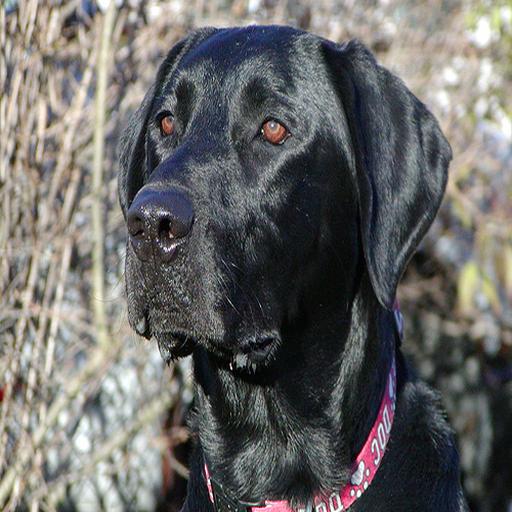}
\end{minipage}
\begin{minipage}{0.19\textwidth}
  \centering
  \includegraphics[height=2.4cm, width=\linewidth, keepaspectratio]{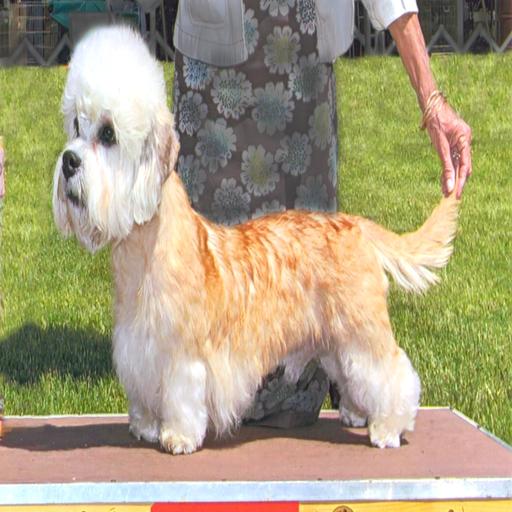}
\end{minipage}
\begin{minipage}{0.19\textwidth}
  \centering
  \includegraphics[height=2.4cm, width=\linewidth, keepaspectratio]{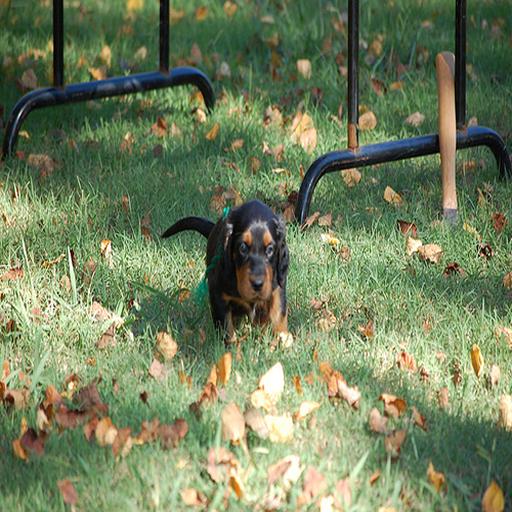}
\end{minipage}
\begin{minipage}{0.19\textwidth}
  \centering
  \includegraphics[height=2.4cm, width=\linewidth, keepaspectratio]{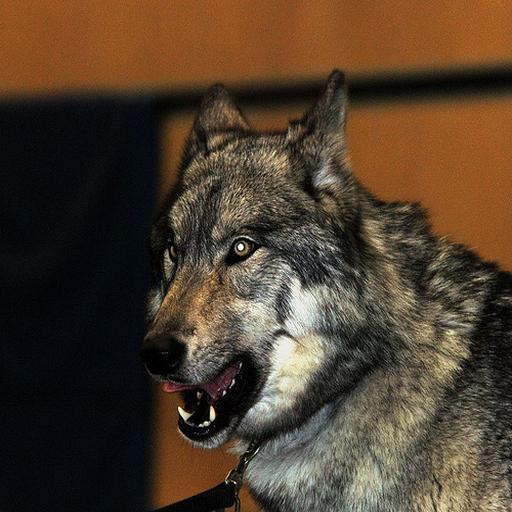}
\end{minipage}


\begin{minipage}{0.19\textwidth}
  \centering
  \includegraphics[height=2.4cm, width=\linewidth , keepaspectratio]{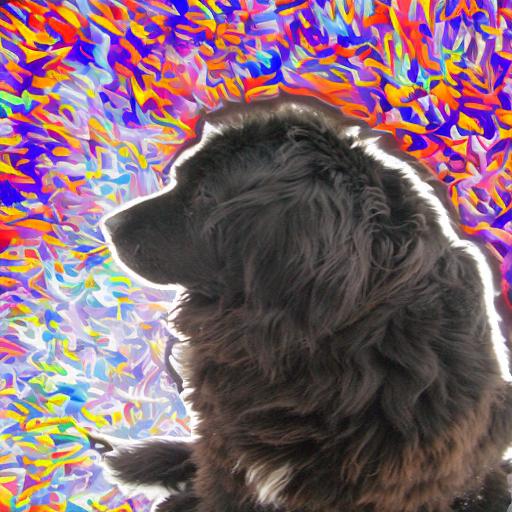}
\end{minipage}
\begin{minipage}{0.19\textwidth}
  \centering
  \includegraphics[height=2.4cm, width=\linewidth, keepaspectratio ]{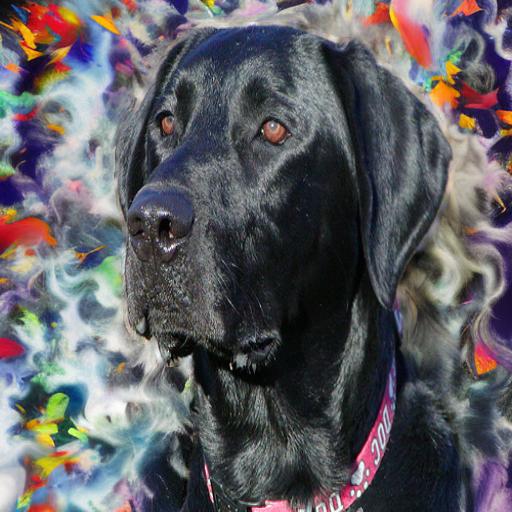}
\end{minipage}
\begin{minipage}{0.19\textwidth}
  \centering
  \includegraphics[height=2.4cm, width=\linewidth, keepaspectratio]{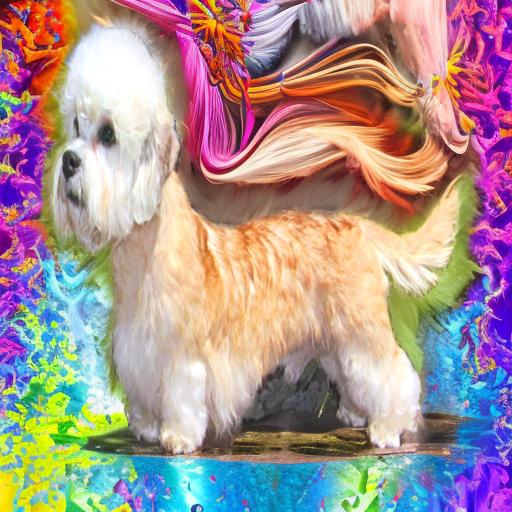}
\end{minipage}
\begin{minipage}{0.19\textwidth}
  \centering
  \includegraphics[height=2.4cm, width=\linewidth, keepaspectratio]{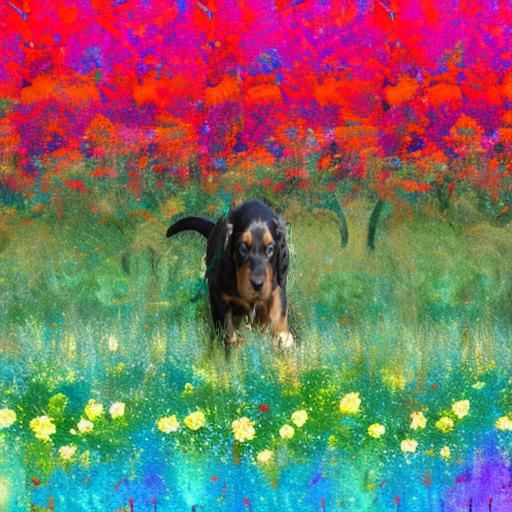}
\end{minipage}
\begin{minipage}{0.19\textwidth}
  \centering
  \includegraphics[height=2.4cm, width=\linewidth, keepaspectratio]{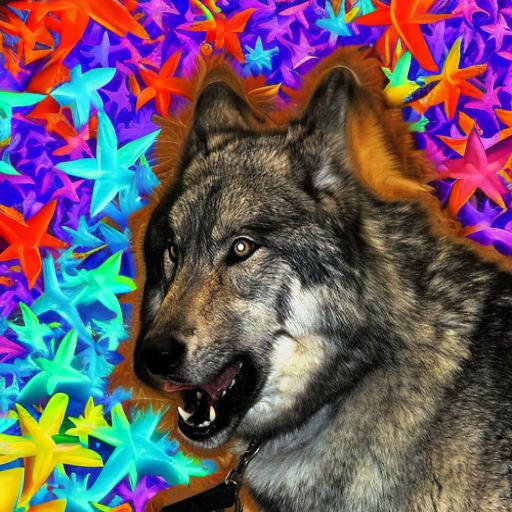}
\end{minipage}


\begin{minipage}{0.19\textwidth}
  \centering
  \includegraphics[height=2.4cm, width=\linewidth , keepaspectratio]{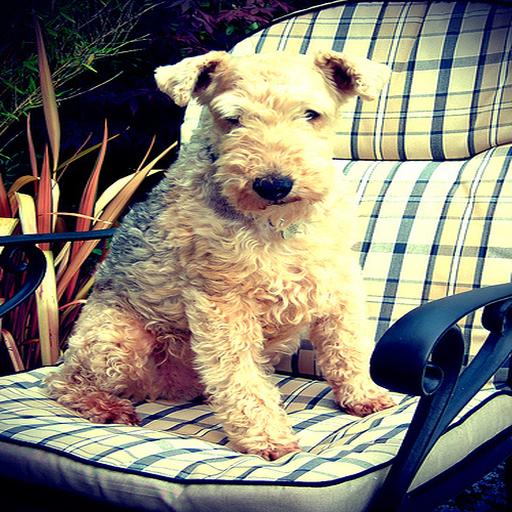}
\end{minipage}
\begin{minipage}{0.19\textwidth}
  \centering
  \includegraphics[height=2.4cm, width=\linewidth, keepaspectratio ]{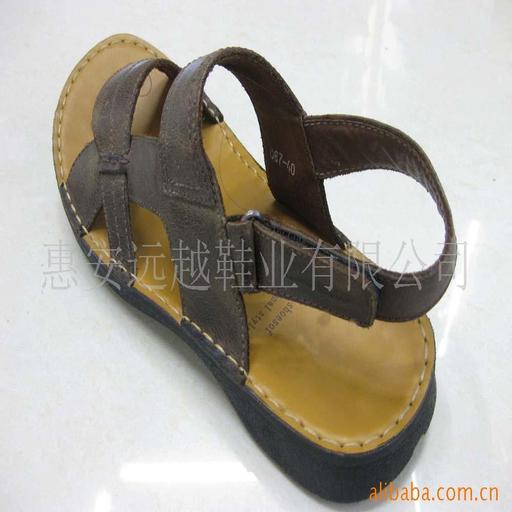}
\end{minipage}
\begin{minipage}{0.19\textwidth}
  \centering
  \includegraphics[height=2.4cm, width=\linewidth, keepaspectratio]{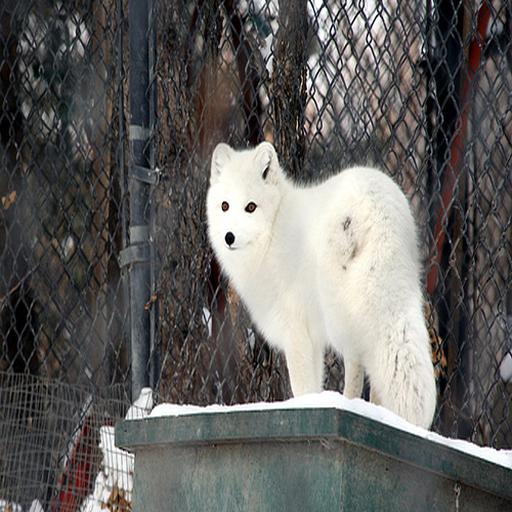}
\end{minipage}
\begin{minipage}{0.19\textwidth}
  \centering
  \includegraphics[height=2.4cm, width=\linewidth, keepaspectratio]{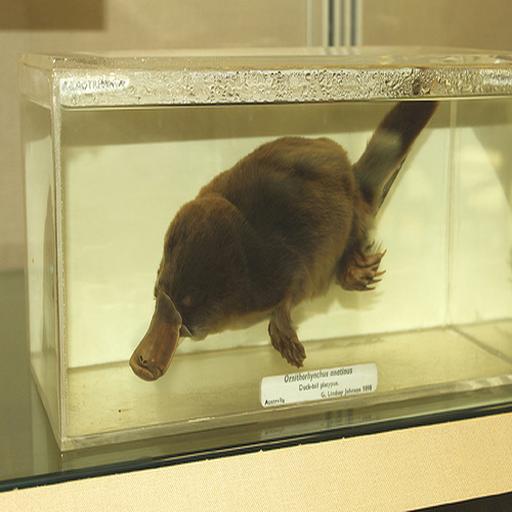}
\end{minipage}
\begin{minipage}{0.19\textwidth}
  \centering
  \includegraphics[height=2.4cm, width=\linewidth, keepaspectratio]{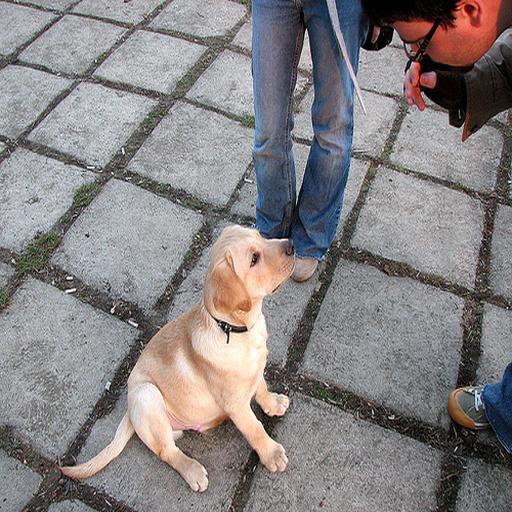}
\end{minipage}

\begin{minipage}{0.19\textwidth}
  \centering
  \includegraphics[height=2.4cm, width=\linewidth , keepaspectratio]{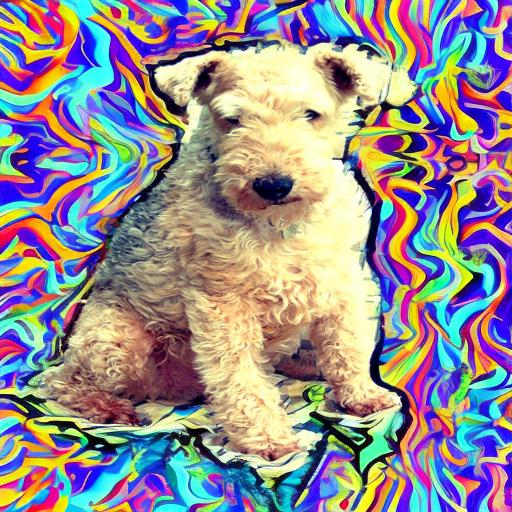}
\end{minipage}
\begin{minipage}{0.19\textwidth}
  \centering
  \includegraphics[height=2.4cm, width=\linewidth, keepaspectratio ]{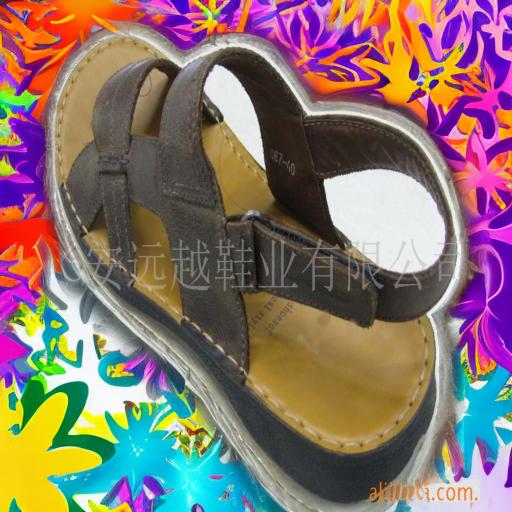}
\end{minipage}
\begin{minipage}{0.19\textwidth}
  \centering
  \includegraphics[height=2.4cm, width=\linewidth, keepaspectratio]{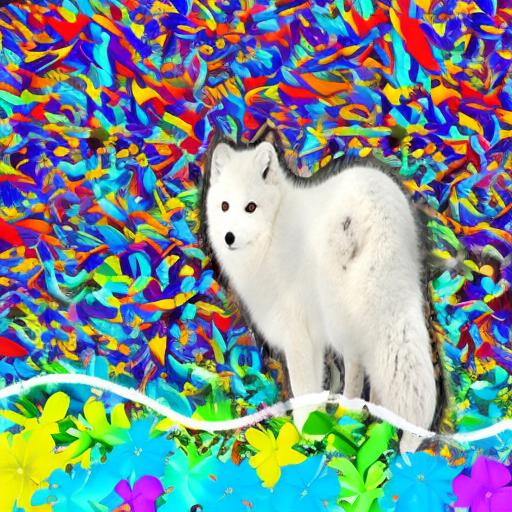}
\end{minipage}
\begin{minipage}{0.19\textwidth}
  \centering
  \includegraphics[height=2.4cm, width=\linewidth, keepaspectratio]{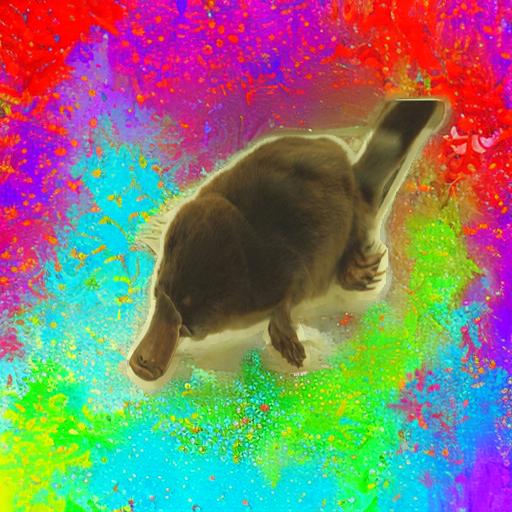}
\end{minipage}
\begin{minipage}{0.19\textwidth}
  \centering
  \includegraphics[height=2.4cm, width=\linewidth, keepaspectratio]{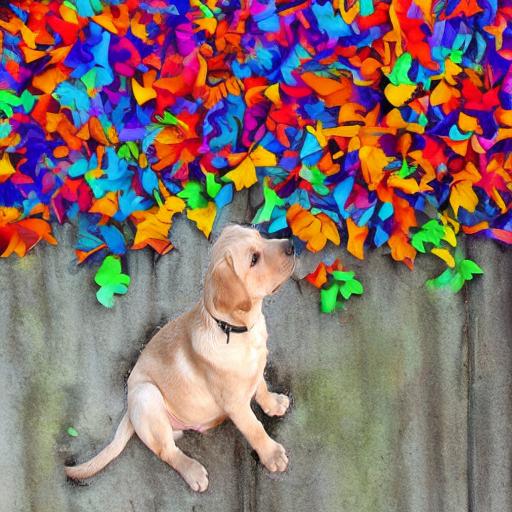}
\end{minipage}


\begin{minipage}{0.19\textwidth}
  \centering
  \includegraphics[height=2.4cm, width=\linewidth , keepaspectratio]{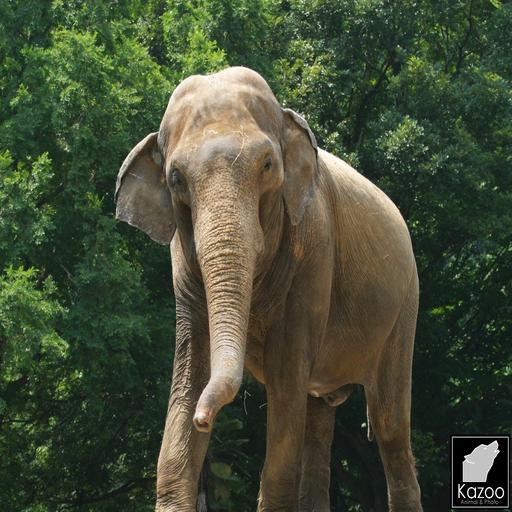}
\end{minipage}
\begin{minipage}{0.19\textwidth}
  \centering
  \includegraphics[height=2.4cm, width=\linewidth, keepaspectratio ]{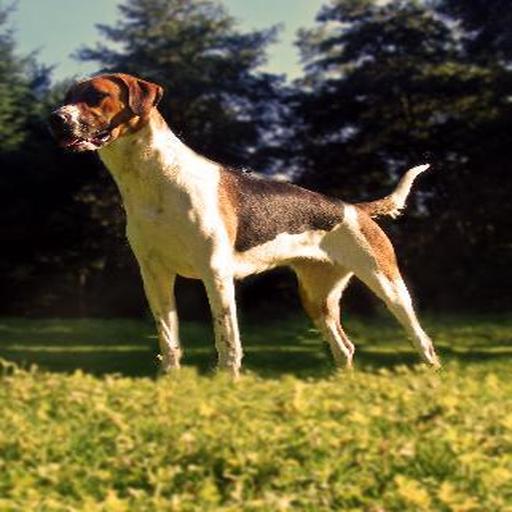}
\end{minipage}
\begin{minipage}{0.19\textwidth}
  \centering
  \includegraphics[height=2.4cm, width=\linewidth, keepaspectratio]{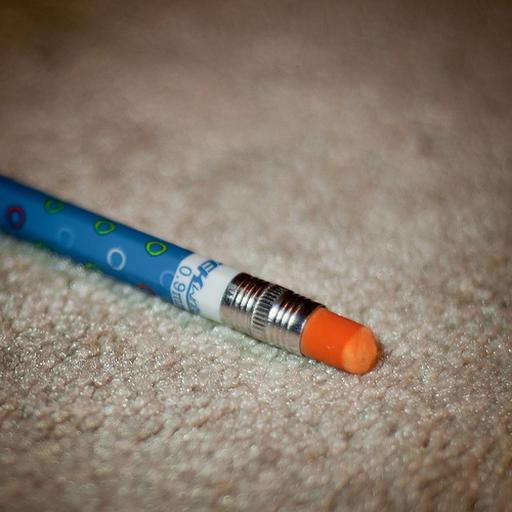}
\end{minipage}
\begin{minipage}{0.19\textwidth}
  \centering
  \includegraphics[height=2.4cm, width=\linewidth, keepaspectratio]{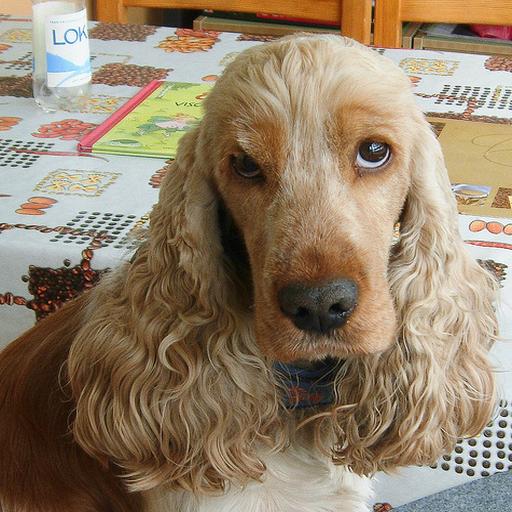}
\end{minipage}
\begin{minipage}{0.19\textwidth}
  \centering
  \includegraphics[height=2.4cm, width=\linewidth, keepaspectratio]{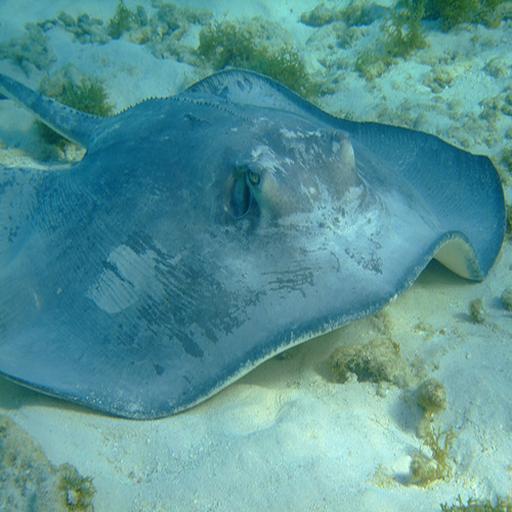}
\end{minipage}


\begin{minipage}{0.19\textwidth}
  \centering
  \includegraphics[height=2.4cm, width=\linewidth , keepaspectratio]{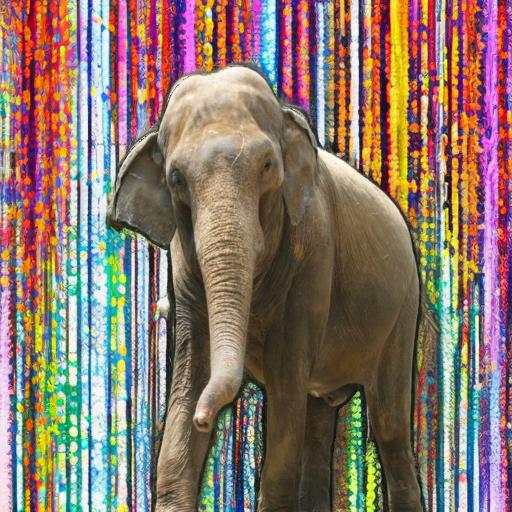}
\end{minipage}
\begin{minipage}{0.19\textwidth}
  \centering
  \includegraphics[height=2.4cm, width=\linewidth, keepaspectratio ]{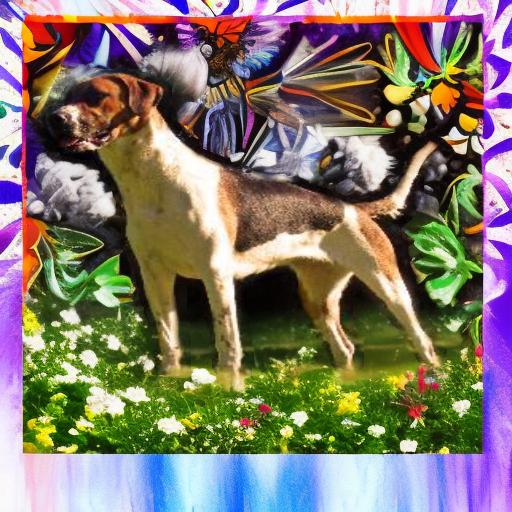}
\end{minipage}
\begin{minipage}{0.19\textwidth}
  \centering
  \includegraphics[height=2.4cm, width=\linewidth, keepaspectratio]{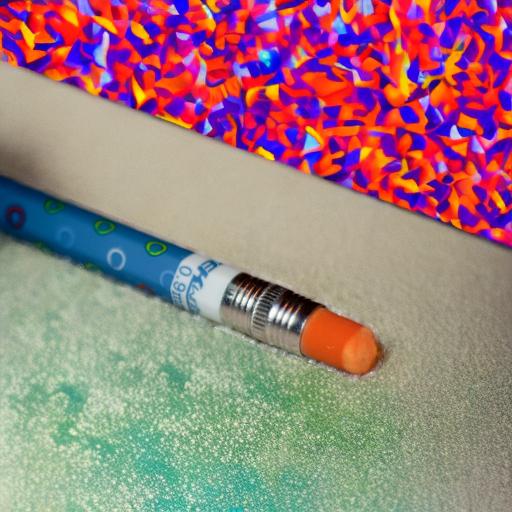}
\end{minipage}
\begin{minipage}{0.19\textwidth}
  \centering
  \includegraphics[height=2.4cm, width=\linewidth, keepaspectratio]{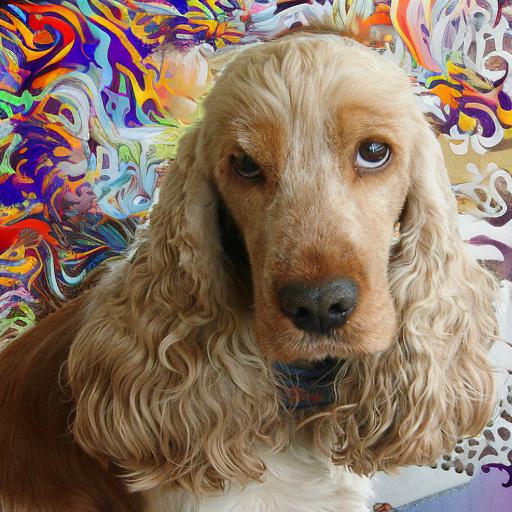}
\end{minipage}
\begin{minipage}{0.19\textwidth}
  \centering
  \includegraphics[height=2.4cm, width=\linewidth, keepaspectratio]{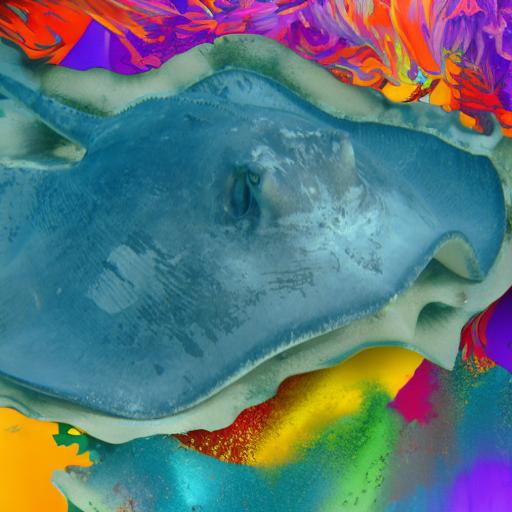}
\end{minipage}

\end{minipage}
\hfill
  \caption{Visual illustration of misclassified samples on color background and corresponding clean image samples. In two adjacent rows, \textit{first row} represent the clean images and the \textit{second row} represent the corresponding colorful background images}
  \label{fig:color-11}
\end{figure*}

\begin{figure*}[!h]

\begin{minipage}{\textwidth}

\centering

\begin{minipage}{0.19\textwidth}
  \centering
  \includegraphics[height=2.4cm, width=\linewidth , keepaspectratio]{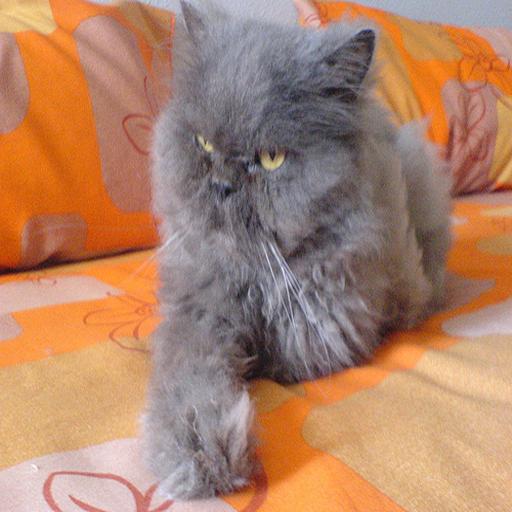}
\end{minipage}
\begin{minipage}{0.19\textwidth}
  \centering
  \includegraphics[height=2.4cm, width=\linewidth, keepaspectratio ]{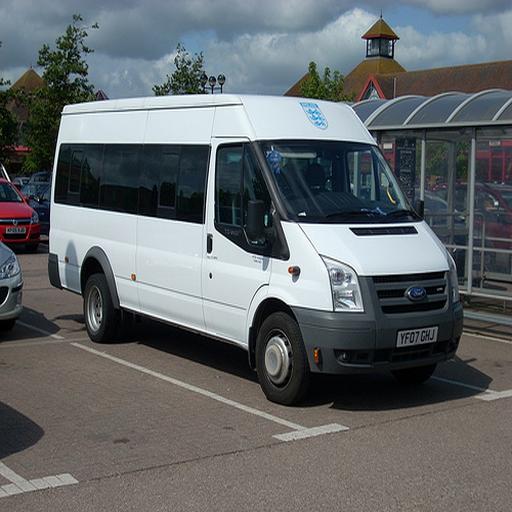}
\end{minipage}
\begin{minipage}{0.19\textwidth}
  \centering
  \includegraphics[height=2.4cm, width=\linewidth, keepaspectratio]{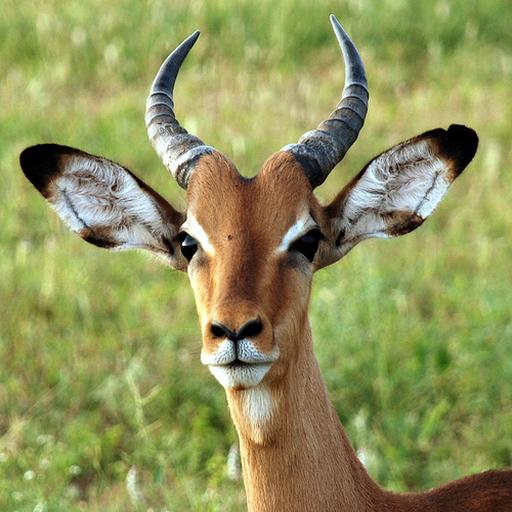}
\end{minipage}
\begin{minipage}{0.19\textwidth}
  \centering
  \includegraphics[height=2.4cm, width=\linewidth, keepaspectratio]{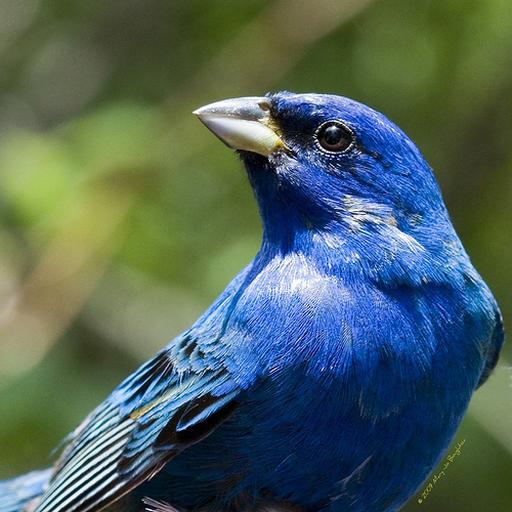}
\end{minipage}
\begin{minipage}{0.19\textwidth}
  \centering
  \includegraphics[height=2.4cm, width=\linewidth, keepaspectratio]{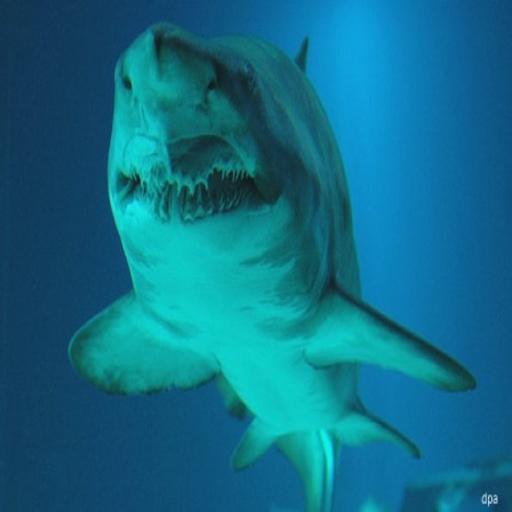}
\end{minipage}


\begin{minipage}{0.19\textwidth}
  \centering
  \includegraphics[height=2.4cm, width=\linewidth , keepaspectratio]{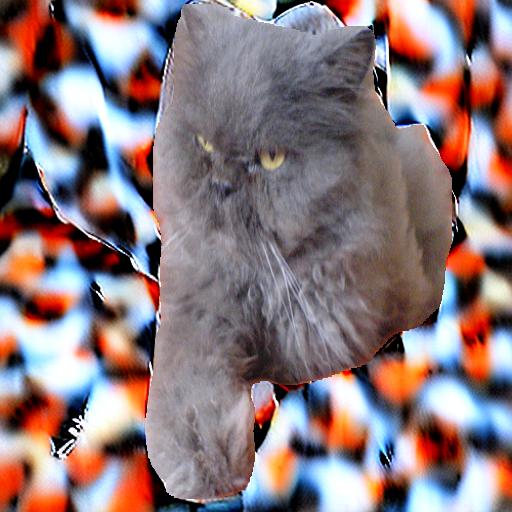}
\end{minipage}
\begin{minipage}{0.19\textwidth}
  \centering
  \includegraphics[height=2.4cm, width=\linewidth, keepaspectratio ]{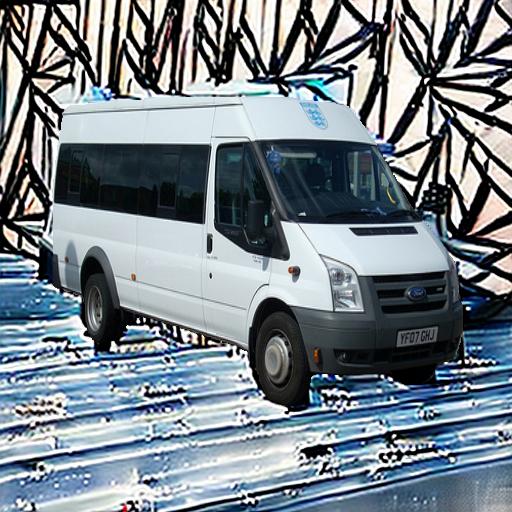}
\end{minipage}
\begin{minipage}{0.19\textwidth}
  \centering
  \includegraphics[height=2.4cm, width=\linewidth, keepaspectratio]{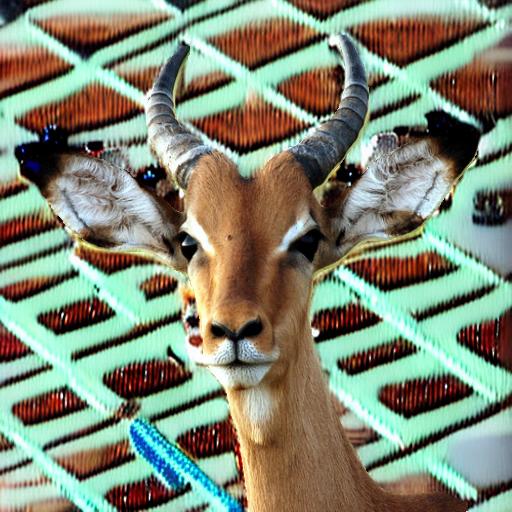}
\end{minipage}
\begin{minipage}{0.19\textwidth}
  \centering
  \includegraphics[height=2.4cm, width=\linewidth, keepaspectratio]{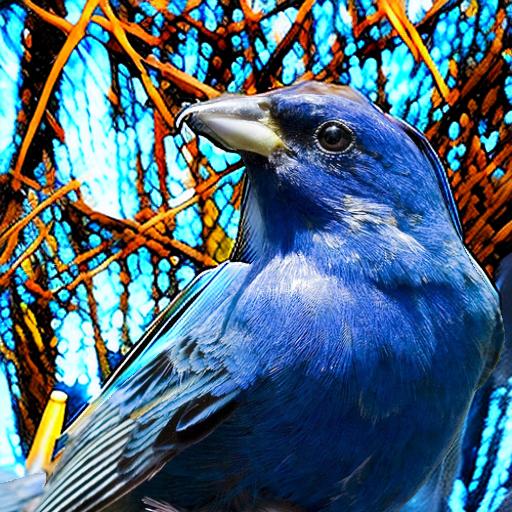}
\end{minipage}
\begin{minipage}{0.19\textwidth}
  \centering
  \includegraphics[height=2.4cm, width=\linewidth, keepaspectratio]{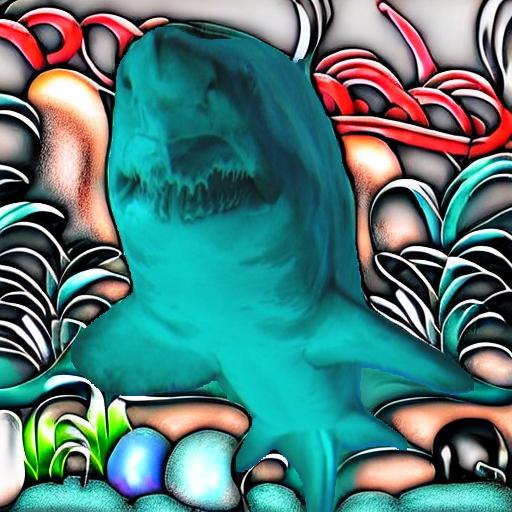}
\end{minipage}


\begin{minipage}{0.19\textwidth}
  \centering
  \includegraphics[height=2.4cm, width=\linewidth , keepaspectratio]{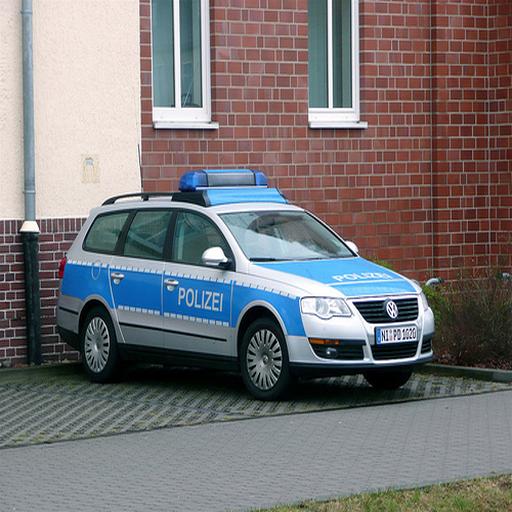}
\end{minipage}
\begin{minipage}{0.19\textwidth}
  \centering
  \includegraphics[height=2.4cm, width=\linewidth, keepaspectratio ]{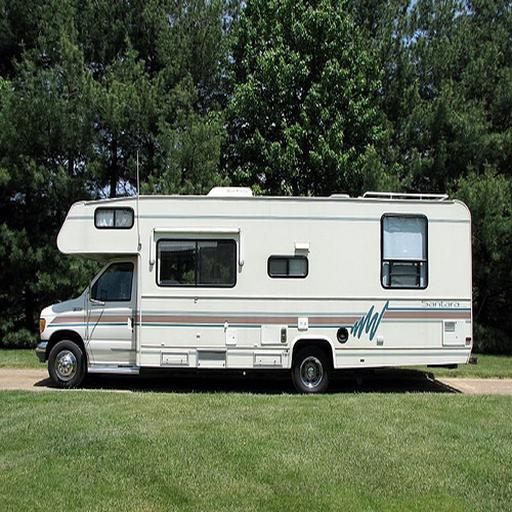}
\end{minipage}
\begin{minipage}{0.19\textwidth}
  \centering
  \includegraphics[height=2.4cm, width=\linewidth, keepaspectratio]{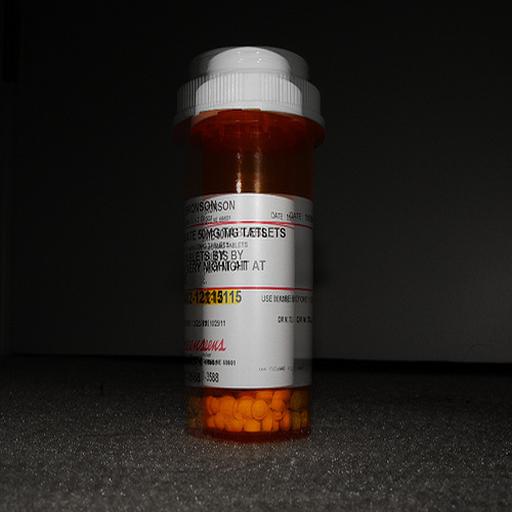}
\end{minipage}
\begin{minipage}{0.19\textwidth}
  \centering
  \includegraphics[height=2.4cm, width=\linewidth, keepaspectratio]{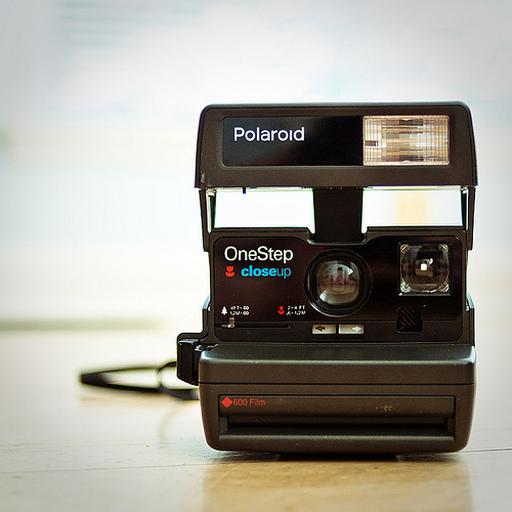}
\end{minipage}
\begin{minipage}{0.19\textwidth}
  \centering
  \includegraphics[height=2.4cm, width=\linewidth, keepaspectratio]{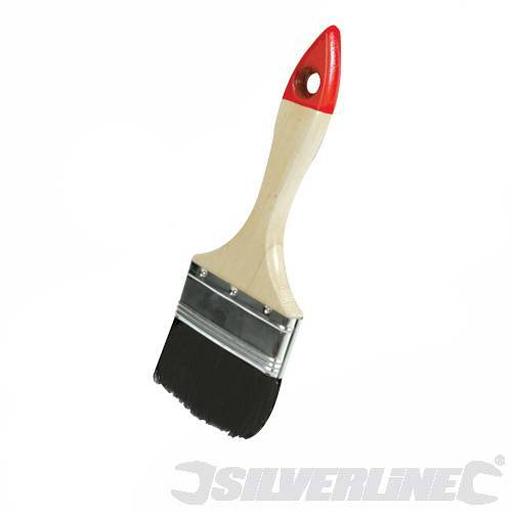}
\end{minipage}

\begin{minipage}{0.19\textwidth}
  \centering
  \includegraphics[height=2.4cm, width=\linewidth , keepaspectratio]{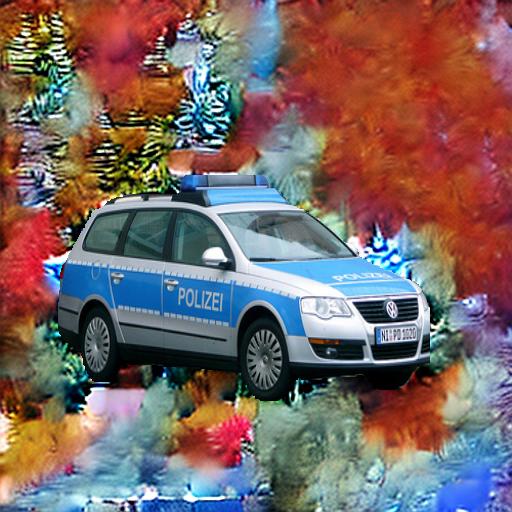}
\end{minipage}
\begin{minipage}{0.19\textwidth}
  \centering
  \includegraphics[height=2.4cm, width=\linewidth, keepaspectratio ]{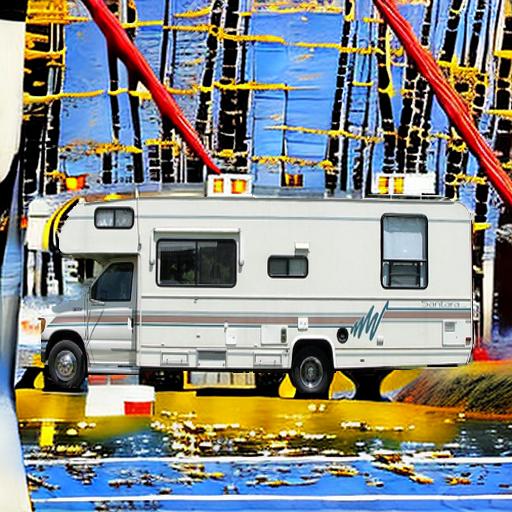}
\end{minipage}
\begin{minipage}{0.19\textwidth}
  \centering
  \includegraphics[height=2.4cm, width=\linewidth, keepaspectratio]{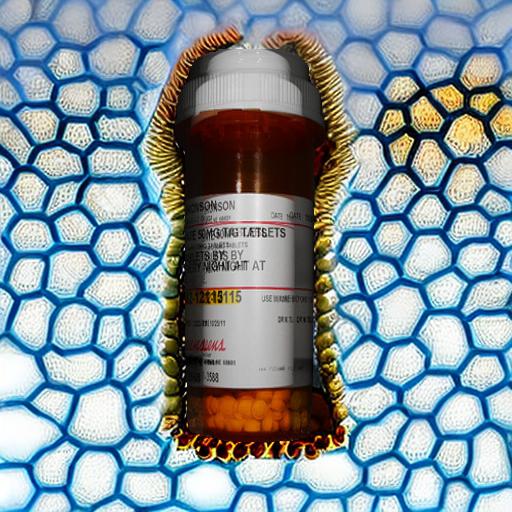}
\end{minipage}
\begin{minipage}{0.19\textwidth}
  \centering
  \includegraphics[height=2.4cm, width=\linewidth, keepaspectratio]{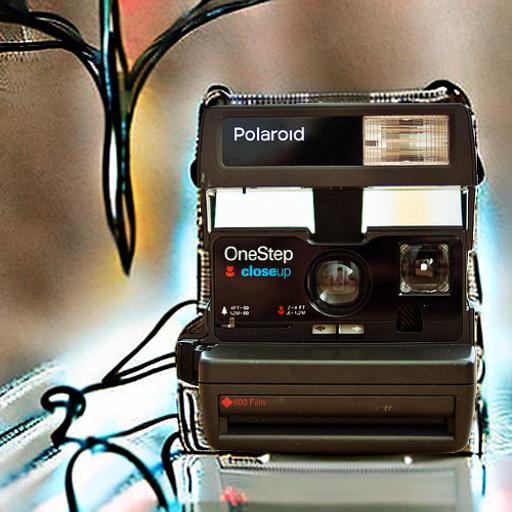}
\end{minipage}
\begin{minipage}{0.19\textwidth}
  \centering
  \includegraphics[height=2.4cm, width=\linewidth, keepaspectratio]{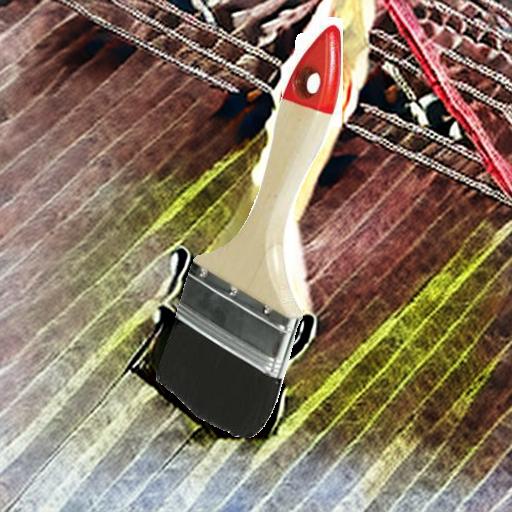}
\end{minipage}


\begin{minipage}{0.19\textwidth}
  \centering
  \includegraphics[height=2.4cm, width=\linewidth , keepaspectratio]{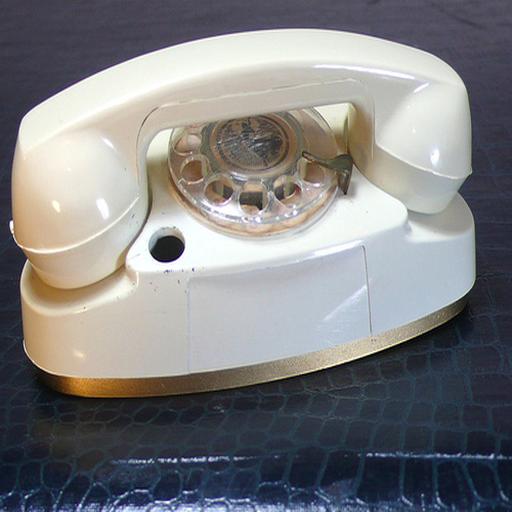}
\end{minipage}
\begin{minipage}{0.19\textwidth}
  \centering
  \includegraphics[height=2.4cm, width=\linewidth, keepaspectratio ]{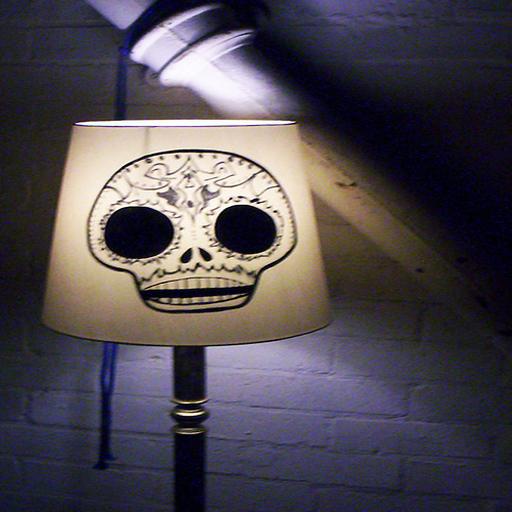}
\end{minipage}
\begin{minipage}{0.19\textwidth}
  \centering
  \includegraphics[height=2.4cm, width=\linewidth, keepaspectratio]{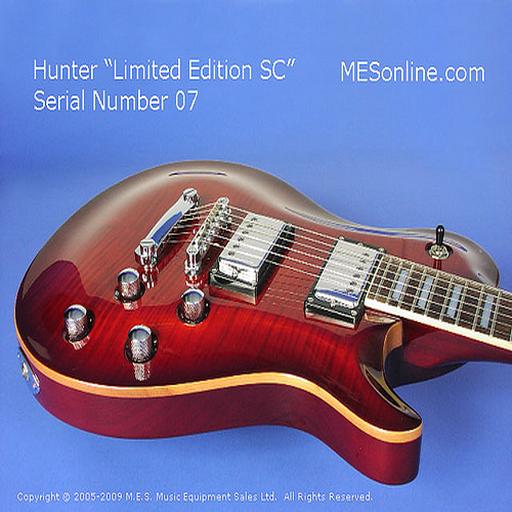}
\end{minipage}
\begin{minipage}{0.19\textwidth}
  \centering
  \includegraphics[height=2.4cm, width=\linewidth, keepaspectratio]{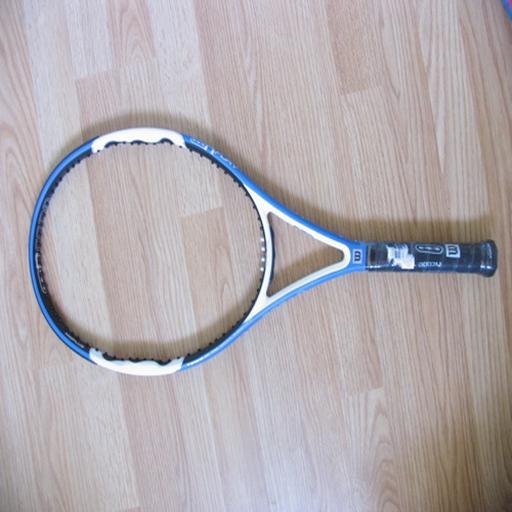}
\end{minipage}
\begin{minipage}{0.19\textwidth}
  \centering
  \includegraphics[height=2.4cm, width=\linewidth, keepaspectratio]{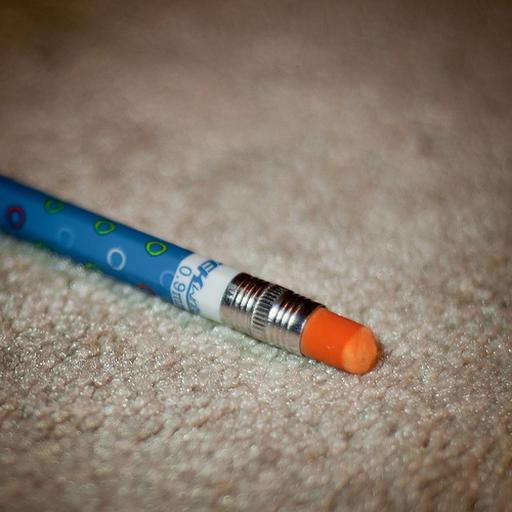}
\end{minipage}


\begin{minipage}{0.19\textwidth}
  \centering
  \includegraphics[height=2.4cm, width=\linewidth , keepaspectratio]{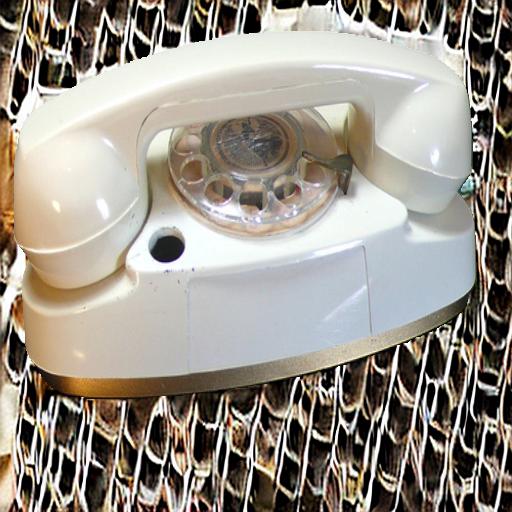}
\end{minipage}
\begin{minipage}{0.19\textwidth}
  \centering
  \includegraphics[height=2.4cm, width=\linewidth, keepaspectratio ]{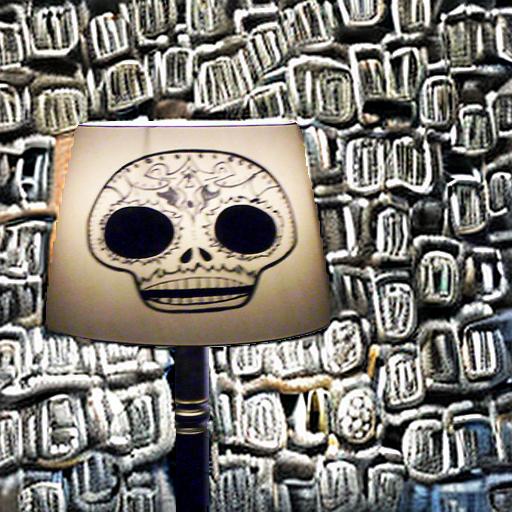}
\end{minipage}
\begin{minipage}{0.19\textwidth}
  \centering
  \includegraphics[height=2.4cm, width=\linewidth, keepaspectratio]{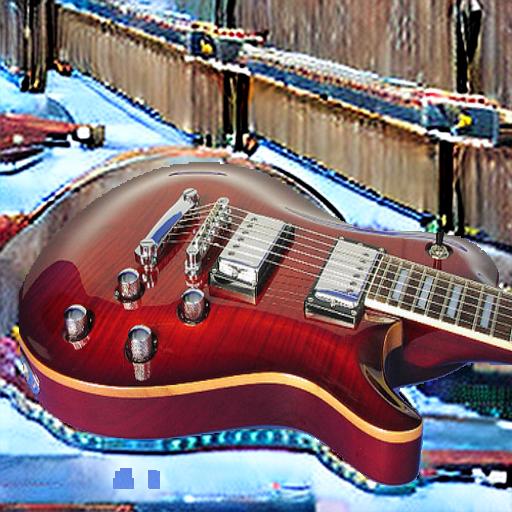}
\end{minipage}
\begin{minipage}{0.19\textwidth}
  \centering
  \includegraphics[height=2.4cm, width=\linewidth, keepaspectratio]{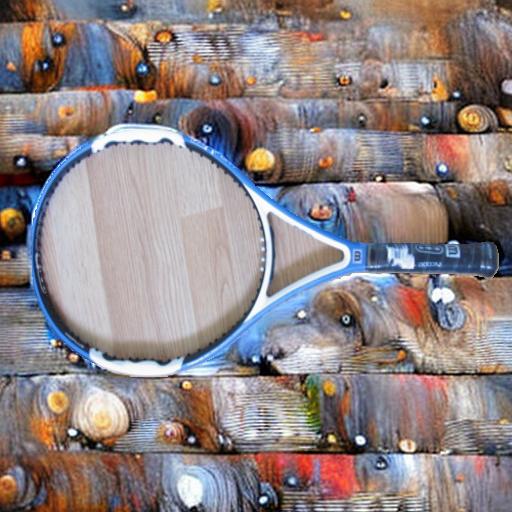}
\end{minipage}
\begin{minipage}{0.19\textwidth}
  \centering
  \includegraphics[height=2.4cm, width=\linewidth, keepaspectratio]{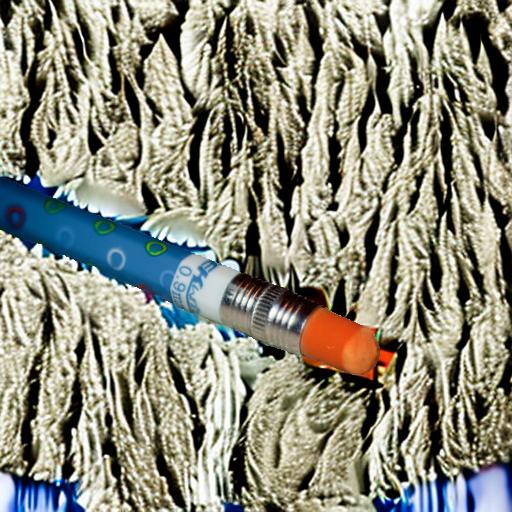}
\end{minipage}

\end{minipage}
\hfill
  \caption{Visual illustration of misclassified samples on adversarial background and corresponding clean image samples. In two adjacent rows, \textit{first row} represents the clean images and the \textit{second row} represents the corresponding adversarial images }
  \label{fig:adv-samples}
\end{figure*}

\begin{figure*}[!h]
\begin{minipage}{\textwidth}

\centering

\begin{minipage}{0.19\textwidth}
  \centering
  \includegraphics[height=2.4cm, width=\linewidth , keepaspectratio]{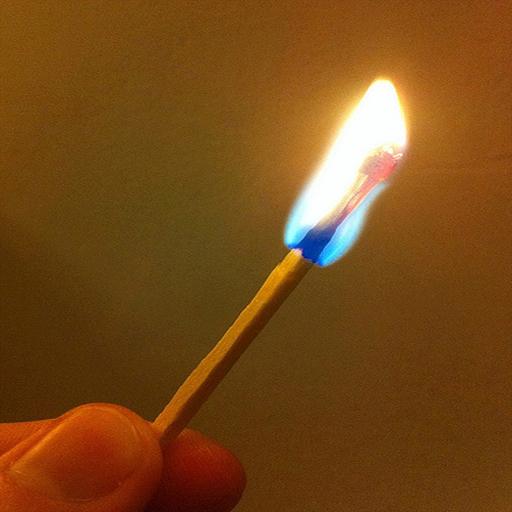}
\end{minipage}
\begin{minipage}{0.19\textwidth}
  \centering
  \includegraphics[height=2.4cm, width=\linewidth, keepaspectratio ]{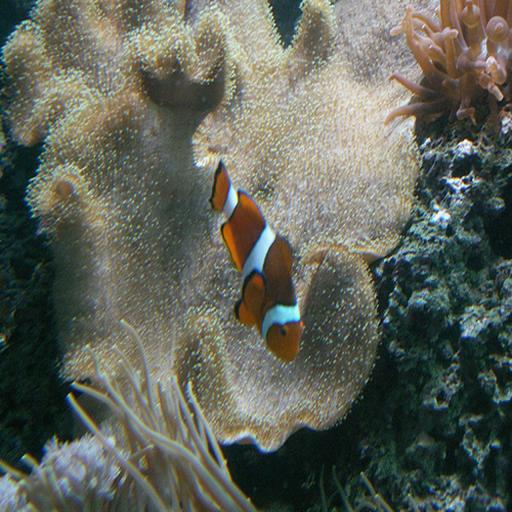}
\end{minipage}
\begin{minipage}{0.19\textwidth}
  \centering
  \includegraphics[height=2.4cm, width=\linewidth, keepaspectratio]{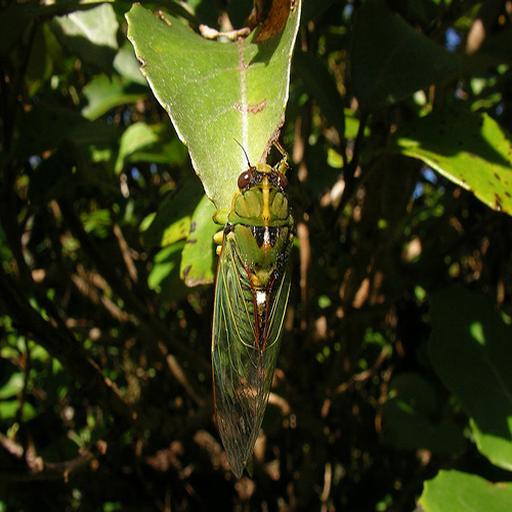}
\end{minipage}
\begin{minipage}{0.19\textwidth}
  \centering
  \includegraphics[height=2.4cm, width=\linewidth, keepaspectratio]{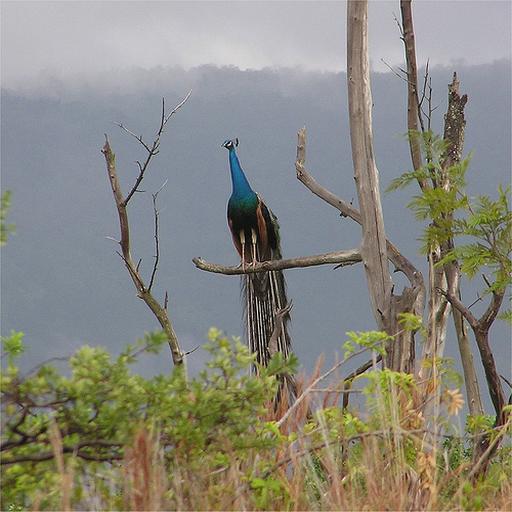}
\end{minipage}
\begin{minipage}{0.19\textwidth}
  \centering
  \includegraphics[height=2.4cm, width=\linewidth, keepaspectratio]{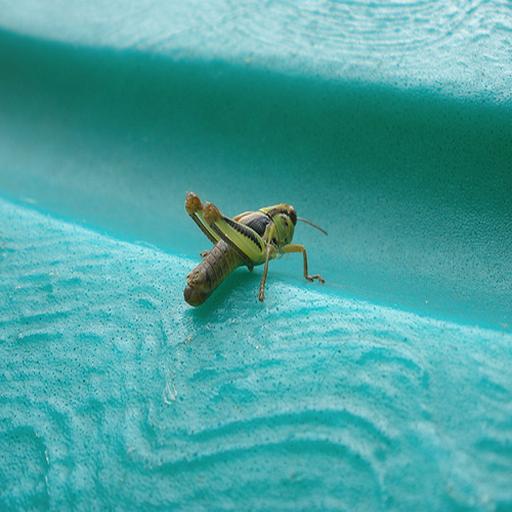}
\end{minipage}


\begin{minipage}{0.19\textwidth}
  \centering
  \includegraphics[height=2.4cm, width=\linewidth , keepaspectratio]{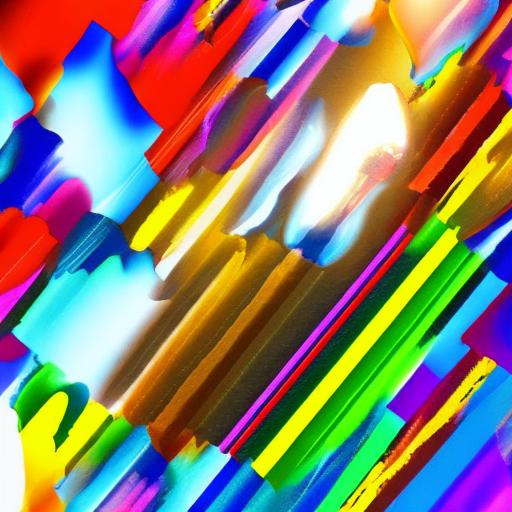}
\end{minipage}
\begin{minipage}{0.19\textwidth}
  \centering
  \includegraphics[height=2.4cm, width=\linewidth, keepaspectratio ]{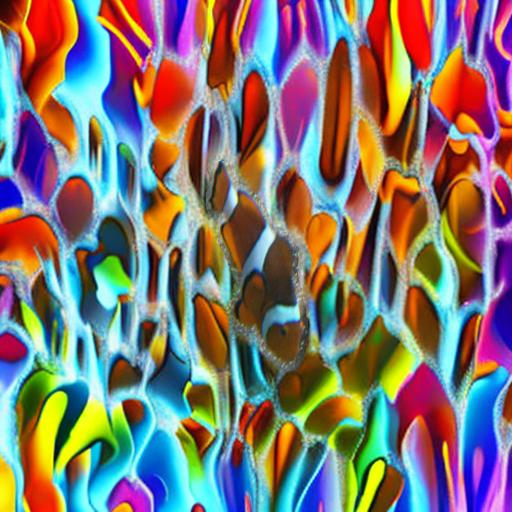}
\end{minipage}
\begin{minipage}{0.19\textwidth}
  \centering
  \includegraphics[height=2.4cm, width=\linewidth, keepaspectratio]{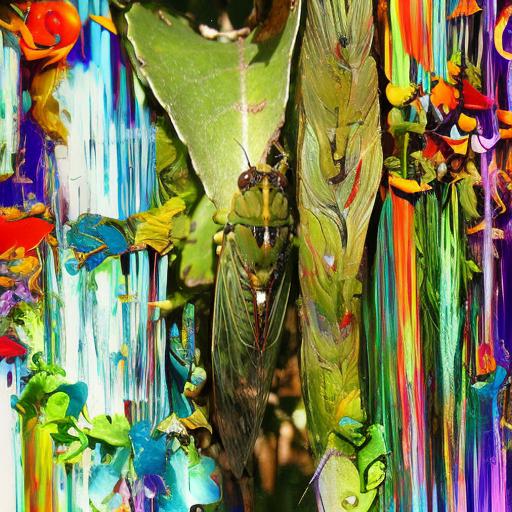}
\end{minipage}
\begin{minipage}{0.19\textwidth}
  \centering
  \includegraphics[height=2.4cm, width=\linewidth, keepaspectratio]{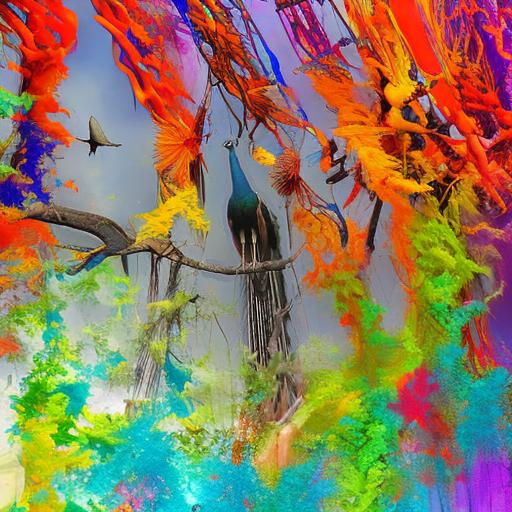}
\end{minipage}
\begin{minipage}{0.19\textwidth}
  \centering
  \includegraphics[height=2.4cm, width=\linewidth, keepaspectratio]{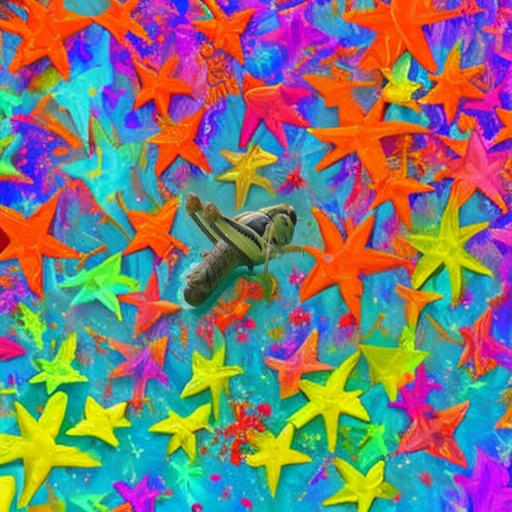}
\end{minipage}


\end{minipage}
\hfill
  \caption{Visual illustration of \textit{hard} samples on color background}
  \label{fig:hard-samples}
\end{figure*}

We observe that there exist images which get misclassified \emph{(by ResNet-50)} across several background alterations as can be seen from from Figure \ref{missclassified}. In Figure \ref{fig:color-11} we show examples on which the highly robust \emph{EVA-CLIP} ViT-E/14+ model fails to classify the correct class. After going through the misclassified samples, we visualize some of the \emph{hard} examples in Figure \ref{fig:hard-samples}. Furthermore, we also provide visualisation of images misclassified with adversarial background changes in Figure \ref{fig:adv-samples}.

\newpage

\FloatBarrier

\subsection{Potential External Factors}
\label{sec:external Factors}
When composing object-to-background change with texture, color, or adversarial patterns, the target models can perceive those as some other class if that pattern or composition is dominant in that class during the training of the models.   We discuss the potential external factors and how our proposed method minimizes the effect of those external factors fo object-to-background compositional changes.

\noindent \textbf{Preserving Object Semantics:} We preserve object semantics by using strong visual guidance via SAM for precise object delineation. 

\noindent \textbf{Possibility of extra objects in the Background:}  Kindly note that \emph{a)} we use a pretrained diffusion model that is conditioned on a pretrained CLIP text encoder, this means that the generated output follows the latent space of the CLIP text encoder which is aligned with CLIP visual encoder. Therefore, we can measure the faithfulness of the generated sample w.r.t the textual prompt used to generate it. We can measure this by encoding the generated output and its corresponding text prompt within CLIP latent space. For a given sample, CLIP or EVA-CLIP performs zero-shot evaluation by measuring the similarity between embedding of class templates (e.g. 1000 templates of ImageNet class) with a given image. Thus, if we  add the template for a textual prompt used to generate the object-to-background changes, then we can measure its alignment with the background changes. For instance, instead of using a “a photo of a {fish}” template for zero-shot classification, we add the relevant template that is with background change, such as “a photo of a {fish} in the vivid colorful background”.  In other words, the relevant template represents the object and background change we introduced. We validate this observation on the EVA-CLIP ViT-E/14+, a highly robust model. Using the class templates such as “a photo of a {}”, the model achieves $95.84\%$ accuracy on the original images (\texttt{ImageNet-B} dataset), which decreases to $88.33\%$ when our color background changes are applied (see Table \ref{tab:appendix_clip_classification} in Appendix \ref{sec:prompt_evaluation}). However, when using the relevant template,  the performance improves to 92.95\%, significantly reducing the gap between the performance on the original and color background changes from $7.51\%$ to $2.89\%$. These results show that accuracy loss from background changes isn't due to unwanted background objects of other classes. Furthermore, we manually assess  2.89\% of misclassified samples (very few samples, see Figure \ref{fig:color-11} and \ref{fig:hard-samples} in Appendix \ref{sec:misclassified}). These can be considered the hardest examples in our dataset. We observe that even in such hard cases the model's confusion often stemmed from the complex background patterns instead of the addition of unwanted objects. We observe a similar trend in the case of adversarial patterns as well (see Figure \ref{fig:adv-samples} in Appendix \ref{sec:misclassified}). b) Another empirical evidence of how our generated output closely follows the given textual prompts can be observed with BLIP-2 Caption of the original image. In this case, object-background change has similar results as compared to original images across different vision models (Table 2 in the main paper).

\noindent \textbf{Extension of Objects:}  As already detailed in Appendix \ref{sec:limitation}, we encountered challenges when dealing with objects that occupy a small region in the image, sometimes leading to certain unwanted extensions to objects. To mitigate this, we filtered our dataset to focus on images where the object covers a significant area. Additionally, we slightly expand object masks computed using SAM to better define boundaries and prevent object shape distortion in the background.

The design choices discussed above, such as strong visual guidance and class-agnostic textual guidance, contribute to the well-calibrated results of our study. This indicates that our results using the conventional metrics such as classification accuracy are well calibrated as well in the context of our high quality of generated data as mentioned above. We note that these choices ensure that the models are primarily challenged by diverse changes in the background, rather than being misled by the presence of unwanted objects. This careful approach underlines the reliability of our findings and highlights the specific factors influencing model performance.

\subsection{Dataset Distribution and Comparison}
\label{sec:Dataset}
\begin{figure}[h!]
    \centering
    \includegraphics[width=\textwidth]{
    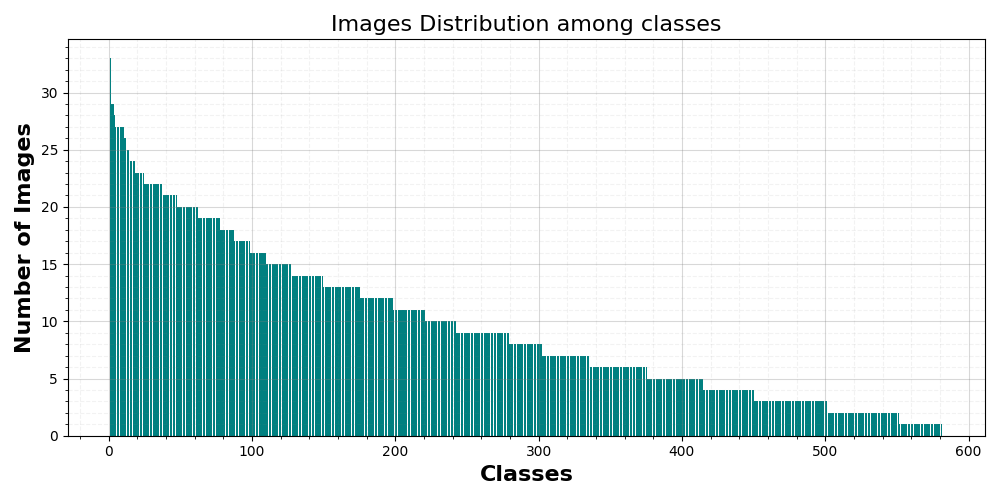
    }
    \caption{Our \texttt{ImageNet-B} dataset encompasses a diverse variety of images spanning 582 distinct classes. In this illustration, we showcase images distribution among all the classes. The figure is plotted in decreasing order of images present in each class.}
    \label{dataset}

\end{figure}

\texttt{ImageNet-B} dataset comprises a wide variety of objects belonging to different classes, as illustrated in Figure \ref{dataset}. Our dataset maintains a clear distinction between the background and objects, achieved through a rigorous filtering process applied to the ImageNet validation dataset. Additionally, we provide the list of prompts in Table \ref{prompt} utilized for  the experiments.  

As shown in Tab. \ref{tab:dataset}, our curated \textbf{ImageNet-B} dataset is the largest in terms of both the number of images \textit{and} classes compared to closely related works \cite{prabhu2023lance, li2023imagenet, zhang2024imagenet}. In contrast to \cite{prabhu2023lance, li2023imagenet, zhang2024imagenet}, we extend our analysis to object detection by introducing the \textbf{COCO-DC} dataset. Our proposed background changes on \textbf{ImageNet-B} \& \textbf{COCO-DC}, enable us to evaluate 
on more than 70k samples for classification \& 5k samples for detection. Our automated framework of delineating between foreground \& background facilitates future dataset expansion.

\begin{table}[t]
    \caption{\small \textbf{Dataset Comparison}}
    \label{tab:dataset}
    \centering
    \setlength{\tabcolsep}{10pt}
    \resizebox{1\linewidth}{!}{
    \begin{tabular}{ll|cccc}
    \toprule
        & Dataset & \textbf{\#}Classes & \textbf{\#}Images & Classification & Detection \\
        \toprule
           \multirow{2}{*}{Baseline} & LANCE(NeurIPS 2023)\cite{prabhu2023lance} & 15 & 750 & $\checkmark$ & $\times$ \\
            & ImageNet-E(CVPR 2023)\cite{li2023imagenet} & 373 & 47872 & $\checkmark$ & $\times$ \\
            & ImageNet-D(CVPR 2024)\cite{zhang2024imagenet} & 113 & 4835 & $\checkmark$ & $\times$ \\
            \midrule
           \multirow{2}{*}{Ours}&\texttt{ImageNet-B} & 582 & 77070 & $\checkmark$ & $\times$ \\ 
           &\texttt{COCO-DC} &  66 & 5635 & $\checkmark$ & $\checkmark$ \\ 
        \midrule
    \end{tabular}
    }
\end{table}

\FloatBarrier

\subsection{Evaluation on Background/Foreground Images}

In this section, we systematically evaluate vision-based models by focusing on background and foreground elements in images. This evaluation involves masking the background of the original image, allowing us to assess the model's performance in recognizing and classifying the foreground without any cues from the background context. Conversely, we also mask the object or foreground from the image. This step is crucial to understand to what extent the models rely on background information for classifying the image into a specific class. This dual approach provides a comprehensive insight into the model's capabilities in image classification, highlighting its reliance on  foreground and background elements.

\label{sec: rmasked}

\begin{table*}[h]
\fontsize{7pt}{6pt}\selectfont
\centering
\caption{Evaluation of Zero-shot CLIP Models on \texttt{ImageNet-B} dataset while masking the object or the background of the image. Top-1(\%) accuracy is reported. The accuracy drop is observe when we remove the object clues from the background such as in texture or color background}
\resizebox{1\linewidth}{!}{%
\begin{tabular}{lcccccccc}
\toprule

\multirow{2}{*}{Background} & 
\multicolumn{6}{c}{\textbf{Foreground}} 

\\  \cmidrule(lr){2-9}
& Res50 & Res101 & Res50x4 & Res50x16 & ViT-B/32 & ViT-B/16 & ViT-B/14  & \cellcolor{gray!20} Average\\
 \midrule
Original & 54.76 & 58.89   & 64.86&70.80&59.47   & 69.42&79.12&\cellcolor{gray!20}  65.33\\
 \toprule
 \multirow{2}{*}{} & 
\multicolumn{6}{c}{\textbf{Background}}
\\  \cmidrule(lr){2-9}

Original & 15.84 & 17.74   & 18.47&20.67&17.72   & 21.28&28.99&\cellcolor{gray!20}  20.10\\
Class label & 27.17 & 29.08  & 33.02 & 35.93 & 31.35    & 38.74&46.88&\cellcolor{gray!20} 34.59 \\
BLIP-2 Caption & 19.05 & 21.39  & 23.37 & 24.57 & 22.39  &27.21 &34.42 &\cellcolor{gray!20} 24.62 \\
Color & 3.92 & 5.46  & 5.64 & 6.53 & 5.64 &6.95    & 10.28&\cellcolor{gray!20} 6.34 \\
Texture & 3.65 & 5.12  & 5.12 & 5.84 & 5.43   & 6.68 & 10.04&\cellcolor{gray!20} 5.98 \\

\bottomrule
\end{tabular}%
}
\label{tab:clip}
\end{table*}

\begin{table*}[h]
\fontsize{6pt}{5pt}\selectfont
\centering
\caption{DINOv2 model evaluation by masking either the object or the background within the \texttt{ImageNet-B} dataset. The integration of the additional token in the DINOv2 model proves beneficial, contributing to enhanced accuracy. However, our observations reveal that these models remain susceptible to background cues, particularly evident in class labels and the BLIP-2 Caption dataset. Interestingly, as we transition towards more generic texture or color backgrounds, a discernible drop in accuracy is observed.}
\resizebox{1\linewidth}{!}{%
\begin{tabular}{lcccccccc}
\toprule

\multirow{2}{*}{Background} & 
\multicolumn{8}{c}{\textbf{Foreground}} 

\\  \cmidrule(lr){2-9}
& ViT-S & ViT-B & ViT-L &\cellcolor{gray!20} Average& ViT-S$_{\text{reg}}$ & ViT-B$_{\text{reg}}$ & ViT-L$_{\text{reg}}$  & \cellcolor{gray!20} Average\\
 \midrule
Original & 88.73 & 93.86   & 94.89&\cellcolor{gray!20} 92.49&96.34&89.95   & 97.25&\cellcolor{gray!20} 94.51 \\
 \toprule
 \multirow{2}{*}{} & 
\multicolumn{8}{c}{\textbf{Background}}
\\  \cmidrule(lr){2-9}

Original & 27.72 & 37.78  & 51.44 &\cellcolor{gray!20} 38.98& 30.10 & 42.08    & 55.18&\cellcolor{gray!20}  42.45\\

Class label & 42.70 & 54.73  & 66.68 &\cellcolor{gray!20}54.70 & 46.88 & 58.81    & 68.97&\cellcolor{gray!20} 58.22 \\
BLIP-2 Caption & 30.51 & 40.74  & 50.57 &\cellcolor{gray!20}40.60 & 33.62 & 42.48  &52.40 &\cellcolor{gray!20} 42.83 \\
Color & 2.96 & 5.03  & 8.39 &\cellcolor{gray!20} 5.46& 3.68 & 5.75    & 9.50&\cellcolor{gray!20} 6.31 \\
Texture & 2.83 & 4.92  & 7.88 &\cellcolor{gray!20}5.21 & 3.45 & 5.57   & 9.28&\cellcolor{gray!20} 6.10 \\

\bottomrule
\end{tabular}%
}
\label{tab:dinov2}
\end{table*}

\newpage

\subsection{Insights}
\label{sec:insights}

\textbf{Across Architectures.} Our results (Tab. \ref{tab:base_class_comparison},\ref{tab:base_zs_comparison}, \ref{tab:stylised}, \& \ref{tab:sota-models}) show that CNNs perform better than transformers across various background changes. 
We note that the mixing of background \& foreground token features through the global attention mechanism may result in the reliance of transformer models prediction on outlier/background tokens. This is validated when we evaluate transformers which are trained to prioritize learning more salient features\cite{darcet2023vision} (Tab. \ref{tab:dinov2_reg} \& \ref{tab:dinov2}), resulting in improved performance under background changes. 

\noindent \textbf{Across Training Methods.} Analysis of adversarially trained models (Fig. \ref{fig:adv_results} \& Tab. \ref{tab:adv-model_comparison-2}, \ref{tab:adv-model_comparison-1},  \ref{tab:adv-model_comparison-3}) reveal their robustness is confined to adversarial background changes, leaving them vulnerable to other background variations.  Similar behavior is observed for models trained on stylized ImageNet dataset(Tab. \ref{tab:stylised}). However, self-supervised training of uni-modal models on large extensively curated datasets shows performance gain across background changes(Fig. \ref{fig:dinov2}). Similarly, for multi-modal models, we observe that stabilizing training on large-scale datasets (as in EVA-CLIP) leads to significant improvement in zero-shot performance across all background changes(Tab. \ref{tab:base_zs_comparison},\ref{tab:appendix_clip_classification}, \& \ref{tab:appendix_zs_classification}). 

\noindent \textbf{Across Vision Tasks.} The absence of object-to-background context during classification model training creates a significant vulnerability to background changes. In contrast, object detection \& segmentation models (Tab. \ref{tab:AP}, Fig. \ref{detr} \& \ref{fig:diversity_detection}), which explicitly incorporate object-to-background context during training, show notably better resilience to background variations. Based on the above insights, we discuss current limitations and future directions next.

\subsubsection{Limitations.}
\label{sec:limitation}
\begin{figure}[h]
    \centering
    \includegraphics[height=6cm, width=\textwidth, keepaspectratio]{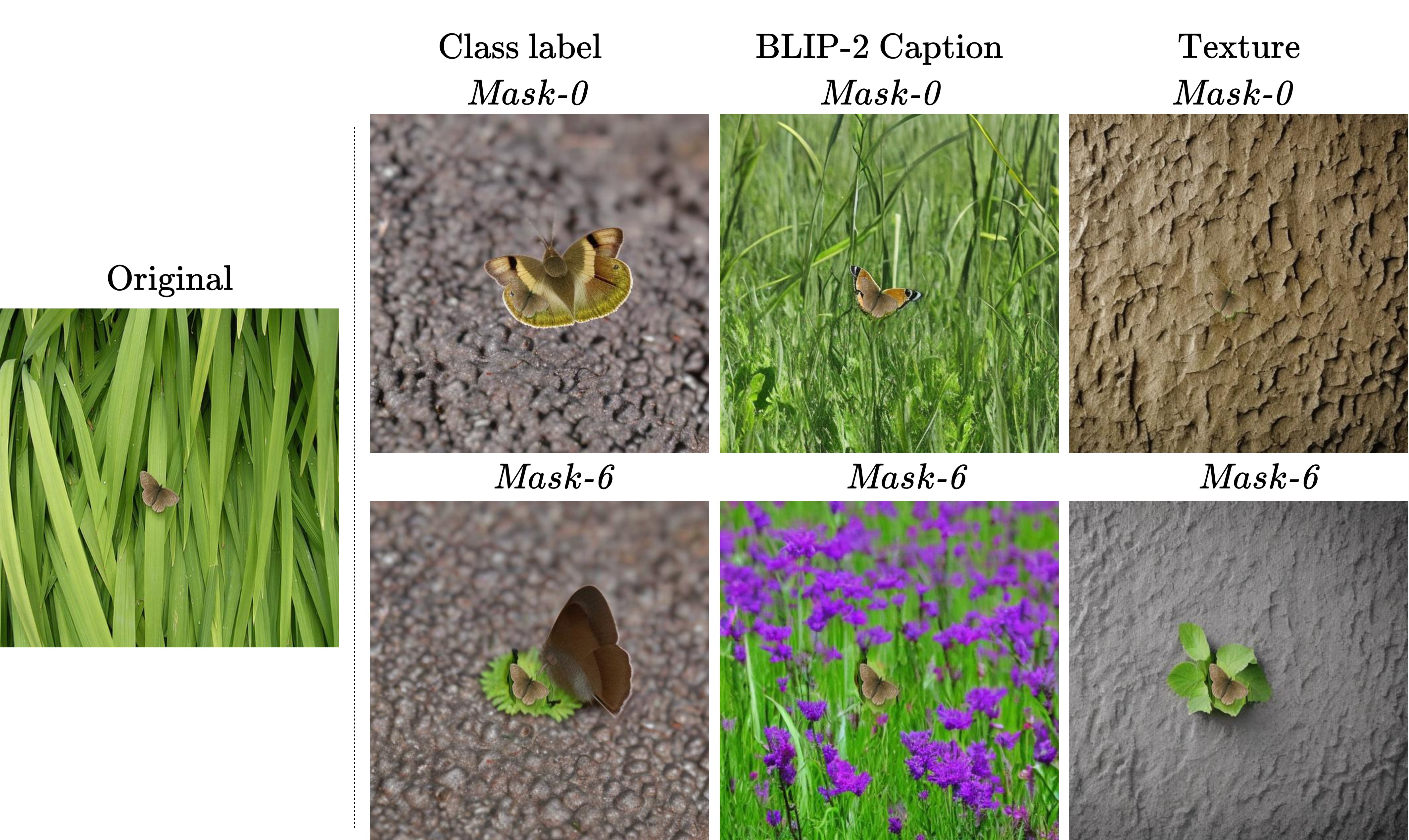}
    
    \caption{Limitation: Background changes on small objects in the scene. Enlarging the mask (here by 6 pixels) helps in mitigating the issue to some effect. }
    \label{limitation}

\end{figure}

In Figure \ref{limitation}, we observe that for objects covering a small region in the image,  relying solely on the class name to guide the diffusion model can  result in alterations of the object shape, expanding the influence of the class name semantics to larger image regions. However, by supplementing with descriptive captions that encompass the object-to-background context, we partially mitigate this effect.  Furthermore, the generated textured background can inadvertently camouflage the object. To address this concern, we slightly expand the object mask to clearly delineate the object boundaries.

\subsubsection{Future Directions.}
Our current work represents one of the preliminary efforts in utilizing diffusion models to study the object-to-background context in vision-based systems. Based on our observations and analysis, the following are the interesting future directions.

\begin{itemize}
    \item Since large capacity models in general show better robustness to object-to-background compositions, coming up with new approaches to effectively distill knowledge from these large models could improve how small models cope with background changes. This can improve resilience in small models that can be deployed in edge devices.
    \item Another direction is to set up object-to-background priors during adversarial training to expand robustness beyond just adversarial changes. To some extent, successful examples are recent works \cite{sitawarin2022part, darcet2023vision} where models are trained to discern the salient features in the image foreground. This leads to better robustness. 
    \item Our work can be extended to videos where preserving the semantics of the objects across the frames while introducing changes to the background temporally will help understand the robustness of video models.
    \item Additionally, the capabilities of diffusion models can be explored to craft complex changes in the object of interest while preserving the semantic integrity. For instance, in \cite{yuan2023customnet}, diffusion models are employed to generate multiple viewpoints of the same object. Additionally, in \cite{kawar2023imagic}, non-rigid motions of objects are created while preserving their semantics. By incorporating these with our approach, we can study how vision models maintain semantic consistency in dynamic scenarios. 
\end{itemize}

\subsection{Calibration Metrics}
\label{sec:calibration}
\begin{figure}
\centering
\begin{minipage}{\textwidth}

\begin{minipage}{0.19\textwidth}
  \centering
  \includegraphics[trim= 5mm 0mm 5mm 5mm, clip, width=\linewidth , keepaspectratio]{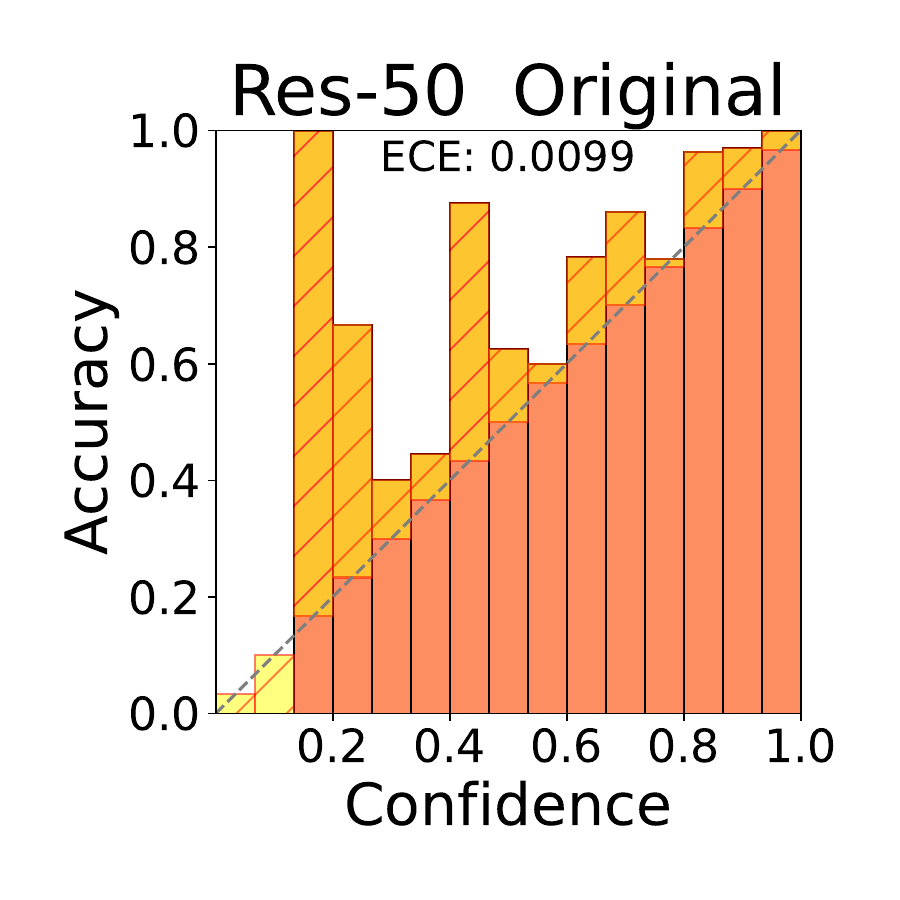}
\end{minipage}
\hfill
\begin{minipage}{0.19\textwidth}
  \centering
  \includegraphics[trim=  5mm 0mm 5mm 5mm, clip, width=\linewidth , keepaspectratio]{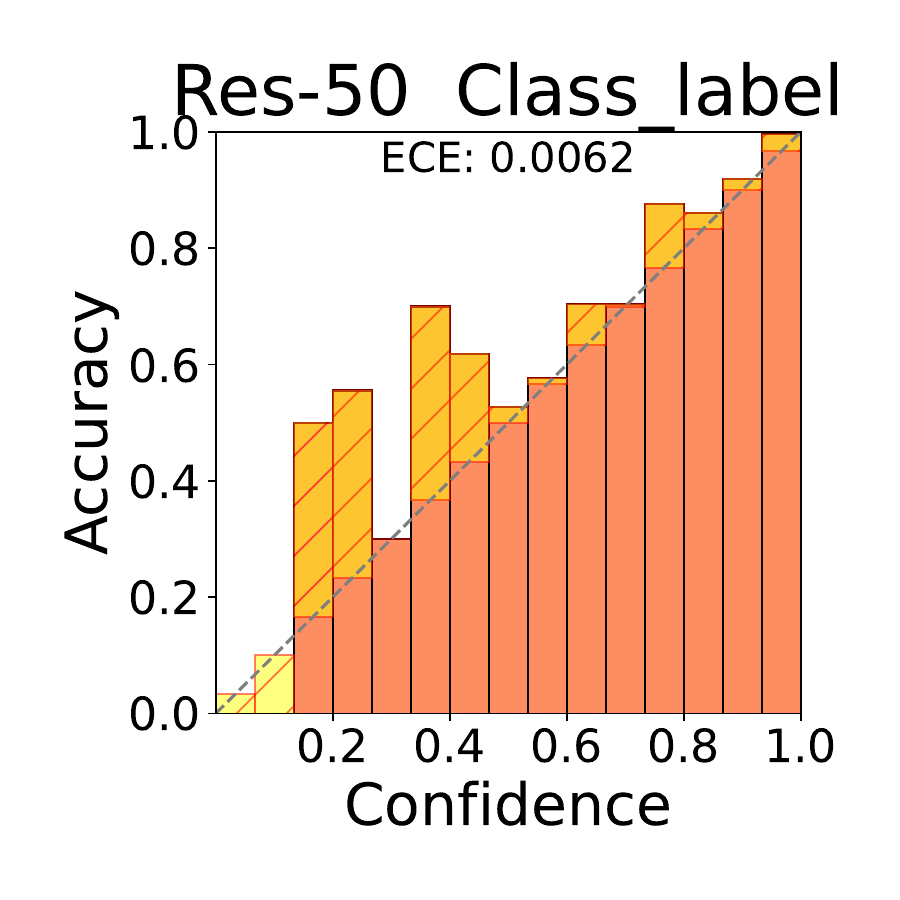}
\end{minipage}
\hfill
\begin{minipage}{0.19\textwidth}
  \centering
  \includegraphics[trim=  5mm 0mm 5mm 5mm, clip, width=\linewidth , keepaspectratio]{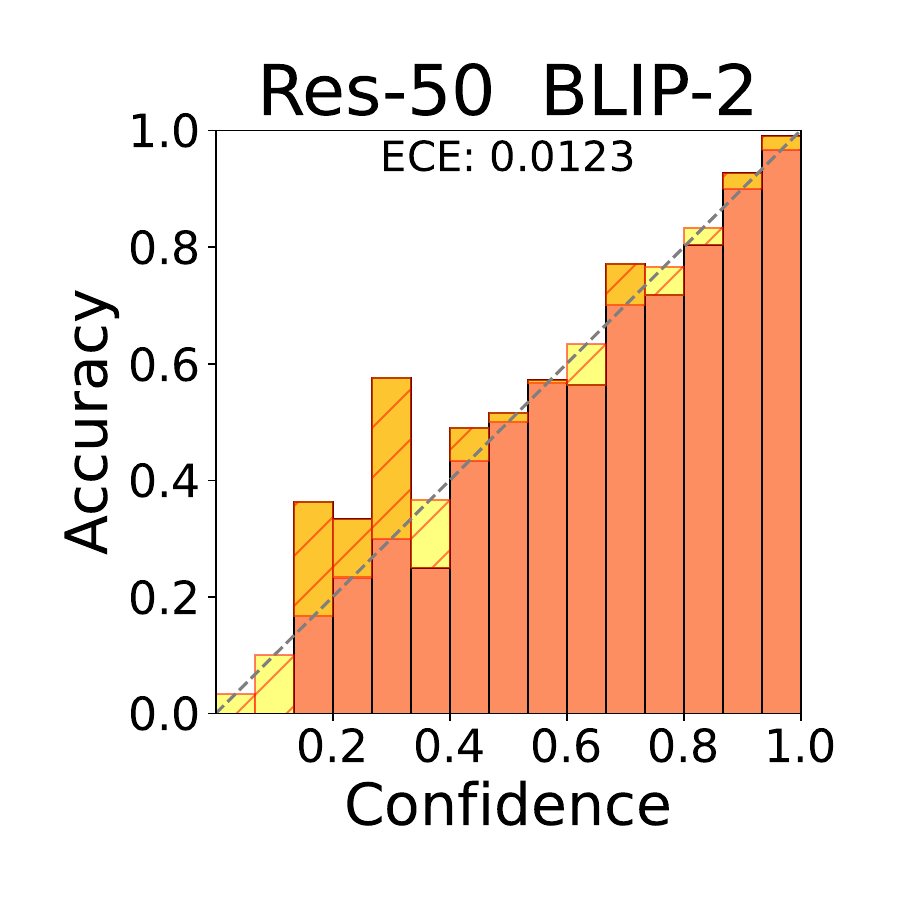}
\end{minipage}
\hfill
\begin{minipage}{0.19\textwidth}
  \centering
  \includegraphics[trim=  5mm 0mm 5mm 5mm, clip, width=\linewidth , keepaspectratio]{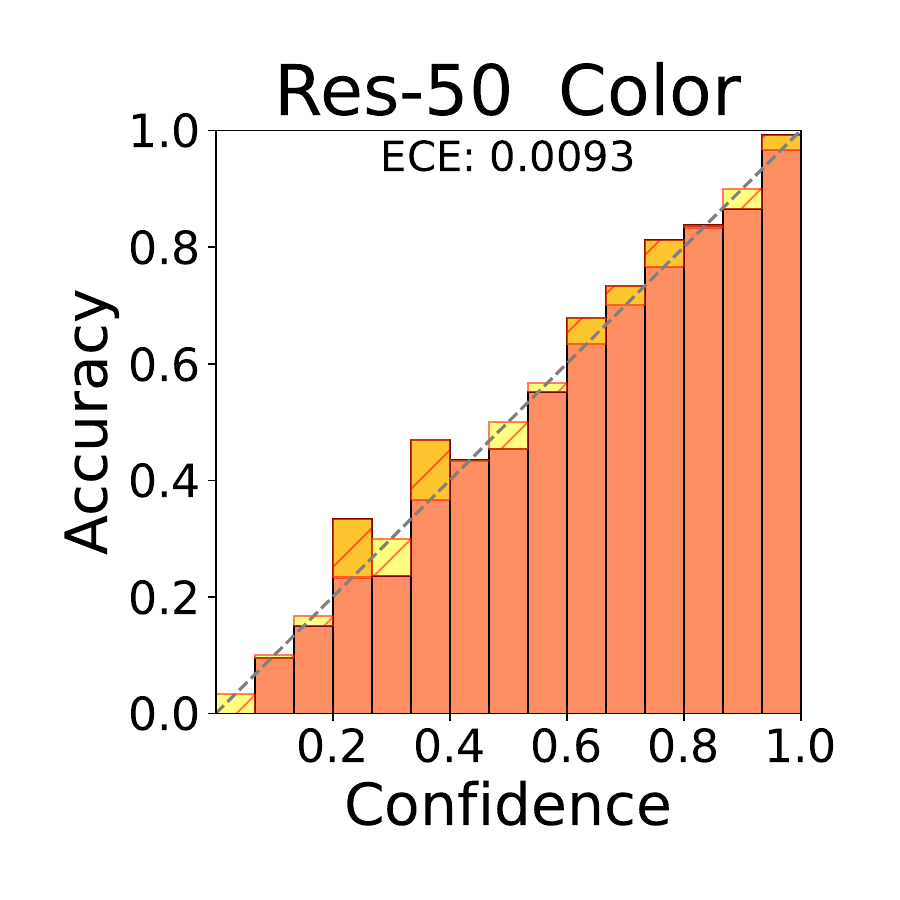}
\end{minipage}
\hfill
\begin{minipage}{0.19\textwidth}
  \centering
  \includegraphics[trim=  5mm 0mm 5mm 5mm, clip, width=\linewidth , keepaspectratio]{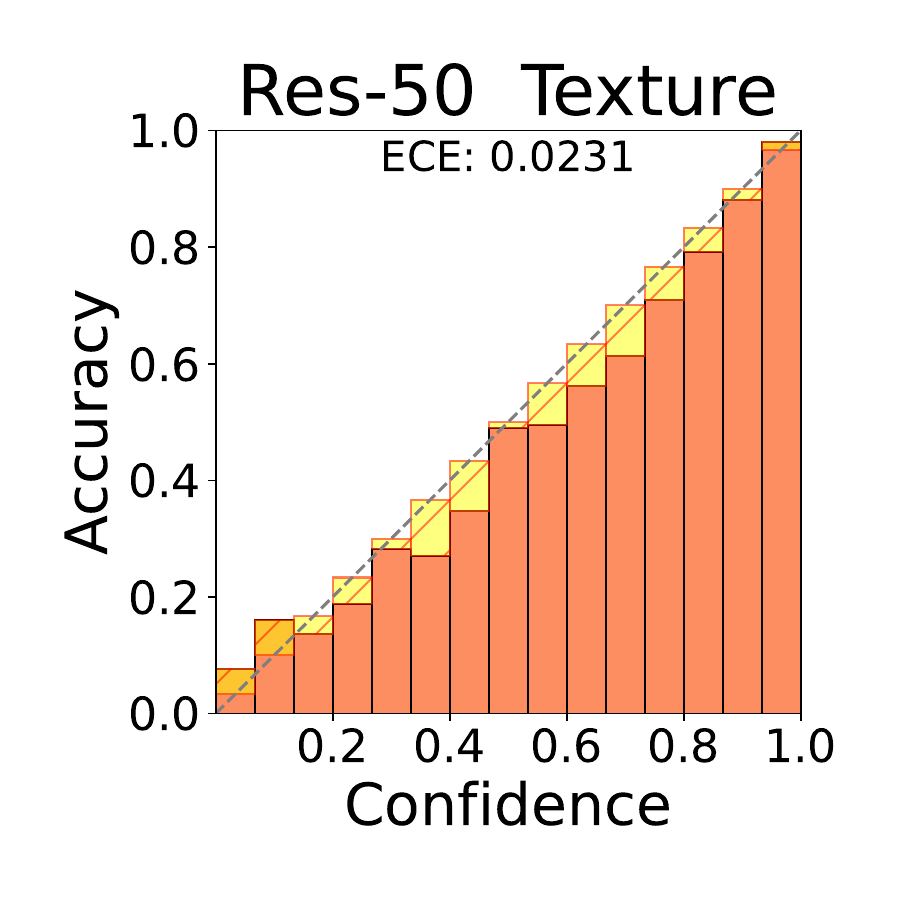}
\end{minipage}

\vspace{10pt}  

\begin{minipage}{0.19\textwidth}
  \centering
  \includegraphics[trim=  5mm 0mm 5mm 5mm, clip, width=\linewidth , keepaspectratio]{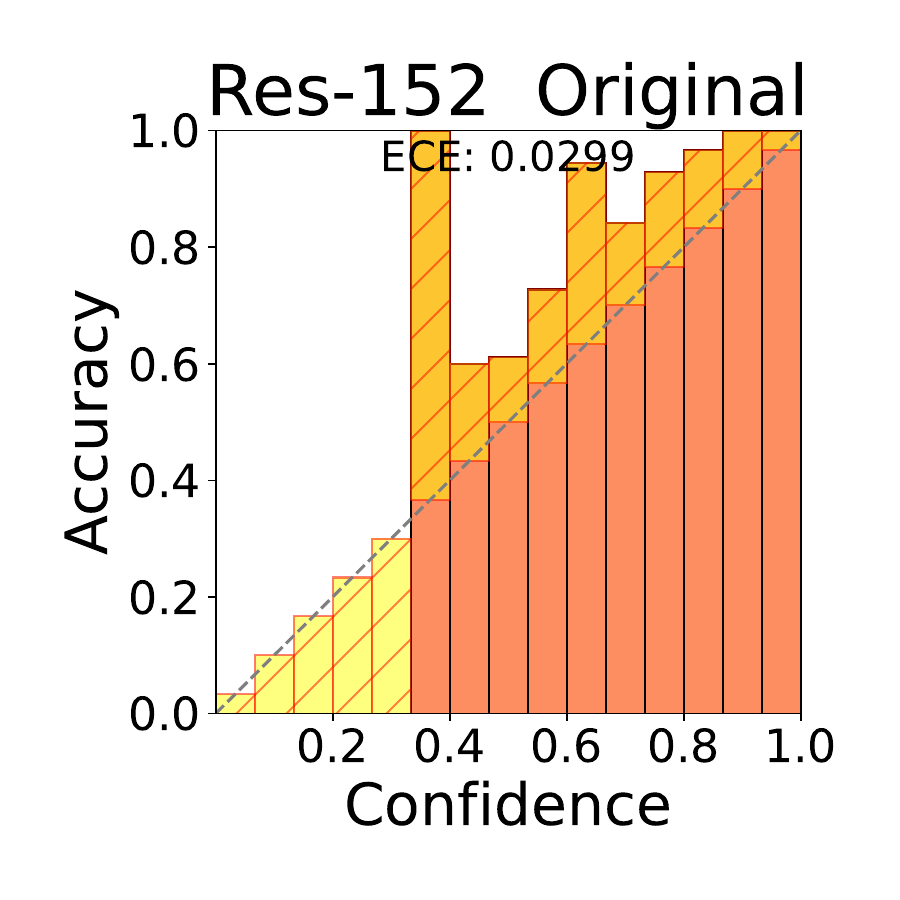}
\end{minipage}
\hfill
\begin{minipage}{0.19\textwidth}
  \centering
  \includegraphics[trim=  5mm 0mm 5mm 5mm, clip, width=\linewidth , keepaspectratio]{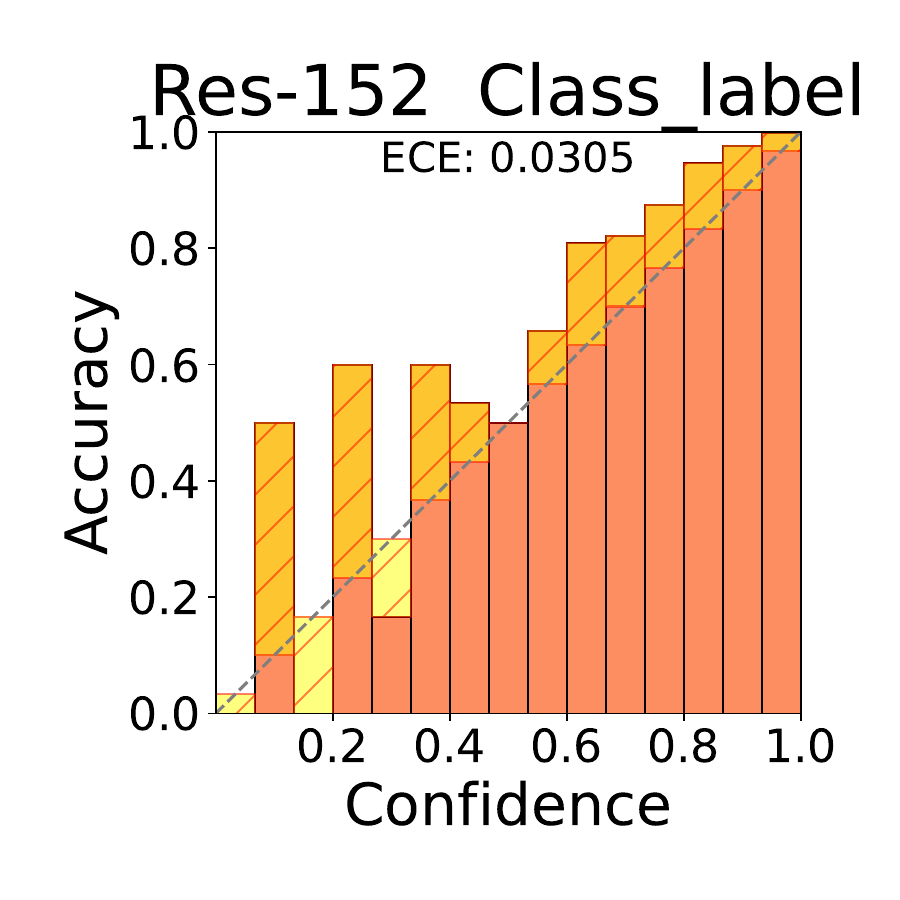}
\end{minipage}
\hfill
\begin{minipage}{0.19\textwidth}
  \centering
  \includegraphics[trim=  5mm 0mm 5mm 5mm, clip, width=\linewidth , keepaspectratio]{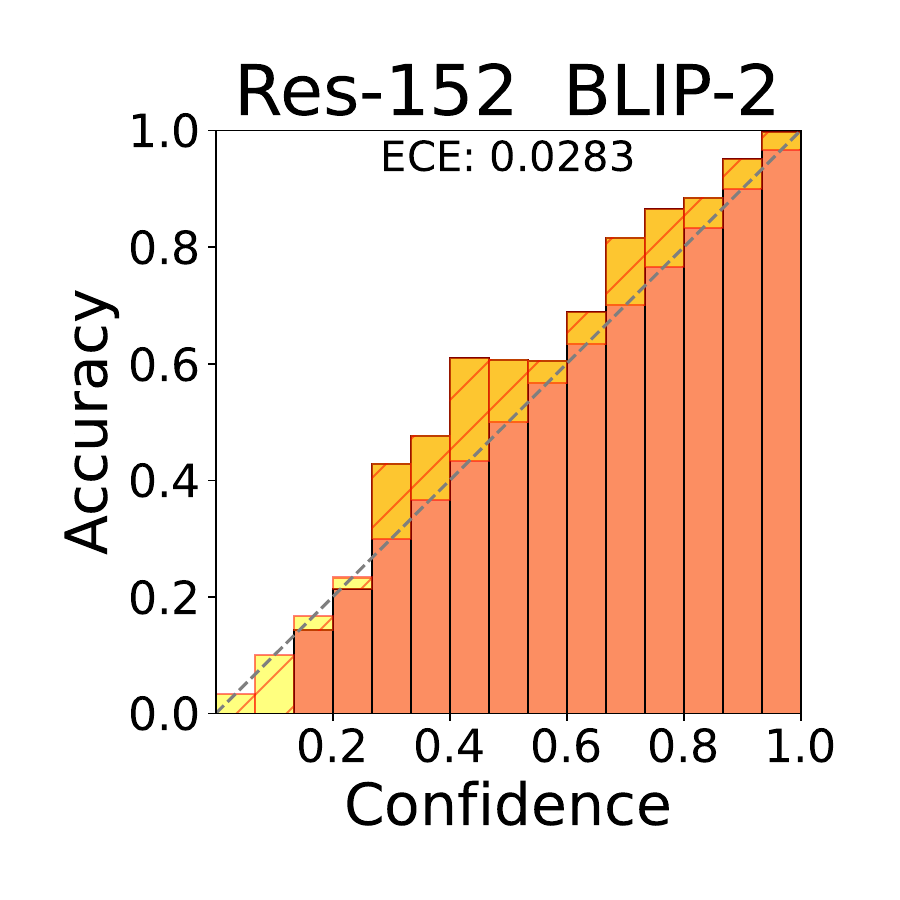}
\end{minipage}
\hfill
\begin{minipage}{0.19\textwidth}
  \centering
  \includegraphics[trim=  5mm 0mm 5mm 5mm, clip, width=\linewidth , keepaspectratio]{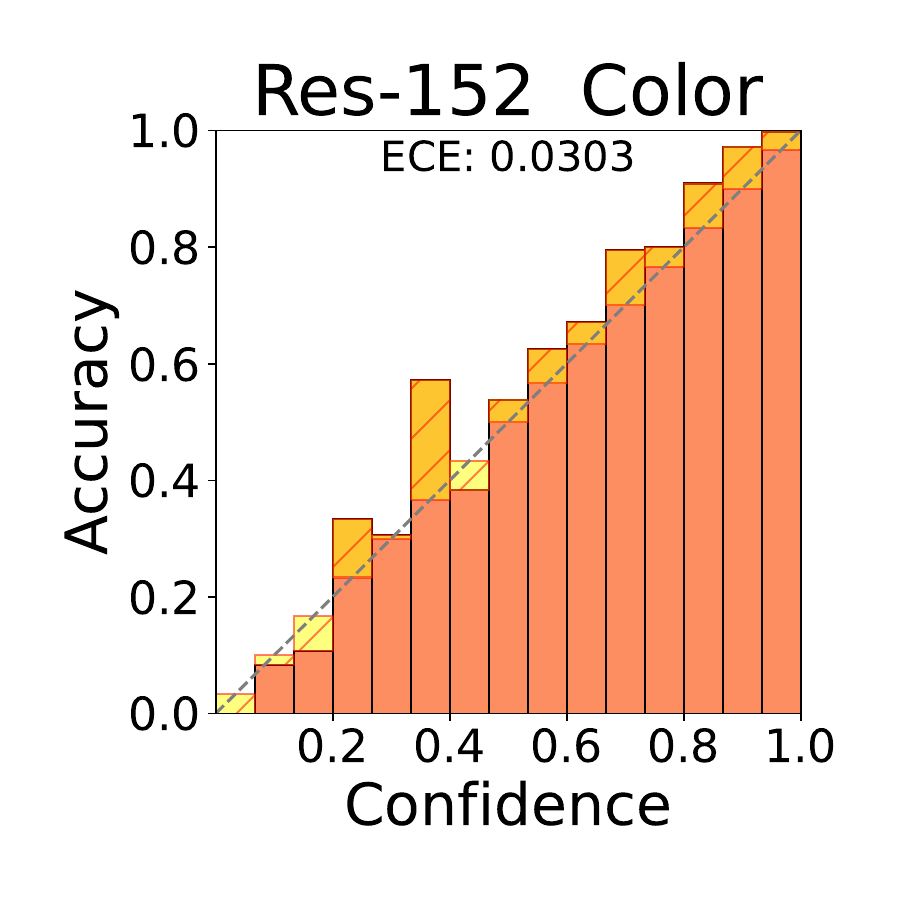}
\end{minipage}
\hfill
\begin{minipage}{0.19\textwidth}
  \centering
  \includegraphics[trim=  5mm 0mm 5mm 5mm, clip, width=\linewidth , keepaspectratio]{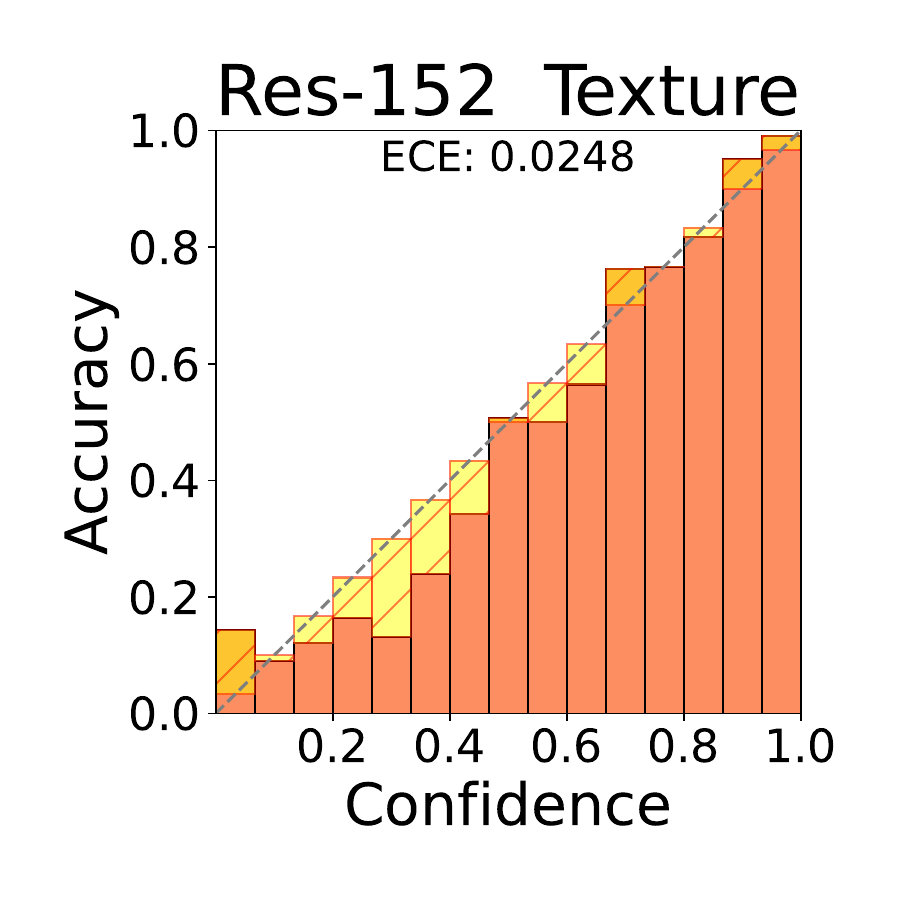}
\end{minipage}

\end{minipage}
\caption{Calibration results comparison of CNN-based models.}
\label{fig:calib_cnn}
\end{figure}

\begin{figure}
\begin{minipage}{\textwidth}


\begin{minipage}{0.189\textwidth}
  \centering
  \includegraphics[ trim=  5mm 0mm 5mm 5mm, clip, width=\linewidth , keepaspectratio]{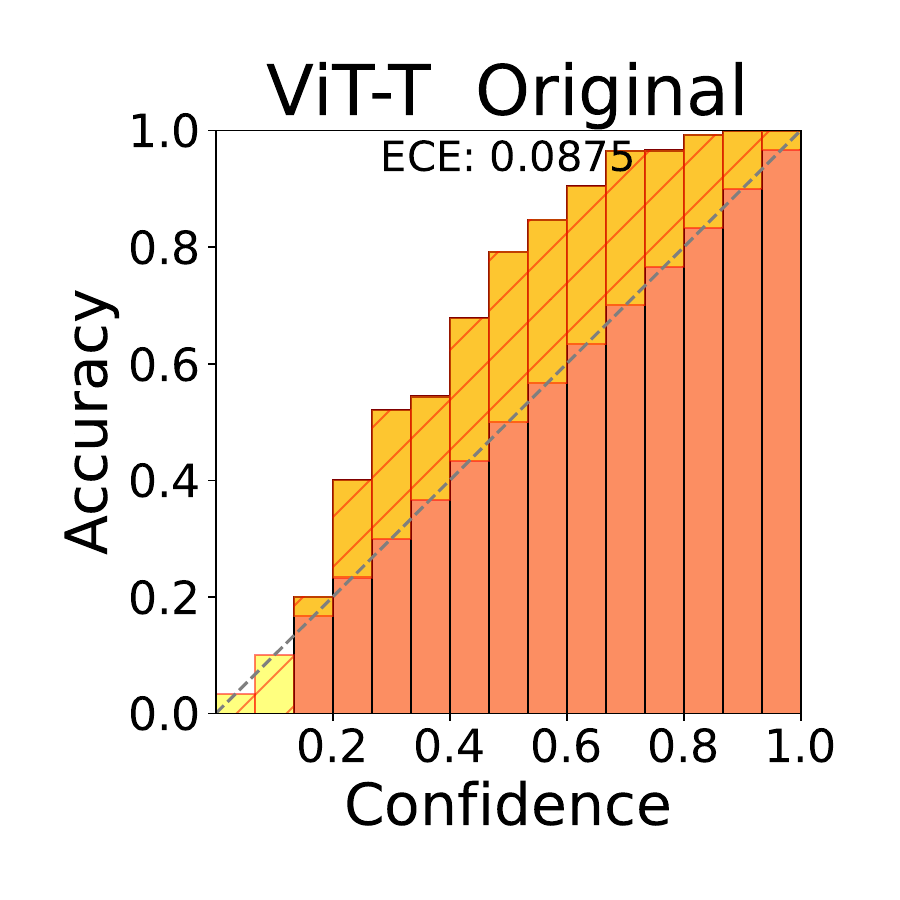}
\end{minipage}
\begin{minipage}{0.189\textwidth}
  \centering
  \includegraphics[trim=  5mm 0mm 5mm 5mm, clip, width=\linewidth , keepaspectratio]{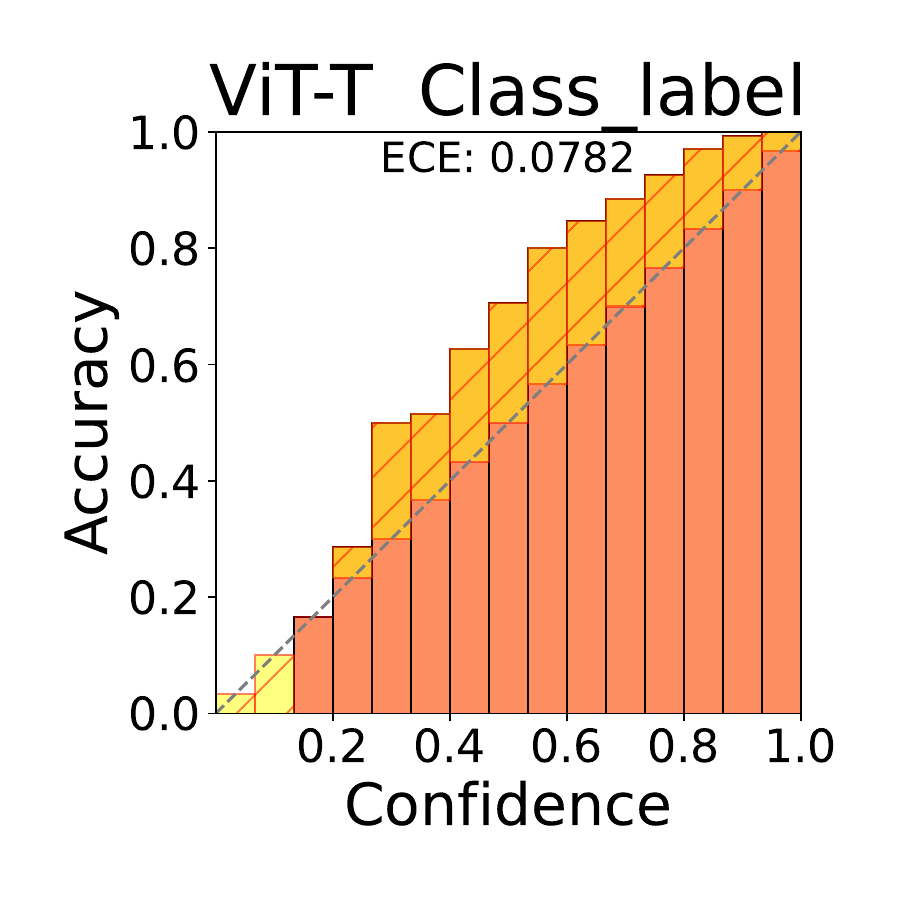}

\end{minipage}
\begin{minipage}{0.189\textwidth}
  \centering
  \includegraphics[trim=  5mm 0mm 5mm 5mm, clip, width=\linewidth , keepaspectratio]{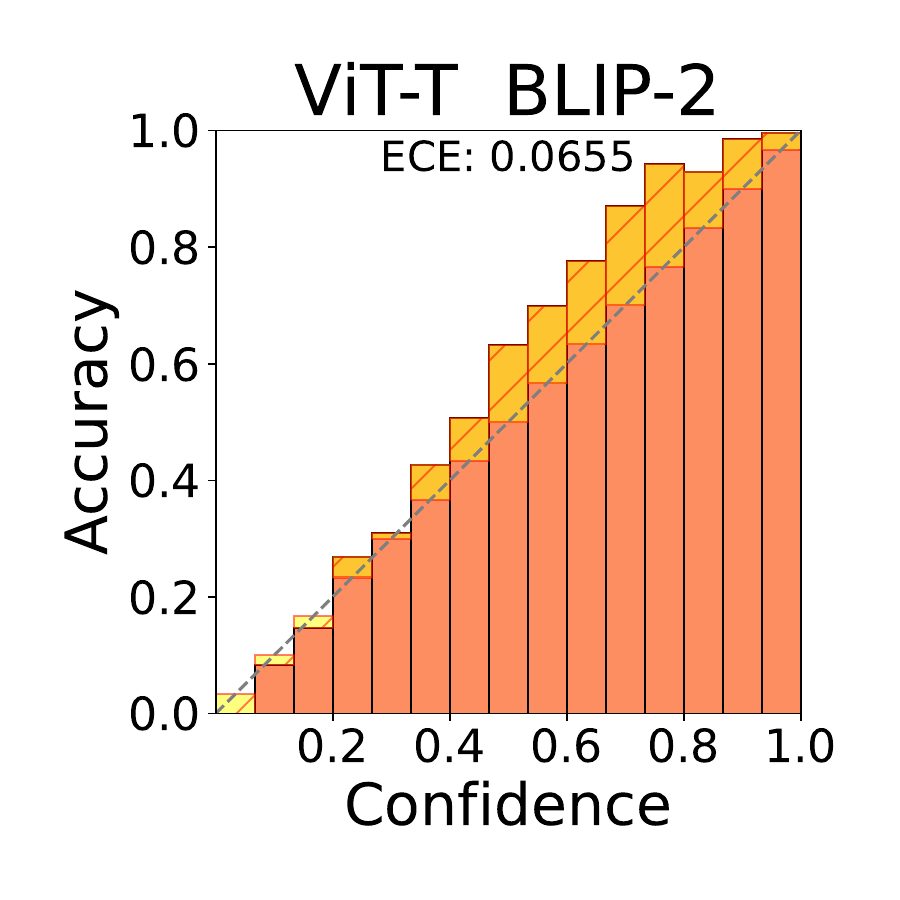}

\end{minipage}
\begin{minipage}{0.189\textwidth}
  \centering
  \includegraphics[trim=  5mm 0mm 5mm 5mm, clip, width=\linewidth , keepaspectratio]{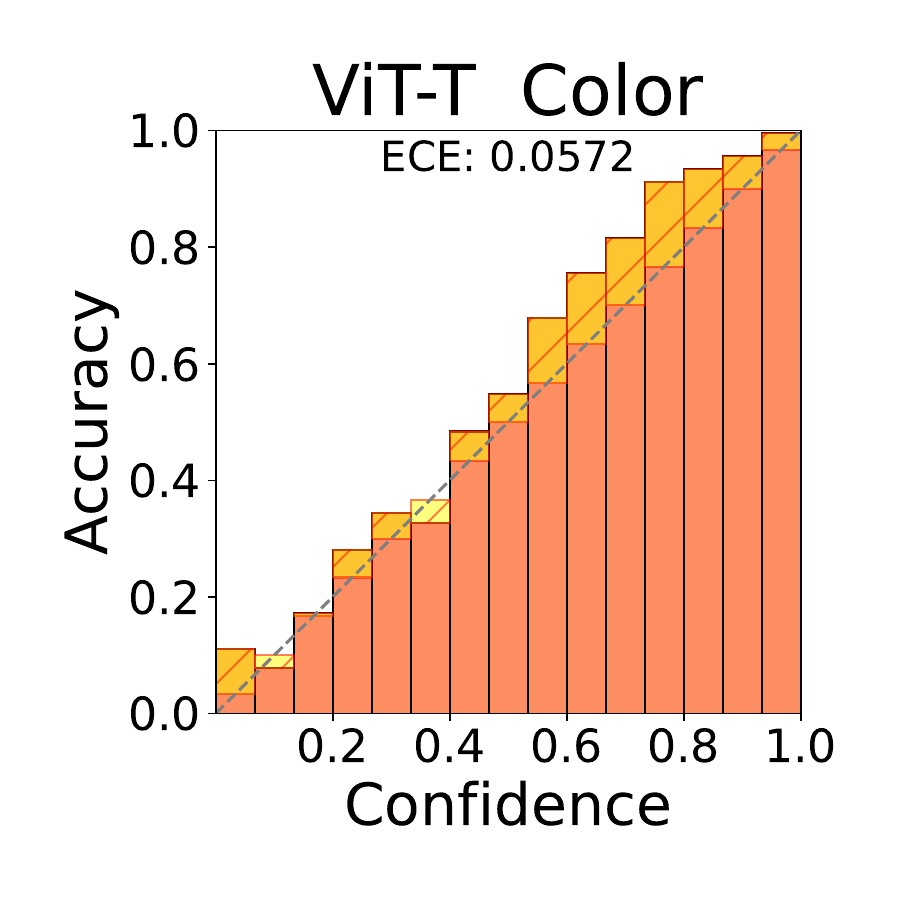}
\end{minipage}
\begin{minipage}{0.189\textwidth}
  \centering
  \includegraphics[trim=  5mm 0mm 5mm 5mm, clip, width=\linewidth , keepaspectratio]{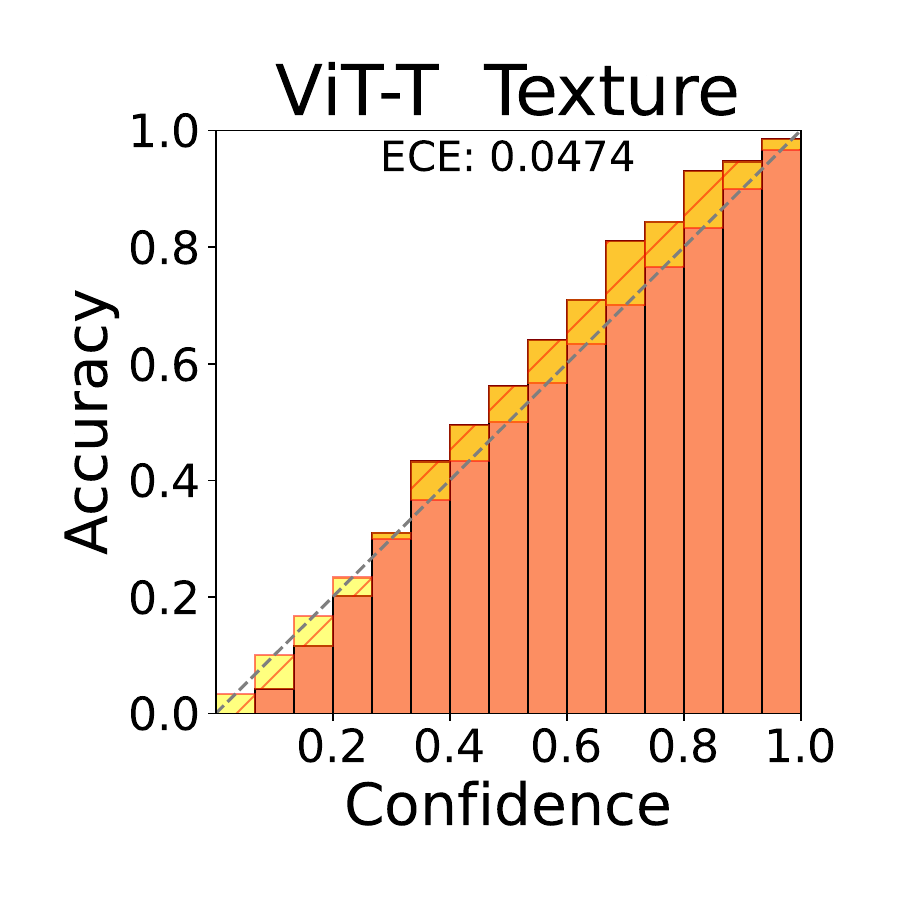}

\end{minipage}

\begin{minipage}{0.189\textwidth}
  \centering
  \includegraphics[ trim=  5mm 0mm 5mm 5mm, clip, width=\linewidth , keepaspectratio]{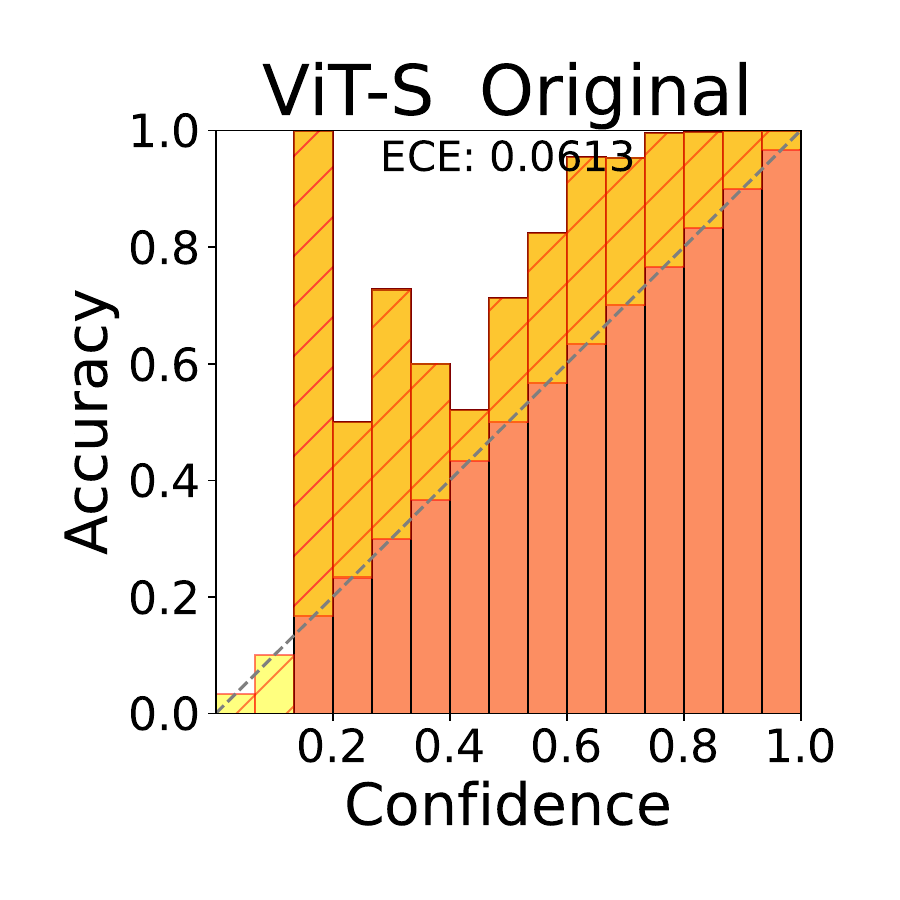}
\end{minipage}
\begin{minipage}{0.189\textwidth}
  \centering
  \includegraphics[trim=  5mm 0mm 5mm 5mm, clip, width=\linewidth , keepaspectratio]{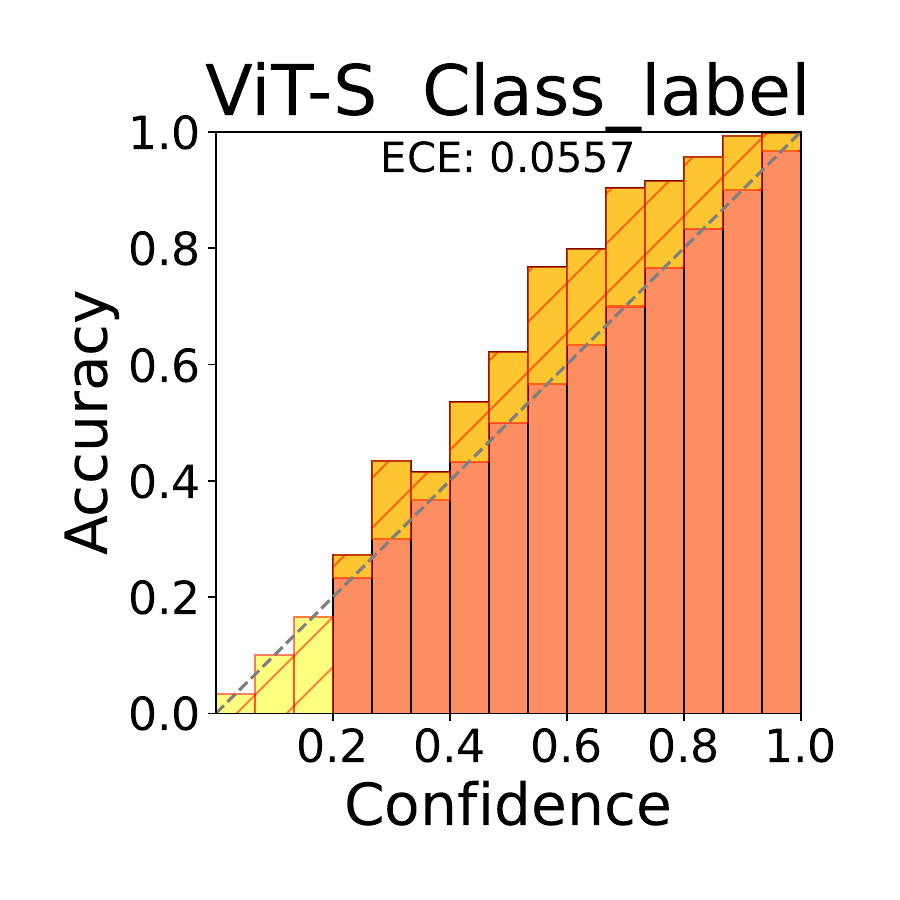}

\end{minipage}
\begin{minipage}{0.189\textwidth}
  \centering
  \includegraphics[trim=  5mm 0mm 5mm 5mm, clip, width=\linewidth , keepaspectratio]{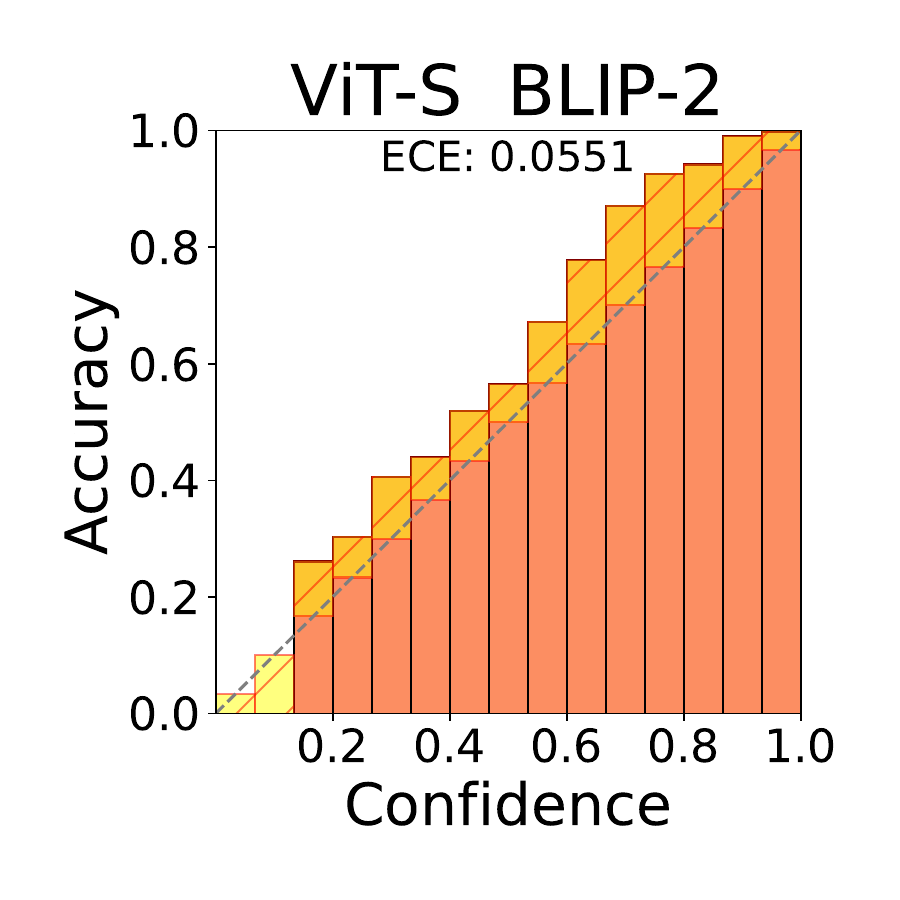}

\end{minipage}
\begin{minipage}{0.189\textwidth}
  \centering
  \includegraphics[trim= 5mm 0mm 5mm 5mm, clip, width=\linewidth , keepaspectratio]{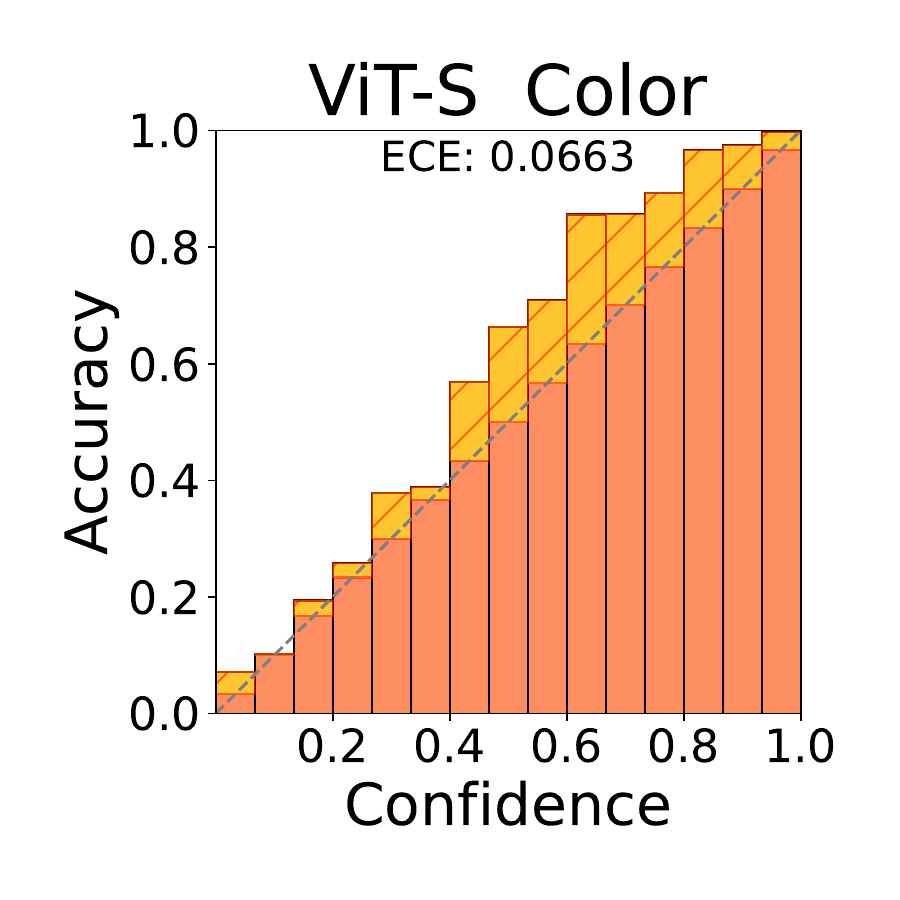}
\end{minipage}
\begin{minipage}{0.189\textwidth}
  \centering
  \includegraphics[trim=  5mm 0mm 5mm 5mm, clip, width=\linewidth , keepaspectratio]{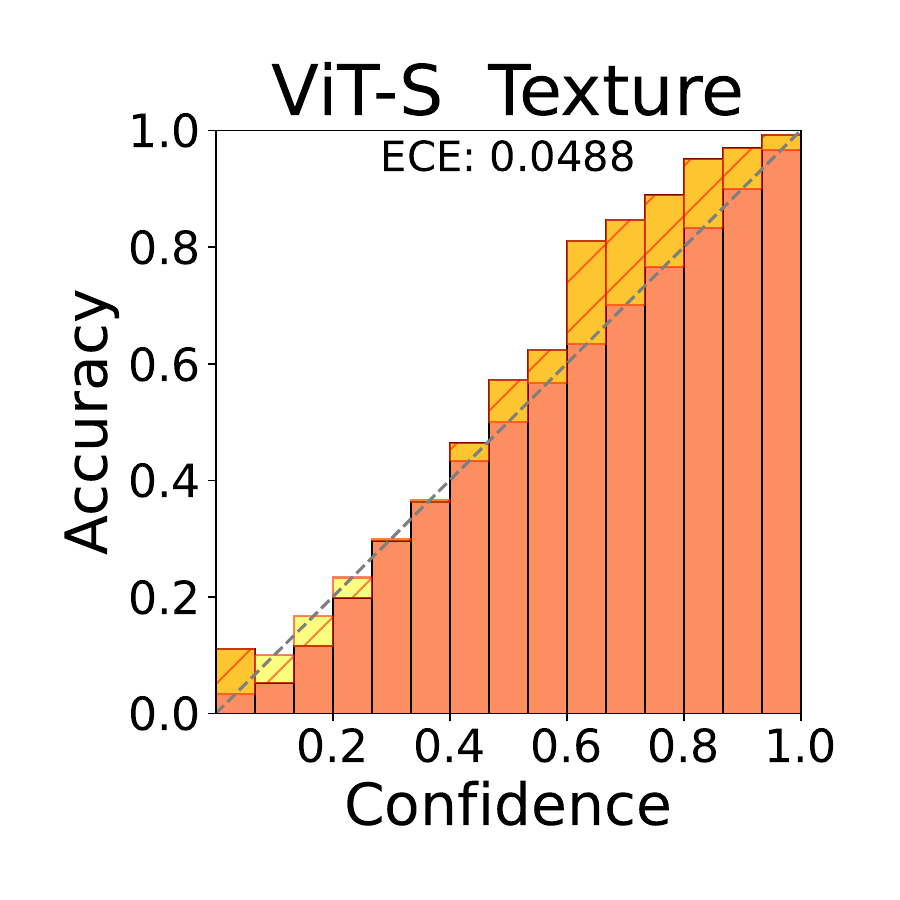}

\end{minipage}

\begin{minipage}{0.189\textwidth}
  \centering
  \includegraphics[ trim=  5mm 0mm 5mm 5mm, clip, width=\linewidth , keepaspectratio]{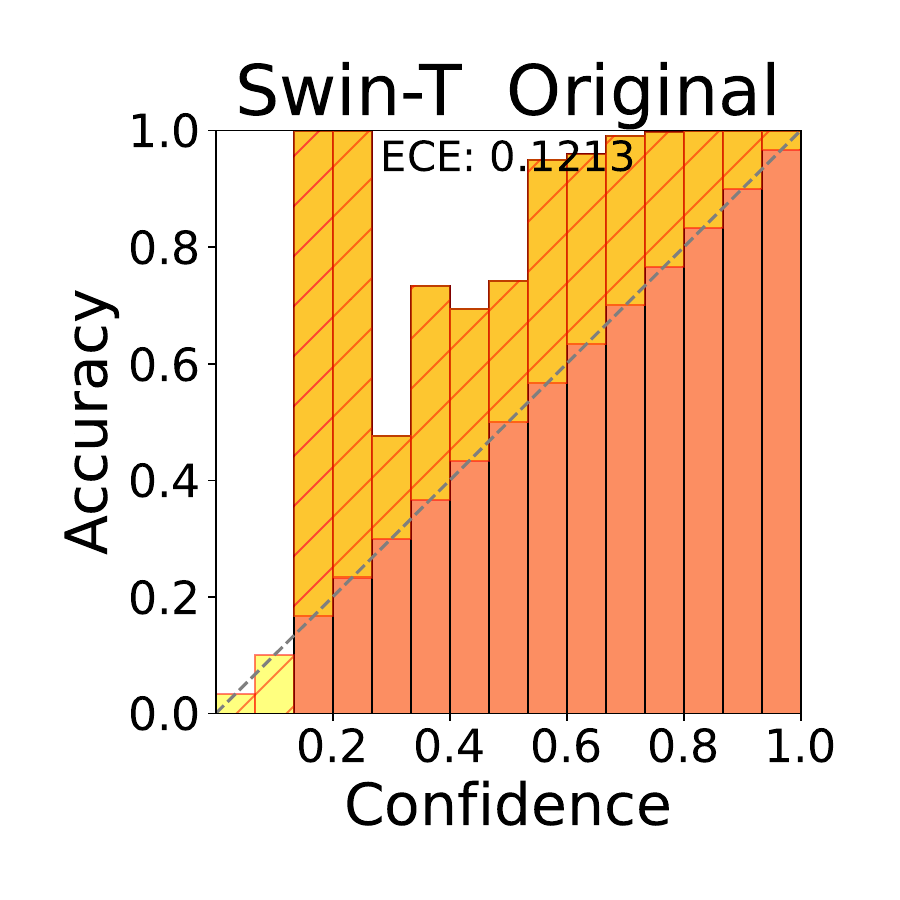}
\end{minipage}
\begin{minipage}{0.189\textwidth}
  \centering
  \includegraphics[trim=  5mm 0mm 5mm 5mm, clip, width=\linewidth , keepaspectratio]{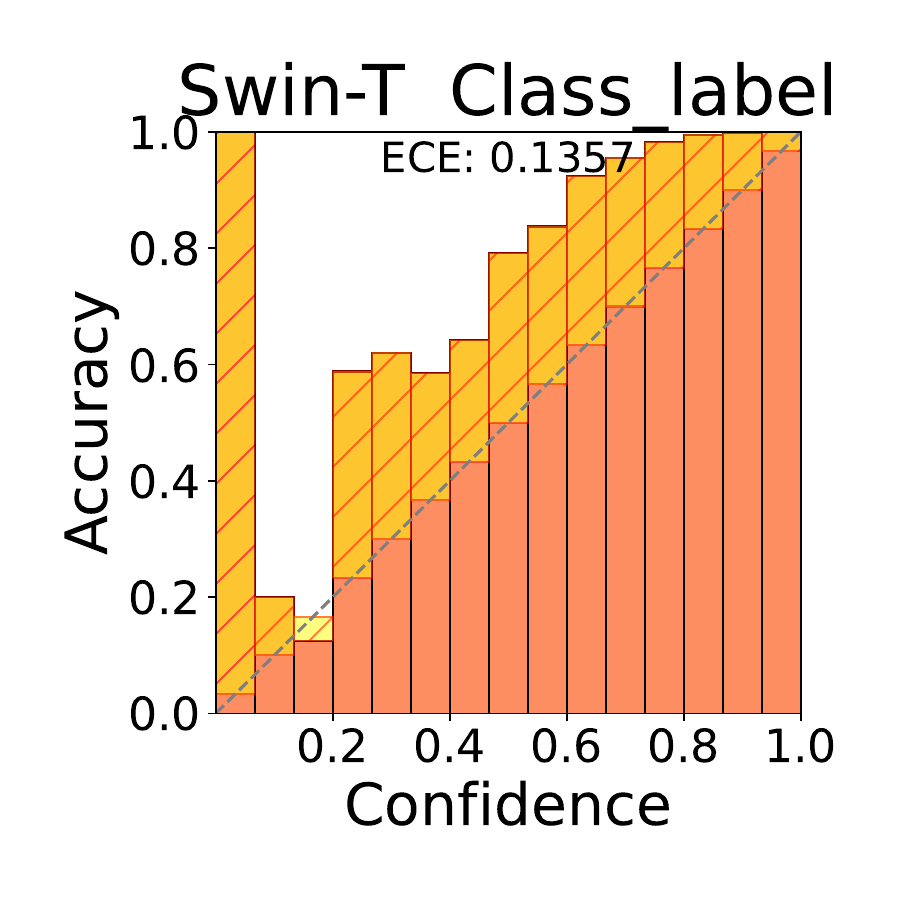}

\end{minipage}
\begin{minipage}{0.189\textwidth}
  \centering
  \includegraphics[trim=  5mm 0mm 5mm 5mm, clip, width=\linewidth , keepaspectratio]{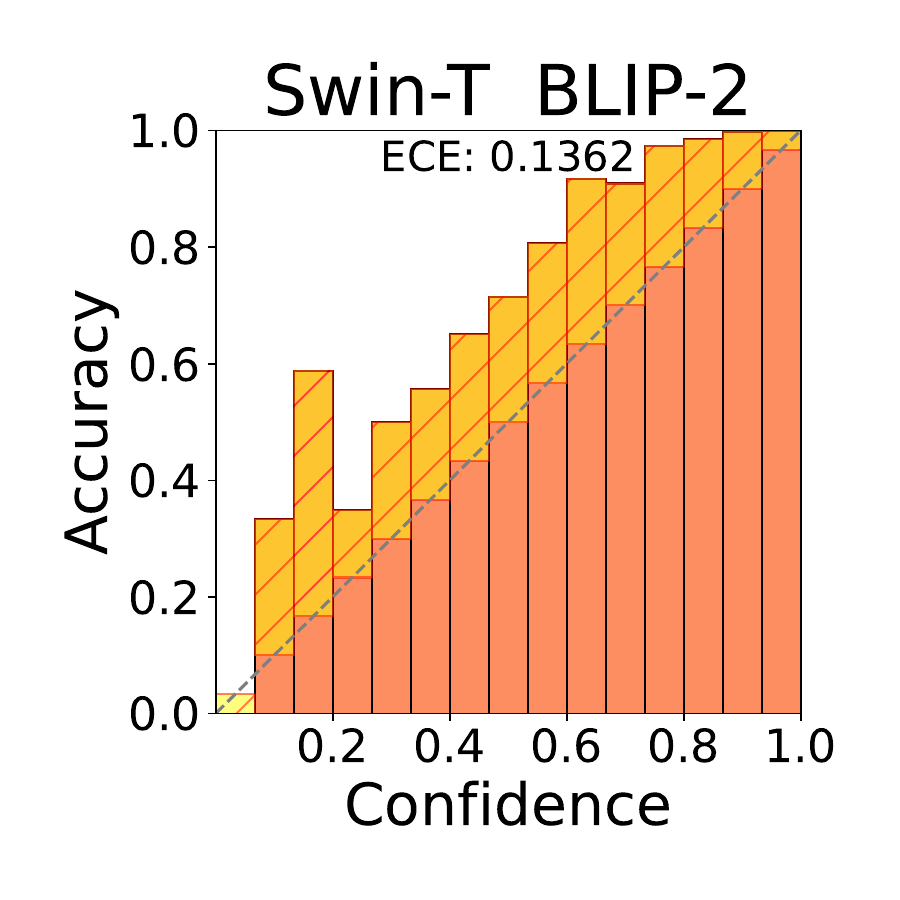}

\end{minipage}
\begin{minipage}{0.189\textwidth}
  \centering
  \includegraphics[trim=  5mm 0mm 5mm 5mm, clip, width=\linewidth , keepaspectratio]{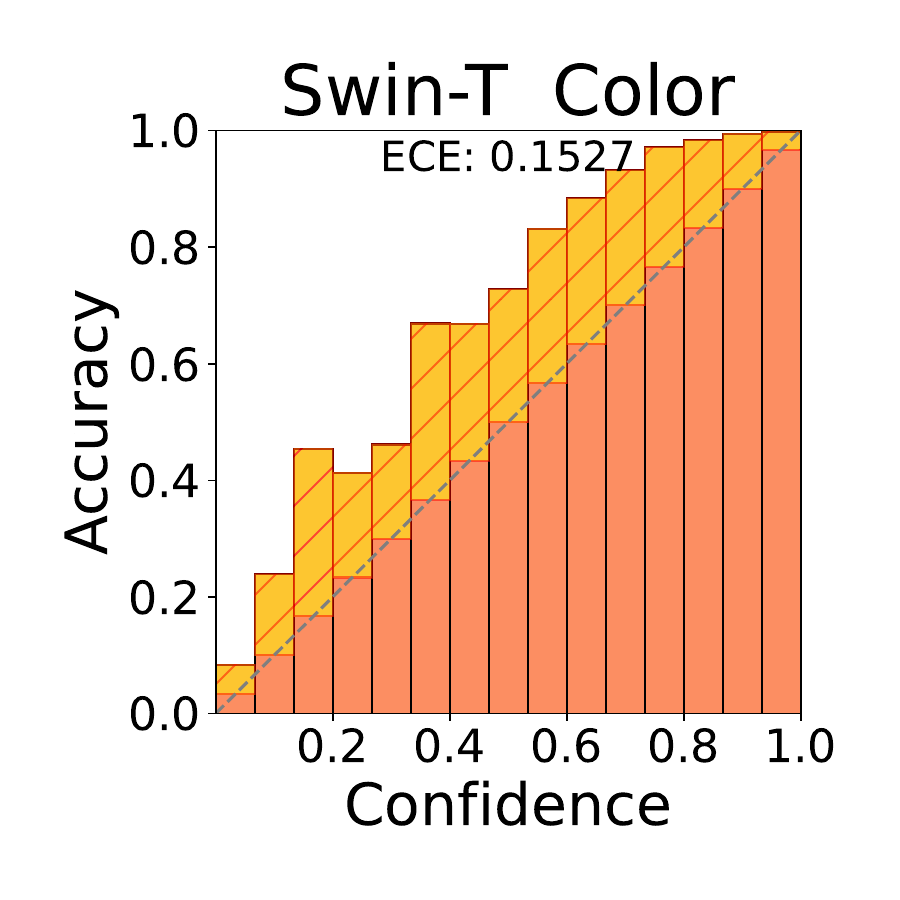}
\end{minipage}
\begin{minipage}{0.185\textwidth}
  \centering
  \includegraphics[trim= 5mm 0mm 5mm 5mm, clip, width=\linewidth , keepaspectratio]{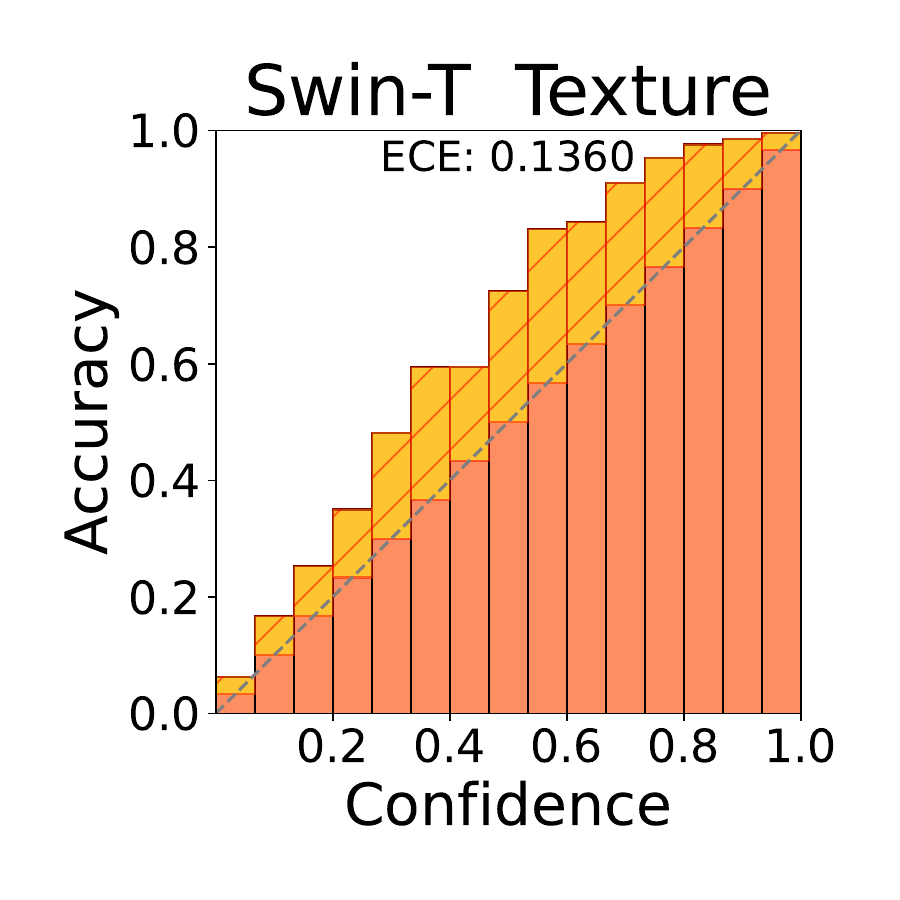}

\end{minipage}

\begin{minipage}{0.189\textwidth}
  \centering
  \includegraphics[ trim=  5mm 0mm 5mm 5mm, clip, width=\linewidth , keepaspectratio]{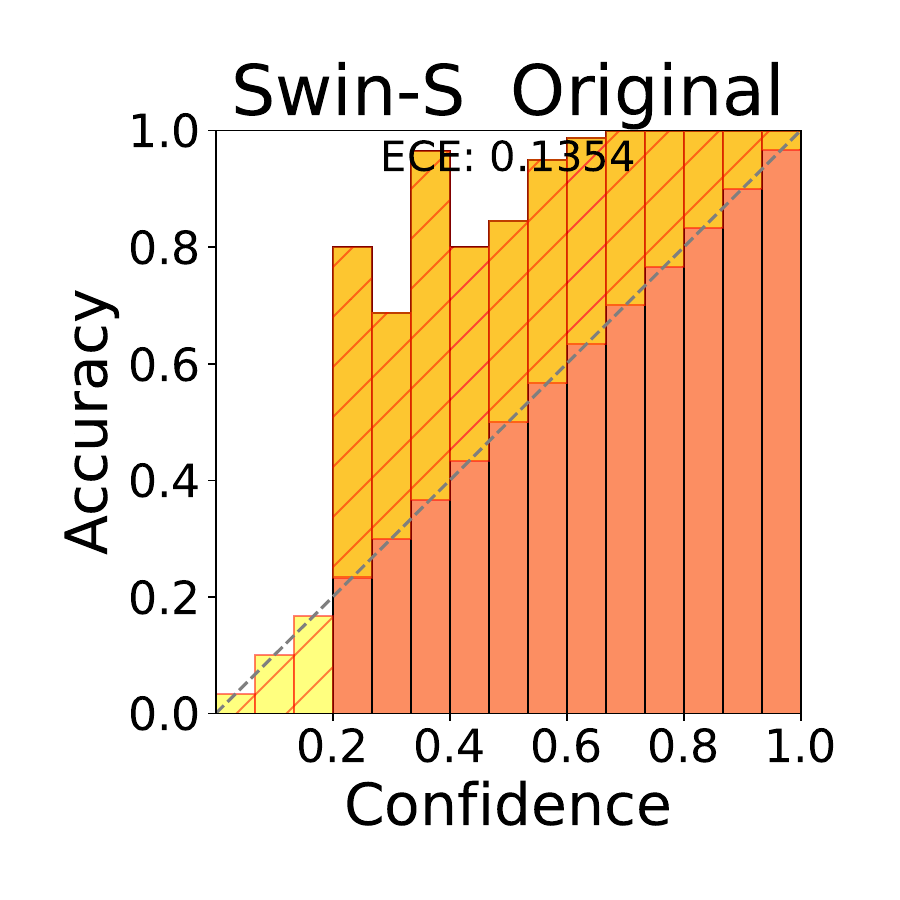}
\end{minipage}
\begin{minipage}{0.189\textwidth}
  \centering
  \includegraphics[trim=  5mm 0mm 5mm 5mm, clip, width=\linewidth , keepaspectratio]{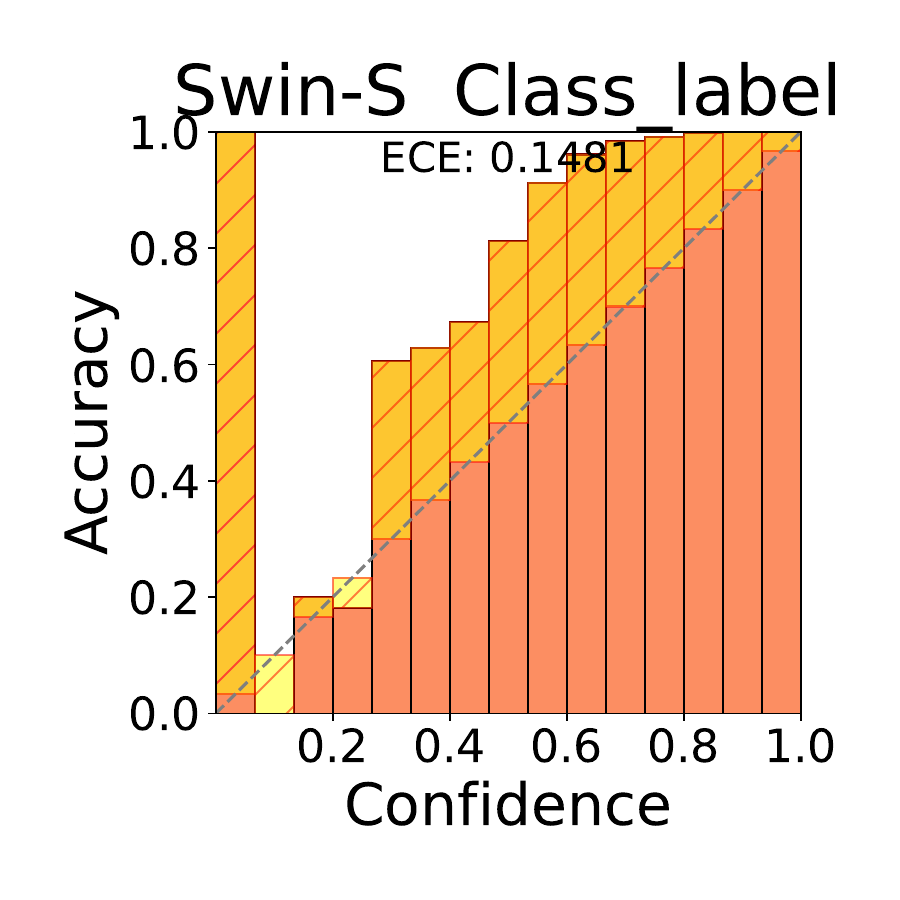}

\end{minipage}
\begin{minipage}{0.189\textwidth}
  \centering
  \includegraphics[trim=  5mm 0mm 5mm 5mm, clip, width=\linewidth , keepaspectratio]{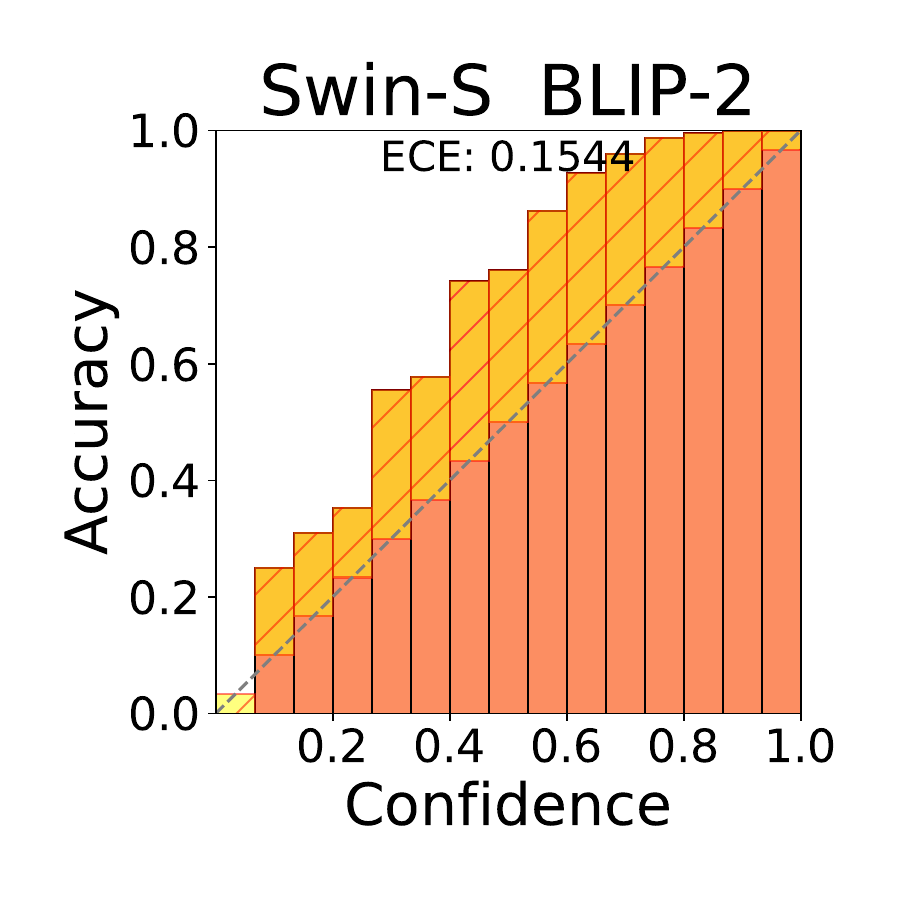}

\end{minipage}
\begin{minipage}{0.189\textwidth}
  \centering
  \includegraphics[trim=  5mm 0mm 5mm 5mm, clip, width=\linewidth , keepaspectratio]{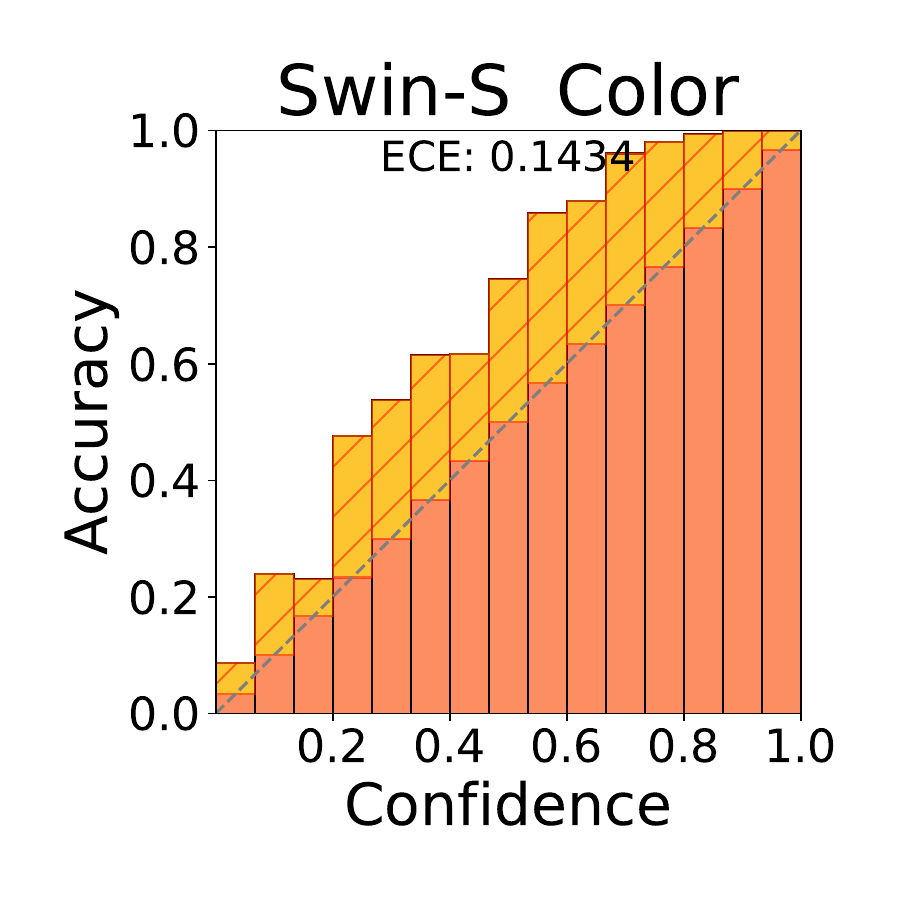}
\end{minipage}
\begin{minipage}{0.189\textwidth}
  \centering
  \includegraphics[trim=  5mm 0mm 5mm 5mm, clip, width=\linewidth , keepaspectratio]{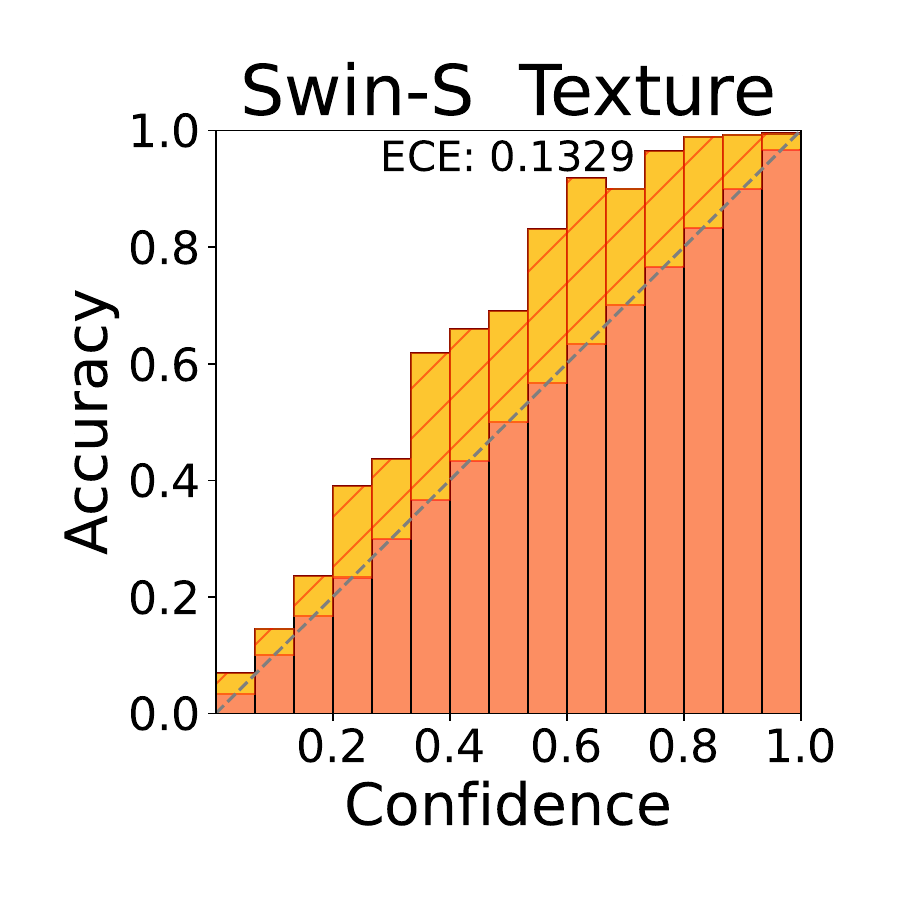}

\end{minipage}

\end{minipage}

\hfill
  \caption{Caliberation results comparison of ViT model}
  \label{fig:calib_transf}
\end{figure}

Model calibration refers to how well a model's predicted confidence levels correspond to its actual accuracy. Confidence represents the probability a model assigns to its predictions, while accuracy measures how often those predictions are correct. For example, if a model predicts with 70\% confidence, a well-calibrated model should have an actual accuracy close to 70\%. To quantify this, we use the Expected Calibration Error (ECE). ECE works by dividing the predictions into \( M \) bins based on confidence intervals (e.g., 60\%-70\%, 70\%-80\%). Within each bin, the average confidence and accuracy are calculated, and the ECE is the weighted average of the differences between these values. To visually evaluate model calibration, reliability diagrams are used. These diagrams plot predicted confidence against actual accuracy, allowing us to compare different models. A well-calibrated model will show points that lie close to the diagonal on these plots.  In Figures \ref{fig:calib_cnn} and \ref{fig:calib_transf}, we plot the reliability diagrams for different convolutional and transformer-based models, respectively.

\FloatBarrier

\subsection{Ablation on Adversarial Loss}
\label{sec:adv_loss_ablation}
Tab.\ref{loss} presents an ablation study on adversarial loss, showing optimizing both vision \& text based emebedding, leads to the most performance drop.

\begin{table}[h]
    \caption{Ablation on different losses}
    \small \centering
 \setlength{\tabcolsep}{8pt}
    \scalebox{0.60}[0.65]{
    \begin{tabular}{lcccccccc}
    \toprule
      & ViT-T & ViT-S & Swin-T & Swin-S & Res-50 &Res-152& Dense-161 & Average \\
    \midrule
\textit{Text} & 28.4 & 46.5 & 40.9 & 47.2 & 19.0  & 46.5 & 31.5 & \textbf{37.1} \\
\textit{Latent} & 32.3 & 47.5 & 42.8 & 50.2 & 1.6  & 44.5 & 26.4 & \textbf{35.1} \\
\textit{Combined} & 18.4 & 32.1 & 25.0 & 31.7 & 2.0  & 28 & 14.4 & \textbf{21.6} \\

    \bottomrule
    \end{tabular}}
         
    \label{loss}
\end{table}

\subsection{Reproducibility and Ethics Statement}
\label{sec:statements}

\textbf{Reproducibility Statement}:
Our method uses already available pre-trained models and the codebase is based on several open source implementations.
 We highlight the main components used in our framework for reproducing the results presented in our paper,
  \texttt{\textbf{a) Diffusion Inpainting Implementation:}} We use the open-source implementation of Stable-Diffusion-Inpainting method (\url{https://github.com/huggingface/diffusers/blob/main/src/diffusers/}) with available pretrained weights \textit{(Stable-Diffusion-v-1-2)} for background generation. 
    \texttt{\textbf{b) Image-to-Segment Implementation:}} We use the official open-source implementation of FastSAM (\url{https://github.com/CASIA-IVA-Lab/FastSAM}) to get the segmentation masks of filtered ImageNet dataset. 
      \texttt{\textbf{c) Image-to-Text Implementation:}} We use the official open-source implementation of BLIP-2(\url{https://github.com/salesforce/LAVIS/tree/main/projects/blip2}) to get the captions for each image. We will also provide captions for each image in our dataset.
            \texttt{\textbf{d) Adversarial Attack:}} We intent to open-source our codebase and release the script for crafting adversarial examples.   \texttt{\textbf{e) Dataset:}} In the paper, we describe the procedure of filtering the images from ImageNet and COCO val. set. Furthermore, we will  provide the filtered datasets, object masks as well as prompts used to generate different backgrounds.

\textbf{Ethics Statement}: Our work focuses on evaluating resilience of current vision and language models against natural and adversarial background changes in real images. This work can be utilized by an attacker to generate malicious backgrounds on real images as well as generate adversarial backgrounds which can fool the deployed computer-vision systems. Nevertheless, we believe that our research will pave the way for improved evaluation protocols to assess the resilience of existing models. This, in turn, is likely to drive the development of enhanced techniques for bolstering the resilience of deployed systems. Since we are benchmarking vision and vision-language models using a subset of images from publicly available ImageNet and COCO datasets, it's
relevant to mention that these datasets are known to have images of people which poses a privacy risk and further it is known to have biases which can encourage social stereotypes. In the future, we intend to benchmark our models on a less biased dataset to mitigate these concerns and ensure a fair evaluation.

\end{document}

%% file: math_commands.tex
\usepackage{amsmath,amsfonts,bm}

\def\eqref#1{equation~\ref{#1}}

\def\1{\bm{1}}

\DeclareMathAlphabet{\mathsfit}{\encodingdefault}{\sfdefault}{m}{sl}
\SetMathAlphabet{\mathsfit}{bold}{\encodingdefault}{\sfdefault}{bx}{n}

